\documentclass[Afour,sageh,times]{sagej}

\usepackage{moreverb,url}
\usepackage[colorlinks,bookmarksopen,bookmarksnumbered,citecolor=red,urlcolor=red]{hyperref}

\newcommand\BibTeX{{\rmfamily B\kern-.05em \textsc{i\kern-.025em b}\kern-.08em
T\kern-.1667em\lower.7ex\hbox{E}\kern-.125emX}}
\def\volumeyear{2016}

\usepackage{subcaption}	       								  %% For sub-figures
\usepackage[nameinlink,capitalise]{cleveref}                  %% For ranges of citations
\crefname{figure}{}{}                                         %% To remove extra "Figs." in cref ranges of figure references

\usepackage{tabularx}         								  %% For adaptive column widths (X)
\usepackage{multirow}          								  %% For merging columns/rows
\usepackage{colortbl}          								  %% For colored cells
\newcolumntype{Y}{>{\centering\arraybackslash}X}              %% Adaptive-width and centered columns
\newcommand{\lightgraycell}{\cellcolor[cmyk]{0, 0, 0, 0.12}}  %% Named cell color
\usepackage{tcolorbox}         								  %% For colored box
\usepackage{bm}                								  %% For boldface for some math symbols
\usepackage[frozencache=true,cachedir=minted-cache]{minted}            								  %% For Python code listings
\newcommand{\vect}[1]{\boldsymbol{\mathbf{#1}}}               %% For bold math symbols

\begin{document}

\runninghead{Abdelrahman et al.}
\title{A Neuromorphic Approach to Obstacle Avoidance in Robot Manipulation}

\author{Ahmed Faisal Abdelrahman\affilnum{1}, Matias Valdenegro-Toro\affilnum{2}, Maren Bennewitz\affilnum{3} and Paul G. Plöger\affilnum{1}}

\affiliation{\affilnum{1} Bonn-Rhein-Sieg University, Germany\\
\affilnum{2} University of Groningen, Netherlands\\
\affilnum{3} University of Bonn, Germany}

\corrauth{Ahmed Abdelrahman,
Munich Institute of Robotics and Machine Intelligence,
Technical University of Munich,
80992 Munich,
Germany}

\email{ahmed.abdelrahman@outlook.de}

\begin{abstract}
Neuromorphic computing mimics computational principles of the brain \textit{in silico} and motivates research into event-based vision and spiking neural networks (SNNs). Event cameras (ECs) exclusively capture local intensity changes and offer superior power consumption, response latencies, and dynamic ranges. SNNs replicate biological neuronal dynamics and have demonstrated potential as alternatives to conventional artificial neural networks (ANNs), such as in reducing energy expenditure and inference time in visual classification. Nevertheless, these novel paradigms remain scarcely explored outside the domain of aerial robots.

To investigate the utility of brain-inspired sensing and data processing, we developed a neuromorphic approach to obstacle avoidance on a camera-equipped manipulator. Our approach adapts high-level trajectory plans with reactive maneuvers by processing emulated event data in a convolutional SNN, decoding neural activations into avoidance motions, and adjusting plans using a dynamic motion primitive. We conducted experiments with a Kinova Gen3 arm performing simple reaching tasks that involve obstacles in sets of distinct task scenarios and in comparison to a non-adaptive baseline.

Our neuromorphic approach facilitated reliable avoidance of imminent collisions in simulated and real-world experiments, where the baseline consistently failed. Trajectory adaptations had low impacts on safety and predictability criteria. Among the notable SNN properties were the correlation of computations with the magnitude of perceived motions and a robustness to different event emulation methods. Tests with a DAVIS346 EC showed similar performance, validating our experimental event emulation. Our results motivate incorporating SNN learning, utilizing neuromorphic processors, and further exploring the potential of neuromorphic methods.

\end{abstract}


\maketitle

\section{Introduction}
\label{introduction}

%\paragraph{}
Modern autonomous systems excel at specific tasks, but lack capabilities that the average human exemplifies, including rapidly learning for and accomplishing various unstructured tasks.
A potential reason is the fundamental differences in human and artificial intelligence (AI) due to their respective physical substrates (biological vs. silicon-based) (\cite{korteling2021human}).
AI systems can have superior information propagation speeds, communication bandwidths, and raw computational power.
Though this may imply a greater capacity for multi-sensory processing, robotic agents embodying AI still struggle with everyday tasks that our brains facilitate with relative ease.
This poses the question of whether the information propagation mechanisms and computational architectures present in our brains could be key factors in the advancement of reliable autonomy.

%\paragraph{}
The human brain maintains multiple, complex cognitive processes while being more energy-efficient than contemporary computers.
It simultaneously regulates essential processes and controls numerous high-level functions while consuming $\sim$20W of power (\cite{drubach2000brain}).
Comparable and less-capable computers require power inputs that are many orders of magnitudes higher.
This is evident when simulating cortical neural networks on conventional computers; at the scale of a mouse, such a simulation could run at 40,000 more power and 9,000 times less speed, while projections from the Human Brain Project underscore the colossal power requirement for simulating a human brain (\cite{thakur2018large}). 
Such neuroscientific studies indicate the vastly superior performance-to-efficiency ratio of the brain and motivate the consideration of biologically-inspired circuitry, sensors, and algorithms in intelligent robot design, which is the aim of this work.

%\paragraph{}
Biological inspiration has frequently driven practical innovations, such as IR detectors and gyroscopes (\cite{wicaksono2008learning}), learning paradigms such as evolutionary algorithms and reinforcement learning (RL) (\cite{sutton1998introduction}), and locomotion control using elementary neural circuits (\cite{ijspeert2008central}).
Artificial neural networks (ANNs) are based on their biological counterparts: the earliest models consist of interconnected layers of simple computational units with adjustable synaptic connections, while later convolutional neural networks (CNNs) aim to mimic connectivity patterns observed in animal visual cortices.

%\paragraph{}
CNNs have been largely successful in some visual processing tasks (see \cite{goodfellow2016deep}) and are commonly deployed on robots.
Nevertheless, these models are crude approximations at best.
Biological neurons asynchronously aggregate inputs over time and propagate discrete, sparse spikes (action potentials), whose precise timings are thought to encode useful information.
Conversely, ANN neurons synchronously propagate real-valued signals, abandoning an additional temporal dimension afforded by relative spike timings.
These discrepancies have become relevant following observations that deep neural networks (DNNs), despite their notable successes, still suffer from ever-growing numbers of parameters, correspondingly sizable data requirements, poor generalization to unobserved yet similar inputs, and catastrophic failures in response to minor perturbations (\cite{serre2019deep}).
Our brains seem better-equipped to handle an extraordinary range of problems while exhibiting superior generalization and robustness than present AI models.

\subsection{Neuromorphic Computing}
\label{introduction:neuromorphic_computing}

%\paragraph{}
\textit{Neuromorphic engineering/computing} reproduces characteristics of the brain in hardware.
The field was established in the 1980s by Carver Mead, who suggested analogies between  neuronal dynamics and the physics of sub-threshold regions of transistor operation (\cite{mead1990neuromorphic}).
This has given rise to various models of \textit{spiking neural networks}, \textit{neuromorphic processors}, and neuromorphic sensors.
This research is motivated by the pursuit of brain-like computation to improve efficiency, parallelization, and energy consumption (\cite{thakur2018large}) through a fundamental paradigm shift, which could benefit various applications including autonomous vehicles, wearable devices, and IoT\endnote{IoT: Internet of Things} sensors (\cite{rajendran2019low}).
Naturally, it holds promise for robotic systems as well.

%\paragraph{}
Various studies provide empirical evidence of the advantages of neuromorphic computing: 10 and 1000 factor increases in speed and energy-efficiency when learning on a neuro-processor compared to a conventional processor (\cite{wunderlich2019demonstrating}), a four-fold increase in the energy-efficiency of a spiking network vs. a DNN for speech recognition (\cite{blouw2020event}), better energy-per-classification ratios in deep learning problems when comparing to a Tesla P100 GPU (\cite{goltz2021fast}), and speed and efficiency gains in various classical problems (\cite{davies2021advancing}). 
These results have coincided with, or perhaps fostered, an ongoing interest in the field, typified by the recent establishment of the "Neuromorphic Computing and Engineering" journal (\cite{indiveri2021introducing}).

\subsection{Event Cameras}
\label{introduction:event_cameras}

\begin{figure*}
    \centering
    \begin{subfigure}{0.24\linewidth}
        \centering	
        \includegraphics[width=1.\columnwidth]{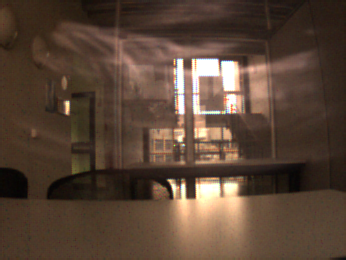}
        \caption{\centering Fast object (RGB)}
        \label{dvs_images_intro:fast_object_rgb}
    \end{subfigure}%
    \hspace{0.1em}
    \begin{subfigure}{0.24\linewidth}
        \centering	
        \includegraphics[width=1.\columnwidth]{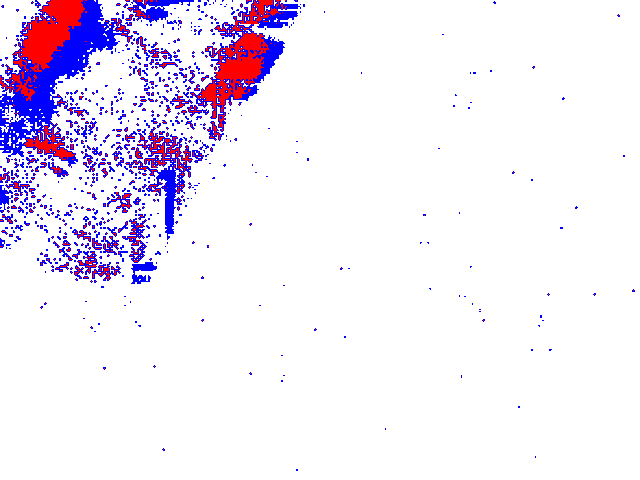}
        \caption{\centering Fast object (events)}
        \label{dvs_images_intro:fast_object_events}
    \end{subfigure}%
    \hspace{0.1em}
    \begin{subfigure}{0.24\linewidth}
        \centering
        \includegraphics[width=1.\columnwidth]{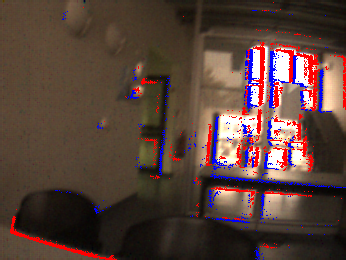}
        \caption{\centering Motion 1 (RGB + events)}
        \label{dvs_images_intro:ego_motion_hybrid_1}
    \end{subfigure}%
    \hspace{0.1em}
    \begin{subfigure}{0.24\linewidth}
        \centering	
        \includegraphics[width=1.\columnwidth]{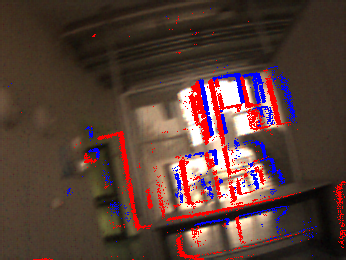}
        \caption{\centering Motion 2 (RGB + events)}
        \label{dvs_images_intro:ego_motion_hybrid_2}
    \end{subfigure}%
    \caption{Images captured with a DAVIS346 EC. \ref{dvs_images_intro:fast_object_rgb} shows an RGB image of a fast-moving object and \ref{dvs_images_intro:fast_object_events} visualizes the captured events \ref{dvs_images_intro:ego_motion_hybrid_1} and \ref{dvs_images_intro:ego_motion_hybrid_2} show events superimposed on images captured while moving the camera.
    }
    \label{dvs_images_intro}
\end{figure*}

%\paragraph{}
Event cameras (ECs) are neuromorphic sensors that are modelled after biological retinas.
ECs exclusively record per-pixel \textit{events} at which the change in intensity crosses a  threshold, mimicking retinal photoreceptor cells (\cite{posch2014retinomorphic}).
Consequently, they only capture significant intensity changes, often due to motion (Figure \ref{dvs_images_intro}), in contrast to frame-based cameras, whose pixels synchronously and continuously transmit absolute values, much of which is often redundant information.

%\paragraph{}
ECs offer lower power consumption, lower transmission latencies, higher dynamic ranges, and more robustness to motion blur than traditional cameras  (\cite{gallego2022event}, \cite{chen2020event}), which have been experimentally validated (\cite{sun2021autonomous}).
These properties can address limitations of frame-based vision in power consumption and bandwidth due to transmitting larger amounts of data (\cite{dubeau2020rgb}).
External clock-driven data acquisition naturally leads to redundant image information and the potential loss of inter-frame information (\cite{risi2020spike}).
Instead, ECs selectively acquire data based on scene dynamics.
ECs have most often been deployed in applications requiring rapid reaction speeds, such as drone flight, due to characteristically high temporal resolutions.

%\paragraph{}
Event-based data necessitates correspondingly novel methods and algorithms; the artificial analogs of visual cortical cells, spiking neurons, are prime candidates.

\subsection{Spiking Neural Networks}
\label{introduction:spiking_neural_networks}

\begin{figure}
    \centering
    \includegraphics[width=0.5\textwidth, trim={35pt 0 0 0},clip]{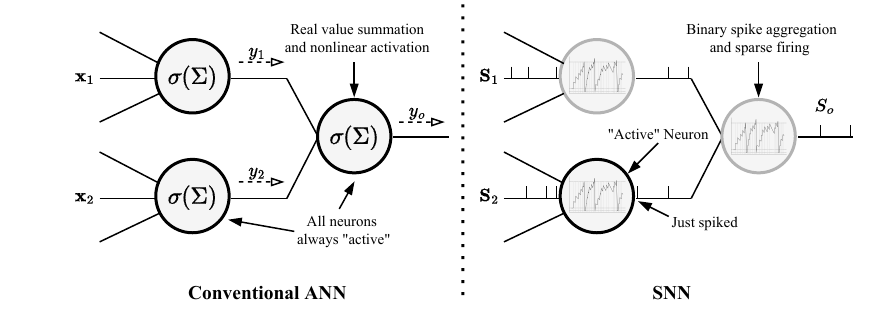}
    \caption{
    Neuronal dynamics in conventional ANNs (left) and SNNs (right). 
    Gray/black borders signify inactivity/activity; only the bottom SNN neuron is active here, since it had just spiked.}
    \label{snn_vs_ann_intro_illustration}
\end{figure}

%\paragraph{}
Spiking neural networks (SNNs) are a neuromorphic alternative to ANNs in which computational units propagate sparse sequences of \textit{spikes} (Figure \ref{snn_vs_ann_intro_illustration}).
Input spikes contribute to a neuron's decaying internal aggregate of past inputs: its \textit{membrane potential}.
When a threshold is exceeded, the neuron emits a spike (or \textit{action potential}) which resets its potential.
Spiking neurons operate asynchronously, such that information flows through the network through trains of distinctly-timed spikes.
These neuronal dynamics match those observed in biology (particularly in the primary visual cortex; \cite{chen2020event}), but raise challenges concerning the encoding/decoding and processing of spiking data as well as learning methods, which remain open areas of research.

%\paragraph{}
Similar to the event/frame-based camera dichotomy, SNNs have stimulated interest due to their potential advantages over ANNs.
SNNs can be as, or potentially more, expressive than ANNs (\cite{maass1997networks}), and successful applications in common visual tasks have shown that they can consume less power and exhibit faster classification inferences (\cite{neil2016learning}), as well as outperform ANNs in energy-delay-product\endnote{A measure of energy efficiency that considers inference delays.} (\cite{davies2021advancing}).
This energy-efficiency is due to neurons emitting outputs only when significantly stimulated, as opposed to constant computations, though this advantage is fully realized with dedicated neuromorphic hardware.
This is also reflected in response latencies; in classification problems, SNNs could produce a correct inference before a frame-based approach could fully process all input pixels (\cite{neil2016learning}).

%\paragraph{}
Although often associated with embodied agents, SNNs have been applied in other areas of research, such as for detecting Covid-19 from CT scans (\cite{garain2021detection}).

\subsection{Contribution}
\label{introduction:contribution}

%\paragraph{}
In this paper, we investigate the viability and utility of neurologically-inspired sensors and algorithms in the context of robotics by implementing and experimentally evaluating a neuromorphic approach to a common problem.
We specifically address real-time, online obstacle avoidance on a camera-equipped manipulator by designing a neuromorphic pipeline that enables end-to-end processing of visual data into motion trajectory adaptations, and which relies on event-based vision and an SNN.

%\paragraph{}
In our pipeline, an EC emulator transforms RGB data into event data, which is in turn processed by a convolutional SNN. 
The SNN output is then decoded into avoidance velocities by a dedicated component that employs a potential fields (PF) method.
The result is used to adapt a pre-planned end-effector trajectory using a dynamic motion primitive (DMP) formulation in a motion control component.
These components, collectively referred to as an \textit{SNN-based obstacle avoidance module}, operate in a closed-loop, online procedure for transforming a trajectory into an adapted version that avoids obstacles through SNN feedback. 

%\paragraph{}
ECs excel at capturing rapid motions, while SNNs are best-suited to process event data, thus motivating their utilization for deriving corrective avoidance motions.
A significant amount of research on obstacle avoidance deals with UAVs and mobile robots, while relatively few works address manipulation and take a similar neuromorphic approach.
This work is unique in addressing manipulator obstacle avoidance by utilizing event data from an onboard camera, SNN processing, and an adaptive trajectory representation.
Most solutions rely on external cameras and classical methods for filtering images, removing backgrounds, object segmentation, etc. (refer to section \ref{related_work:obstacle_avoidance} for examples of classical approaches).
Our approach may show that event data and SNN processing could eliminate the necessity of manual operations that preclude generalization over environments, lighting conditions, and platforms.

%\paragraph{}
Due to the scarcity of comparable works, proposed evaluation methodologies, and performance metrics, we formalize a set of quantitative metrics and qualitative criteria with which we evaluate our implementation.
We conduct experiments first in simulations and then on a real Kinova Gen3 arm. 
These experiments constitute repeated executions in a range of obstacle \textit{scenarios} drawn from a task distribution.
During simulation tests, we use different task scenarios to tune, validate, and finally test candidate sets of parameter values.
The results demonstrate consistent success in avoiding obstacles, which a non-adaptive baseline is incapable of.
We also analyze statistical performance across many trials, showing that the adapted trajectories do not drastically increase execution times, trajectory lengths, and velocities.
Moreover, we assess how reliable, predictable, and safe the trajectories are in our qualitative evaluation, and find at least moderately positive results in each.

%\paragraph{}
In further analyses, we additionally explore certain properties of the neuromorphic elements.
We implement and compare different event emulation strategies, based on their event outputs, SNN responses, and ultimate task performances; this yields the insight that the SNN exhibits robustness to variations in event data.
To validate its utility, we test the exclusion of the SNN and the derivation of obstacle avoidance accelerations directly from raw events, which we show adversely affects the success of obstacle avoidance.
We also investigate the effect of varying SNN weights, showing a slight variance in performance.
Finally, we show the successful integration of a real EC, a DAVIS346 and present results of preliminary experiments.
Most notably, we find that the resultant performance is fairly similar, validating our experimental emulation method and the compatibility of the pipeline with a real camera.

%\paragraph{}
The paper is structured as follows.
Section \ref{preliminaries} provides additional background information on event-based vision, SNNs, and DMPs.
Section \ref{related_work} contains a review of related work on the relevant areas of research. 
In section \ref{proposed_approach}, we present our proposed approach, elaborating on the design principles and the pipeline components.
Section \ref{evaluation} is dedicated for a detailed description of our evaluation methodology, including evaluation tasks, metrics and criteria, and experiment design and procedures for both the simulation and real experiments.
Section \ref{results_and_discussion} discusses the results of these experiments.
Section \ref{further_analyses} presents the analyses concerning the event emulation and SNN components, as well as the real EC tests. 
We conclude in section \ref{conclusion}.

\section{Preliminaries}
\label{preliminaries}

\subsection{Event-Based Vision}
\label{preliminaries:event_based_vision}

%\paragraph{}
Silicon retinas were developed to imitate the neural architectures of biological retinas in analog VLSI circuits, giving rise to neuromorphic computing (\cite{mahowald1994silicon}).
Contemporary models are known as \textit{neuromorphic retinas}, \textit{dynamic vision sensors (DVS)}, or \textit{event cameras (ECs)}.
EC pixels mimic retinal ganglion cells by asynchronously emitting a binary signal, i.e. an event, only when the incident light intensity significantly changes.
Using an address event representation (AER), the pixel arrays provide streams of  spatio-temporally-registered events which encode typically interesting information, such as motion.

%\paragraph{}
An event can be represented as a tuple of pixel position, emission timestamp, and polarity: $e_k=(\textbf{x}_k, t_k, p_k)$.
Pixels are independently monitored to compute a measure of the difference in successive intensities, the simplest example of which is the difference in raw intensities:
\begin{equation}
    \Delta L(\textbf{x}_k, t_k) = L(\textbf{x}_k, t_k) - L(\textbf{x}_k, t_{k-1})
    \label{event_intensity_generic_diff_equation}
\end{equation}
When this difference crosses threshold $\theta$, event $e_k$ is emitted and designated an ON (+1) or OFF (-1) event:
\begin{equation}
    p_k = 
    \begin{cases}
    +1,	& \text{if } \Delta L(\textbf{x}_k, t_k) > \theta \\
    -1,	& \text{if } \Delta L(\textbf{x}_k, t_k) < -\theta \\
    \end{cases}
    \label{ece_default_strategy_equation_1}
\end{equation}
An EC transmits a stream or vector of such $e_k$ tuples.

%\paragraph{}
It follows that event data can be emulated from conventional camera data, a method often used due to the scarce availability of or relative expense of acquiring ECs (\cite{garcia2016pydvs}, \cite{hu2021v2e}, \cite{rebecq2018esim}).

\subsection{Spiking Neural Networks}
\label{preliminaries:spiking_neural_networks}
 
\begin{figure}
    \centering
    \includegraphics[scale=0.7]{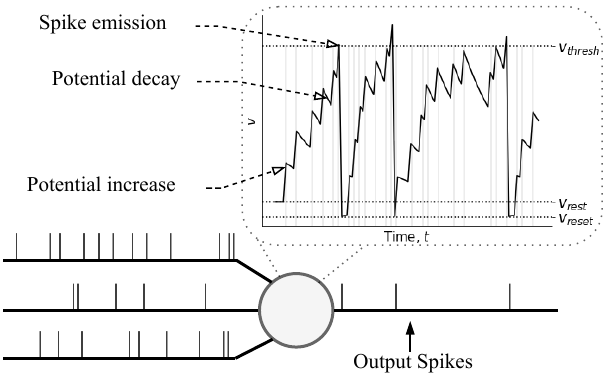}
    \caption{An illustration of spiking neuron dynamics.}
    \label{snn_dynamics_illustration}
\end{figure}

%\paragraph{}
SNNs are more biologically-plausible models in which neurons communicate via asynchronous spikes to replicate the complex neuronal dynamics observed in the brain.
Figure \ref{snn_dynamics_illustration} illustrates these spiking dynamics.

%\paragraph{}
Neurons propagate sequences of sparse, binary \textit{spike trains}, across their synapses.
Each pre-synaptic spike contributes to the post-synaptic potential (PSP) or \textit{membrane potential}, $v$, of a post-synaptic neuron: an internal analog representation of neuronal activation that decays over time, and an approximation of ionic concentrations across a nerve cell's membrane.
A neuron whose potential exceeds $v_{thresh}$ emits a spike and enters a short refractory period in which it is inhibited from spiking (not depicted).
Following a spike, the potential is reset to a baseline value, $v_{reset}$. 
Individual neurons therefore fire or are \textit{active} only in response to a significant aggregate of recent inputs.
Information can be contained in average spiking rates and/or the relative timings of spikes, a concept drawn from neuroscientific evidence (\cite{fairhall2001efficiency}).
The time-varying potential signal inherent to every neuron and the independent spiking latencies create an additional temporal dimension in SNNs.

%\paragraph{}
Various mathematical approximations and abstractions of neuronal dynamics have been employed in SNN research to produce a variety of neuron models, including the Hodgkin-and-Huxley model (\cite{hodgkin1952quantitative}), Spike Response Model (SRM) (\cite{gerstner1995time}), Izhikevich model (\cite{izhikevich2003simple}), and various probabilistic models (\cite{jang2019introduction}).
The most commonly used model is the leaky integrate-and-fire (LIF) model (\cite{rajendran2019low}, \cite{lee2020enabling}, \cite{dupeyroux2021neuromorphic})

%\paragraph{}
A LIF neuron represents a leaky integrator modeled as an RC circuit with an equation describing membrane voltage:
\begin{equation}
    \tau_v \frac{dv}{dt} = -(v-v_{rest}) + I(t)
    \label{lif_model_voltage_equation}
\end{equation}
where $v_{rest}$ is a resting potential, $\tau_v$ is a potential decay constant, and $I(t)$ is the sum of input currents.
$I(t)$ is a sum of input spikes, $S(t)$, arriving from pre-synaptic neurons indexed by $i$, multiplied by synaptic weights, $w$:
\begin{equation}
    I(t) = \sum_{i=1}^{n_l} w_i S_i(t)
    \label{lif_model_current_equation}
\end{equation}
where $S_i(t)$ is 1 if a spike occurs at time $t$ and 0 otherwise.
If $v$ exceeds $v_{thresh}$, a spike is emitted and $v$ is reset to $v_{reset}$:
\begin{equation}
    v(t) \leftarrow v_{reset}\text{, if } v(t) > v_{thresh}
    \label{lif_model_voltage_reset_equation}
\end{equation}
The neuron is prevented from firing again for a refractory period, $T_{refrac}$.
The LIF neuron thus maintains a decaying memory of past inputs and implements fundamental spiking and refractoriness properties.
The model is popular for its computational simplicity.
In our implementation, we use a variant that was presented in \cite{diehl2015unsupervised}.

\subsection{Dynamic Motion Primitives (DMP)}
\label{preliminaries:dynamic_motion_primitives}

%\paragraph{}
Dynamic motion primitives capture and reproduce discrete or rhythmic motions using a set of differential equations that produce stable global attractor dynamics (\cite{ijspeert2013dynamical}).
For discrete motions, DMPs model the evolution of a position variable, $\mathbf{y}$, from an initial to a goal position, $\mathbf{g}$, through equations describing a linear spring-damper system augmented with an additional non-linear term:
\begin{align}
    \tau \ddot{\mathbf{y}} &= \alpha_{\mathbf{y}} (\beta_{\mathbf{y}} (\mathbf{g} - \mathbf{y}) - \dot{\mathbf{y}}) + \mathsf{f(s)} \label{dmp_equation_1}\\
    \tau \dot{\mathsf{s}} &= -\alpha_{\mathsf{s}} \mathsf{s}
    \label{dmp_equation_2}
\end{align}
where $\alpha_{\mathbf{y}}$ and $\beta_{\mathbf{y}}$ control spring and damping characteristics and $\tau$ is a time scaling constant. 
$\mathsf{f(s)}$ represents a \textit{forcing term} which describes trajectory shape through a superposition of basis functions and depends on a phase variable, $\mathsf{s}$:
\begin{equation}
    \mathsf{f(s)} = \frac{\sum_{i}\mathsf{w}_i \psi_i(\mathsf{s}) \mathsf{s}}{\sum_{i}\psi_i(\mathsf{s})}
    \label{dmp_forcing_term}
\end{equation}
where $\psi_i$ are basis functions (usually of a Gaussian kernel) and $\mathsf{w}_i$ are learnable weights that capture the desired trajectory shape.
A demonstrated trajectory can be learned by sampling $\mathbf{y}$, $\dot{\mathbf{y}}$, and $\ddot{\mathbf{y}}$ then solving for a least-squares solution $\mathsf{w}_i$ using linear optimization.
Equation \ref{dmp_equation_2} describes the evolution of $\mathsf{s}$ from 1 to 0 (the start to the end of the motion).
Function $\mathsf{f}$ depends on $\mathsf{s}$ instead of a time variable, which enables scaling the trajectory in time by modifying $\tau$.

\section{Related Work}
\label{related_work}

\subsection{Event-Based Vision}
\label{related_work:event_based_vision}

%\paragraph{}
Event cameras (ECs) are inspired by biological retinas' 
superior efficiency compared to conventional, frame-based cameras.
The silicon retina (\cite{mahowald1994silicon}) paved the way for modern ECs including the DVS (\cite{lichtsteiner2008128}), ATiS (\cite{posch2010qvga}), and DAVIS (\cite{brandli2014240})\endnote{DVS: Dynamic Vision Sensor; ATiS: Asynchronous Time-Based Image Sensor; DAVIS: Dynamic and Active-pixel Vision Sensor}.
Refer to \cite{gallego2022event} for an extensive survey of event-based vision.

%\paragraph{}
Recent publications highlight applications of event-based vision in various domains. 
A survey of bio-inspired sensing for autonomous driving, presented in \cite{chen2020event}, provides a review of EC signal processing techniques and successful implementations for segmentation, recognition, optical flow (OF) estimation, image reconstruction, visual odometry, and drowsiness detection.
In a more resource-constrained scenario, event-based vision has been employed in a gesture recognition smartphone application in \cite{maro2020event}.
In \cite{dubeau2020rgb}, a combination of RGB-D and EC data for DNN-based 6-DOF object tracking is shown to outperform an RGB-D-only DNN.

%\paragraph{}
In the context of robotics, ECs have most often been utilized on aerial robots, such as for high-speed obstacle avoidance (\cite{falanga2020dynamic}) powerline tracking (\cite{dietsche2021powerline}), and fault-tolerant control (\cite{sun2021autonomous}).
UAVs particularly benefit from the advantages of ECs on account of frequent high-speed motions and motion blur, energy consumption constraints, and rapid changes in illumination.
We focus here on applications that do not involve flight, where ECs may similarly provide opportunities for improving robot capabilities.

%\paragraph{}
Bečanović et. al. used precursory optical analog VLSI (aVLSI) sensors in soccer robots, harnessing the faster reactivity properties to improve ball control and goalkeeping (\cite{becanovic2002reactive}, \cite{bevcanovic2002silicon}).
Simultaneous localization and mapping (SLAM) has been addressed with stereo visual odometry using DAVIS346 ECs in \cite{zhou2021event}, facilitating scene mapping and ego-motion estimation and matching the performance of mature frame-based approaches on benchmark datasets, while demonstrating robustness in difficult lighting conditions.
In \cite{arakawa2020exploration}, EC data was used for visual RL of tracking and avoidance policies.
The authors train in simulations using emulated data and deploy the model on a robot equipped with a DAVIS240.
In \cite{chen2019fast}, event data was used to train DNNs for Atari gameplay and action recognition, outperforming RGB image-based networks.
While these works adapt DNNs to process event frames, our approach utilizes SNNs, which are naturally suited to processing the spike-like event data.

%\paragraph{}
For this work, we implemented a software component that emulates event data from a stream of conventional images in a live camera feed or a ROS topic (see section \ref{proposed_approach:event_camera_emulation}).
\textit{pydvs} is a python-based DVS emulator that can similarly convert image intensity differences to rate- or time-encoded spikes/events (\cite{garcia2016pydvs}), but does not natively support ROS.
The prominent \textit{ESIM} simulator provides a framework for simulating 3D scenes (in OpenGL and the Unreal rendering engines) and user-defined camera motions as well as event outputs (\cite{rebecq2018esim}).
Microsoft’s \textit{AirSim} offers an EC emulation component which is similarly coupled to its rendering engine.
Naturally, such a coupling limits applicability to the associated simulation.
The \textit{v2e} toolbox (\cite{hu2021v2e}) was designed to address assumptions of the ESIM simulator that deviate from real cameras, but is limited to video files.
In \cite{joubert2021event}, the \textit{ICNS} emulator for video files and Blender scenes was qualitatively compared to ESIM, v2e, and a real DVS.
Despite some limitations (coupling to rendering engines and the absence of out-of-the-box support for live feeds and ROS data), all reviewed emulators are publicly available and provide useful tools for research and development in event-based vision.

\subsection{Spiking Neural Networks}
\label{related_work:spiking_neural_networks}

%\paragraph{}
Research into SNNs is motivated by attempts to approach the computational capability coupled with energy efficiency of the brain (as expressed by the most works reviewed in this section), compared to the more specialized yet drastically less efficient computer systems of today (\cite{roy2019towards}).

%\paragraph{}
The so-called third generation of neural networks can be potentially more expressive than first and second-generation NNs in addition to requiring significantly less neurons to represent some functions (\cite{maass1997networks}).
The authors of \cite{neil2016learning} converted ANNs pre-trained for MNIST digit recognition into SNNs and analyzed the comparative performance.
The SNNs achieved comparable accuracy using significantly less computational operations (42-58\% less) and in less time.
These and similar results demonstrate that SNNs could be as expressive/accurate as ANNs, while consuming less power and exhibiting faster inference.

%\paragraph{}
The prevalence of deep learning with ANNs has ignited research into the same in the spiking domain, not least because of prospective improvements in energy efficiency.
\cite{pfeiffer2018deep} and \cite{tavanaei2019deep} provide extensive overviews of SNNs and focus on methods for training deep SNNs, reviewing spiking analogs of CNNs, RNNs, LSTMs\endnote{\label{nn_abbrevs}CNNs: Convolutional Neural Networks; RNNs: Recurrent Neural Networks; LSTM: Long Short-Term Memory}, and echo state networks.
The authors of \cite{bouvier2019spiking} reviewed hardware implementations that leverage SNN characteristics and associated challenges.
\cite{jang2019introduction} presents a probabilistic view of SNNs, the main advantage of which is to facilitate gradient-based learning and other well-known statistical methods.

%\paragraph{}
SNN implementations have been demonstrated for solving common AI problems, particularly involving vision.
Diehl et. al. trained two-layer SNNs with STDP\endnote{STDP: Spike-Timing-Dependent Plasticity} for MNIST digit recognition and achieved state-of-the-art classification accuracy among unsupervised methods (95\%) (\cite{diehl2015unsupervised}).
In \cite{mirsadeghi2021stidi}, a supervised learning algorithm, STiDi-BP, is shown to achieve an accuracy of 99.2\% on MNIST.
Spike-YOLO was created by converting a pre-trained Tiny-YOLO model to an SNN, achieving comparable results on the PASCAL and COCO datasets, while being 2000 times more energy efficient (\cite{kim2020spiking}).
Similarly, the supervised STBP-tdBN algorithm achieved state-of-the-art results on the CIFAR and ImageNet datasets and was among the few to successfully train relatively deep SNNs (of 50+ layers)
(\cite{zheng2020going}).
Zhou et. al. presented object recognition in datasets of DVS (N-MNIST, DVS-CIFAR10, etc.) and LIDAR (KiTTi) data, demonstrating the applicability of SNNs to different data modalities (\cite{zhou2021spike}).

%\paragraph{}
Various works have successfully applied SNNs in robotics.
An extensive survey of SNNs for robot control is presented in \cite{bing2018survey}, underscoring significant potential for improving speeds, energy efficiency, and computational capabilities.
In \cite{bing2018end}, SNNs were trained using R-STDP on DVS event data for lane-keeping on a mobile robot, outperforming a conventional Braitenberg controller.
Zahra et. al. utilized shallow SNNs to learn a differential sensorimotor mapping for a UR3 robot that supports reliable Cartesian control (\cite{zahra2021differential}).
In a fully-embedded applications of SNNs (on an Intel Loihi neuro-processor), Dupeyroux et. al. designed a neuromorphic vertical thrust controller for landing a quadrotor (\cite{dupeyroux2021neuromorphic}).
The input is a spike representation of OF divergence that is estimated using data from an onboard CMOS camera.
The SNN was trained (with an evolutionary algorithm) and evaluated on a neural simulator, PySNN, achieving consistent landing behaviour.
A limitation on the generality of this approach is the controller’s dependence on a pre-set visual pattern for OF estimation.
In the present work, we similarly utilize a neural simulation tool.

%\paragraph{}
An event-driven, SNN-based PD thrust controller was designed to achieve high-speed orientation adjustment on a dual copter in \cite{vitale2021event}.
The authors demonstrated superior control speeds and reductions in latencies, particularly when running on a neuro-processor (vs. a CPU).
In \cite{risi2020spike}, reliable stereo matching of event data from two DAVIS sensors was achieved using an SNN architecture designed with neuronal populations implementing coincidence and disparity detectors, and running on a DYNAP neuro-processor.
This approach was particularly favoured for the temporal dimension of the asynchronous event and SNN spike data, which enable exploiting temporal coincidences and thus improve stereo matching.
Similar to our approach, no learning is involved; the SNN architecture's inherent properties are shown to be beneficial in realizing the desired behaviour.

%\paragraph{}
Most publications in robotics address relatively constrained navigation and flight tasks (e.g. lane-keeping and 1D thrust control). 
Our work demonstrates a less common application in obstacle avoidance for manipulation.

\subsection{Neuromorphic Computing}
\label{related_work:neuromorphic_computing}

%\paragraph{}
Neuromorphic computing/engineering mimics the fundamental neural architectures and dynamics of the brain \textit{in silico}, aiming to replicate its superior energy efficiency, compute, and robust learning capabilities in computer architectures and engineered systems.
Common approaches incorporate asynchronous event-driven communication, spike-based neural processing, analog neuronal dynamics, and local synaptic adjustments.
This research serves a dual purpose: enhancing AI systems with lessons from neuroscientific research, and advancing our understanding of the brain by experimenting with neurologically-inspired platforms.
Among the most prominent results are ECs and neuro-processors designed to run SNN architectures.
A review of neuromorphic hardware and applications can be found in \cite{rajendran2019low}, which highlights the role of neuro-processors in realizing the full potential of SNNs for exploiting event-based sensing, learning and inference.

%\paragraph{}
Neuro-processors model membrane potential evolution using voltages across capacitors or transistor sub-/supra-threshold dynamics, and transfer spikes via an AER.
\cite{furber2016large} provides a survey of pioneering neuro-processors, namely IBM’s TrueNorth, Stanford's Neurogrid, BrainScaleS and SpiNNaker, and compares their performances.
Other notable surveys provide similar statistical comparisons of these processors, in addition to the more recent Intel Loihi, DYNAP, PARCA, Braindrop, ODIN, and Deepsouth (\cite{bouvier2019spiking}, \cite{thakur2018large}, \cite{rajendran2019low}).
The Loihi and its applications have been extensively discussed in \cite{davies2021advancing}.
The variety of domains the Loihi was shown to be successfully implemented in demonstrates the general applicability of SNNs, while the quantified gains in energy efficiency validate the benefits of the neuro-processor.
In addition, the authors find that conventional DNNs exhibit little to no benefits when run on the Loihi, but SNNs achieve orders of magnitude less energy consumption and latency in some applications.

%\paragraph{}
Several publications investigated the effects of neuro-processors on speed and energy efficiency.
SNNs trained with R-STDP on a BrainScaleS2 processor to control an agent in playing the game of Pong and achieved one and three orders of magnitude improvements in speed and energy efficiency when compared to a CPU simulation (\cite{wunderlich2019demonstrating}).
The authors of \cite{ceolini2020hand} achieved hand-gesture recognition with a neuromorphic sensor fusion approach, where DVS streams and EMG signals (converted to spikes/events) were used to train SNNs running on a Loihi or ODIN.
With respect to a GPU-based implementation, this was at least 30 times more energy-efficient, though inference was 20\% slower.
\cite{taunyazov2020event} presented a visual-tactile SNN (VT-SNN) which fuses data from an EC and a spike-based tactile sensor to accomplish robot manipulation tasks requiring object classification and slip detection.
The SNN classifiers running on a Loihi performed similar to state-of-the-art DNNs run on GPUs while consuming 1900 times less power and exhibiting lower latency.
In \cite{goltz2021fast}, backpropagation was used to train SNNs for MNIST classification on a BrainScaleS and then compared to the performance of a conventional CNN running on an NVidia Tesla P100 GPU.
The authors show that the neuromorphic implementation is (approx. 100 times) more energy-efficient, at the cost of a slight drop in accuracy and the number of classifications per second (since the GPU implementation utilizes parallelization, while individual images must be processed sequentially on the SNN).

%\paragraph{}
The related field of \textit{neurorobotics} studies the design of computational structures that are inspired by the human and animal nervous systems in robots (\cite{van2016neurorobotics}) and has lead to various interesting applications (\cite{dumesnil2016robotic}, \cite{lobov2020spatial}, \cite{falotico2017connecting}).
A review in \cite{chen2020neurorobots} demonstrated the utility of neurorobotics for explaining how neural activity gives rise to intelligence, as a form of \textit{computational neuroethology}.
Robotics could thus similarly benefit from ongoing research and development efforts aimed at drawing inspiration from the brain.

%\paragraph{}
In the present work, we do not run SNNs on neuromorphic hardware, instead aiming to investigate the utility of an event-based SNN approach on conventional hardware.
Nevertheless, the reviewed research motivates exploiting the potential gains in energy efficiency and latency in a subsequent study.
On a broader scale, this may additionally inspire the pursuit of neuroethology-based robot designs that advance further into the realm of biological realism.

\subsection{Obstacle/Collision Avoidance}
\label{related_work:obstacle_avoidance}

%\paragraph{}
Obstacle avoidance is a critical feature for planning robot motions in evolving, dynamic environments, where obstacles may appear during task execution and invalidate motion plans. 
We discuss seminal works in this domain, focussing on relevant approaches and especially those that address manipulation problems and employ camera sensors.

%\paragraph{}
Research on the rudimentary issue of reactively computing collision-free paths has lead to established methods such as vector field histograms (VFH) (\cite{borenstein1991vector}), the Dynamic Window Approach (DWA) (\cite{fox1997dynamic}), and the elastic strips framework (\cite{brock2002elastic}).
The artificial potential fields (PF) method represents task criteria in the form of attractive and repulsive forces acting on an agent moving within a virtual force field (\cite{khatib1986real}).
PF techniques have been extensively applied and improved upon, and are utilized in the present work. 
A helpful review of classical methods is provided in \cite{minguez2016motion}.

%\paragraph{}
Optical flow (OF) estimation, a bio-inspired computer vision technique that is applicable to obstacle avoidance, shares similarities to the methods proposed here.
OF quantifies the motion of light intensity patterns observed on a sensor as it moves relative to observable objects, and is used by organisms, such as honeybees, for navigation (\cite{van2016neurorobotics}).
This can provide estimates of ego- or object motion, which facilitate tracking and collision avoidance.
In \cite{schaub2016reactive}, OF is computed from an autonomous car's monocular camera data and used to optimize maneuvers to avoid obstacles.
OF estimation approaches are split into two categories.
In estimating object motion, \textit{dense} OF tracks changes to every pixel between consecutive frames and \textit{sparse} OF relies on tracking identified features.
The former imposes high computational and memory costs, while the latter depends on the reliability of feature matching algorithms and may not generalize if object models have to be specified a priori (\cite{lee2021deep}).
In comparison, our approach is designed to eliminate unnecessary computations, by virtue of the event-based processing, and be independent of specific obstacle features.

%\paragraph{}
Compared to 3D sensors, IMUs, and laser sensors, cameras are less frequently used for obstacle avoidance. 
\cite{lee2021deep} demonstrated DNN-based obstacle recognition and avoidance on a UAV navigating a plantation, where trees are recognized, distances are estimated, and free regions are determined for simple heading adjustments. Limitations include a restriction to obstacles that the DNN is trained to recognize and the entailed computational costs.
In \cite{hua2019small}, semantic segmentation DNNs recognize roads and obstacles that a mobile robot encounters, which are incorporated in PF-based local path planning.
More rudimentary approaches involve classical methods such as detecting contours for obstacle detection (\cite{martins2018computer}) or using feature extractors like SURF to recognize known obstacles (\cite{aguilar2017obstacle}), followed by searching for free regions and applying corrective motions.
These approaches may be susceptible to changes in lighting conditions, where ECs are expected to perform better.
In addition, purely reactive approaches require rectifying velocity commands that return the agent to its original path. 
In our work, we propose using DMPs for adaptive path plans that obviate the need for extra corrective computations.

%\paragraph{}
For robot manipulation, obstacle detection and avoidance could be crucial in safety-critical human-robot collaboration settings.
The authors of \cite{chiriatti2021adaptive} designed a control law for a UR5 manipulator that incorporates collision cylinders instantiated from estimates of obstacle geometries, positions and velocities, demonstrating constrained avoidance behaviours.
The method was tested in simulation with prior obstacle information; extending this to real environments requires dedicated sensors and algorithms to estimate the pose of every person/object.
In \cite{safeea2019line}, collision bounds on a person are incorporated in a PF approach to controlling a KUKA LBR iiwa. 
These are obtained using IMUs placed on persons in an industrial workspace.
Another approach relies on proximity sensors placed on the manipulator, which provide time-of-flight, IMU, and gyroscope readings (\cite{escobedo2021contact}). 
While reliable behaviours are achievable by deploying arrays of sensors, this may limit generalization to different scenarios, especially when a robot is not confined to a controlled workspace or when arbitrary sensor placement is not possible.
In contrast, our approach relies on a single onboard camera.

%\paragraph{}
Other notable implementations integrate vision-based sensing, leveraging RGB-D cameras in particular. 
Mronga et. al. used pointcloud data to extract convex hulls of obstacles that are incorporated as constraints in an optimization problem whose solution leads to avoidance motions on a dual-arm system (\cite{mronga2020constraint}). 
The optimizer leads to task-compliant avoidance, but depends on cameras covering the workspace and several pointcloud processing steps. 
A similar strategy is applied in \cite{song20193d}, where obstacles in a bin-picking task are avoided by detecting moving objects in a pointcloud and accordingly adjusting motions through a PF algorithm.
These implementations benefit from depth perception that monocular RGB or event cameras do no provide.
However, event-based processing could impose a significantly lower computational overhead than pointcloud processing, which may be a concern in applications that require rapid reactivity.

%\paragraph{}
A few publications are particularly relevant for their similar approaches and thus merit mentions in the remainder of this section.
In \cite{park2008movement}, a DMP obstacle avoidance term was first introduced and used within a PF formulation.
We similarly utilize DMPs and draw insights from the authors' mathematical integration of obstacle avoidance information for the method presented in section \ref{proposed_approach}.
Scoccia et. al. presented offline planning of trajectories along with online adjustments using a formulation of PFs that operates on the Jacobian matrices for null space control (\cite{scoccia2021collision}).

%\paragraph{}
Another implementation of manipulator obstacle avoidance combined PFs and elastic bands for adaptive trajectory planning (\cite{tulbure2020closing}).
Here, an RGB-D camera capturing the workspace provided pointcloud data which is processed to estimate obstacle positions that affect the PF.
The authors augment a PF algorithm with an elastic bands planner, which enables adjusting a global plan with minimum deviations, thus addressing the susceptibility of PFs to local minima.
This resembles our application of PFs for local velocity corrections, while our DMP maintains a high-level plan.
Again, pointcloud processing introduces a computational expense and a dependence on the camera position and/or robot platform, thus placing theoretical limits on generality and applicability to different environments.

%\paragraph{}
The utility of event-based vision for obstacle avoidance has been demonstrated in the past.
\cite{milde2015bioinspired} presents event-based collision avoidance on a mobile robot by computing OF from DVS data and deriving velocity commands.
Here, the usage of ECs is motivated by the redundancy in data and wastage of computations associated with processing conventional camera images, particularly when the robot is stationary.
Furthermore, the authors suggest extending their work with a "neuromorphic circuit" and SNNs to address limitations, including the significant amount of data required by their PCA-based method for computing OF.
In \cite{sanket2020evdodgenet}, dynamic avoidance on a quadrotor was achieved by training CNNs on event data to estimate the OF of moving objects, while placing priors on obstacle shape (sphere) for tractability.
Our work differs in addressing tasks where the robot must avoid obstacles whilst actively moving towards a goal and employing SNNs.
In a navigation scenario, Yasin et. al. utilized a DVS for car obstacle avoidance in low-light settings, demonstrating superior reaction times when compared to standard cameras (\cite{yasin2020night}).
Objects in the event image are obtained through denoising, corner detection, segmentation, and filtering procedures, and used to recompute plans.
While these efforts present viable applications of event data, much of the requisite pre-processing may be obviated by utilizing SNNs: the natural complement to event-based vision.

%\paragraph{}
This usage of SNNs has nevertheless also been shown in recent years.
\cite{salvatore2020neuro} demonstrated neuro-inspired UAV collision avoidance by running event data on an SNN that was converted from a trained deep Q-Learning (DQN) ANN.
Successful behaviours were achieved in AirSim simulations after training the DQN agent on emulated event data, transferring weights to an equivalent SNN, and further training the SNN with data from successful trials.
Of greatest similarity to our work is the feasibility study of a neuromorphic approach to obstacle avoidance presented in \cite{milde2017obstacle}. 
Their method involves processing event data from a DVS mounted on a mobile robot in SNNs that are implemented on a ROLLS neuro-processor, whose output is decoded into avoidance and target-following behaviours by aggregating responses of neuron populations.
As in our approach, SNN connections are non-plastic (i.e. not adjusted through learning); the inherent properties of the SNN architecture are shown to facilitate viable navigation behaviours.
The present work proposes a similar approach in the context of manipulation, which presents a different set of challenges.

\section{Proposed Approach}
\label{proposed_approach}

\begin{figure*}
    \centering
    \includegraphics[width=\linewidth]{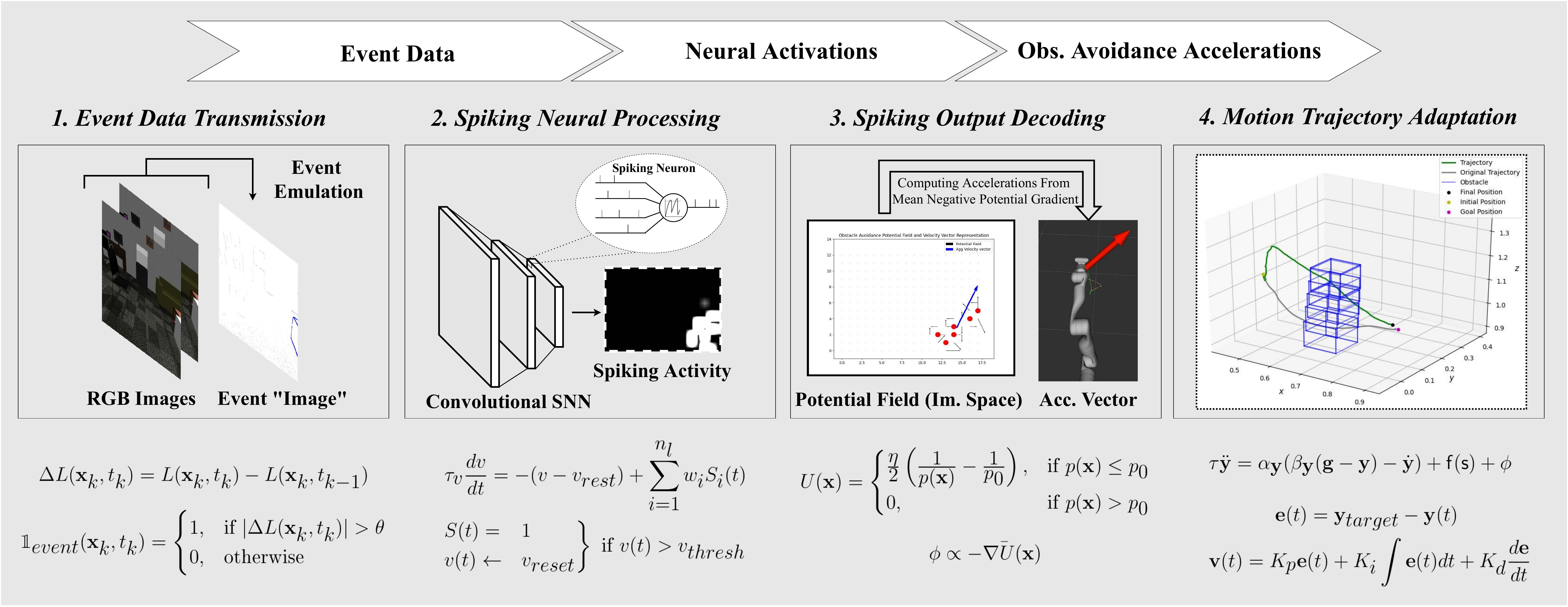}
    \caption{The main components and processing stages of our neuromorphic pipeline. The plot on the right shows a pre-planned trajectory (grey) and a resultant trajectory (green) adapted to avoid an obstacle (blue).}
    \label{proposed_approach_pipeline_illustration}
\end{figure*}

%\paragraph{}
We present a neuromorphic approach that incorporates event-based vision and SNNs for adaptive motion execution.

%\paragraph{}
In this approach, we use DMPs to generate trajectory plans for reaching tasks.
The DMP formulation supports additive acceleration terms which we utilize to inject obstacle avoidance information and therefore adapt to guide the robot’s motion away from perceived obstacles.
This information is obtained by continuously processing visual, event-based data within an SNN, then decoding neural activation maps into avoidance accelerations.
The latter procedure involves a potential fields (PF) method for computing the most favourable avoidance direction.
By utilizing events induced by relative motion and the spatio-temporal filtering properties of SNNs, we can extract reactive motions that modify high-level motion plans to account for obstacles while maintaining progress towards the task goal, thus achieving real-time, online trajectory adaptation.

%\paragraph{}
A core aspect of our approach is the synergy between global planning and local corrections that enables goal-directed, obstacle avoidance.
The common alternative: completely re-planning whenever obstacles are encountered, can impose higher computational expenses and latencies (\cite{d2009learning}, \cite{feng2020obstacle})), which are particularly undesirable in dynamic environments.
Instead, we utilize the DMP as an adaptive planner, where perceived obstacles are handled by appropriately adjusting the next waypoint during execution and the global attractor dynamics ensure a graceful return to the original path.

%\paragraph{}
The choice of SNNs is justified by their natural compatibility with and direct applicability to event-based vision, unlike conventional computer vision algorithms, such as CNNs (\cite{chen2020event}, \cite{vitale2021event}).
SNNs are designed to process discrete, asynchronous signals and may be key in achieving compelling real-world applications of ECs.
The combination of events and spiking neurons particularly holds potential for capturing temporal information relating to obstacle avoidance, such as through the decaying influence (neural activation) of an obstacle that has just been observed.
Furthermore, the analog SNN dynamics can inherently induce temporal filtering properties that negate effects of insignificant events (see section \ref{proposed_approach:convolutional_spiking_neural_networks}).

\subsection{Overview}
\label{proposed_approach:overview}

%\paragraph{}
We designed a modular pipeline of specialized components that enable end-to-end processing of visual data into motion trajectory adaptations:
\begin{enumerate}
    \item \textbf{EC/EC Emulator:} Produces event data. Either an EC or an emulator which transforms RGB data.
    \item \textbf{Convolutional SNN (C-SNN):} Processes event data in a network of spiking neurons. 
    \item \textbf{Obstacle Avoidance Component:} Decodes SNN outputs into obstacle avoidance velocities.
    \item \textbf{Adaptive Motion Planner:} Generates end-effector positions of a pre-planned trajectory while adjusting the plan according to the obstacle avoidance component’s output; a DMP. Positions set by the DMP are followed using a PID controller. 
\end{enumerate}
These components form an \textit{SNN-based obstacle avoidance module} which is designed to easily integrate within an existing robot software stack.

%\paragraph{}
Figure \ref{proposed_approach_pipeline_illustration} depicts the processing stages and the data flow (top arrows) within the pipeline, which we describe in the following sub-sections.

\subsection{Event Camera/Emulator}
\label{proposed_approach:event_camera_emulation}

%\paragraph{}
The sensory input is event data that is derived from RGB camera data through an emulator.
Following the fundamental EC operating principles, events can be generated from the thresholded intensity difference at every pixel between consecutive timesteps.
Section \ref{related_work:event_based_vision} contains a review of existing emulators and their deficits, which motivate developing our \textsf{event\_camera\_emulation}\endnote{https://github.com/AhmedFaisal95/event\_camera\_emulation} component.

%\paragraph{}
An event $e_k=(\textbf{x}_k, t_k, p_k)$ is emitted at pixel position $\textbf{x}_k$ at time $t_k$ with polarity $p_k \in \{+1, -1\}$ if the difference in intensities, $\Delta L(\textbf{x}_k, t_k)$\endnote{An arbitrary difference measure: $d(L(\textbf{x}_k, t_k), L(\textbf{x}_k, t_{k-1}))$.}, exceeds a threshold, $\theta$.

\begin{figure}
    \centering
    \begin{subfigure}{.33\linewidth}
        \centering
        \frame{\includegraphics[width=.9\columnwidth]{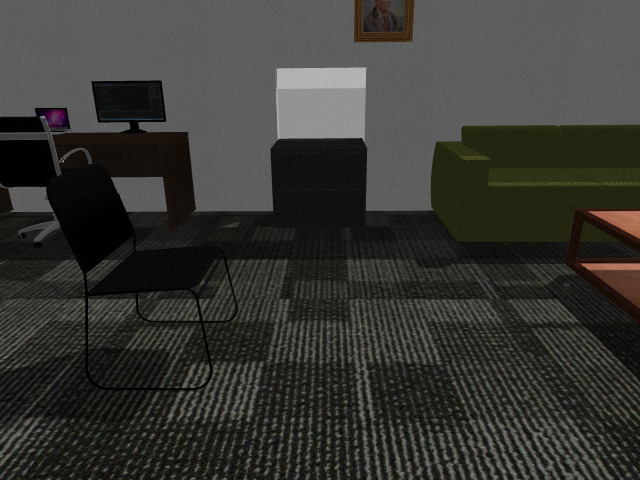}}
        \caption{\centering RGB Image 1}
    \end{subfigure}%
    \begin{subfigure}{.33\linewidth}
        \centering
        \frame{\includegraphics[width=.9\columnwidth]{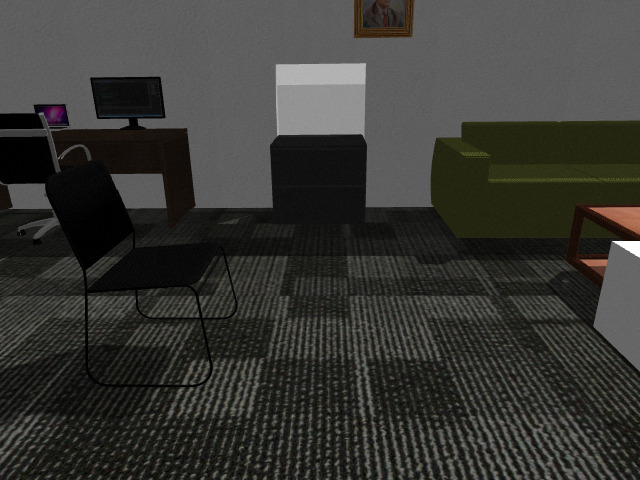}}
        \caption{\centering RGB Image 2}
    \end{subfigure}%
    \begin{subfigure}{.33\linewidth}
        \centering
        \frame{\includegraphics[width=.9\columnwidth]{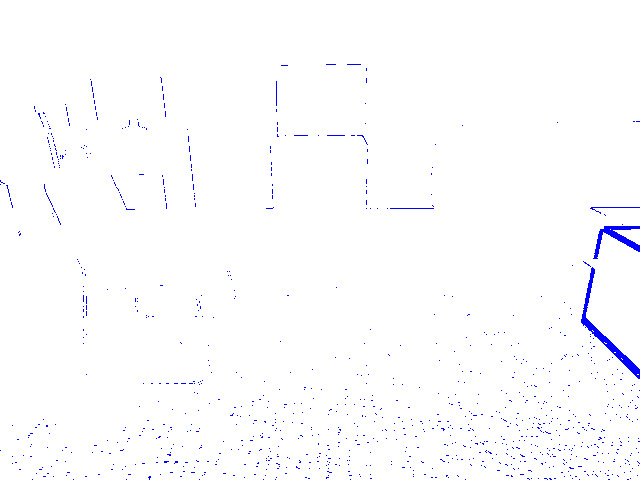}}
        \caption{\centering Event "Image"}
    \end{subfigure}
    \caption{Emulated events due to object motion on the right. Here, any event (ON or OFF) is indicated by a blue pixel on the event image. This was captured as the camera moved forward, inducing some events on the edges of distant objects.}
    \label{ece_event_image_example}
\end{figure}

%\paragraph{}
We form an \textit{event image} representation by placing $p_k$ of every event at the respective pixel location.
The resulting event image, $I_e$, is a single-channel analog of the source RGB images that contains values, $i_k \in \{+1, 0, -1\}$.
An event image derived from two consecutive RGB images is shown in Figure \ref{ece_event_image_example}.
The distinction of event polarity is often of little importance and OFF (-1) events are either ignored or treated the same as ON (+1) events, as we do here (and similar to \cite{dubeau2020rgb} and \cite{maro2020event}).

\subsubsection{Limitations of Emulation}

%\paragraph{}
Note that deriving events from differences in RGB frames places an upper bound on the event generation rate: the camera frame rate.
This effectively negates the advantages of event asynchrony and the theoretically higher transmission rates in comparison to conventional camera pixels.
Nevertheless, it provides a reasonable approximation for demonstrating and presenting elementary arguments for our approach.
We verify this by comparing the output and consequent task performance of our emulator to those of a real EC in section \ref{further_analyses:real_ec_tests}.

\subsubsection{Filtering Event Noise: Binary Erosion}
\label{binary_erosion}

%\paragraph{}
We have found that applying a binary erosion filter produces cleaner event images in cases where certain textures or surfaces (e.g. rough carpets) induce too many insignificant background events that may lead to over-reactive responses.
The filter effectively removes events at which the local region does not contain sufficiently many other events.

\subsection{Convolutional Spiking Neural Network}
\label{proposed_approach:convolutional_spiking_neural_networks}

%\paragraph{}
The resulting event data is processed by a C-SNN.

\subsubsection{Input Data Encoding}

%\paragraph{}
We utilize a Poisson process spike generation model to induce spikes in the input layer from incoming event images.
A Poisson process presents a plausible stochastic approximation of biological neuron firing activity, in which the generation of each spike is assumed to depend on some firing rate, $r$, and be independent of all other spikes (\cite{heeger2000poisson}).

%\paragraph{}
Given a $w \times h$ event image $I_e$, we assign rate values according to:
\begin{equation}
    I_e'(\textbf{x}) = 
    \begin{cases}
        r_{ON},	    & \text{if } I_e(\textbf{x}) = +1 \\
        r_{OFF},	& \text{if } I_e(\textbf{x}) = -1 \\
        0,          & otherwise
    \end{cases}
\end{equation}
where $r_{ON}$ and $r_{OFF}$ represent firing rates in $Hz$. 
In our experiments, these parameters are set such that OFF events induce half the stimulation that ON events do, which is based on insights from similar works and observed performance. In particular, the same ratio is applied in \cite{salvatore2020neuro}, while the authors of \cite{maro2020event} and \cite{dubeau2020rgb} do not consider event polarity. 
Similarly, we do not require a strict distinction between positive and negative brightness changes for the purpose of detecting a moving obstacle, and these values have been adequate in practice.

\begin{figure}
    \centering
    \includegraphics[width=\columnwidth]{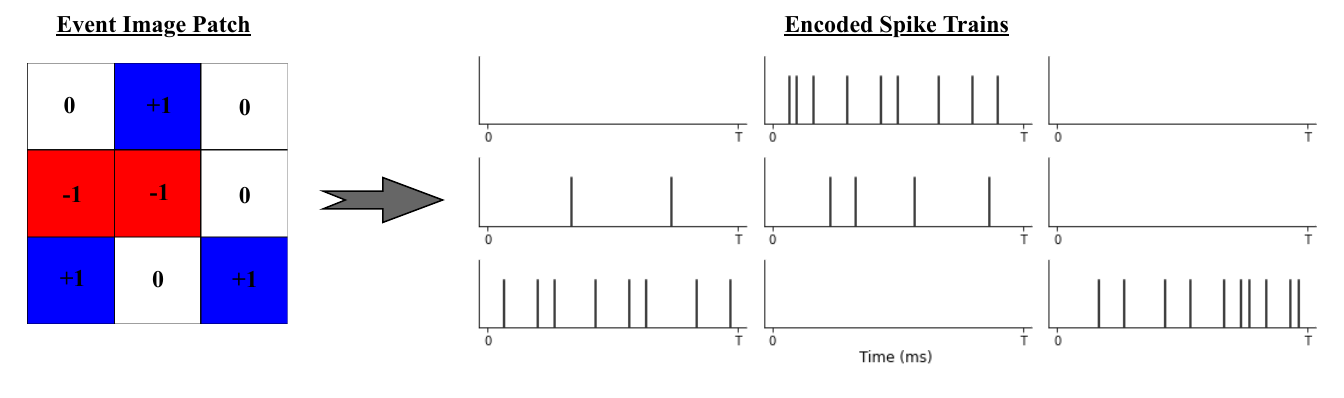}
    \caption{Poisson spike trains generated from events at 9 input neurons. Positive and negative events are shown in blue and red. Note the lower spiking frequency at negative events.}
    \label{event_image_poisson_spike_gen_illustration}
\end{figure}

Subsequently, the spike train entering each input neuron across $T$ is drawn from a Poisson process, resulting in sequences of spikes that follow an average firing rate but exhibit random spike timings.
Figure \ref{event_image_poisson_spike_gen_illustration} illustrates an example of a $3 \times 3$ event image patch and the spike trains generated at every pixel/input neuron.
The result is a $w \times h \times T$ binary matrix $S$, where each element $s^{ij}_t$
defines whether neuron $ij$ receives an input spike at timestep $t$\endnote{Note that this assumes a constant firing rate over $T$.}.

\subsubsection{Convolutional Network Topology}

\begin{figure}
    \centering
    \includegraphics[width=\columnwidth, trim={0pt 0 0pt 4},clip]{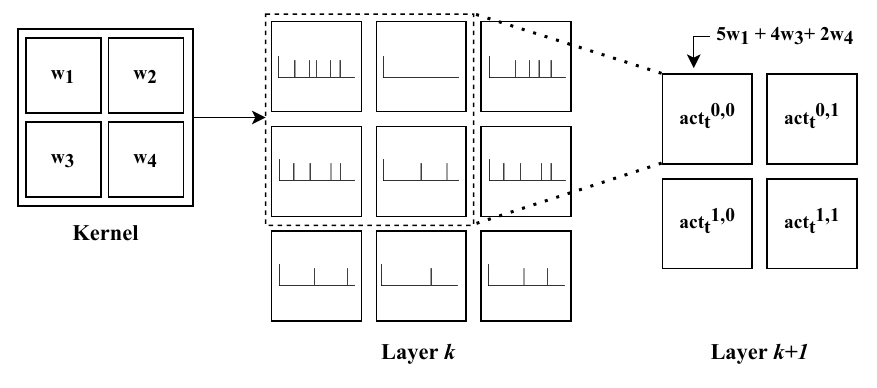}
    \caption{An illustration of convolving a $2 \times 2$ kernel with a $3 \times 3$ layer ($k$) of spiking neurons, whose output spike trains are depicted within each cell. Assuming a stride of $1 \times 1$, the result is the pre-activations of neurons in layer $k+1$ (see Equation \ref{csnn_convolution_formula}. The summation resulting in $act^{0,0,k+1}_t$ is shown above the cell.)
    }
    \label{snn_conv_op_illustration}
\end{figure}

%%\paragraph{}
The C-SNN resembles a CNN in architecture and weight sharing principles; neurons are arranged in two-dimensional layers and information is propagated through convolution operations.
Here, the input to each neuron is a pre-activation computed by convolving the spiking output of neurons in the previous layer that occupy the target neuron’s receptive field and a kernel matrix of shared, real-valued weight parameters, $K$.
This is encapsulated in the following adaptation of the convolution equation provided in \cite{goodfellow2016deep}, where the pre-activation of neuron $(i, j)$ in layer $k+1$, at time-step $t$, is computed by convolving a kernel with spike trains arriving from the preceding layer, $k$:
\begin{align}
    act^{i,j,k+1}_t &= (K * S)(i, j)\\
    &= \sum_{m}\sum_{n}S(i-m, j-m)K(m, n)
    \label{csnn_convolution_formula}
\end{align}
The pre-activation value adds to the neuron's membrane potential.
Figure \ref{snn_conv_op_illustration} depicts a convolution operation.

%\paragraph{}
Spatio-temporal distributions of spikes within an SNN lead to novel dynamics from which interesting properties may arise. 
The convolutional operator and the analog dynamics of the neurons create a form of \textit{spatio-temporal filter}, where signals that are particularly persistent in space and time are selectively propagated.
Moreover, spiking neurons possess a form of memory in the decaying potential, which reflects recent levels of stimulation.
The spike trains emitted at the output neurons are used to derive obstacle avoidance in the next component of the pipeline.

%\paragraph{}
We set the synaptic weights to fixed, random values.
Non-plastic SNN connections have often been employed in SNN applications, such as in \cite{risi2020spike} (see section \ref{related_work:spiking_neural_networks}) and \cite{milde2017obstacle} (see section \ref{related_work:obstacle_avoidance}), which demonstrated the achievement of desired behaviours solely due to the properties of spiking dynamics.
Similarly, we presently involve no \textit{learning}, and instead investigate how robust task performance is to different random, but fixed, “features” that manifest through the randomly sampled weights.
Future extensions will incorporate weight adjustment strategies through, for example, STDP and supervised variants or RL.
The weight values are initialized by sampling from a standard uniform distribution, scaled by a weight factor, $w_c$:
\begin{equation}
    W \sim \mathcal{U}(0, w_c)
    \label{snn_weight_sampling_equation}
\end{equation}

\subsection{Obstacle Avoidance Component}
\label{proposed_approach:obstacle_avoidance_component}

%\paragraph{}
The obstacle avoidance component decodes the SNN's output into meaningful avoidance signals.
This step involves extracting indications of obstacle presence or motion from the spiking activity and deriving velocities/accelerations that can adapt the planned motion trajectory. 
To that end, we utilize a first-spike-time representation of  spiking activity and a PF method.

\subsubsection{SNN Output Representation}

\begin{figure}
    \centering
    \includegraphics[width=\columnwidth, trim={0 10 65 0},clip]{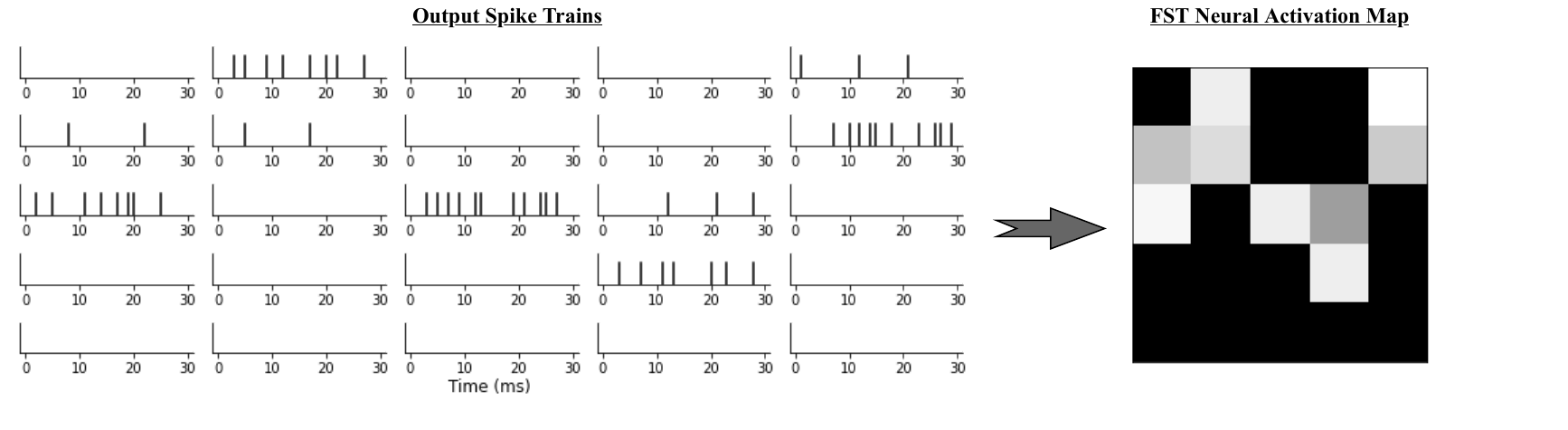}
    \caption{A depiction of the FST code applied to 25 output neurons. On the right, cell brightness corresponds to each neuron's FST and thus activation magnitude. Note: the top-right neuron has a higher activation due to an early first spike, although the neuron just under exhibits a higher spike count.}
    \label{snn_5x5_output_neural_activation_illustration}
\end{figure}

%\paragraph{}
Spike trains can be represented in various temporal and rate coding schemes.
We use first-spike-time (FST) temporal coding to interpret the output spiking activity (\cite{tuckwell2005time}, \cite{goltz2021fast}, \cite{liu2021sstdp}).
Within this scheme, the time until a neuron’s first spike after stimulus presentation fully determines the magnitude of stimulation: the earlier a neuron \textit{first} spikes, the more stimulated it is.
When applied to the output spike trains, the FST code provides a \textit{neural activation map}, as illustrated in Figure \ref{snn_5x5_output_neural_activation_illustration}.
Significantly intense neural activation at a given neuron is expected to indicate the persistent presence and (relative) motion of a perceived object in regions of the input image for which that neuron is in the effective receptive field.
This provides indications of avoidance directions.

%\paragraph{}
Other common codes include the absolute or average number of spikes.
Like either, the time to the first spike can indicate a neuron’s level of stimulation; however, FST necessitates flagging only the first spike, as opposed to waiting until all spikes in a given time window are accumulated.
Consequently the FST code could reduce \textit{time-to-solution} or \textit{energy-to-solution} and thus be more efficient (\cite{goltz2021fast}).
Note that the FSTs are calculated with respect to the time point at which the last event image was presented.

\subsubsection{Computing Obstacle Avoidance Direction Using Potential Fields}

%\paragraph{}
The neural activation map can be interpreted as a downscaled version of the input event image, filtered in the space and time dimensions to indicate approximate locations of persistent obstacles (while removing potential noise).
Therefore, we regard high activation regions as \textit{obstacle points} in this space.
In order to derive an avoidance motion vector, we utilize a method that can aggregate the obstacle points’ spatial influences and compute a direction that maximizes movement away from these points.

\begin{figure}
    \centering
    \begin{subfigure}{0.7\linewidth}
        \centering
        \includegraphics[width=\columnwidth]{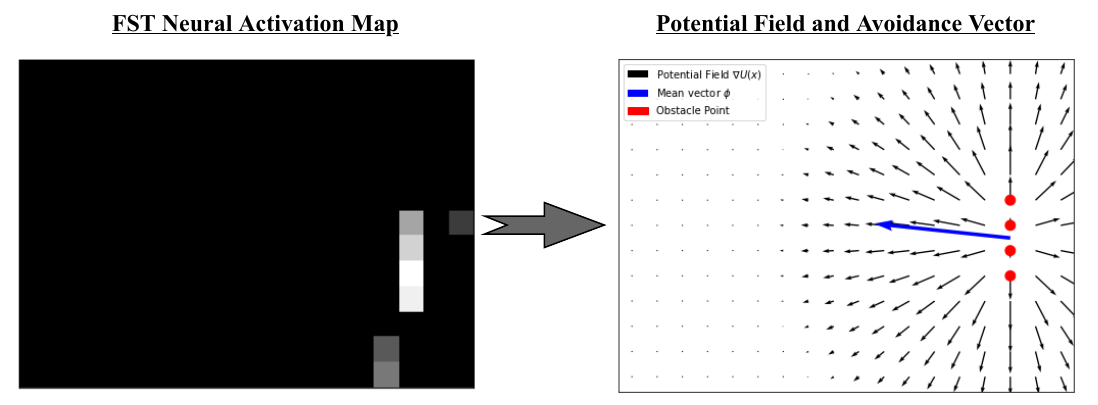}
        \caption{\centering Neural activation map to potential field}
        \label{snn_13x18_neural_activation_to_PF_example:neural_activation_to_PF}
    \end{subfigure}%
    \begin{subfigure}{0.3\linewidth}
        \centering
        \includegraphics[width=\columnwidth, trim={140pt 20 0pt 30},clip]{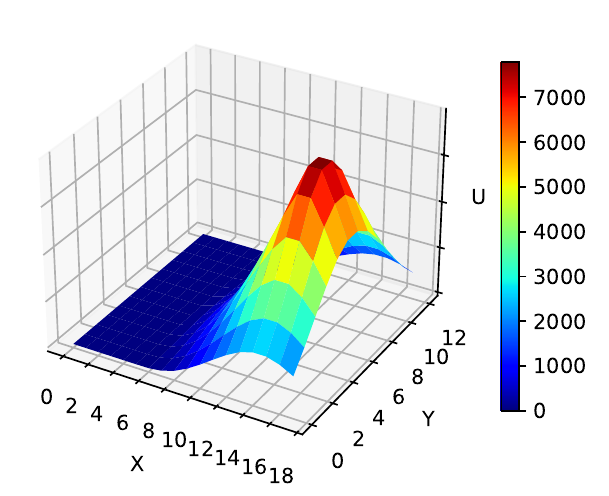}
        \caption{\centering 3D potential field}
        \label{snn_13x18_neural_activation_to_PF_example:potential_landscape}
    \end{subfigure}%
    \caption{The PF computed from an SNN's output neural activation map, visualized in 2D (\ref{snn_13x18_neural_activation_to_PF_example:neural_activation_to_PF}) and 3D (\ref{snn_13x18_neural_activation_to_PF_example:potential_landscape}), the latter showing the slope that determines the motion vector (blue).}
    \label{snn_13x18_neural_activation_to_PF_example}
\end{figure}

%\paragraph{}
Artificial potential fields (PFs) are fields of attractive and repulsive forces overlaid on a robot’s environment to drive goal reaching and avoidance behaviours, respectively.
We compute a PF from the neural activation map, with obstacle points set to exert repulsive forces, using the formulation presented in \cite{park2008movement}.
For an arbitrary point on the field, $\textbf{x}$, the potential is determined by the distance to an obstacle point, $p(\textbf{x})$, according to:
\begin{equation}
    U(\textbf{x}) = 
        \begin{cases}
            \frac{\eta}{2}\left(\frac{1}{p(\textbf{x})} - \frac{1}{p_0}\right),	& \text{if } p(\textbf{x}) \le p_0 \\
            0,	& \text{if } p(\textbf{x}) > p_0 \\
        \end{cases}
    \label{pf_equation}
\end{equation}
where $p_0$ denotes obstacles' radius of influence and $\eta$ is a constant gain.
If multiple obstacle points are perceived, the potential at point $\textbf{x}$ is simply the aggregate of their contributions, i.e. for $n$ obstacle points:
$U(\textbf{x}) = \sum_{i=1}^{n} U_i(\textbf{x})$.
The negative potential gradient, $-\nabla U(\textbf{x})$, provides estimates of directions leading away from high potential regions.
Figure \ref{snn_13x18_neural_activation_to_PF_example} shows a PF derived from a neural activation map, yielding a gradient field (black arrows) which points away from the aggregated obstacle points (red).
The mean negative potential gradient, $\vect{\tilde{\phi}}=\overline{-\nabla U(\textbf{x})}$, then provides an average motion vector that incorporates all potential across the field and estimates an optimal direction for avoiding the perceived obstacles.
In Figure \ref{snn_13x18_neural_activation_to_PF_example:neural_activation_to_PF}, this is visualized as a blue arrow.
(Note: it is not necessary to map the neural activation map back to the input image space; in practice, the resulting motion vector leads to effective avoidance motions.)

%\paragraph{}
The SNN output is thus decoded into motion vector $\vect{\tilde{\phi}}$, which provides the information needed to most adequately adapt the current motion plan.

\subsection{Motion Planning and Control}
\label{proposed_approach:motion_planning_and_control}

%\paragraph{}
The aforementioned components operate alongside the planning and control component, which generates and executes trajectories.
It also utilizes feedback from the obstacle avoidance component to adapt motion plans online by deviating to avoid obstacles while maintaining progress towards the goal; this is accomplished through a DMP.
The robot follows the positions specified by the DMP planner through velocity commands generated by a PID controller.

\subsubsection{Dynamic Motion Primitive (DMP)}
\label{dynamic_motion_primitives}

%\paragraph{}
DMPs can model the evolution of a point’s position over time in a set of differential equations that produce stable global attractor dynamics (see section \ref{preliminaries:dynamic_motion_primitives}). 
A particularly useful property is the extensibility of the transformation system equations with task-related acceleration terms.

%\paragraph{}
A secondary objective can be accomplished by adding appropriate acceleration values to Equation \ref{dmp_equation_1} during the evolution of variable $\mathbf{y}$.
The stable attractor dynamics guarantee eventual convergence to the goal, thus enabling the plan to be adjusted through deliberate perturbations.
We utilize this property to adapt DMP-planned trajectories through the obstacle avoidance component's output: $\vect{\tilde{\phi}}$. 

%\paragraph{}
Here, the DMP controls the end-effector's position, $\mathbf{y} = [x, y, z]^T$ as it progresses towards goal $\mathbf{g}$.
The trajectory shape, i.e. $\mathsf{w}_i$, is fixed, because i) the capability of generalizing any trajectory shape to arbitrary initial and goal positions provides sufficient adaptivity and ii) adaptation is externally achieved by incorporating obstacle avoidance feedback.
Specifically, we augment equation \ref{dmp_equation_1} with an additive obstacle avoidance acceleration term, $\vect{\phi}$:
\begin{equation}
    \tau \ddot{\mathbf{y}} = \alpha_{\mathbf{y}} (\beta_{\mathbf{y}} (\mathbf{g} - \mathbf{y}) - \dot{\mathbf{y}}) + \mathsf{f(s)} + \vect{\phi} 
    \label{dmp_equation_1_adapted}
\end{equation}
The instantaneous value of $\vect{\phi}$ is directly derived from $\vect{\tilde{\phi}}$.
The latter expresses a motion vector in the image space which is transformed to the end-effector's operational space:
\begin{equation}
    \vect{\phi} = T_{camera}^{ee} \vect{\tilde{\phi}} 
    \label{phi_transform_1}
\end{equation}

%\paragraph{}
Note that $\vect{\tilde{\phi}}$ is computed from a camera image and is thus strictly two-dimensional; as a result, $\vect{\phi}$ inherits the same constraint.
For a camera mounted at the end-effector, such that the image plane is parallel to the end-effector’s \textit{y}-\textit{z} plane and perpendicular to the forward-facing \textit{x}-axis, the consequence is that avoidance vectors are constrained to the \textit{y}-\textit{z} plane.
This reflects the fact that the sensor lacks depth perception and thus can not be used to compute avoidance motions that are perpendicular to the image plane.
Nevertheless, avoidance behaviours resulting from this approach are sufficient for the class of tasks that we target, where the end-effector moves forwards towards a goal and the robot is expected to avoid obstacles within the field of view (FOV) of a forward-facing camera.

%\paragraph{}
The resulting obstacle-avoiding trajectory is finally executed by following positions integrated from Equation \ref{dmp_equation_1_adapted}.

\subsubsection{PID Controller}    
\label{pid_controller}

\begin{figure}
    \centering
    \includegraphics[width=\columnwidth]{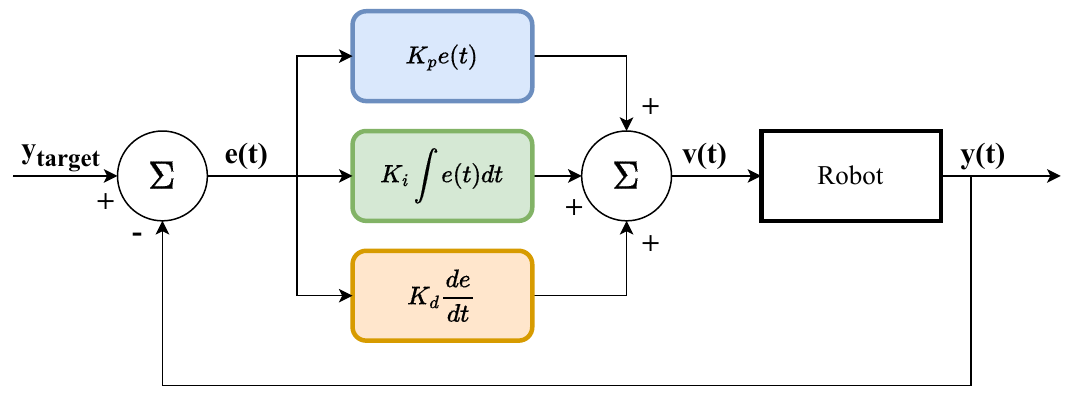}
    \caption{A block diagram depicting the PID controller.}
    \label{pid_control_block_diagram}
\end{figure}

%\paragraph{}
We use a PID controller to compute velocity commands that move the robot’s end-effector between DMP positions during task execution.
Given current and target positions, $\mathbf{y}(t)$ and $\mathbf{y}_{target}$, the error is: 
\begin{equation}
    \mathbf{e}(t) = \mathbf{y}_{target} - \mathbf{y}(t)
    \label{pid_error}
\end{equation}
The PID velocity control is then described by:
\begin{equation}
    \mathbf{v}(t) = K_p \mathbf{e}(t) + K_i \int \mathbf{e}(t)dt + K_d \frac{d\mathbf{e}}{dt}
    \label{pid_control}
\end{equation}
$K_p$, $K_d$, and $K_i$ represent constant gains.
Figure \ref{pid_control_block_diagram} depicts a block diagram of the control system.

%\paragraph{}
By tuning the respective gains, we can optimize for properties such as the smoothness and stability of motions.
In addition, the tight integration between motion and sensor-based obstacle avoidance corrections requires motions to be responsive and reliable or risk not adequately utilizing the feedback provided by the obstacle avoidance component.

%\paragraph{}
The neuromorphic pipeline described in this section was implemented using the ROS framework. Refer to Appendix \hyperref[appendix:implementation_details]{D} for implementation details.

\section{Evaluation Methodology}
\label{evaluation}

%\paragraph{}
Our implementation was evaluated in experiments of simple reaching tasks on a Kinova Gen3 arm (shown in Figure \ref{kinova_arm_configuration_experiments}) that involve static and dynamic obstacles.
These were designed to compare outcomes of task executions with and without the presented SNN-based obstacle avoidance module and thus verify the merits of our neuromorphic approach.
We initially tune and evaluate performance in simulations before running experiments on the robot for a final validation of the module and how well it transfers to the real world.
In the following, we describe these experiments, including the formalized task scenarios, metrics, and performance criteria.

\subsection{Simulation Experiments}
\label{evaluation:simulation_experiments}

%\paragraph{}
We use an adapted version of the Gen3 Gazebo simulation provided in the \textsf{ros\_kortex}\endnote{https://github.com/Kinovarobotics/ros\_kortex} software package (see Appendix \hyperref[appendix:implementation_details]{D}).
By varying task variables such as the visual background and obstacle properties, we utilize sets of distinct task \textit{scenarios} in a tuning $\rightarrow$ validation $\rightarrow$ testing procedure:
\begin{enumerate}
    \item \textbf{Tuning}: manually tune pipeline parameters to establish candidate sets of parameter values.
    \item \textbf{Validation}: evaluate tuned parameter sets to identify the best-performing set.
    \item \textbf{Testing}: perform final experiments with the best-performing set.
\end{enumerate}

\subsubsection{Evaluation Tasks}
\label{simulation_evaluation_tasks}

%\paragraph{}
We formulated four goal-directed \textit{tasks} for evaluating obstacle avoidance capabilities.
In each task, a regular course of action leads to an imminent collision, unless a deliberate avoidance action is taken.
Therefore, we address situations in which na\"ive motion planning is certain to fail and investigate how well our obstacle-aware manipulation trajectories solve the problem.

%\paragraph{}
The set of tasks in our simulation experiments consists of:
\begin{itemize}
    \item \textbf{Task 1}: The arm must reach a goal position that lies behind a \textbf{static} obstacle.
    \item \textbf{Task 2}: The arm must reach a goal position as a \textbf{dynamic} obstacle enters the FOV and crosses the end-effector’s path.
    \item \textbf{Task 3}: The arm must reach for an object on a table that is partially occluded by a \textbf{static} obstacle.
    \item \textbf{Task 4}: The arm must maintain its initial position as a \textbf{dynamic} obstacle moves directly towards it; the arm is free to move but must constantly minimize distance to the initial position.
\end{itemize}

%\paragraph{}
Refer to Figures \crefrange{task_1_demonstrative}{task_4_demonstrative} in Appendix \hyperref[appendix:scenario_variables]{A} for visual illustrations of the tasks.

%\paragraph{}
Within these tasks, we vary the background and the obstacle type, color, and speed. These \textit{task variables} and their values are listed in Table \ref{task_variables} and visualized in Appendix \hyperref[appendix:scenario_variables]{A}.
We define a \textit{scenario} as a task instance that is characterized by a unique set of variable values.
Apart from examining the robustness of our method, testing in various scenarios is useful in identifying limitations and can provide a validation of applicability in various environmental conditions before transferring to a real robot.

\begin{table*}
    \small\sf
    \centering
    \caption{Task variables that define experimental scenarios. Example scenario (gray cells): \{Task 4, Bookstore, Box, Brick Pattern, Medium\}. Approximate obstacle speeds: Low = $0.09m/s$, Medium = $0.17m/s$, High = $0.36m/s$.}
    \begin{tabularx}{\textwidth}[t]{p{0.1\linewidth}p{0.15\linewidth}XXXX}
        \toprule
        Task & Variable & Values & & &\\
        \midrule
        
        \multirow{1}{*}{Task 1} & \multicolumn{1}{l}{Background} & Empty & Office & Bookstore & Kitchen \\
        & \multicolumn{1}{l}{Obstacle Type} & Box & Buckyball & Spiky Sphere & Rock \\
        & \multicolumn{1}{l}{Obstacle Color} & White & Red & Yellow-Black & Brick Pattern\\
        \hline
        
        \multirow{1}{*}{Task 2} & \multicolumn{1}{l}{Background} & Empty & Office & Bookstore & Kitchen \\
        & \multicolumn{1}{l}{Obstacle Type} & Box & Buckyball & Spiky Sphere & Rock \\
        & \multicolumn{1}{l}{Obstacle Color} & White & Red & Yellow-Black & Brick Pattern\\
        & \multicolumn{1}{l}{Obstacle Speed} & Low & Medium & High\\
        \hline
        
        \multirow{1}{*}{Task 3} & \multicolumn{1}{l}{Background} & Empty & Office & Bookstore & Kitchen \\
        & \multicolumn{1}{l}{Obstacle Color} & White & Red & Yellow-Black & Brick Pattern\\
        \hline
        
        \multirow{1}{*}{Task 4} & \multicolumn{1}{l}{Background} & Empty & \lightgraycell Bookstore & Kitchen \\
        & \multicolumn{1}{l}{Obstacle Type} & \lightgraycell Box & Buckyball & Spiky Sphere & Rock \\
        & \multicolumn{1}{l}{Obstacle Color} & White & Red & Yellow-Black & \lightgraycell Brick Pattern\\
        & \multicolumn{1}{l}{Obstacle Speed} & Low & \lightgraycell Medium & High\\
        
        \bottomrule
    \end{tabularx}
    \label{task_variables}
\end{table*}

\subsubsection{Tuning, Validation and Testing}
\label{tuning_validation_testing}

%\paragraph{}
The task variables form a distribution of 416 task scenarios which we utilize for parameter tuning in addition to the final evaluation.
Inspired by a common methodology in machine learning (ML) research, we draw disjoint sets of scenarios that we use to initially optimize and ultimately evaluate through a \textit{tuning}, \textit{validation}, and \textit{testing} strategy.
The \textit{tuning set} is used to tune sets of parameter values to achieve adequate performance, while the \textit{validation set} is used to evaluate the degree to which a given set (or "model") generalizes to unseen conditions and to select a candidate set.
Our experiments involve testing with the selected values on the \textit{testing set} and comparing to executions that do not utilize the avoidance module.
This eliminates biases that could result from optimizing and evaluating on the same data and ensures that we do not "overfit" to specific environmental conditions.

%\paragraph{}
The tunable parameters that we consider govern the behaviour of each pipeline component outlined in section \ref{proposed_approach} and are listed in Table \ref{pipeline_component_parameters} of Appendix \hyperref[appendix:component_parameters]{E}.

%\paragraph{}
From the scenario distribution, we allocated 3 in the tuning set, 8 in the validation set, and 20 in the testing set.
The first two were manually selected to guarantee some variation in scenario properties (with the validation set containing more variations, some unobserved during tuning) and to ensure that at least one task and one background are never observed in either phase.
Testing scenarios were sampled randomly from the remaining pool, with the only restriction being at least $N_s=5$ scenarios per task, resulting in a set that includes a novel task and a background/environment.
In this case, these were \textit{Task 2} and the \textit{Kitchen} environment (which is characterized by dim ambient lighting).

%\paragraph{}
Table \ref{tuning_validation_testing_scenarios} in Appendix \hyperref[appendix:experiment_scenarios]{B} lists the scenarios of each set.
Note the inclusion of an additional scenario at the top (Task 1, Empty, Cracker Box), which was used to establish baseline parameter values in initial tests, termed a \textit{pre-tuning} phase.

\subsubsection{Evaluation Metrics and Criteria}
\label{evaluation_metrics_and_criteria}

%\paragraph{}
The relevant literature presents scarce results of obstacle avoidance performance and limited consensus on the appropriate quantitative metrics thereof.
We address this deficit by establishing a set of quantitative metrics and qualitative criteria for our evaluation.

%\paragraph{}
The quantitative metrics we use to analyze the performances of parameter sets and to ultimately evaluate our approach are:
\begin{itemize}
    \item Task execution time, $T$
    \item Trajectory length, $l_{\mathbf{Y}}$
    \item Number of collisions, $N_{collisions}$
    \item Final distance to goal, $d_G$
    \item Success (reaching the goal and never colliding), $S$
    \item End-effector velocities and accelerations, $\mathbf{\dot{y}}$ and $\mathbf{\ddot{y}}$
\end{itemize}

%\paragraph{}
Refer to Appendix \hyperref[appendix:evaluation_metrics_and_criteria_details]{C} for detailed descriptions of each.
These metrics facilitate a holistic evaluation of the SNN-based obstacle avoidance module and comparisons to baseline task executions on the basis of task success and along other dimensions that describe trajectory properties.

%\paragraph{}
In addition, we define qualitative criteria to describe trajectory properties to examine the extent to which they deviate from the baseline.
Ideally, adapted trajectories would be successful while remaining qualitatively similar.
In order to conduct a formal evaluation, we ground each criterion in terms of quantitative measures, where possible, to avoid subjective assessments.
These criteria are presented in Table \ref{qualitative_criteria} and described in more detail in Appendix \hyperref[appendix:evaluation_metrics_and_criteria_details]{C}.

\begin{table}
    \small\sf
    \centering
    \caption{Qualitative performance criteria.}
    \begin{tabularx}{\columnwidth}[t]{@{}l>{\raggedright\arraybackslash}X@{}}
        \toprule
        Criterion & Evaluated Through\\
        \midrule
        Reliability
        & Ratio of successful trials to total no. of executions in imminent collision cases\\
        Predictability
        & Frequency/magnitudes of heading changes\\
        Safety
        & Magnitudes of velocities and accelerations\\
        \bottomrule
    \end{tabularx}
    \label{qualitative_criteria}
\end{table}

\subsubsection{Experiment Procedure}

%\paragraph{}
Our experiments involve running $N_{trials}$ executions in all testing scenarios in each of two cases:
\begin{enumerate}
    \item Baseline executions that do not involve the obstacle avoidance module.
    \item Executions that incorporate the obstacle avoidance module, using the selected parameter set.
\end{enumerate}

We then analyse and compare performance using the described metrics and criteria.

\subsection{Real Experiments}
\label{evaluation:real_experiments}

\begin{figure}
    \centering
    \begin{subfigure}{0.47\linewidth}
        \centering
        \includegraphics[width=\columnwidth, trim={0 165pt 0 0},clip]{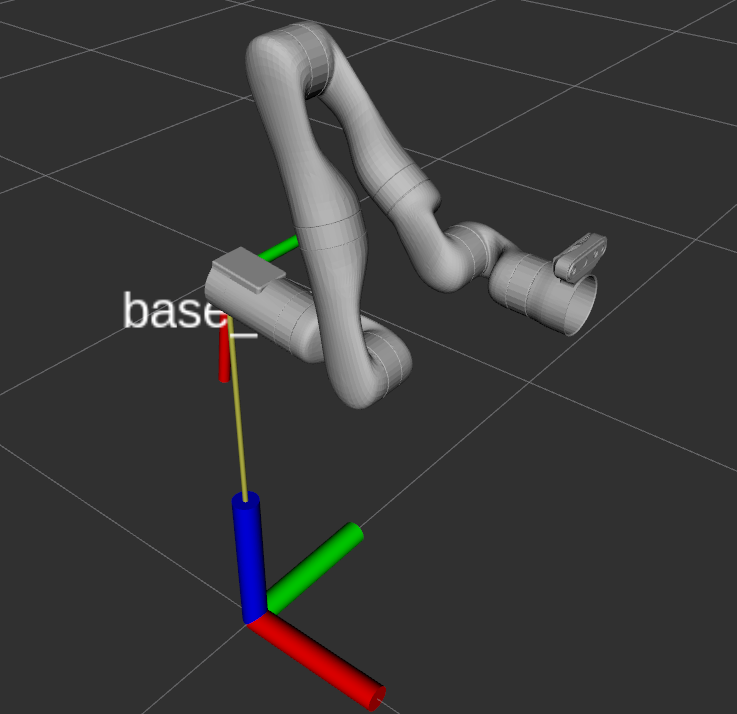}
        \caption{\centering In simulation.}
        \label{kinova_arm_configuration_experiments:sim}
    \end{subfigure}%
    \hspace{1em}
    \begin{subfigure}{0.47\linewidth}
        \centering
        \includegraphics[width=\columnwidth]{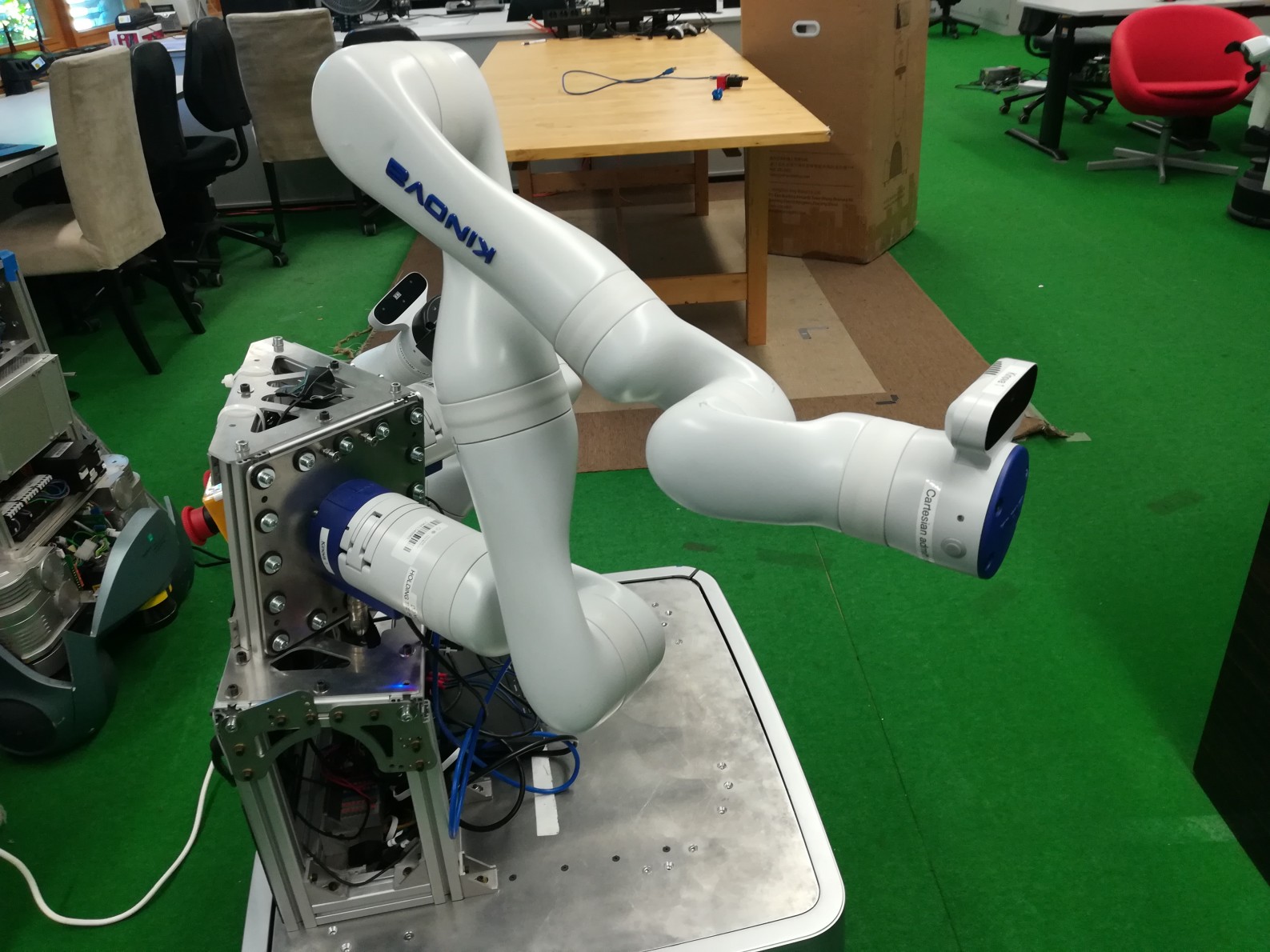}
        \caption{\centering The robot platform.}
        \label{kinova_arm_configuration_experiments:real}
    \end{subfigure}%
    \caption{The Kinova Gen3 arm configuration used in the experiments.}
    \label{kinova_arm_configuration_experiments}
\end{figure}

%\paragraph{}
The real experiments are conducted on a Kinova Gen3 arm attached to a mobile platform (see Figure \ref{kinova_arm_configuration_experiments:real}), to determine how well the tuned parameters
transfer to the real world and for a more concrete validation of our approach.
These experiments contain fewer variations of scenarios from a subset of the tasks defined in section \ref{simulation_evaluation_tasks} but share the same evaluation metrics/criteria and experimental procedure.

\subsubsection{Evaluation Tasks}
\label{real_evaluation_tasks}

\begin{figure}
    \centering
    \hspace{-1em}
    \begin{subfigure}{0.5\linewidth}
        \centering
        \includegraphics[width=\columnwidth]{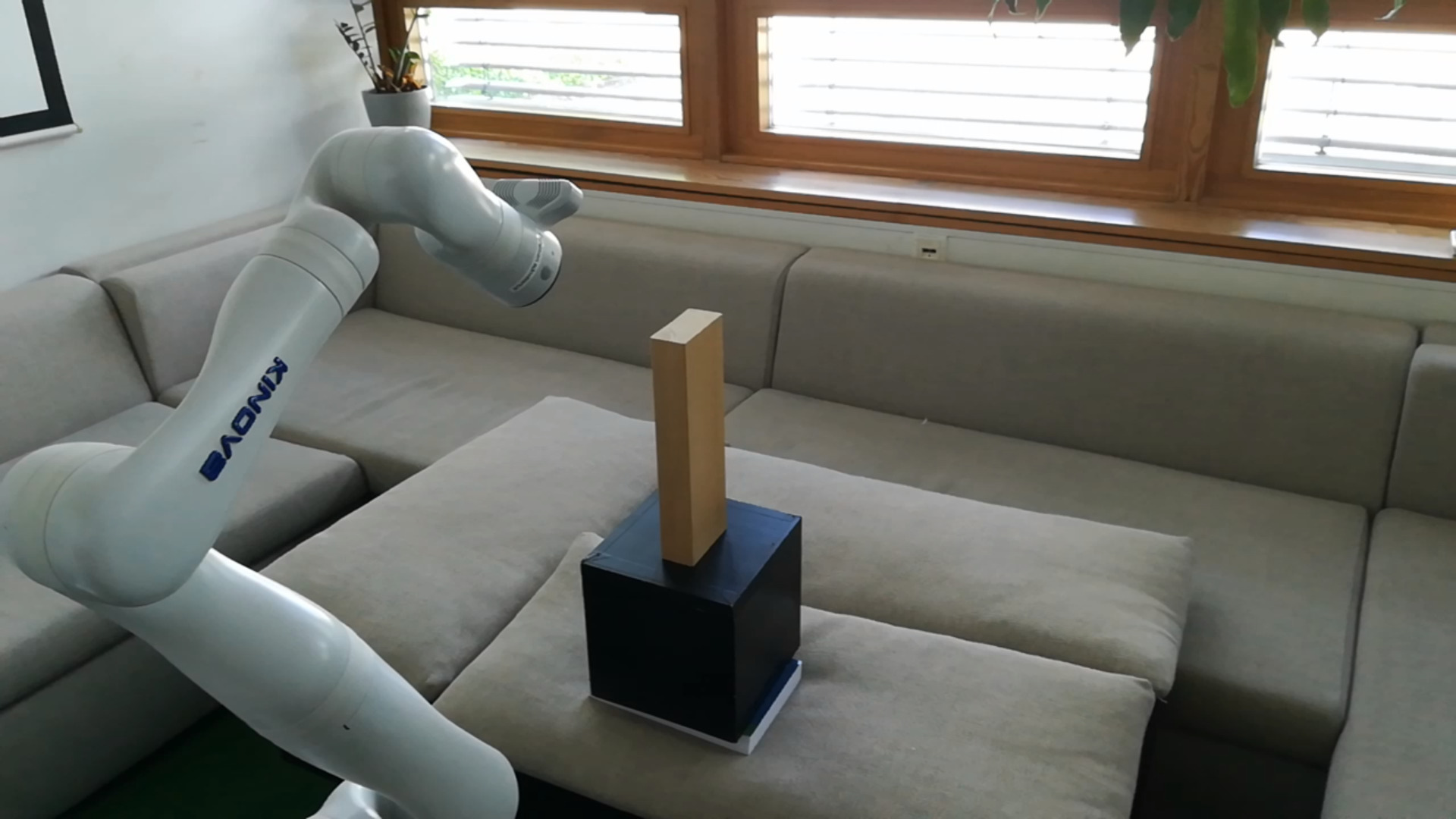}
        \caption{\centering Start}
        \label{}
    \end{subfigure}%
    \hspace{0.5em}
    \begin{subfigure}{0.5\linewidth}
        \centering
        \includegraphics[width=\columnwidth]{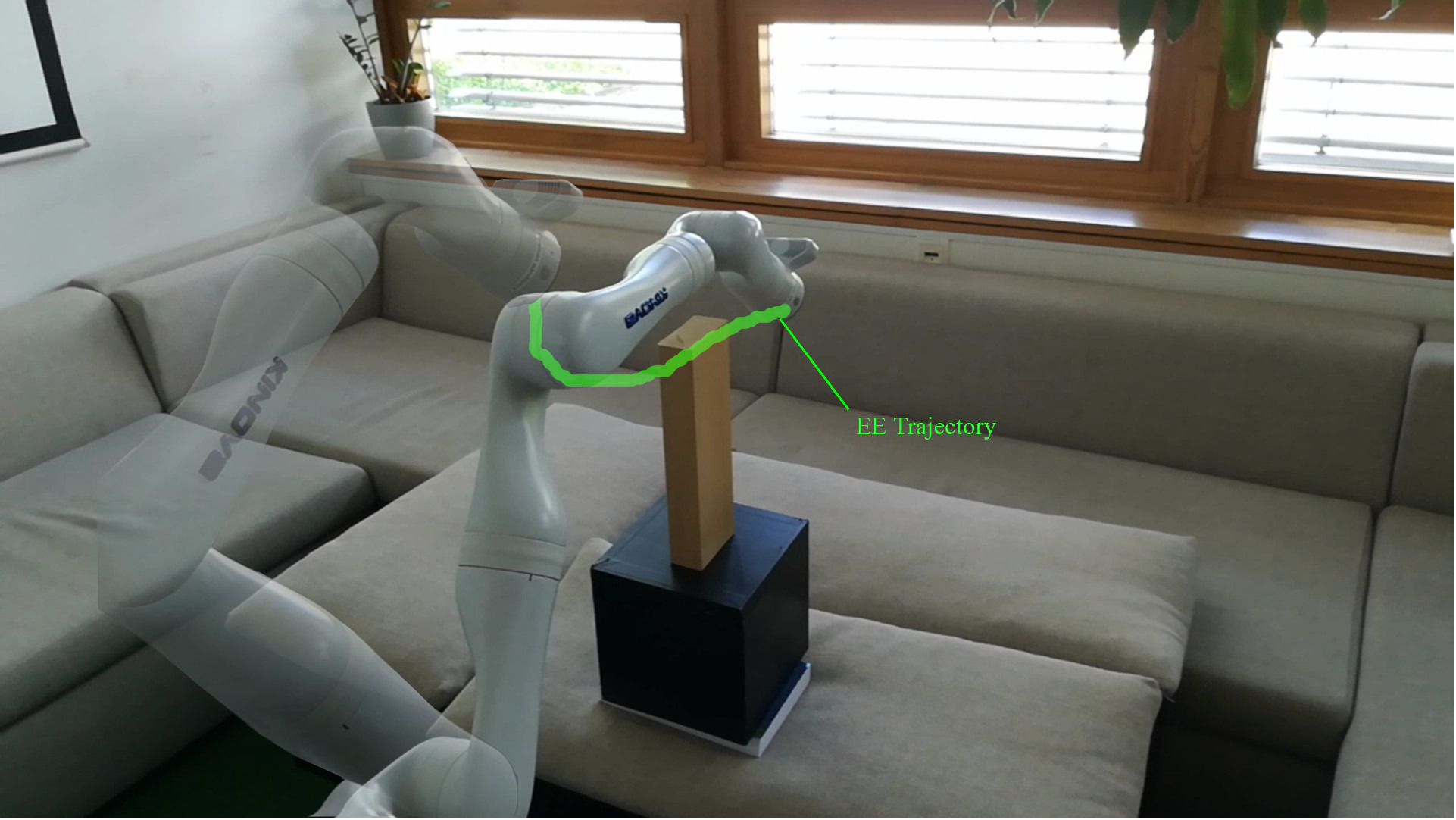}
        \caption{\centering End}
        \label{}
    \end{subfigure}%
    \hspace{-1em}
    \caption{Real experiment Task 1 setup. A baseline end-effector trajectory is illustrated in green.}
    \label{real_robot_task_1_demonstrative}
\end{figure}

\begin{figure}
    \centering
    \hspace{-1em}
    \begin{subfigure}{0.5\linewidth}
        \centering
        \includegraphics[width=\columnwidth]{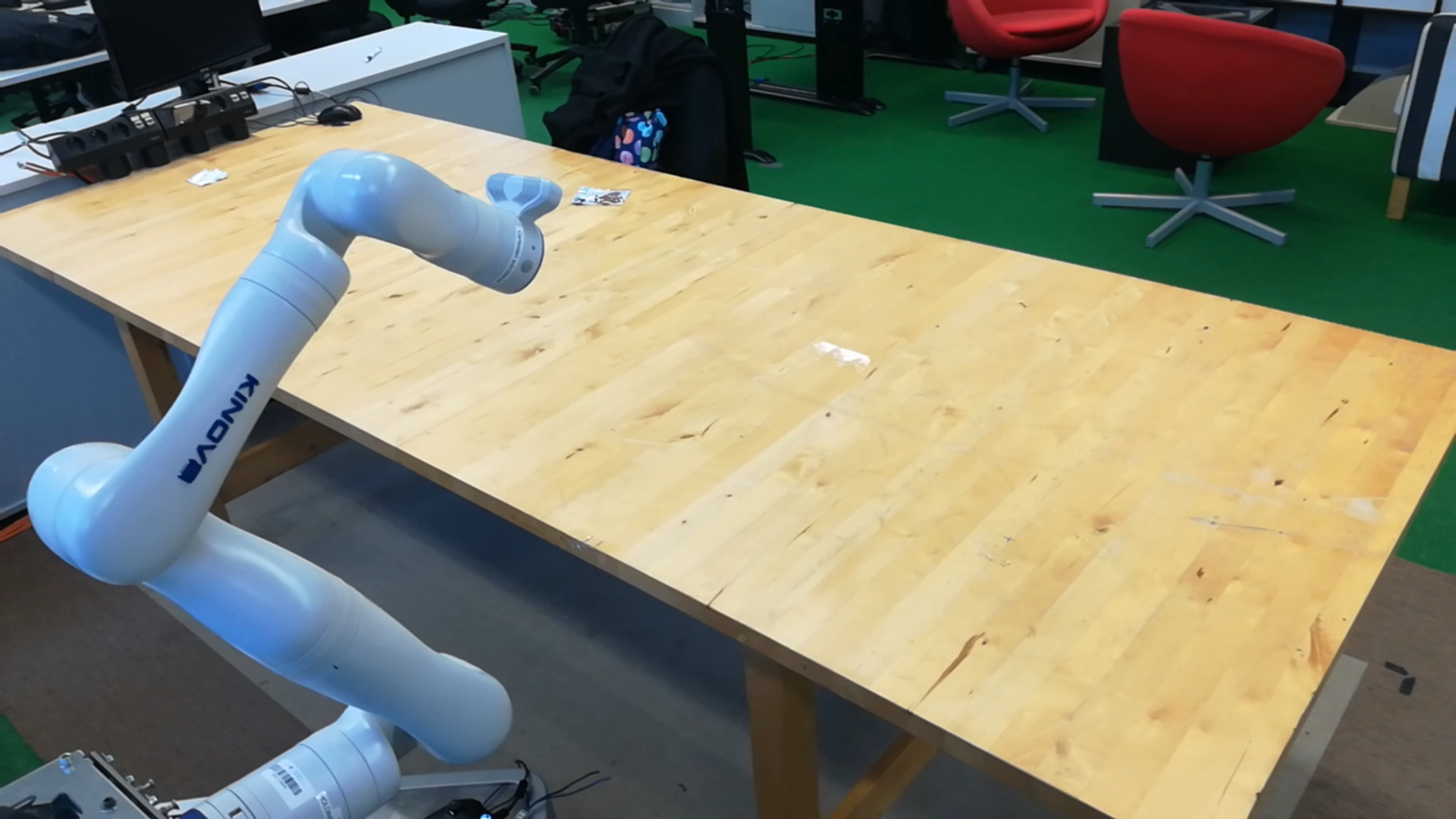}
        \caption{\centering Start}
        \label{}
    \end{subfigure}%
    \hspace{0.5em}
    \begin{subfigure}{0.5\linewidth}
        \centering
        \includegraphics[width=\columnwidth]{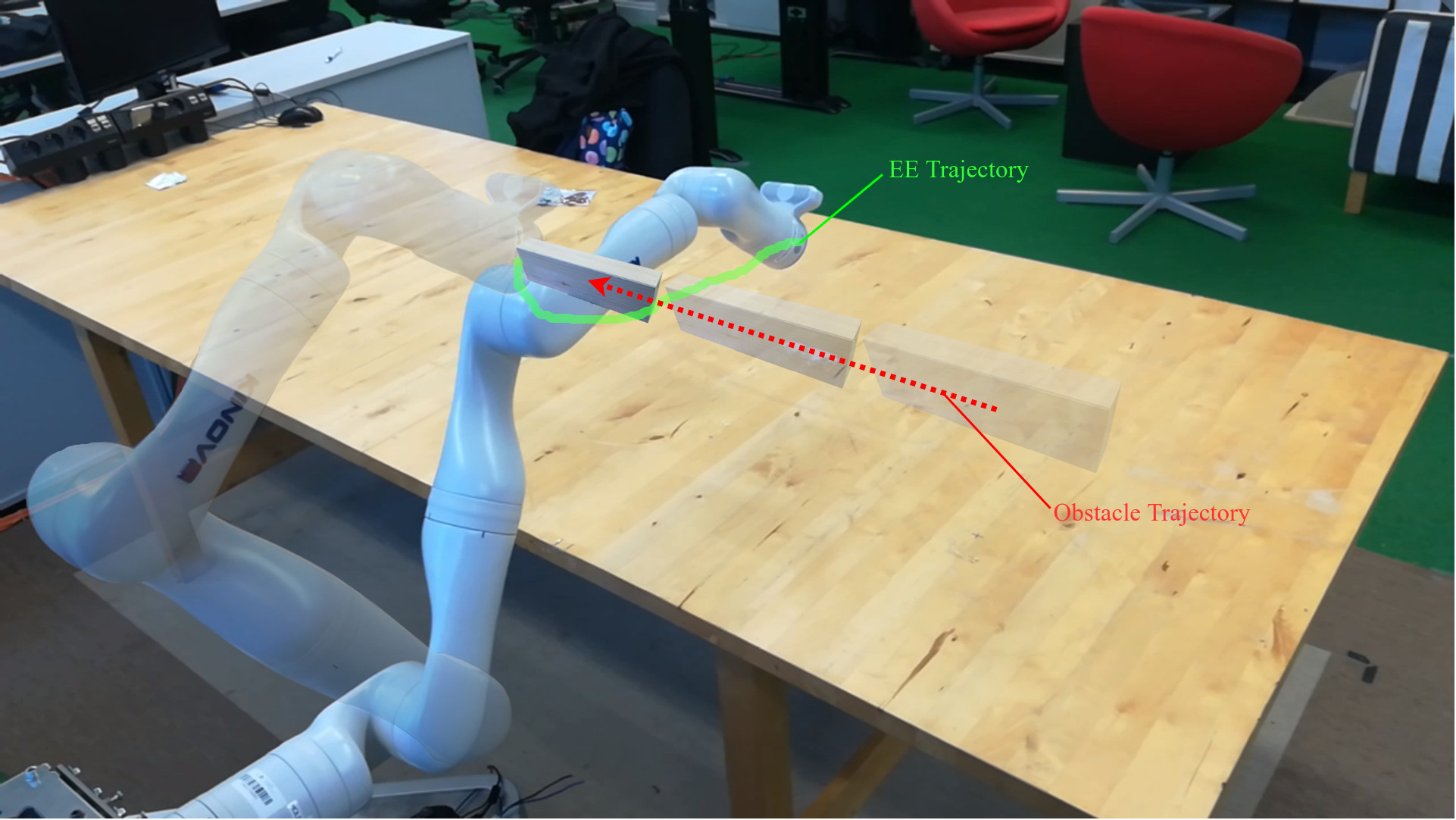}
        \caption{\centering End}
        \label{}
    \end{subfigure}%
    \hspace{-1em}
    \caption{Real experiment Task 2 setup. The obstacle's trajectory and a baseline trajectory are illustrated in red and green.}
    \label{real_robot_task_2_demonstrative}
\end{figure}

\paragraph{}
We consider two of the tasks defined in section \ref{simulation_evaluation_tasks}:
\begin{itemize}
    \item \textbf{Task 1}: The arm must reach a goal position that lies behind a \textbf{static} obstacle.
    \item \textbf{Task 2}: The arm must reach a goal position as a \textbf{dynamic} obstacle enters the FOV and crosses the end-effector’s path.
\end{itemize}
The task setups are shown in Figures \ref{real_robot_task_1_demonstrative} and \ref{real_robot_task_2_demonstrative}.

%\paragraph{}
These tasks are equivalent to the simulated versions with a few exceptions.
Firstly, the object in Task 1 was suspended but is now placed on a surface.
For Task 2, the simulation offers accurate control of the obstacle’s trajectory and thus a high degree of consistency across trials.
On the other hand, the object’s trajectory is controlled manually in the real experiments, which may introduce inter-trial inconsistencies, although preventative steps such as guiding markers on surfaces and periodic measurements to verify the object’s initial position were taken.
While the simulated experiments were designed to gather statistical evidence from highly repeatable trials, which rarely occur in the real world, the primary objective here is to transition into real-life conditions and investigate how well the implementation adapts. 
Furthermore, for reasonably small variations in conditions, results from multiple trials can eliminate any variance due to these imprecisions.

%\paragraph{}
The real experiment task scenarios are varied exclusively along the “Background” and “Obstacle Type” variables.
“Lab Background 1” is situated in an area near a window providing a natural but dim light (see Figure \ref{real_robot_task_1_demonstrative}) while “Lab Background 2” has a much brighter artificial lighting and contains a large table and other objects (see Figure \ref{real_robot_task_2_demonstrative}).
The obstacles are a wooden block, a metal bar, and a person’s hand.
The latter is a particularly relevant case for human-robot collaborative scenarios\endnote{The end-effector speed posed negligible safety risks. Nevertheless, executions were carefully monitored and an emergency stop was available to interrupt the execution at any point.}.
Table \ref{real_experiments_scenarios} in Appendix \hyperref[appendix:experiment_scenarios]{B} lists the four scenarios we conduct our real robot experiments in.

\subsubsection{Evaluation Metrics and Criteria}
\label{evaluation:real_experiments:evaluation_metrics_and_criteria}

%\paragraph{}
The performance is evaluated using the same metrics and criteria of the simulation experiments.
However, the $N_{collisions}$ metric was excluded since it was only possible to evaluate in simulations by disabling collision dynamics and counting the number of instances in which the end-effector intersects an obstacle model.
Instead, collisions are only reflected in the execution success, as before, which is true only if the end-effector never collides and also reaches its goal.

\section{Results and Discussion}
\label{results_and_discussion}

\subsection{Simulation Experiments}
\label{results_and_discussion:simulation_experiments}

%\paragraph{}
Parameter tuning, validation, and testing in the simulation experiments was preceded by a \textit{pre-tuning} phase, where an initial set of values was established.
Next, we tuned several variants of this set on the tuning scenarios, validated their performance on the validation scenarios, and finally selected the best-performing set to run on the testing scenarios.

\begin{figure*}
    \centering
    \begin{subfigure}{0.27\linewidth}
        \centering
        \includegraphics[width=\textwidth, trim={90pt 10pt 83pt 25pt},clip]{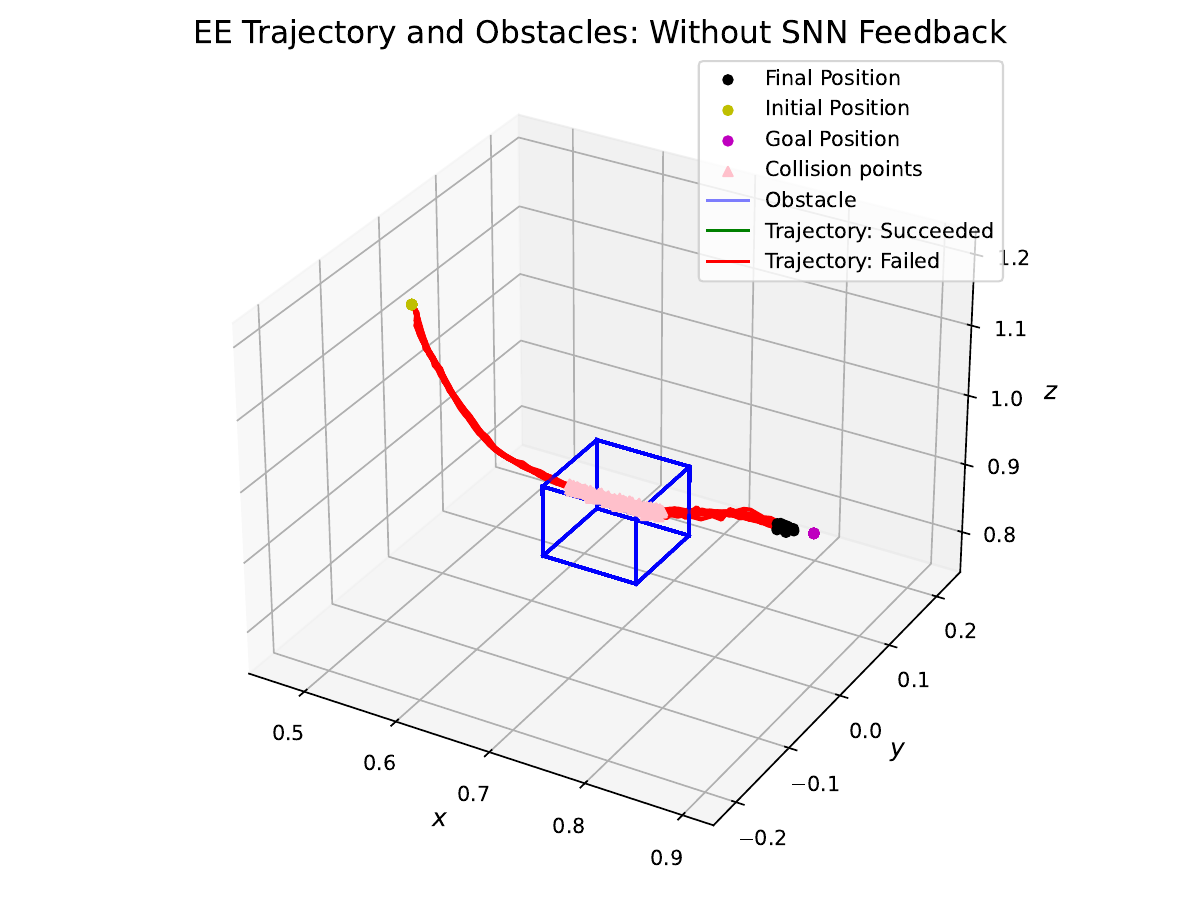}
        \caption{\centering Without SNN feedback}
        \label{trial_trajs:without_obstacle_avoidance}
    \end{subfigure}%
    \hspace{2em}
    \begin{subfigure}{0.27\linewidth}
        \centering
        \includegraphics[width=\linewidth, trim={90pt 10pt 83pt 25pt},clip]{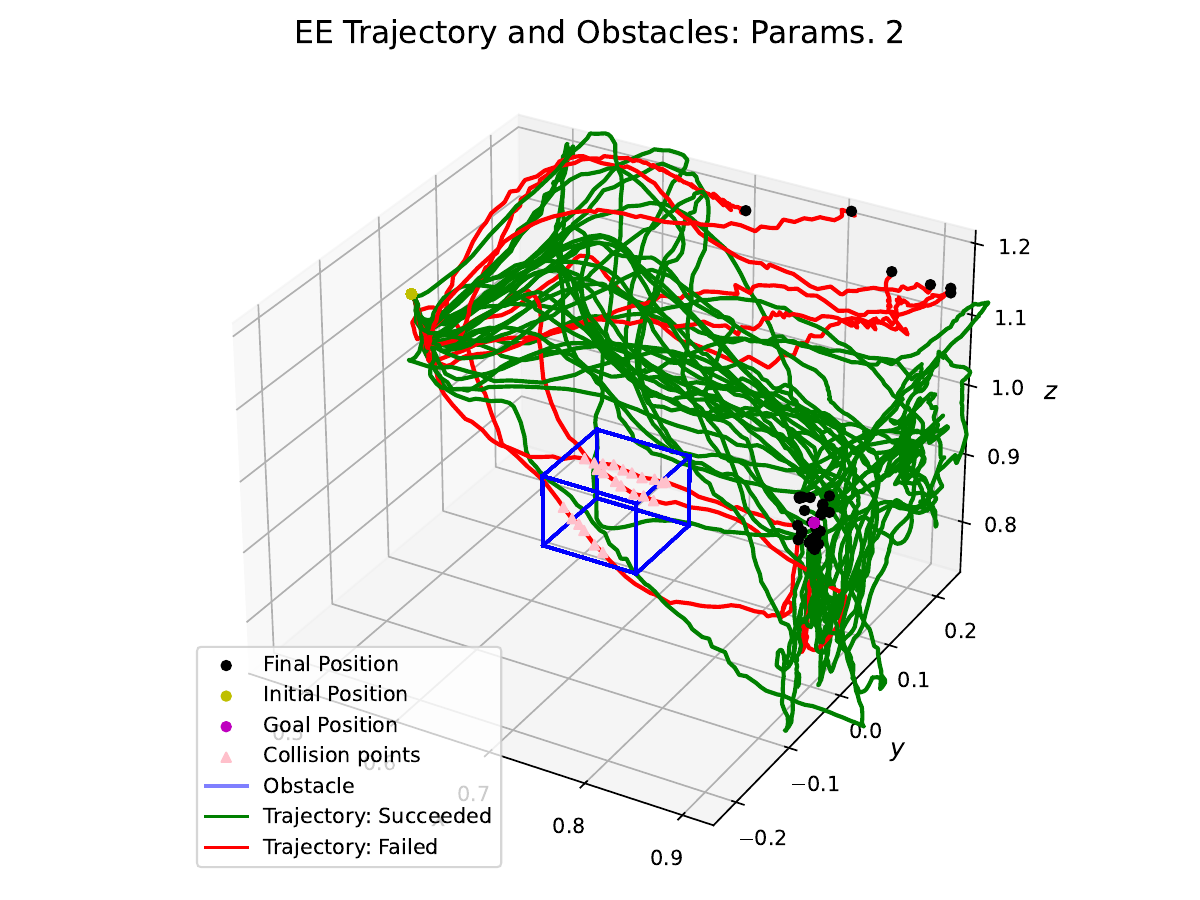}
        \caption{\centering With pre-tuning parameter set}
        \label{trial_trajs:pre_tuning}
    \end{subfigure}%
    \hspace{2em}
    \begin{subfigure}{0.27\linewidth}
        \centering
        \includegraphics[width=\linewidth, trim={90pt 10pt 83pt 25pt},clip]{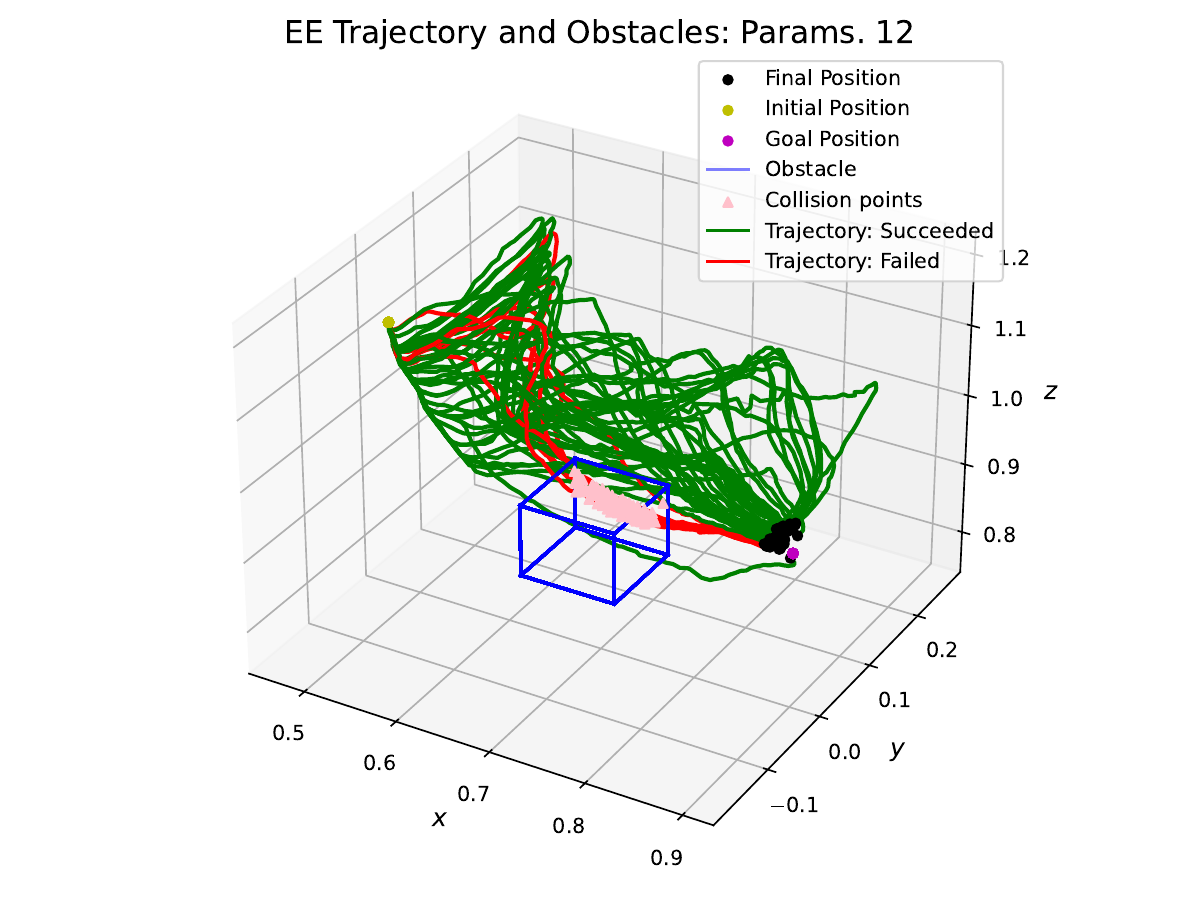}
        \caption{\centering With post-tuning parameter set (12)}
        \label{trial_trajs:post_tuning_param_set_12}
    \end{subfigure}%
    \\
    \caption{Trajectories executed during pre-tuning and tuning trials.}
    \label{trial_trajs}
\end{figure*}

%\paragraph{}
Figure \ref{trial_trajs} illustrates the evolution of executed trajectories during this process in a Task 1 scenario.

\subsubsection{Initial Parameterization (Pre-Tuning Phase)}
\label{initial_parameters}

%\paragraph{}
This phase consisted of iterative testing to search for a region in the parameter space that leads to acceptable performance.
The main targets of this optimization were motion control parameters, such as the PID gains, distance tolerances, and motion loop frequency; PF parameters; SNN parameters, including the architecture, weight initialization, and SNN dynamics variables; and event emulation variables, such as RGB vs. grayscale inputs and the event emission threshold ($\theta$).
These tests were conducted in a pre-tuning scenario: a variant of Task 1 containing a “Cracker Box” object that is excluded from the task distribution defined in Table \ref{task_variables}.

\begin{table*}
    \small\sf
    \centering
    \caption{The SNN architecture that was selected during the pre-tuning phase.}
    \begin{tabularx}{\textwidth}[t]{p{0.15\linewidth}p{0.1\linewidth}YYYY}
        \toprule
        Layer & Type & Kernel Size & Stride Size & Input Size & Output Size\\
        \midrule
        Input & - & - & - & - & 120 $\times$ 160\\
        Layer 1 & LIF (conv) & 8 $\times$ 8 & 4 $\times$ 4 & 120 $\times$ 160 & 29 $\times$ 39\\
        Layer 2 (Output) & LIF (conv) & 4 $\times$ 4 & 4 $\times$ 4 & 29 $\times$ 39 & 13 $\times$ 18\\
        \bottomrule
    \end{tabularx}
    \label{snn_architecture}
\end{table*}

%\paragraph{}
For the SNN architecture, we chose a two-layer network of the LIF neurons presented in \cite{diehl2015unsupervised}. 
It consists of a layer that propagates input spike trains and two layers of spiking neurons, whose specifications are shown on Table \ref{snn_architecture}.
We set $v_{thresh}$, $v_{reset}$, $v_{rest}$, $T_{refrac}$, and $\tau_v$ to the default values of the original paper, which are intended to match biologically plausible ranges.
However, we increased $v_{reset}$ and $v_{rest}$ (from -65.0 to -62.0) to encourage more frequent spiking.
The simulation time ($T_{sim}$), which controls the time period of a single pass, was set to 20 (ms), while the weight initialization factor, $w_c$, was set to 7.0.

%\paragraph{}
The pre-tuning parameters were evaluated by running $N_{trials}=30$ trials in both testing cases.
As expected, the trajectories executed in the baseline case all fail by colliding with the obstacle, (see Figure \ref{trial_trajs:without_obstacle_avoidance}).
On the other hand, utilizing the module leads to trajectories that are adapted to avoid the obstacle while moving towards the goal (Figure \ref{trial_trajs:pre_tuning}), which were 80\% successful in these trials.

%\paragraph{}
The obstacle avoiding trajectories tended to have higher execution times, $T$, and trajectory lengths, $l_{\mathbf{Y}}$, approximately by factors of 2 and 1.5, respectively.
In addition, these values had higher variances, reflecting variations in the executed trajectories across trials.
The average number of collisions per trial, $N_{collisions}$, was reduced from approximately 7.8 to 0.6.
A significant fraction of failures occurred due to executions ending before the end-effector had reached the goal, i.e. $d_G > \delta_{\mathbf{g}}$\endnote{If the goal is not reached before a timeout period (60 seconds by default), the trial is aborted and the position of the end-effector at that point in time is recorded as its final position.}, because the applied velocities occasionally caused the arm to move into singular positions that complicate returning back to the intended path.

\subsubsection{Tuning Results}
\label{tuning_results}

%\paragraph{}
Using the tuning scenarios, we derived and iteratively optimized 12 parameter sets that were expected to improve results.
Furthermore, we extended our implementation and parameters to address some failures that we had observed in pre-tuning and initial tuning trials.

%\paragraph{}
In order to address the problem of reaching singular arm configurations, which could lead to failures and potentially unsafe executions, we implemented a safety strategy that discourages excessive motions away from the pre-planned trajectory by reducing velocities and accelerations that move the end-effector further beyond a definite \textit{safety boundary}.
This method is described in the colored box below.
Note that for Task 4, the safety strategy penalizes motions that deviate a certain distance from the initial position.
\begin{tcolorbox}[boxsep=6pt,left=2pt,right=2pt,top=2pt,bottom=2pt]
    \textbf{Safety Strategy:}
    If the distance of the current end-effector position from the nearest point on the pre-planned trajectory exceeds a safety threshold:
    \vspace{-0.5em}
    \begin{equation}
    || \mathbf{y}(t) - \hat{\mathbf{y}}_{ref} ||_2 < \delta_{safety, 1}
    \label{safety_strategy_condition_1}
    \end{equation}
    and the current position is further away from the trajectory point than the last recorded position:
    \vspace{-.5em}
    \begin{equation}
    || \mathbf{y}(t) - \mathbf{y}_{ref} ||_2 > || \mathbf{y}(t-1) - \mathbf{y}_{ref} ||_2
    \label{safety_strategy_condition_2}
    \end{equation}
    slow down by reducing the current commanded velocity and the $\vect{\phi}$ values that define the next acceleration values:
    \vspace{-.5em}
    \begin{equation}
    \mathbf{v}(t) = \gamma_{v, 1} \mathbf{v}(t)
    \label{safety_strategy_equation_1}
    \end{equation}
    \vspace{-1.5em}
    \begin{equation}
    \bm{\phi}(t+1) = \gamma_{a, 1} \bm{\phi}(t)
    \label{safety_strategy_equation_2}
    \end{equation}
\end{tcolorbox}

%\paragraph{}
We also added optional workspace boundaries: positional limits beyond which a planned DMP position is rectified by clipping its value:
\begin{equation}
    \mathbf{y}_i = min(\delta^{+}_{pos, i}, max(\delta^{-}_{pos, i}, \mathbf{y}_i(t))), \forall i \in \{x, y, z\}
\end{equation}
Here, $\delta^{-}_{pos, i}$ and $\delta^{+}_{pos, i}$ represent lower and upper positional limits along dimension $i$ (set to $-\infty$ and $\infty$ by default).
This addresses safety concerns due to the end-effector colliding with the edge or moving under the table in Task 3, but could be set for different 
task conditions.
Otherwise, various tunable parameters (see Table \ref{pipeline_component_parameters} of Appendix \hyperref[appendix:component_parameters]{E}) were optimized to form the 12 candidate parameter sets.

%\paragraph{}
We ran batches of trials without SNN feedback (baseline) and with SNN feedback, the latter with each of the parameter sets, on each of the three tuning scenarios (1, 2, and 3).
The baseline batches consisted of $N_{trials}=30$ trials, while the rest consisted of $N_{trials}=40$ each for a total of 1530 trials.

%\paragraph{}
The trajectories executed in scenario 1 are shown in Appendix \hyperref[appendix:metrics_and_trajectory_plots]{H} in Figure \ref{tuning_scenario_1_trial_trajs_params1_to_params12}.

%\paragraph{}
Scenario 1 trials confirmed that the safety strategy reduced instances of terminal arm configurations.
Some parameter sets (particularly 6-9) exhibited less reactive motions due to stronger event erosion filter parameterizations, which lead to more collisions.
Similar failures were observed in scenario 2 trials for the first nine parameter sets, which were also attributed to a weak response due to parameters that control \textit{sensitivity} to inputs, including the event emission and filter thresholds, neuronal spiking thresholds, and $\vect{\phi}$ acceleration parameters.
Sets 10-12 were tuned to address this limitation and were successful in improving results.
Scenario 3 trials revealed that earlier parameter sets suffered from the opposite: a higher reactivity lead to unstable motions.
These observations point to a trade-off: increasing sensitivity to inputs at the perceptual, motion, or intermediate levels in the pipeline may lead to excessive, dangerous or oscillatory motions, while decreasing sensitivity may risk not reacting fast enough to avoid collisions.

\subsubsection{Validation Results}
\label{validation_results}

%\paragraph{}
Next, we selected a subset of the parameter sets on the basis of best average performance and trajectory properties: 5, 8, 10, and 12.

%\paragraph{}
Due to initial observations and unsatisfactory performance in a challenging, high-speed validation scenario (8), we instantiated two additional sets: 13 and 14. 
These were aimed at exploring quicker responses and faster real-time performance, mainly by testing a single-layer SNN and tuning motion loop frequency parameters.

%\paragraph{}
We ran $N_{trials}=30$ baseline trials and $N_{trials}=40$ trials with the avoidance module parameterized by each of the six parameter sets. All trials were repeated in each of the eight validation set scenarios (4-11): a total of 2160 trials.

%\paragraph{}
Figure \ref{validation_scenarios_metrics} in Appendix \hyperref[appendix:metrics_and_trajectory_plots]{H} contains the metrics plots summarizing the quantitative performance in all scenarios.

%\paragraph{}
Results from Task 1 scenarios (4-6) indicated that the module performed worse in the "Office" environment (scenario 4) than in "Store" or "Empty".
A likely cause is the relatively lower illumination, which decreases contrasts between the background and the obstacle, thus leading to less events being generated and in turn more latency in or lower magnitudes of avoidance velocities.
Another potential reason is the relative background clutter which, despite event filtering, may produce more background events that saturate the overall response of the SNN.

%\paragraph{}
Results from Task 4 scenarios (7-9) varied significantly.
Most sets performed well with medium-speed obstacles (scenario 7) but considerably worse in response to the high-speed obstacle of scenario 8 (traveling about twice as fast).
While the module would react to the obstacle, which was visible for a shorter amount of time, the eventual response would not be sufficiently effective.
This was a product of fewer events, a resulting lower SNN activation, and the limited speed of the arm.
Sets 13 and 14 yielded more positive results through stronger responses, leading to less predictable and potentially unsafe trajectories.
In addition, these improvements did not extend to the low-speed case.

%\paragraph{}
All selected parameter sets performed at least better than the baseline and reasonably well in most scenarios.
However, the last findings indicate a limit on how fast we can command the arm to move before approaching dangerous speeds.
While sets 13 and 14 improved performance, they exhibited significantly higher accelerations that enable the necessary sudden reactions. 
This is characteristic of unsafe trajectories and presents an undesirable compromise.
Instead, it is reasonable to acknowledge an inability to reliably react to and avoid obstacles whose speeds exceed a certain threshold.

\subsubsection{Testing Results}
\label{testing_results}

%\paragraph{}
Parameter set 12 was selected for the experiments based on its superior average performance.
Following the same procedure, we ran batches of $N_{trials}=30$ trials \textit{without SNN feedback} (baseline) and $N_{trials}=40$ trials \textit{with SNN feedback} (set 12), on all testing scenarios (12-31) for a total of 1400 trials.
Figure \ref{testing_scenarios_12_to_31_metrics} in Appendix \hyperref[appendix:metrics_and_trajectory_plots]{H} shows the quantitative results from all scenarios.
Table \ref{testing_scenarios_metrics_table} contains the mean metric values in each scenario for the \textit{with SNN feedback} case, along with the task averages (across all scenarios of the same task).

\begin{table*}
    \small\sf
    \centering
    \caption{Quantitative results of the testing phase runs, averaged over the $N_{trials}$ trials of each scenario.}
    \begin{tabularx}{\textwidth}[t]{p{0.1\linewidth}>{\centering\arraybackslash}p{0.11\linewidth}>{\centering\arraybackslash}p{0.12\linewidth}>{\centering\arraybackslash}p{0.12\linewidth}>{\centering\arraybackslash}p{0.125\linewidth}>{\centering\arraybackslash}p{0.125\linewidth}>{\centering\arraybackslash}p{0.125\linewidth}>{\centering\arraybackslash}p{0.125\linewidth}}
        \toprule
        & & \multicolumn{5}{c}{Results (averages over $N_{trials}$ trials)}\\
        \cmidrule{3-7}
        & Scenario No. & Success, $S$ (\%) & Collisions, $N_{collisions}$ & Dist. to Goal, $d_G$ (m) & Traj. Length, $l_{\mathbf{Y}}$ (m) & Execution Time, $T$ (s) \\
        \midrule
        Task 1 & 12 & 95.0 & 0.625 & 0.026 & 0.965 & 12.200\\
        & 13 & 95.0 & 0.175 & 0.027 & 1.008 & 12.875\\
        & 14 & 95.0 & 0.125 & 0.027 & 0.995 & 12.997\\
        & 15 & 90.0 & 0.225 & 0.027 & 0.936 & 12.297\\
        & 16 & 87.5 & 0.450 & 0.027 & 0.926 & 12.260\\
        \midrule
        \multicolumn{2}{r}{\textbf{Task Average:}} & \textbf{92.5} & \textbf{0.320} & \textbf{0.027} & \textbf{0.966} & \textbf{12.526}\\
        
        \midrule
        Task 2 & 17 & 42.5 & 1.175 & 0.027 & 0.665 & 9.524\\
        & 18 & 32.5 & 1.650 & 0.027 & 0.655 & 9.590\\
        & 19 & 77.5 & 0.425 & 0.076 & 1.061 & 14.546\\
        & 20 & 85.0 & 0.250 & 0.028 & 1.157 & 14.734\\
        & 21 & 67.5 & 1.500 & 0.027 & 0.713 & 10.063\\
        \midrule
        \multicolumn{2}{r}{\textbf{Task Average:}} & \textbf{61.0} & \textbf{1.000} & \textbf{0.037} & \textbf{0.850} & \textbf{11.691}\\
        
        \midrule
        Task 3 & 22 & 70.0 & 3.400 & 0.028 & 1.034 & 13.924\\
        & 23 & 57.5 & 3.825 & 0.028 & 1.010 & 13.670\\
        & 24 & 82.5 & 1.450 & 0.028 & 1.317 & 16.500\\
        & 25 & 85.0 & 1.725 & 0.028 & 1.247 & 15.690\\
        & 26 & 77.5 & 1.775 & 0.029 & 1.154 & 14.674\\
        \midrule
        \multicolumn{2}{r}{\textbf{Task Average:}} & \textbf{74.5} & \textbf{2.435} & \textbf{0.028} & \textbf{1.152} & \textbf{14.892}\\
        
        \midrule
        Task 4 & 27 & 95.0 & 0.250 & 0.016 & 0.901 & 17.547\\
        & 28 & 0.0 & 2.375 & 0.027 & 0.421 & 7.780\\
        & 29 & 47.5 & 1.875 & 0.028 & 0.359 & 6.865\\
        & 30 & 95.0 & 0.200 & 0.008 & 0.420 & 10.320\\
        & 31 & 92.5 & 0.425 & 0.007 & 0.433 & 10.179\\
        \midrule
        \multicolumn{2}{r}{\textbf{Task Average:}} & \textbf{66.0} & \textbf{1.025} & \textbf{0.017} & \textbf{0.507} & \textbf{10.538}\\

        \bottomrule
    \end{tabularx}
    \label{testing_scenarios_metrics_table}
\end{table*}%

%\paragraph{}
The avoidance module succeeded in 92.5\% of all Task 1 trials.
The success rate, distance-to-goal, trajectory length, and execution time exhibited low variances, indicating a high level of consistency over the five environments, including the novel, low-light setting of scenario 14.

%\paragraph{}
Performance was less consistent in the novel Task 2, where success rate had an average and standard deviation of 61\% and 22.6\%, respectively.
Most failures were observed in the "Office" scenarios 17 and 18, where the relatively low mean trajectory length and execution time indicated less movement, possibly due to lower neural activations.
The avoidance behaviour succeeded more often in the remaining scenarios and was not affected by obstacle speed.

%\paragraph{}
Success in Task 3 scenarios averaged at 74.5\% and had a standard deviation of 11\%.
While the end-effector always reached the goal, $N_{collisions}$ varied significantly across trials of a given scenario and we observed no correlations between the different task variable values and the frequency of failures.
Occasionally, initial trajectory adaptations moved the end-effector to a region closer to the obstacle from which subsequent corrections were unlikely to effectively steer it away.
These occurrences indicate some uncertainty in the resultant trajectory which may be attributed to the cascade of non-linear operations performed within the pipeline.

%\paragraph{}
The reported mean success rate in Task 4 (66\%) is skewed due to failures in the high-speed obstacle scenario 28 (the median is 92.5\%).
This reinforces the validation phase findings concerning very fast obstacles.

%\paragraph{}
We computed the average time that elapses in a single iteration of each pipeline stage and present the results in Table \ref{computation_times}.
The computation times of the first two stages were also measured when the arm was stationary and no motion was induced in the image.
We observed that SNN computation time decreased in the latter case, i.e. when no input spikes are induced.
This indicates a positive correlation between the amount of SNN computations and the number of input events: a measure of new perceptual information.
Considering only visual changes to be relevant for the avoidance behaviour, the SNN thus expends computations only to process salient information in the context of the task.
Events efficiently encode this information and the SNN avoids unnecessary computations (i.e. computing and propagating spikes) in the absence of input spikes.
This input-dependent computational property distinguishes SNNs from conventional DNNs and has potential for better power and time efficiency in some applications.

%\paragraph{}
Finally, we evaluated executions based on our qualitative criteria (refer to Appendix \hyperref[appendix:evaluation_metrics_and_criteria_details]{C} for descriptions of each).

%\paragraph{}
We correlate \textit{reliability} with the consistency of positive results (i.e. success rates) in consistent task conditions.
Each row in Table \ref{testing_scenarios_metrics_table} represents the success rate in a given set of controlled conditions, the average of which is 74\%, including the scenario 28 outlier.
The median, which is less influenced by the outlier, is 84\%, and success rates were higher in half of the scenarios.
Overall, this indicated the moderate reliability of the implementation and chosen parameter values.

%\paragraph{}
We observed a high level of \textit{predictability} in motions by analyzing the magnitudes of directional changes.
The estimated values of heading angular velocities, $\dot{\zeta}$, had similar distributions in both cases and fell in the [-2,2] $deg/s$ range.
(A plot of these distributions for three scenarios are provided in Appendix \hyperref[appendix:qualitative_evaluation_visualizations]{G} in Figure \ref{angular_vel_dist_plots_testing_phase}.)
This suggests that the evolution of the trajectory is easy to predict from observations.

%\paragraph{}
For evaluating the \textit{safety} of obstacle-avoiding trajectories, we measured the overall end-effector speeds, which averaged at $\sim$0.07$m/s$.
According to the ISO standard for industrial robot safety requirements (\cite{ISO10218}), this is well below a tool speed threshold under which a robot can be considered to operate in a safety-rated, "reduced speed control" mode (\cite{beckert2017online}): $0.25m/s$.

\begin{table}
    \small\sf
    \centering
    \caption{Mean computation times (in seconds).}
    \begin{tabularx}{\columnwidth}[t]{p{0.27\linewidth}>{\centering\arraybackslash}X>{\centering\arraybackslash}X}
        \toprule
        Stage & \multicolumn{2}{c}{Computation Time (s)}\\
        \cmidrule{2-3}
        & During Executions & No Stimuli\\
        \midrule
        Event Emulation & 0.025 $\pm$ 0.001 & 0.025 $\pm$ 0.001\\
        SNN Simulation & \textbf{0.123 $\pm$ 0.014} & \textbf{0.086 $\pm$ 0.007}\\
        Obs. Avoidance Computations & 0.002 $\pm$ 0.001& -\\
        \midrule
        Total & 0.150 $\pm$ 0.016 & 0.111 $\pm$ 0.009\\
        \bottomrule
    \end{tabularx}	
    \label{computation_times}
\end{table}%

\subsection{Real Experiments}
\label{results_and_discussion:real_experiments}

%\paragraph{}
The pipeline implementation and tuned parameter values (12) were directly transferred to a real Kinova Gen3.
Preliminary tests showed acceptable performance except for a slight degradation in motion smoothness, which necessitated minor adjustments of 5 of the 26 parameters:
\begin{itemize}
    \setlength\itemsep{0em}
    \item $K_p: 5.0 \rightarrow 2.0$; $K_d: 10.0 \rightarrow 5.0$; $K_i: 0.0 \rightarrow 5.0$
    \item $\delta_{\mathbf{y}}: 0.01 \rightarrow 0.02$
    \item $\theta: 28 \rightarrow 45$
\end{itemize}
The controller gains were a primary cause, which verifies expected discrepancies between the simulated and real-world dynamics (mainly in the actuators).
Transitions between trajectory positions were less abrupt after slightly increasing the position reaching tolerance, $\delta_{\mathbf{y}}$, contributing to smoother motions, which similarly highlighted a discrepancy in actuator dynamics.
The emulator generated more events for similar motions, visual conditions, and distances to objects, indicating that real-world images contained more colour variations, contrasts and possibly noise.
By increasing emission threshold, $\theta$, we effectively offset this naturally larger variation in RGB data, thus producing more similar event densities and, by extension, SNN activations and avoidance motions to those executed in the simulation.

%\paragraph{}
We ran $N_{trials}=30$ trials in each scenario, which are listed in Table \ref{real_experiments_scenarios} in Appendix \hyperref[appendix:experiment_scenarios]{B}, with and without the module.

\begin{figure}
    \centering
    \includegraphics[width=\columnwidth, trim={12pt 5pt 10pt 10pt},clip]{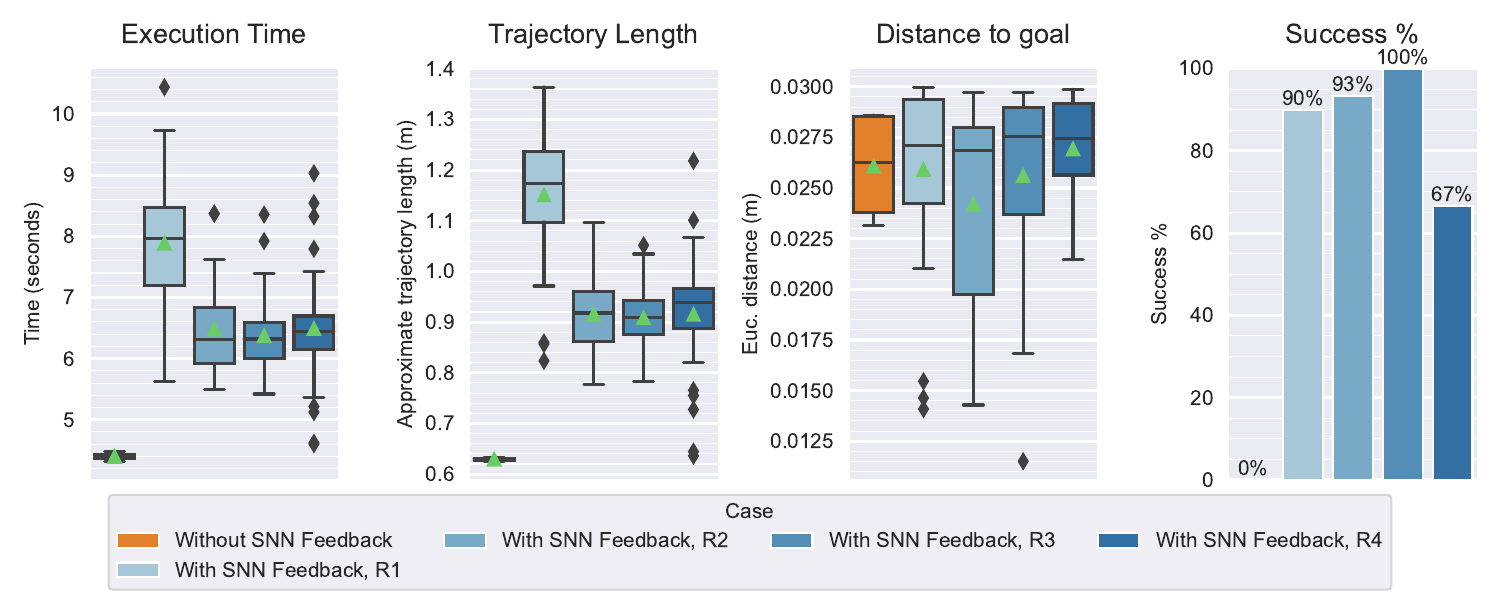}
    \caption{Quantitative metric results without vs. with SNN feedback: real experiment scenarios R1-R4.}
    \label{real_robot_experiments_metrics_scenarios_R1_to_R4}
\end{figure}

%\paragraph{}
Figure \ref{real_robot_experiments_metrics_scenarios_R1_to_R4} contains the quantitative results of these experiments from each scenario (R1-R4) in addition to a single batch of baseline trials\endnote{We aggregate \textit{Without SNN Feedback} executions from all scenarios and present the aggregate as a single case, because the statistics are not expected to differ across scenarios when no obstacle avoidance is used, since we use the same parameter set and the arm always executes the pre-planned motion trajectory.}. Table \ref{real_robot_experiment_scenarios_metrics_table} contains the mean values of each metric and the averages for each task.

%\paragraph{}
The avoidance module succeeded in 87.5\% of all 120 trials.
In the Task 1 scenario, the arm was 90\% successful in avoiding the wooden block.
In Task 2, the arm was most successful with the metal bar obstacle (R3), as it never failed the task, followed by the hand (R2) with a success rate of 93\% and the wooden block (R4) with 67\%. 
Since the goal was always reached, all failures were due to collisions.

\begin{table*}
    \small\sf
    \centering
    \caption{Quantitative results of the real experiment runs, averaged over the $N_{trials}$ trials of each scenario.}
    \begin{tabularx}{\textwidth}[t]{p{0.1\linewidth}>{\centering\arraybackslash}p{0.11\linewidth}>{\centering\arraybackslash}p{0.16\linewidth}>{\centering\arraybackslash}p{0.16\linewidth}>{\centering\arraybackslash}p{0.16\linewidth}>{\centering\arraybackslash}p{0.16\linewidth}}
        \toprule
        & & \multicolumn{4}{c}{Results (averages over $N_{trials}$ trials)}\\
        \cmidrule{3-6}
        & Scenario ID & Success, $S$ (\%) & Dist. to Goal, $d_G$ (m) & Traj. Length, $l_{\mathbf{Y}}$ (m) & Ex. Time, $T$ (s) \\
        \midrule
        Task 1 & R1 & 90.0 & 0.026 & 1.151 & 7.885\\
        \midrule
        Task 2 & R2 & 93.3 & 0.024 & 0.915 & 6.482\\
        & R3 & 100.0 & 0.026 & 0.908 & 6.374\\
        & R4 & 66.7 & 0.027 & 0.915 & 6.492\\
        \midrule
        \multicolumn{2}{r}{\textbf{Task Average:}} & \textbf{87.5} & \textbf{0.026} & \textbf{0.972} & \textbf{6.808}\\
        
        \bottomrule
    \end{tabularx}
    \label{real_robot_experiment_scenarios_metrics_table}
\end{table*}%

\begin{figure}
    \centering
    \begin{subfigure}{0.8\linewidth}
        \centering
        \includegraphics[width=\columnwidth]{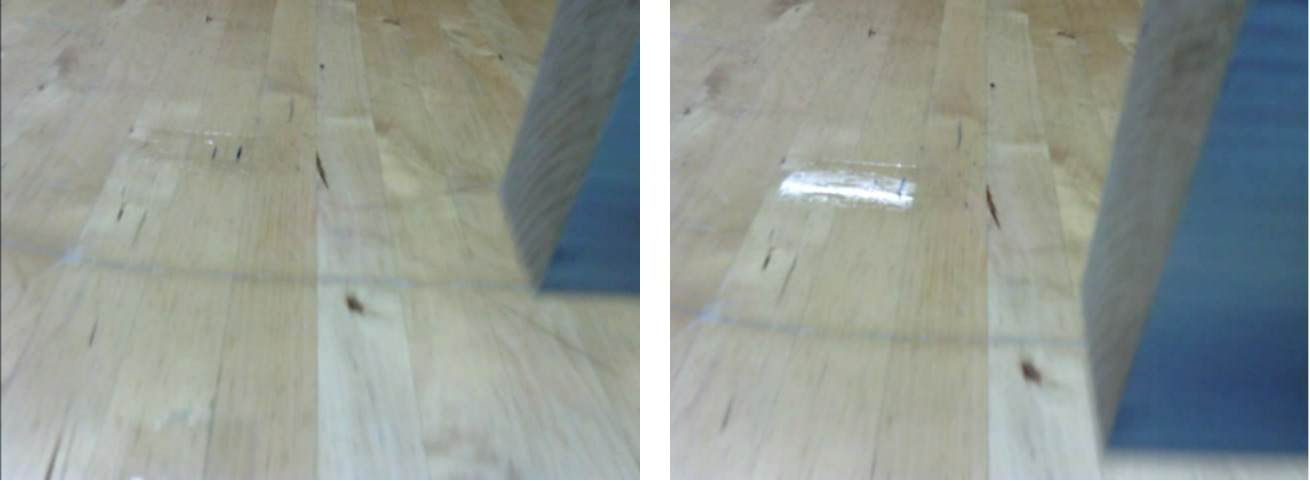}
        \caption{\centering Scenario R4}
        \label{real_robot_task_2_ego_image_obstacle_comparison:wooden_block}
    \end{subfigure}%
    \\
    \begin{subfigure}{0.8\linewidth}
        \centering
        \includegraphics[width=\columnwidth]{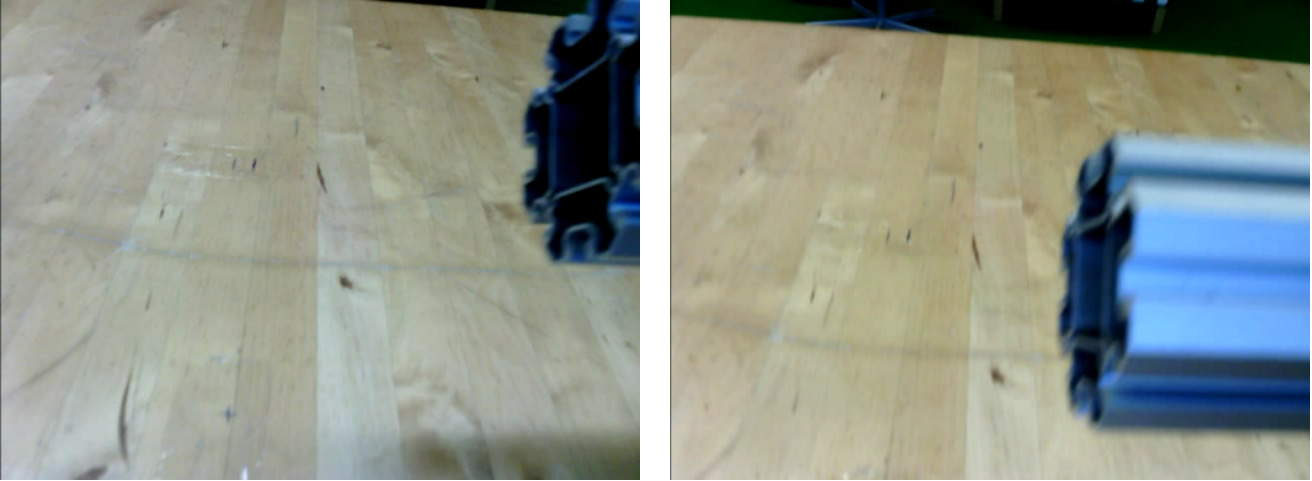}
        \caption{\centering Scenario R3}
        \label{real_robot_task_2_ego_image_obstacle_comparison:metal_bar}
    \end{subfigure}%
    \caption{Images captured by the onboard camera during an execution of scenarios R4 and R3 (Task 2).}
    \label{real_robot_task_2_ego_image_obstacle_comparison}
\end{figure}

%\paragraph{}
The higher failure rate in R4 was due to the visual properties of the wooden block, which blended with the background (see Figure \ref{real_robot_task_2_ego_image_obstacle_comparison:wooden_block}).
As the obstacle comes into view, it generates less events and less neural activation than the more contrasting objects, causing trajectory adaptations to occasionally be too weak or late to effectively avoid the obstacle.
The perfect success in R3 could be explained by the significantly higher color  contrast (see Figure \ref{real_robot_task_2_ego_image_obstacle_comparison:metal_bar}).
Therefore, similar colors induce lower intensity differences that lead to a less effective response for less visible objects.

\begin{figure*}
    \centering
    \begin{subfigure}{0.5\linewidth}
        \centering
        \includegraphics[width=\columnwidth]{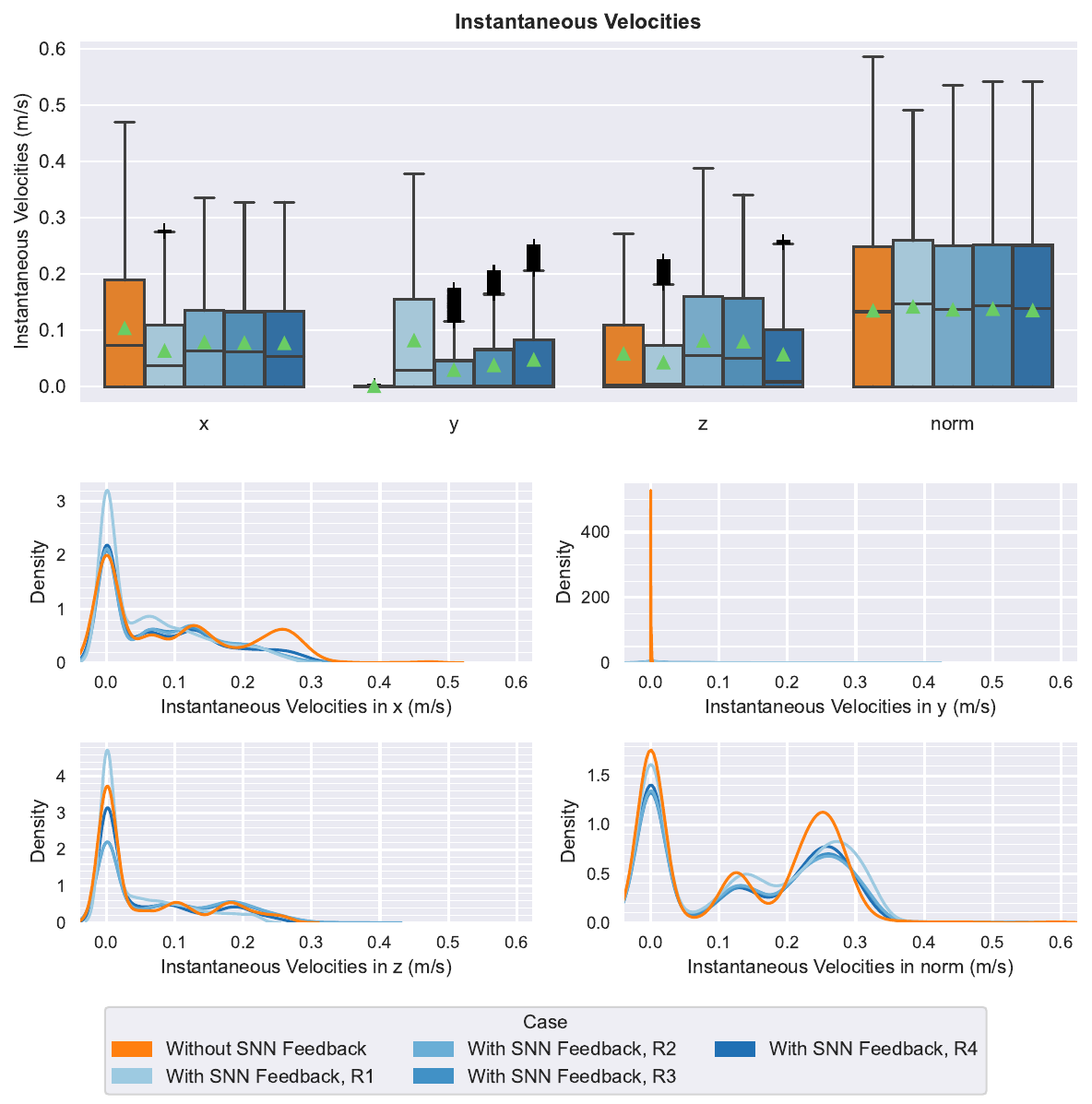}
        \caption{\centering Velocities}
        \label{real_robot_experiments_vel_and_acc_plots:velocities}
    \end{subfigure}%
    \begin{subfigure}{0.5\linewidth}
        \centering
        \includegraphics[width=\columnwidth]{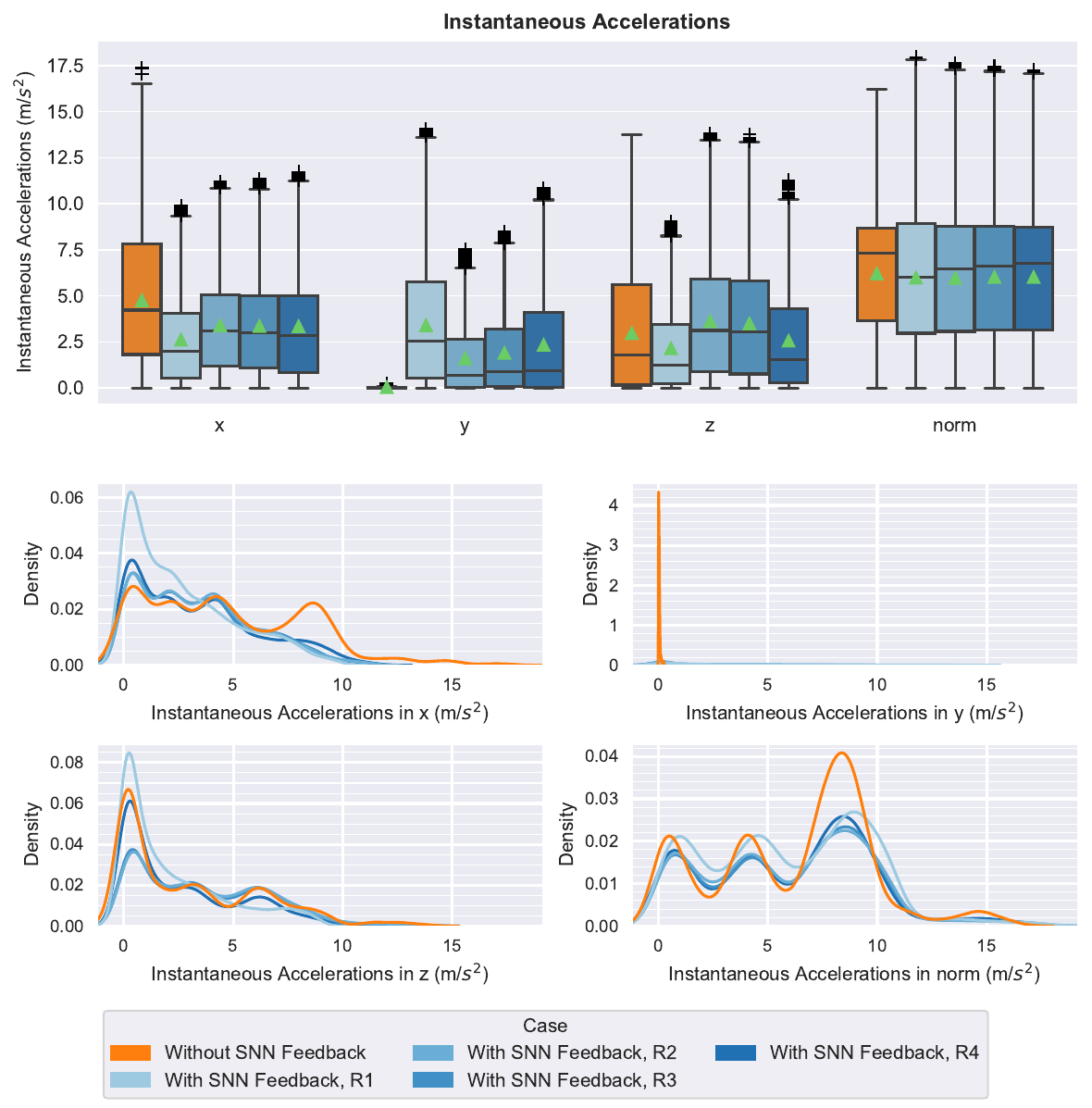}
        \caption{\centering Accelerations}
        \label{real_robot_experiments_vel_and_acc_plots:accelerations}
    \end{subfigure}%
    \caption{Distributions of instantaneous velocities and accelerations in each spatial dimension, measured during real robot experiments. The figure contains data from nominal executions (without SNN Feedback) and scenarios R1-R4.}
    \label{real_robot_experiments_vel_and_acc_plots}
\end{figure*}

%\paragraph{}
When comparing the baseline and adaptive cases, we observed similar distributions of accelerations and velocities, which are shown in Figure \ref{real_robot_experiments_vel_and_acc_plots}.
However, magnitudes in $y$ were marginally higher in the second case.
This indicates a preference for side-ways motions i.e. in the $y$ direction (left-right axis, relative to the camera), which is expected due to the avoidance velocity vectors being computed from the camera’s image space, ruling out motions in $x$ (front-back axis), as expressed in section \ref{dynamic_motion_primitives}.
Accelerations and velocities in $x$ tended to be lower with the avoidance module, indicating that the trajectory adaptations naturally slow down forward motion when avoiding perceived obstacles, which is a desirable effect when aiming to safely clear an obstacle while continuing progress towards a goal.

%\paragraph{}
The trajectories executed in these experiments (plotted in Figure \ref{real_robot_experiments_trial_trajs} in Appendix  \hyperref[appendix:metrics_and_trajectory_plots]{H}) were qualitatively similar to the simulation counterparts and lead to similar conclusions from the qualitative evaluation.

%\paragraph{}
The average success rate was 87.5\% (the median was 91.7\%), indicating a moderately high level of reliability.

%\paragraph{}
The distribution of $\dot{\zeta}$ spanned low values, similar to the baseline, which indicate that the end-effector did not exhibit sudden, large changes in direction.
(The values for R2 are plotted in Figure \ref{angular_vel_dist_plots_real_experiments} in Appendix \hyperref[appendix:qualitative_evaluation_visualizations]{G}.)
Therefore, the module generated \textit{predictable} motions on the real robot as well.

%\paragraph{}
The same \textit{safety} comparison to the reference velocity threshold ($0.25m/s$) indicated that trajectories were fairly safe, albeit to a less degree than in the simulation.
While the overall end-effector velocity averaged at $0.13m/s$, the Task 1 and Task 2 maximums approached $0.5m/s$ and $0.55m/s$.
However, these values lay beyond the third quartile of the distribution (upper plot in Figure \ref{real_robot_experiments_vel_and_acc_plots:velocities}).
We can assume that this is not an effect of the avoidance maneuvers themselves, but rather the controller parameterization (since similar values are observed in the \textit{Without SNN Feedback} case).

%\paragraph{}
Overall, the SNN-based obstacle avoidance module demonstrated 84\% and 92\% median success rates in simulated and real experiments.
Execution times, trajectory lengths, and velocity and acceleration magnitudes indicated that the adaptive trajectories were quantitatively similar to baseline executions, but with the added capability of consistent obstacle avoidance.
The qualitative assessment suggested that these trajectories were adequately reliable, predictable, and safe, although they were directly optimized only for obstacle avoidance.
Finally, we demonstrated comparable performance in the real domain.

%\paragraph{}
\section{Further Analyses}
\label{further_analyses}

We additionally explored properties of our neuromorphic components through further experimentation.
We first compared different event emulation methods in terms of output events, spiking responses, and effects on performance.
To validate the SNN's utility, we excluded the component from the pipeline and tested the decoding of avoidance behaviour from raw events.
Furthermore, we investigated the effects of random SNN weight values on performance by repeating scenario trials with different initializations.
Finally, we substituted our event emulator with a real EC and repeated experiments on the robot.

\subsection{Comparing Event Emulation Methods}
\label{further_analyses:comparison_of_event_emulation_strategies}

%\paragraph{}
For this analysis, we compared five emulation strategies:
\begin{itemize}
    \item \textbf{M1:} Multi-C RGB absolute differences
    \item \textbf{M2:} Multi-C RGB absolute differences, with blur
    \item \textbf{M3:} Multi-C RGB log differences
    \item \textbf{M4:} “Salvatore” method
    \item \textbf{M5:} “pydvs” method
\end{itemize}

%\paragraph{}
Our default strategy, M1, involves computing the differences in absolute intensities at every pixel, $\Delta \mathbf{L}_{abs}(\textbf{x}_k, t_k)$, and emitting an event wherever this exceeds $\theta$ in \textbf{all} three color channels (which we term a \textit{multi-channel} condition).

%\paragraph{}
For the second method (M2), we blur the source images before computing $\Delta \mathbf{L}_{abs}(\textbf{x}_k, t_k)$. This is similarly applied in \cite{zahra2021differential} for removing high-frequency noise; here, we aim to investigate its effects on derived events.
This is accomplished by convolving images with a low-pass filter represented by a normalized $5 \times 5$ box kernel.

%\paragraph{}
M3 emits events according to differences in \textit{log} intensities:
\begin{equation}
    \Delta \mathbf{L}_{log}(\textbf{x}_k, t_k) = log(\mathbf{L}(\textbf{x}_k, t_k)) - log(\mathbf{L}(\textbf{x}_k, t_{k-1}))
    \label{ece_method_3_equation:diff}
\end{equation}
This mimics how most ECs measure irradiance changes, resulting in their characteristically high dynamic ranges (\cite{rebecq2019high}, \cite{gallego2022event}), and is often used in event emulation (\cite{rebecq2018esim}).

%\paragraph{}
The fourth method (M4) is based on a strategy employed in \cite{salvatore2020neuro}, where the log of a sum of weighted channel intensities is used to compute the difference:
\begin{align}
    \Delta \mathbf{L}(\textbf{x}_k, t_k) = & \hspace{0.5em} log(0.299\mathbf{L}_R(\textbf{x}_k, t_k) + 0.587\mathbf{L}_G(\textbf{x}_k, t_k) + \nonumber \\
    & 0.114\mathbf{L}_B(\textbf{x}_k, t_k)) - log(0.299\mathbf{L}_R(\textbf{x}_k, t_{k-1}) + \nonumber \\
    &0.587\mathbf{L}_G(\textbf{x}_k, t_{k-1}) + 0.114\mathbf{L}_B(\textbf{x}_k, t_{k-1}))
    \label{ece_method_4_equation:diff}
\end{align}

%\paragraph{}
Finally, M5 mimics the \textit{pyDVS} emulator (\cite{garcia2016pydvs}), which encodes pixel intensities using Gamma functions, instead of absolute or log values.

%\paragraph{}
While all methods could be run interchangeably, it was necessary to adjust some $\theta$ thresholds.

%\paragraph{}
Firstly, we visually compared events and SNN responses. 
Three pairs of consecutive images that were captured during a scenario R3 trial were used to generate event images using each method. These were then input to the SNN, from which we recorded the spike trains output in a period of $T_{sim}=40ms$. 
Appendix \hyperref[appendix:ece_comparison_visualizations]{F} contains the event images (Figure \ref{ece_strategy_comparison_events_images}) and plots that depict the resulting spike trains and output neuron spike counts across $T_{sim}$ (Figure \ref{ece_strategy_comparison_spikes_plots}).

%\paragraph{}
We observed subtle variations in the emulated event data.
The blurring applied in M2 leads to similar images but additionally eliminates seemingly spurious events.
The event distributions appear notably different in M3; for instance, darker regions produce significantly more events than brighter regions.
This is likely due to the higher sensitivity of the log-based measure to darker
(lower intensity) colors.
The other log-based method, M4, showed less dark/bright differences and its event data was denser in the most relevant regions (containing obstacle motion).
Results from M5 measure look fairly similar to M2's, except for a slight reduction in apparent noise. 
Similar to the first two methods, M5 reacts strongly to reflective surfaces, such as one near the center of the frame, which produce high and unstable intensities.

%\paragraph{}
The SNN responses largely matched the input events.
For example, M1-M4 caused significant spike counts for the first sample in the first few neurons (by index); these neurons correspond to upper regions of the image, which contained strong event activity.
Although the blurring in M2 produced less noisy events than M1, both resulted in very similar spike distributions. 
A likely reason is that the SNN’s dynamics filter out spurious inputs anyway, which indicates that the SNN could obviate the need for such a denoising operation.
Generally, the SNN’s responses were similar across methods (except for M5), showing a degree of robustness to variations in events (see in Figure \ref{ece_strategy_comparison_spikes_plots}, for example, the spike trains concentrated in the top and bottom regions of the plots for sample 3 across methods).
The ultimate obstacle avoidance response would also be similar due to our FST decoding strategy: we consider neurons that fire before $t_{act}$ (depicted as a yellow line in the plots) active and indicative of obstacle points, and neurons that spike first are generally around similar positions across methods.

\begin{figure}
    \centering
    \includegraphics[width=\columnwidth, trim={12pt 5pt 5pt 10pt},clip]{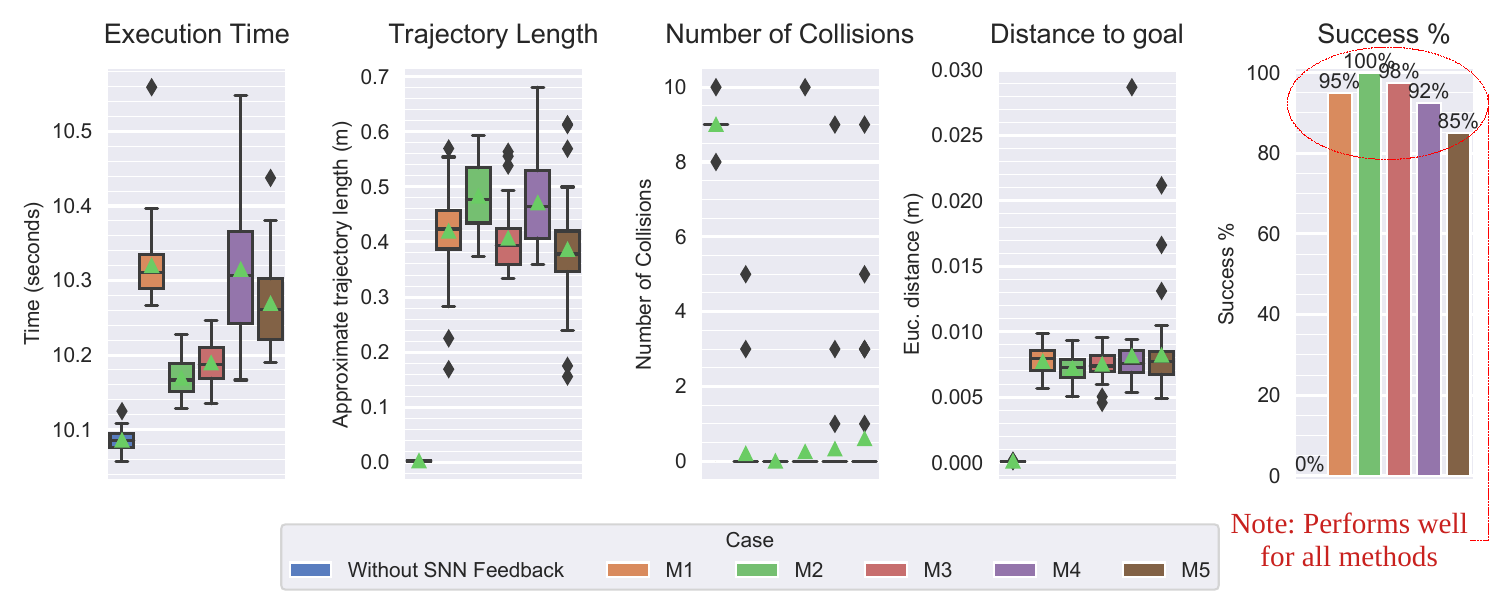}
    \caption{Quantitative metric results for testing scenario 30: without SNN feedback and four batches, each with a different event emulation method: M1 (default) to M5.}
    \label{testing_scenario_30_ece_strategy_comparison_metrics_plots}
\end{figure}

%\paragraph{}
We repeated $N_{trials}=40$ trials in scenario 30 for each method to compare the resultant obstacle avoidance performance through our quantitative metrics, which are plotted in Figure \ref{testing_scenario_30_ece_strategy_comparison_metrics_plots}. 
The results were not significantly affected by the emulation method; M5's success rate of $85\%$ deviated most notably from the mean ($94\%$).
This verifies that the resultant behaviours are fairly similar, as we would expect from the similarity in SNN responses.

%\paragraph{}
The spiking responses and task performance we observed indicate that our SNN is robust to differences in the emulated event data, since they are inherently tolerant to some noise or variance.
A different SNN output decoding strategy that, for example, considers more than neurons' first spike times may result in more nuanced and varied responses and behaviors between the EC emulation methods and is thus worth exploring in future work.

\subsection{Decoding Avoidance Behaviour From Raw Event Data}
\label{further_analyses:decoding_from_raw_events}

%\paragraph{}
We experimented with decoding trajectory adaptations directly from raw events, instead of spikes output by the SNN in response to events.
Additionally, we tested decoding random events, in order to verify that the presented results are indeed due to the information contained in the event data and the subsequent SNN processing.

%\paragraph{}
In our approach, the output first-spike-times define the neural activation map which contains the \textit{obstacle points} in the output feature space.
To decode raw events directly, we resized the event images to match the SNN’s output feature space (using bilinear interpolation) and designated events as the obstacle points.
The PF-based decoding procedure that follows is otherwise identical, resulting in the $\vect{\phi}$ accelerations that in turn dictate trajectory adaptations.

%\paragraph{}
To quantitatively evaluate effects on task performance, we repeated $N_{trials}=40$ trials in scenario 25, while i) decoding from raw events and ii) decoding from random events.

\begin{figure}
    \centering
    \includegraphics[width=\columnwidth, trim={12pt 5pt 5pt 10pt},clip]{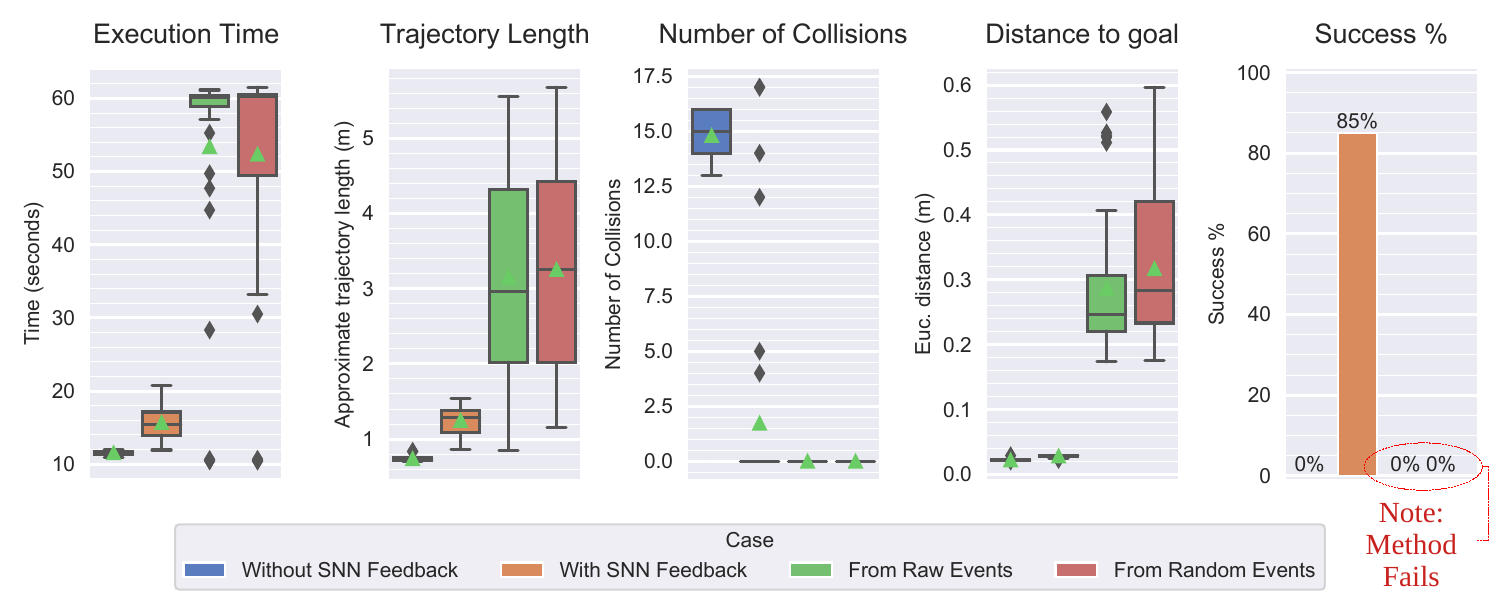}
    \caption{Quantitative metric results for testing scenario 25: without SNN feedback, with SNN Feedback, and decoding obstacle avoidance behaviour from raw or random events.}
    \label{testing_scenario_25_decoding_from_events_comparison_metrics_plots}
\end{figure}

%\paragraph{}
In both cases, the robot failed in all trials and would often oscillate due to the accelerations induced by the raw events and never reach the goal, despite subsequent efforts to re-tune parameters.
We illustrate the metric results in Figure \ref{testing_scenario_25_decoding_from_events_comparison_metrics_plots}.

%\paragraph{}
The inadequacy of raw event data indicates that the neural dynamics of the SNN are integral for processing that data to achieve successful trajectory adaptations.
The similarly negative results obtained from either random events or the raw events further confirm the insufficiency of the event data in our approach and the importance of the SNN.

\subsection{Testing Effects of SNN Weight Initializations}
\label{further_analyses:random_weight_initializations}

%\paragraph{}
In all experiments, we constrained the SNN weights to a set of random values to eliminate the variance in results.
To investigate their influences, we analyzed performance with different random weight initializations.

%\paragraph{}
We ran eight batches of $N_{trials}=60$ trials in scenario 31, initializing the SNN with a different set of weights in each.

\begin{figure}
    \centering
    \includegraphics[width=\columnwidth, trim={12pt 5pt 5pt 10pt},clip]{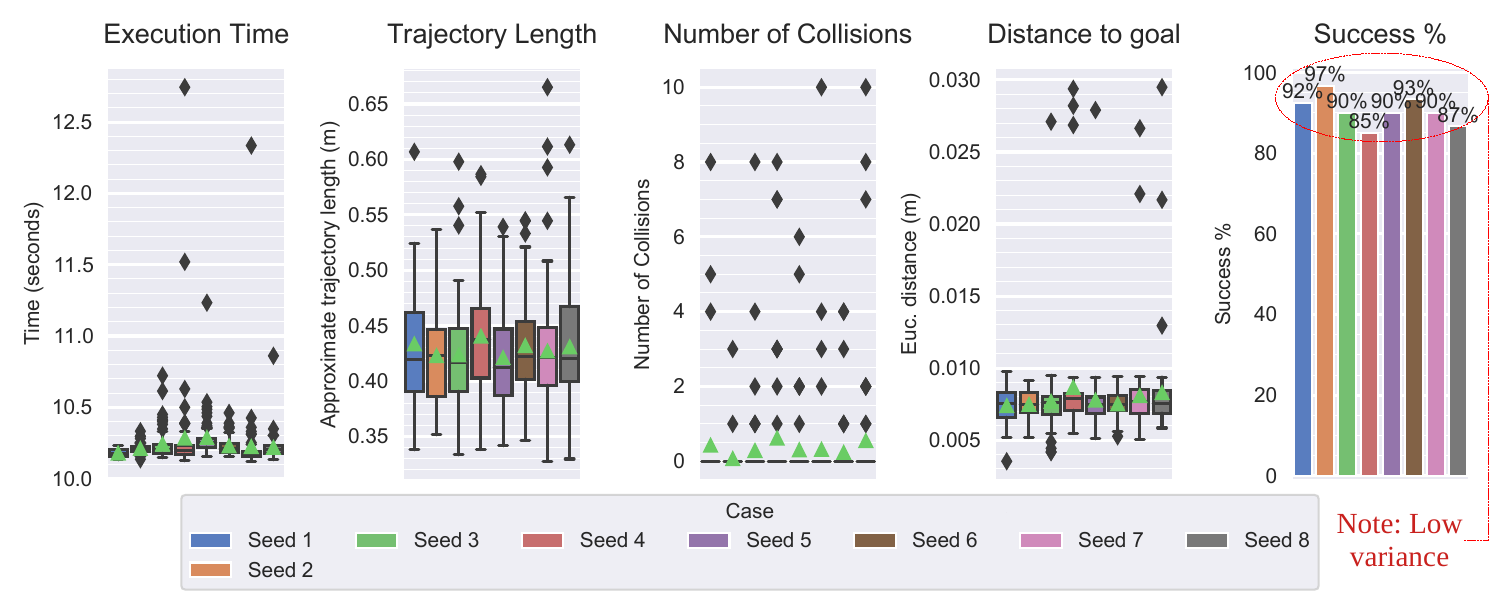}
    \caption{Quantitative metric results for testing scenario 31, repeated with eight random SNN weight initializations.}
    \label{testing_scenario_31_random_weight_init_comparison_metrics_plots}
\end{figure}

%\paragraph{}
Figure \ref{testing_scenario_31_random_weight_init_comparison_metrics_plots} shows the metrics plot (note that “Seed 1” refers to the batch from the original experiments).
The success rate varied around $90.5\%$ with a standard deviation of $3.5\%$, indicating that the weight values have a non-negligible, but not excessive, effect on performance.
Except for occasional outliers, the other metrics' distributions were similar across  weight initializations.
No differences between the resultant trajectories were perceptible from visual observations. 

%\paragraph{}
The limited variation in the results was not significant enough to invalidate previously established conclusions and, in fact, has positive implications for applications of learning.
In particular, the observed relevance of the weights motivates searching for values that could improve performance, e.g. through optimization and learning algorithms to tune SNN weights; an interesting avenue for future work.

\subsection{Testing Real Event Camera Data}
\label{further_analyses:real_ec_tests}

%\paragraph{}
We had utilized event emulation within our pipeline with the hypothesis that our conclusions can be extended to a system equipped with a real EC.
As an initial validation step, we integrated and conducted preliminary tests with an EC.

%\paragraph{}
We used a DAVIS346 EC, which contains a dynamic vision sensor (DVS) and active pixel sensor (APS) that enable capturing event and RGB data, respectively.
The DAVIS346 has a relatively large pixel size of $18.5\mu m^2$ and a resolution of 346x260.
By comparison, the Omnivision OV5640 sensor mounted on the Kinova Gen3 has a pixel size of $1.4\mu m^2$ and a resolution that ranges from 320x240 to 2592x1944.
However, the DAVIS has a significantly higher dynamic range ($120dB$ vs. $68dB$).
In addition, the OV5640 consumes $\sim$700$mW$ while the DAVIS consumes only 10-30$mW$ to transmit event data and an additional 140$mW$ if the APS is active, for a 5V DC supply\endnote{The OV5640’s power consumption was derived from reported consumption in $mA$ and usual voltage supply rating, since a figure in $mW$ was not provided in the specifications.}.
The DAVIS was attached to the top of the end-effector for these tests. 

\begin{figure}
    \centering
    \includegraphics[width=\columnwidth, trim={0pt 60pt 0pt 60pt},clip]{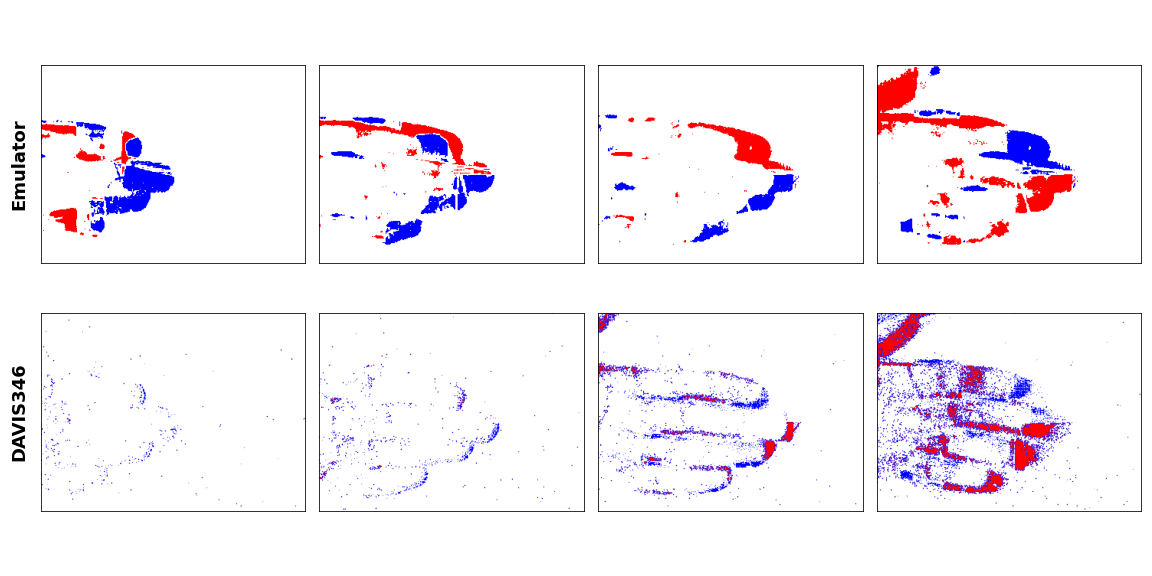}
    \caption{A comparison of events generated by our emulator and the DAVIS346. Note: RGB images captured by the DAVIS346 were used to generate the emulated event images.}
    \label{emulator_vs_davis_hand_motion_comparison}
\end{figure}

%\paragraph{}
Figure \ref{emulator_vs_davis_hand_motion_comparison} depicts the events generated for four frames of a simple hand motion by the emulator and the DAVIS.
Evidently, the events are spatio-temporally similar, but the DAVIS is more sensitive to minute motions and produces significantly more salient events.
On the other hand, its output is notably noisier, but may be alleviated with careful tuning of the camera’s parameters.

%\paragraph{}
For evaluating the DAVIS’s performance, we repeated executions of scenario R3 trials.
The camera parameters, particularly the emission thresholds, were moderately adjusted to induce SNN responses of similar magnitudes to those induced by the emulator.
However, the DAVIS produced denser and noisier event data on average, which necessitated diminishing sensitivity to the event data elsewhere in the pipeline.
To that end, we modified the values of i) the limit on instantaneous avoidance accelerations, $\phi_{max}$ and ii) the binary erosion filter size, $s_{BE}$. 
$N_{trials}=30$ trials were executed and compared to the previous trials.

\begin{figure}
    \centering
    \includegraphics[width=\columnwidth, trim={12pt 5pt 5pt 10pt},clip]{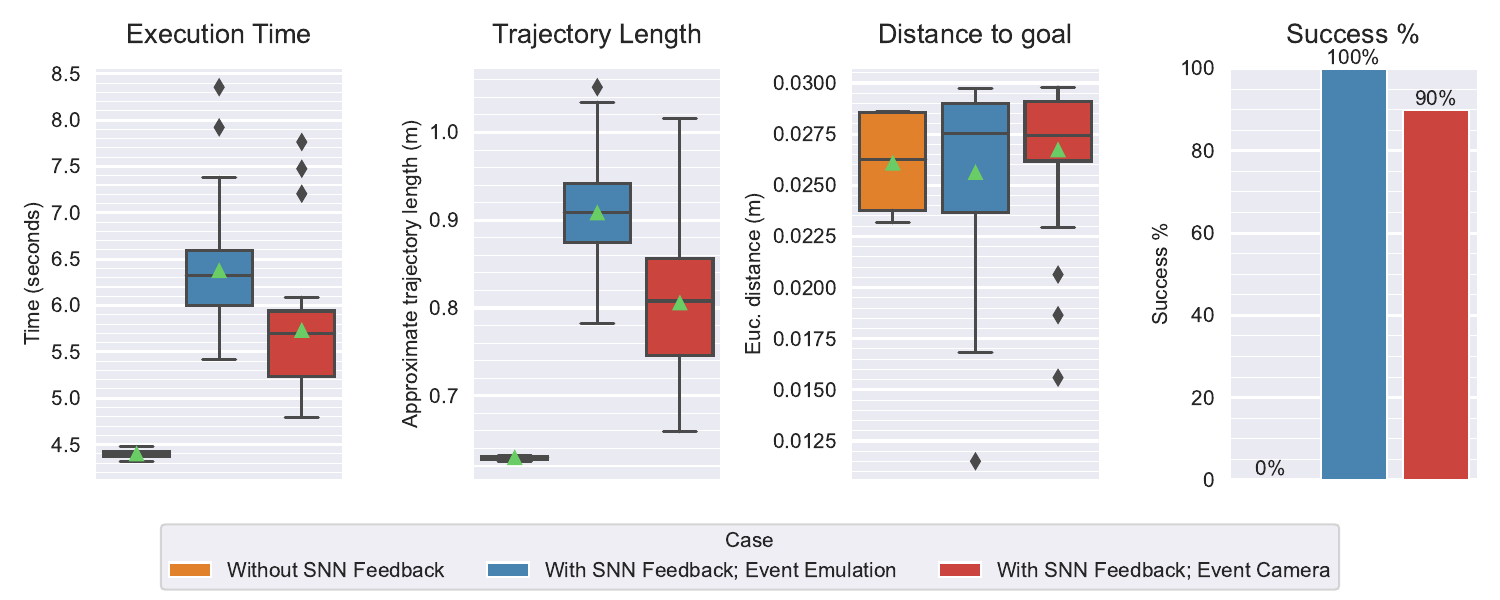}
    \caption{Quantitative metric results for scenario R3 with results when using the event emulation (obtained during the experiments discussed in section \ref{results_and_discussion:real_experiments}) and the real EC.}
    \label{emulator_vs_davis_scenario_r3_metrics_plot}
\end{figure}

%\paragraph{}
Figure \ref{emulator_vs_davis_scenario_r3_metrics_plot} illustrates the quantitative results with event emulation and with the DAVIS346.
Most significantly, we observed a similarly high success rate with the EC, despite three observed failures, two of which were light touches of the obstacles.
Ultimately, the minimal tuning of the DAVIS’s parameters lead to similar results and trajectory shapes.

%\paragraph{}
These tests show the seamless substitution of the emulator with an EC.
The similarity in task performance indicates that our experiment conclusions could be extended to a system incorporating a real EC, and provides some validation for using emulation as a substitute in research.
Nevertheless, testing in more scenarios may yield more insights, especially since the relatively large number of camera parameters have not been fully explored.
The small decrease in task success may be due to a sub-optimal tuning of the DAVIS parameters, in contrast to the emulator’s parameters, which underwent a more rigorous refinement process during our tuning phase.

\section{Conclusion}
\label{conclusion}

%\paragraph{}
We presented and evaluated a neuromorphic approach to obstacle avoidance on a manipulator with a single onboard camera.
Our pipeline transforms visual inputs into corrective obstacle avoidance maneuvers, combining high-level trajectory planning and low-level reactive adjustments using DMPs.
We utilized event-based vision and spiking neural networks for neuromorphic sensing and processing capabilities.
Simulated and real experiments have demonstrated its success in achieving real-time, online obstacle avoidance across various task scenarios, proving its utility over a non-adaptive baseline.
The adaptive trajectories minimally deviated from the baseline and were often predictable, safe, and reasonably smooth.

%\paragraph{}
Further analyses highlighted useful properties of the SNN, such as robustness to variations/noise in event data. 
The proportionality of computations to input magnitudes, as evidenced by the time analysis, validates two suppositions: the SNN's tendency to exclusively process relevant information, which can save energy and time, and the homogeneity of EC data and SNNs.
Since the relevant information for the problem we address is relative input change, events efficiently distill this information and the SNN selectively performs computations only on those portions of the visual inputs.
Additionally, the SNN weight analysis motivated future applications of learning-based optimization.
Finally, we verified the compatibility of our implementation with a real EC.

\subsection{Limitations}
\label{conclusion:limitations}

%\paragraph{}
We noted a limited capability of avoiding high-speed obstacle collisions, which could be addressed by increasing sensitivity but at the cost of compromising safety.
Therefore, we presently acknowledge a fundamental limitation in reliably reacting to fast obstacles due to physical constraints.
Additionally, two factors that seemed to occasionally cause failures were dimmer lighting, which lead to less contrasts between foreground and background, and clutter, which could contribute to more background events that saturate the SNN response, possibly overwhelming localized responses at obstacle positions.

%\paragraph{}
We presented a set of well-formulated qualitative evaluation criteria, for which we selected specific quantitative measures to evaluate each.
However, since concepts like predictability can be interpreted in different ways, a better approach could aggregate multiple measures, including from user data, to evaluate each criterion.

%\paragraph{}
The method discussed in this paper does not provide a mechanism for distinguishing obstacles from objects that the robot must interact with, since it is intended to solely enable fast, reactive obstacle avoidance (which could be considered a low-level behaviour). 
The ability to recognize goal-relevant objects could be delegated to a higher-level planning component that adapts the behaviour of the avoidance module according to task-specific knowledge, such as an object to be grasped. In this case, a simple adaptation could be to dampen or cancel the effects of the avoidance module when the input data corresponds to the location of the task object.
Therefore, we currently assume that this method is augmented by such planning components.

%\paragraph{}
As mentioned in section \ref{dynamic_motion_primitives}, our obstacle avoidance is currently restricted to motions in two dimensions that align with the camera's image plane. While this is sufficient for the targeted type of tasks, other situations may require more general motions. Although it is outside the scope of this work, it is worth noting that this could be addressed through a novel method for recovering depth information from distributions and magnitudes of local events in event images or using stereo vision (such as in \cite{zhou2021event} and \cite{risi2020spike}), which is often applied in conventional camera systems to achieve some depth perception and is biologically plausible in principle.

%\paragraph{}
While this work focuses on the design, implementation, and systematic evaluation of the novel neuromorphic pipeline itself, with obstacle avoidance being a representative application, we do not experimentally compare to conventional obstacle avoidance methods, which could be useful in validating our method. 
Such a comparison to existing, non-neuromorphic methods can be challenging due to the drastically different sensor configurations (refer to section \ref{related_work:obstacle_avoidance} for discussions of examples) and approaches to collision avoidance that they apply.
In particular, most reviewed methods employ multiple sensors on and around the robot, most commonly RGB-D cameras, compared to our single, biologically-inspired, onboard event camera, which presents a novel paradigm.

\subsection{Future Work}
\label{conclusion:future_work}

%\paragraph{}
An interesting extension is to incorporate learning for adjusting the SNN weights and optimizing the hyperparameters of the pipeline.
Given the results of our random weights analysis, we expect synaptic weight tuning to have a positive impact on avoidance behaviour.
The best method to train an SNN remains an open question, but we could consider STDP or surrogate gradient methods.
Other approaches include liquid-state machines (\cite{ponghiran2019reinforcement}) and applying a simple rule such as linear regression on a representation of output spiking activity (\cite{michaelis2020robust}).

%\paragraph{}
Although our manual hyperparameter tuning procedure leads to interpretable results, it could limit applicability to different tasks and situations. This could be substituted by a learning or optimization algorithm, such as RL or evolutionary optimization, with a formalized objective function, thereby avoiding the tedium of manual tuning and potentially yielding better parameter values.
In addition, these optimization algorithms could be deployed online to adapt and adjust parameters to different tasks and situations.

%\paragraph{}
Given preliminary results with the DAVIS346, a next step is to further explore and utilize the capabilities of the event camera.

%\paragraph{}
We presented additional analyses of particular components in section \ref{further_analyses}, which could be extended further with ablation studies. These studies could involve conducting experiments that enable studying and evaluating the neuromorphic components more intensively and potentially drawing more interesting connections to their biological counterparts.

%\paragraph{}
Our qualitative evaluation can be improved by conducting user studies in which executions are evaluated by na\"ive subjects through questionnaires designed to elucidate the general perception of the robot's behaviour.

%\paragraph{}
We noted some uncertainty in trajectory outcomes in one of the simulation tasks (3).
The slight variations in results are due to the complex interactions and data flow within the pipeline stages.
In order to better understand these interactions, we could formulate experiments designed to quantify the effects of input variations, the results of which may explain the observed differences in trajectories.

%\paragraph{}
We have only considered collisions of the end-effector throughout this work, but we could augment our trajectory adaptation approach to incorporate avoidance of collisions with the rest of the robot's body.
For example, one could incorporate information on the robot morphology in the replanning of motion trajectories, which could be achieved more easily in the case of DMPs by representing trajectories in joint space (not end-effector space) or using other motion planners and feeding back obstacle information to ultimately ensure that all parts of the robot body maintain distance from perceived obstacles. In addition, higher-level planning behaviours can be implemented to maintain a memory of perceived objects around the workspace and to execute motions whose purpose is to perceive and "remember" object locations even when they are no longer within the field of view (similar to mapping the environment), particularly since EC data relies on relative motion.

%\paragraph{}    
The full extent to which the SNN provides processing speed, power consumption, and other improvements can be studied only when the networks are run on neuromorphic hardware.
Naturally, a potential extension is thus to explore the integration of a neuromorphic processor as a step towards a more fully neuromorphic processing approach.

%\paragraph{}
Finally, we have demonstrated our approach in the domain of robot manipulation.
In future work, we can apply our implementation to a navigation scenario, for example.
The results of experiments in a different domain can provide further validation of the neuromorphic concept we have developed and evaluated in this paper.

\begin{acks}
    We gratefully acknowledge the support by the b-it International Center for Information Technology.
\end{acks}

\begin{dci}
    The authors declare that there is no conflict of interest.
\end{dci}

\begin{funding}
    The authors received no financial support for the research, authorship, and/or publication of this article.
\end{funding}

\theendnotes

\bibliographystyle{SageH}
\bibliography{bibliography.bib}

\clearpage

%% Note: recommended type-setting method (see Latex FAQs: https://us.sagepub.com/sites/default/files/latex_frequently_asked_questions.pdf)
\appendix
\section*{Appendix A: Experimental Scenario Variables}
\label{appendix:scenario_variables}

%\paragraph{}
This appendix contains visualizations of the tasks and variables (backgrounds, obstacle types, and obstacle colors) that define the scenarios of the simulation experiments.

\subsection*{Tasks}

\begin{figure}
    \centering
    \begin{subfigure}{0.47\linewidth}
        \centering
        \includegraphics[width=\columnwidth]{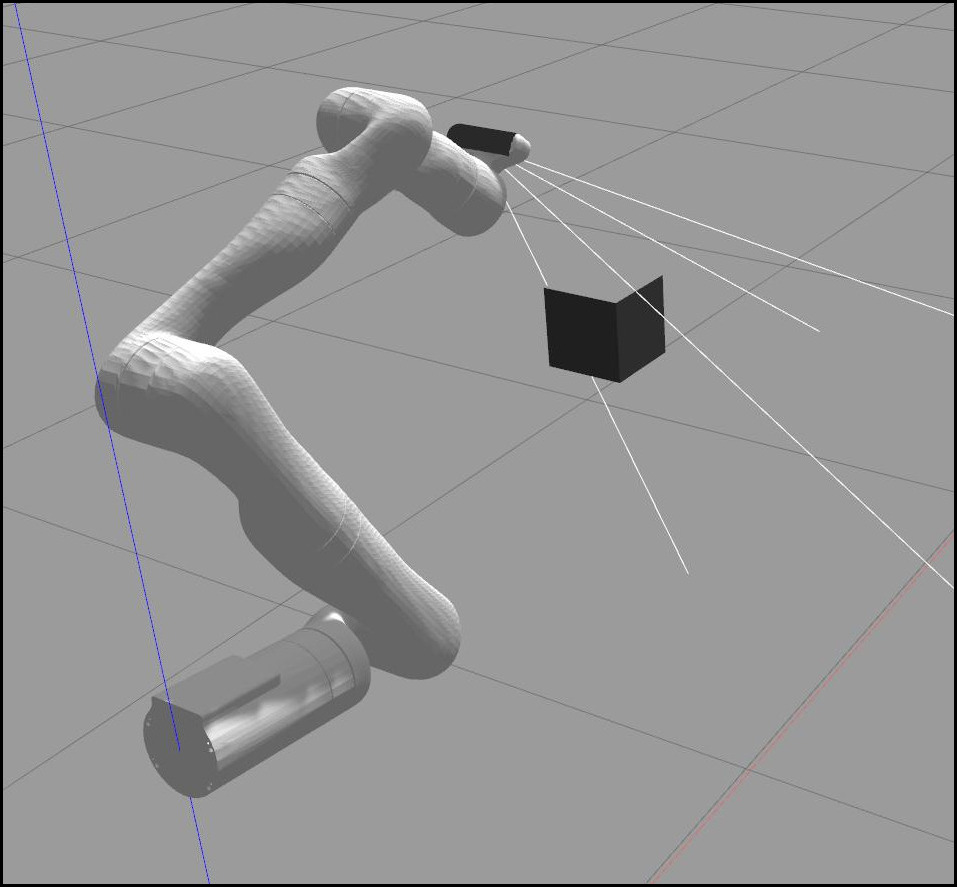}
        \caption{\centering Start}
        \label{}
    \end{subfigure}%
    \hfill
    \begin{subfigure}{0.47\linewidth}
        \centering
        \includegraphics[width=\columnwidth]{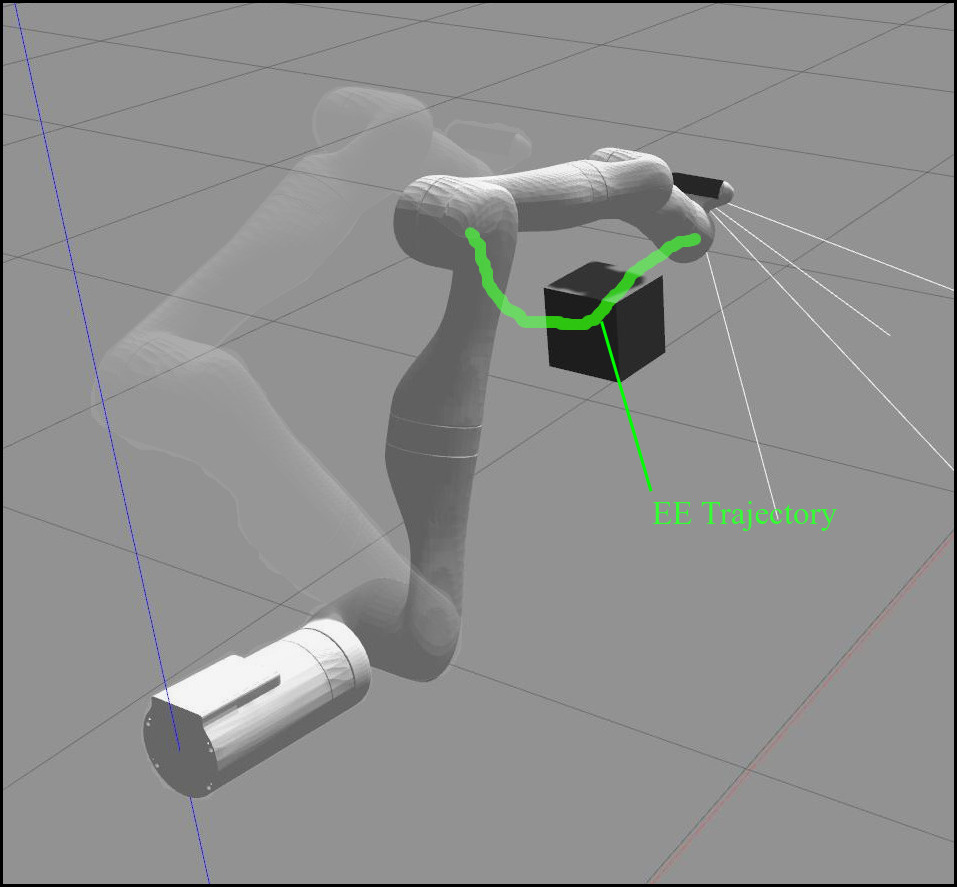}
        \caption{\centering End}
        \label{}
    \end{subfigure}%
    \caption{Simulation Task 1 setup. A baseline trajectory is illustrated in green.}
    \label{task_1_demonstrative}
\end{figure}

\begin{figure}
    \centering
    \begin{subfigure}{0.47\linewidth}
        \centering
        \includegraphics[width=\columnwidth]{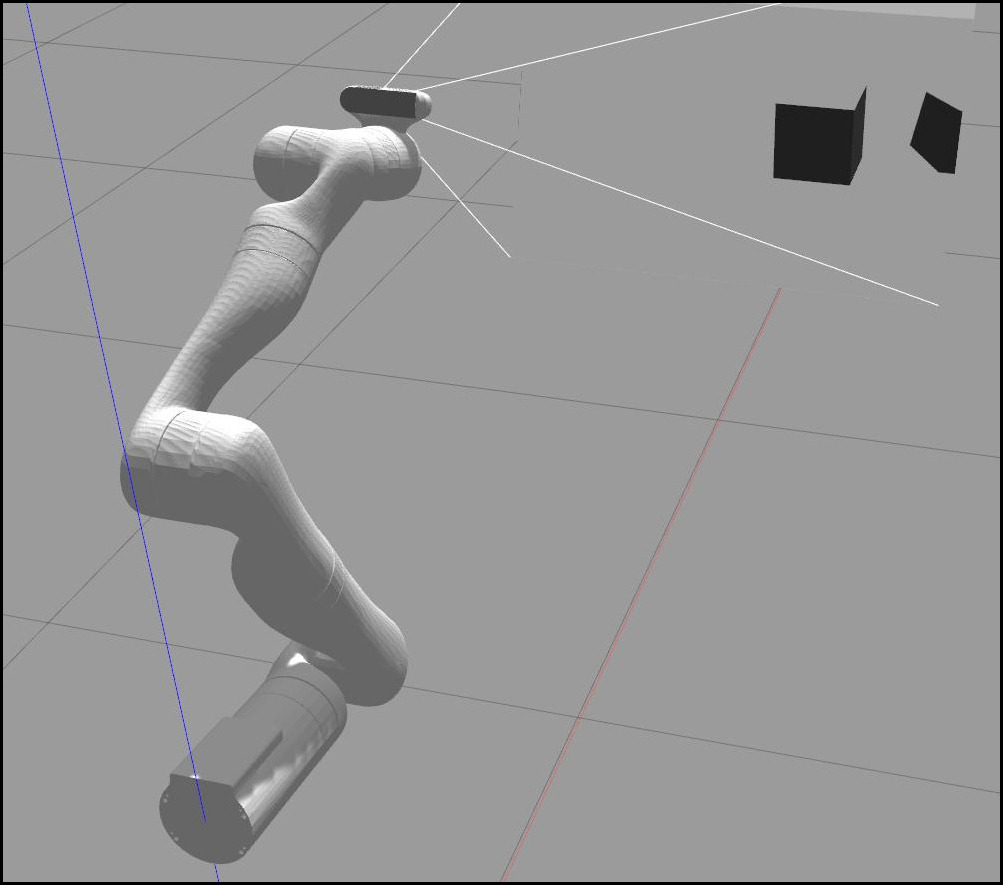}
        \caption{\centering Start}
        \label{}
    \end{subfigure}%
    \hfill
    \begin{subfigure}{0.47\linewidth}
        \centering
        \includegraphics[width=\columnwidth]{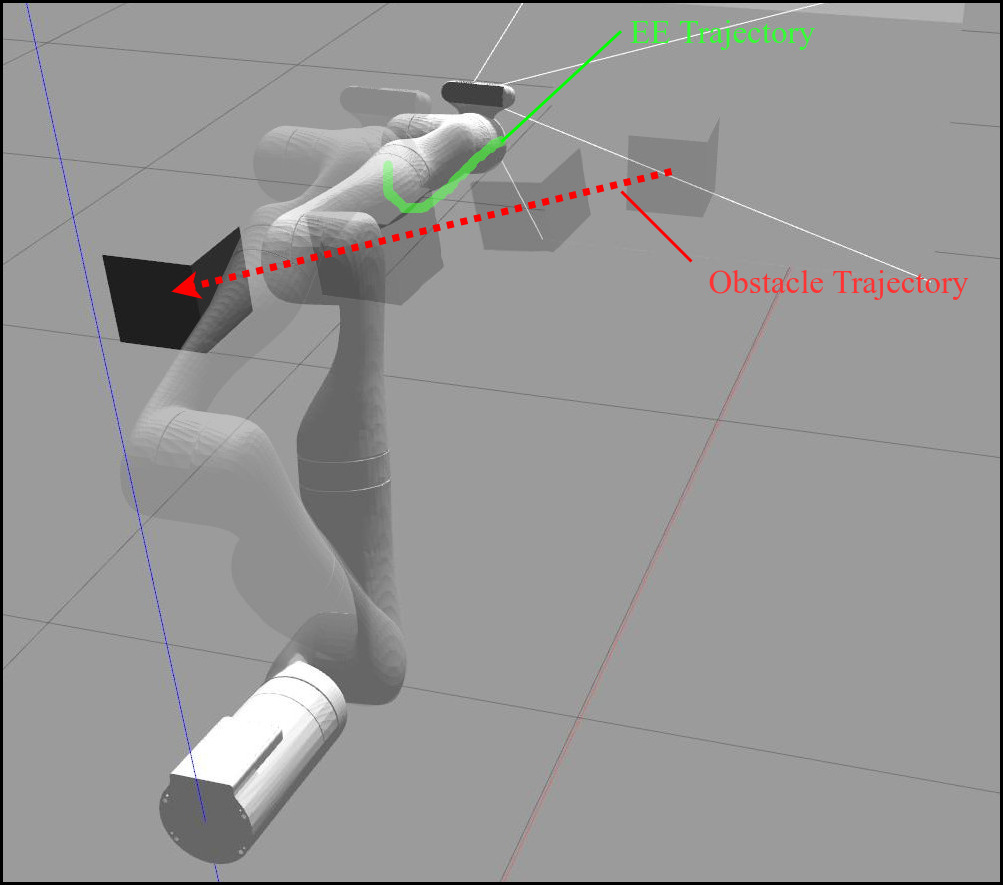}
        \caption{\centering End}
        \label{}
    \end{subfigure}%
    \caption{Simulation Task 2 setup. The obstacle's trajectory and a baseline end-effector trajectory are illustrated in red and green.}
    \label{task_2_demonstrative}
\end{figure}

\begin{figure}
    \centering
    \begin{subfigure}{0.47\linewidth}
        \centering
        \includegraphics[width=\columnwidth]{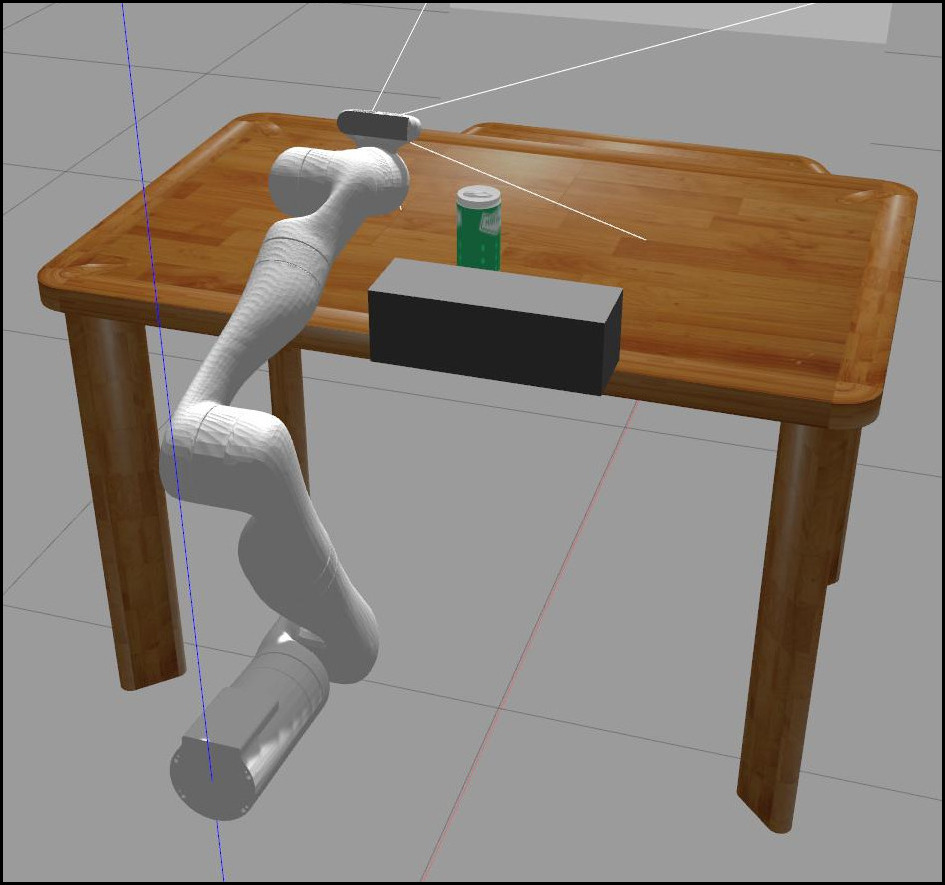}
        \caption{\centering Start}
        \label{}
    \end{subfigure}%
    \hfill
    \begin{subfigure}{0.47\linewidth}
        \centering
        \includegraphics[width=\columnwidth]{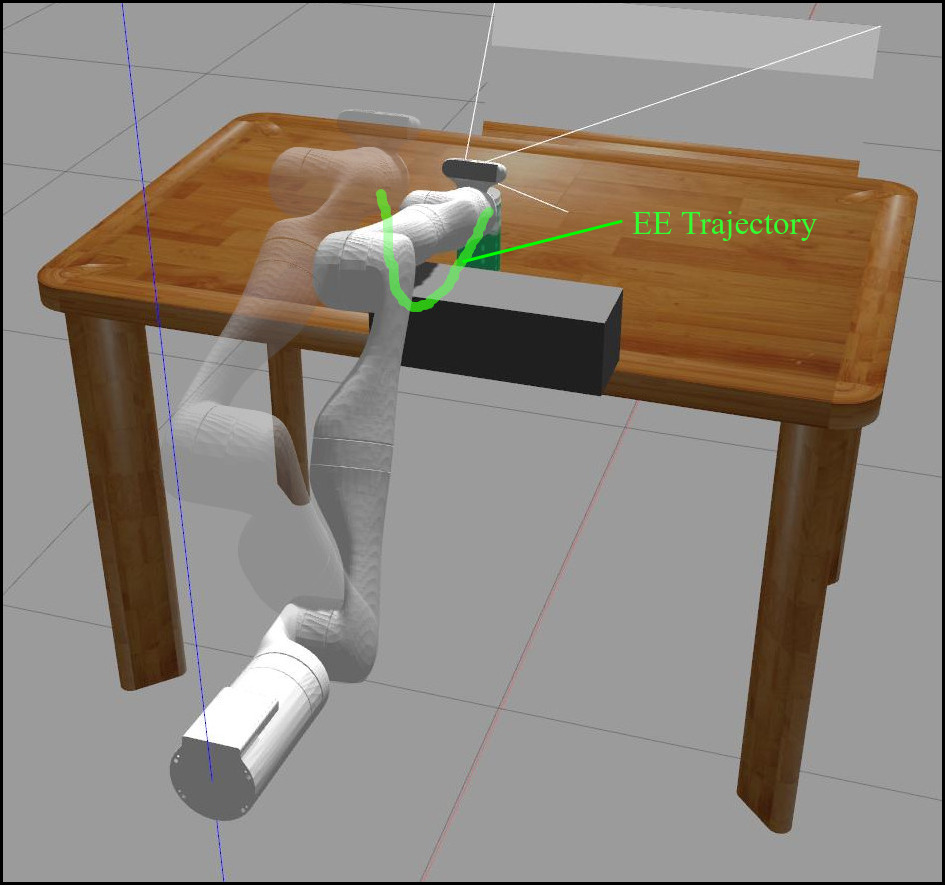}
        \caption{\centering End}
        \label{}
    \end{subfigure}%
    \caption{Simulation Task 3 setup. A baseline trajectory is illustrated in green.}
    \label{task_3_demonstrative}
\end{figure}

\begin{figure}
    \centering
    \begin{subfigure}{0.47\linewidth}
        \centering
        \includegraphics[width=\columnwidth]{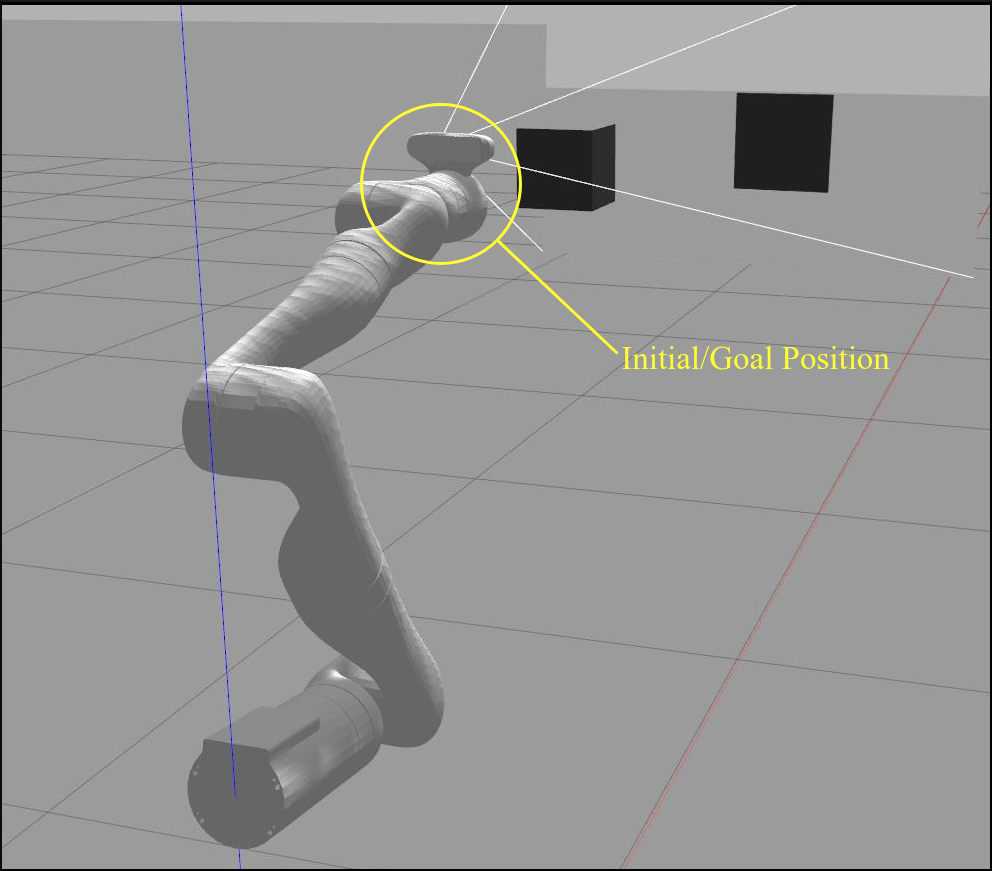}
        \caption{\centering Start}
        \label{}
    \end{subfigure}%
    \hfill
    \begin{subfigure}{0.47\linewidth}
        \centering
        \includegraphics[width=\columnwidth]{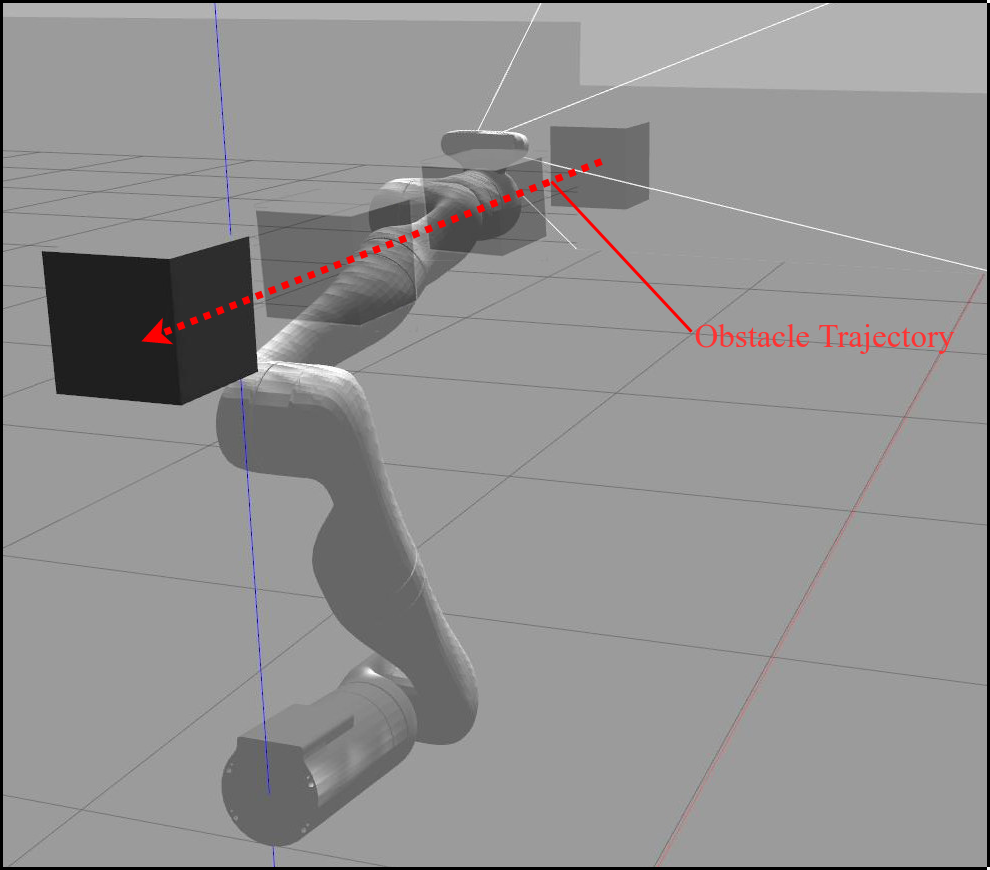}
        \caption{\centering End}
        \label{}
    \end{subfigure}%
    \caption{Simulation Task 4 setup. The obstacle's trajectory and a baseline end-effector trajectory are illustrated in red and green.}
    \label{task_4_demonstrative}
\end{figure}

%\paragraph{}
Figures \crefrange{task_1_demonstrative}{task_4_demonstrative} illustrate the setups of the tasks in Gazebo.
Each figure depicts the start and end of each task, in addition to a baseline end-effector trajectory that does not consider obstacles.

\subsection*{Backgrounds, Object Types and Object Colors/Textures}

\begin{figure}
    \centering
    \begin{subfigure}{0.48\linewidth}
        \centering
        \includegraphics[width=\columnwidth]{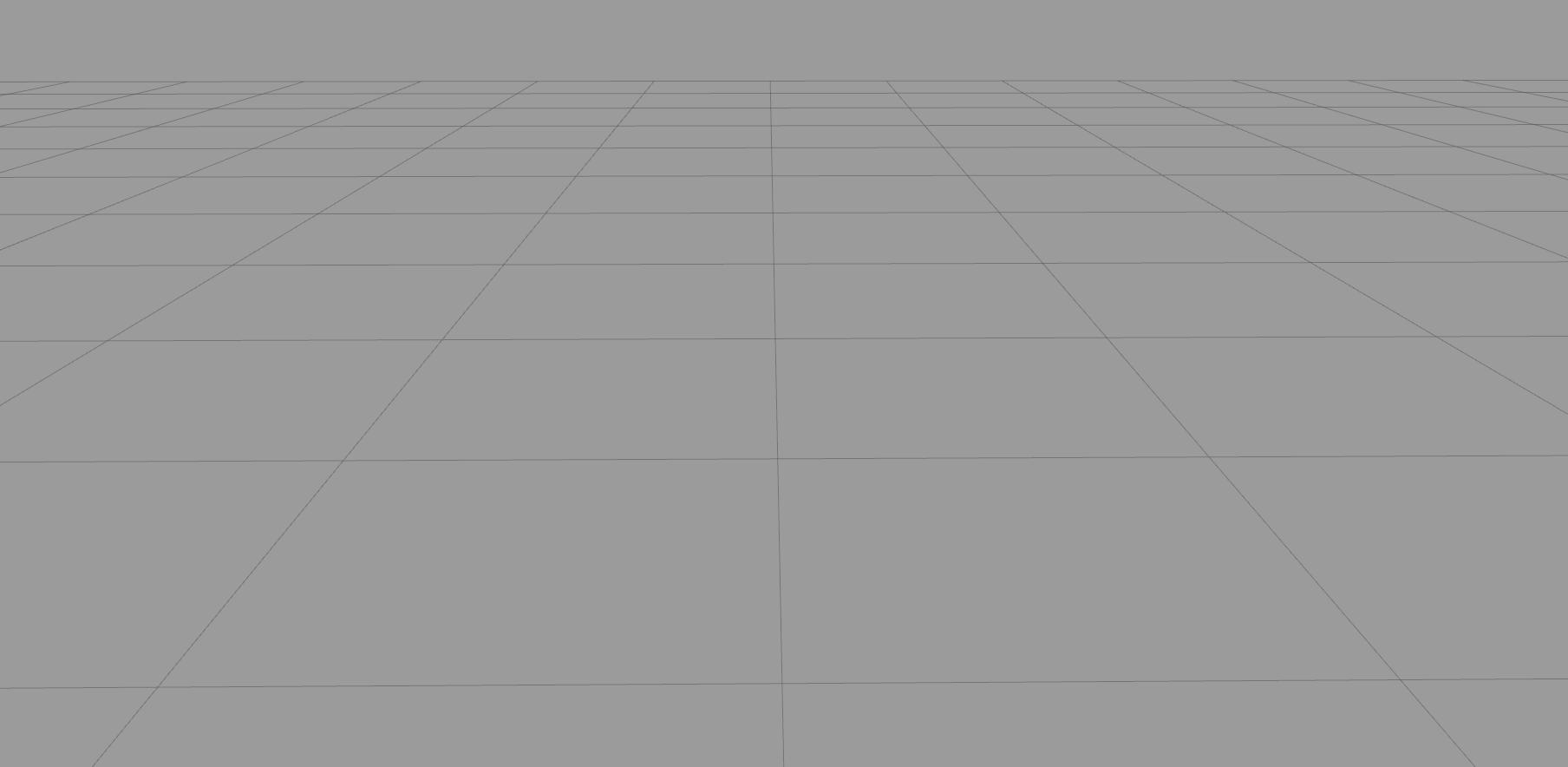}
        \caption{\centering Empty}
        \label{scenario_backgrounds:empty}
    \end{subfigure}%
    \hfill
    \begin{subfigure}{0.48\linewidth}
        \centering
        \includegraphics[width=\columnwidth]{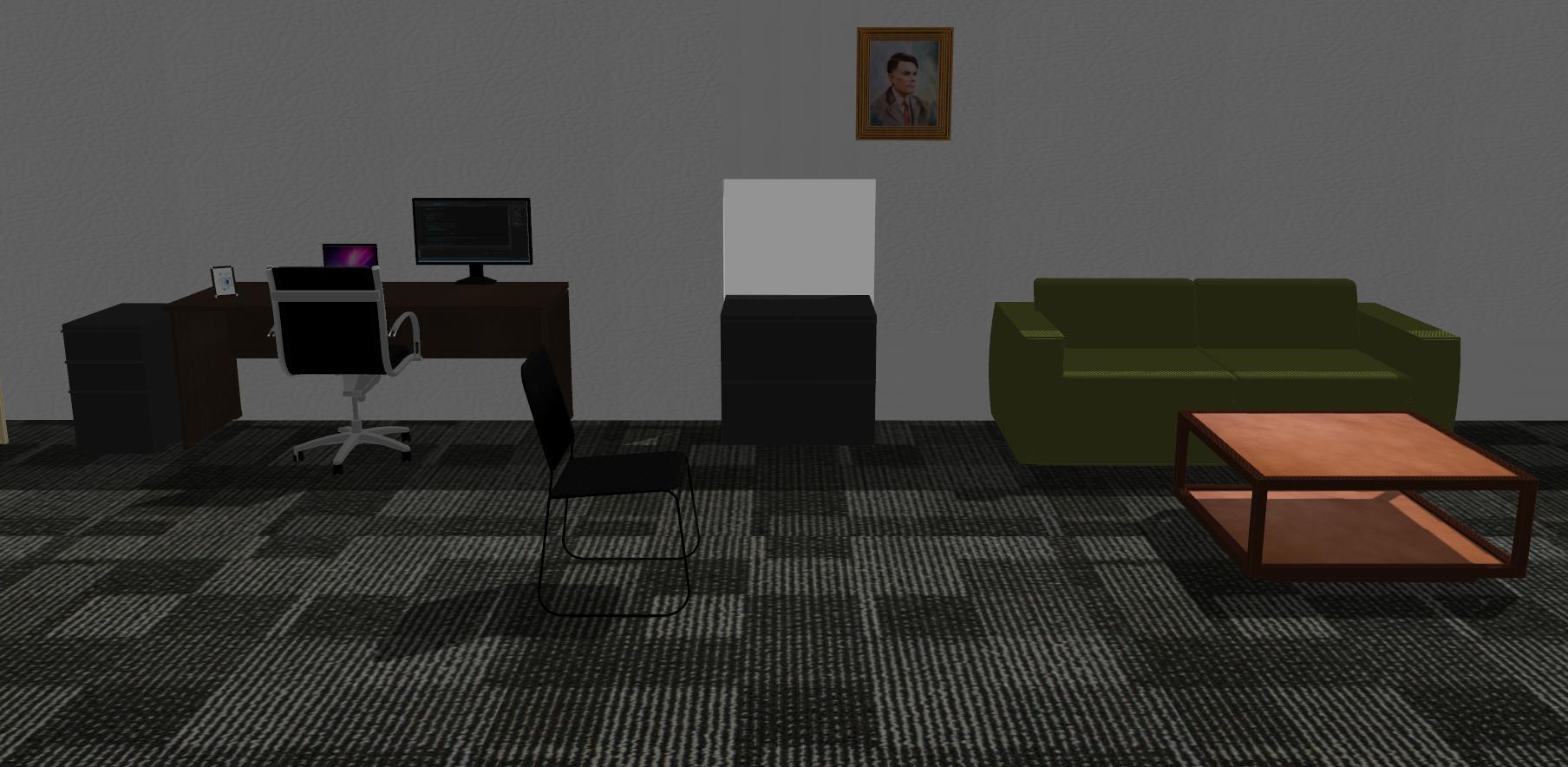}
        \caption{\centering Office}
        \label{scenario_backgrounds:office}
    \end{subfigure}%
    \\
    \begin{subfigure}{0.48\linewidth}
        \centering
        \includegraphics[width=\columnwidth]{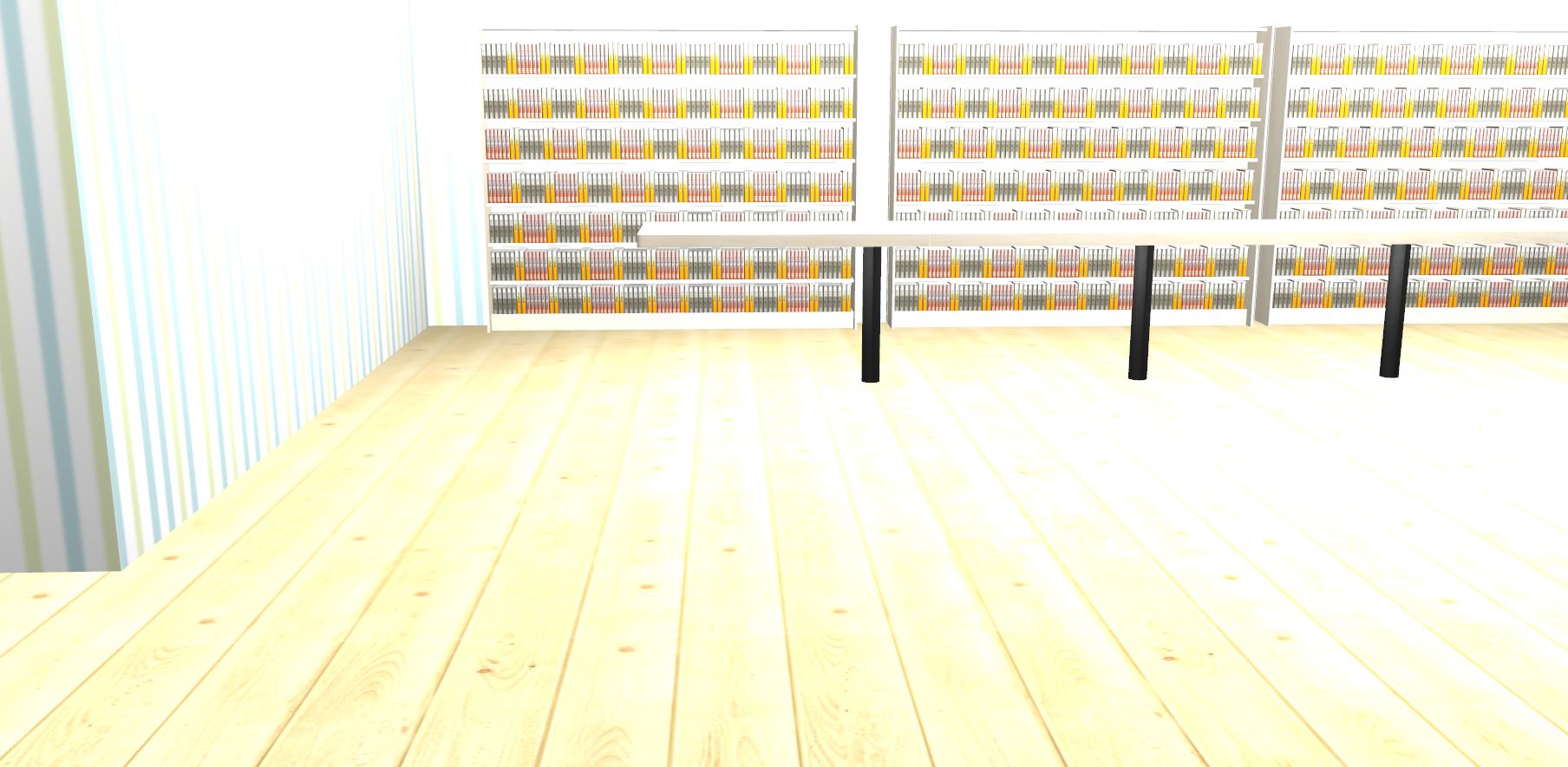}
        \caption{\centering Store (high illumination)}
        \label{scenario_backgrounds:store}
    \end{subfigure}%
    \hfill
    \begin{subfigure}{0.48\linewidth}
        \centering
        \includegraphics[width=\columnwidth]{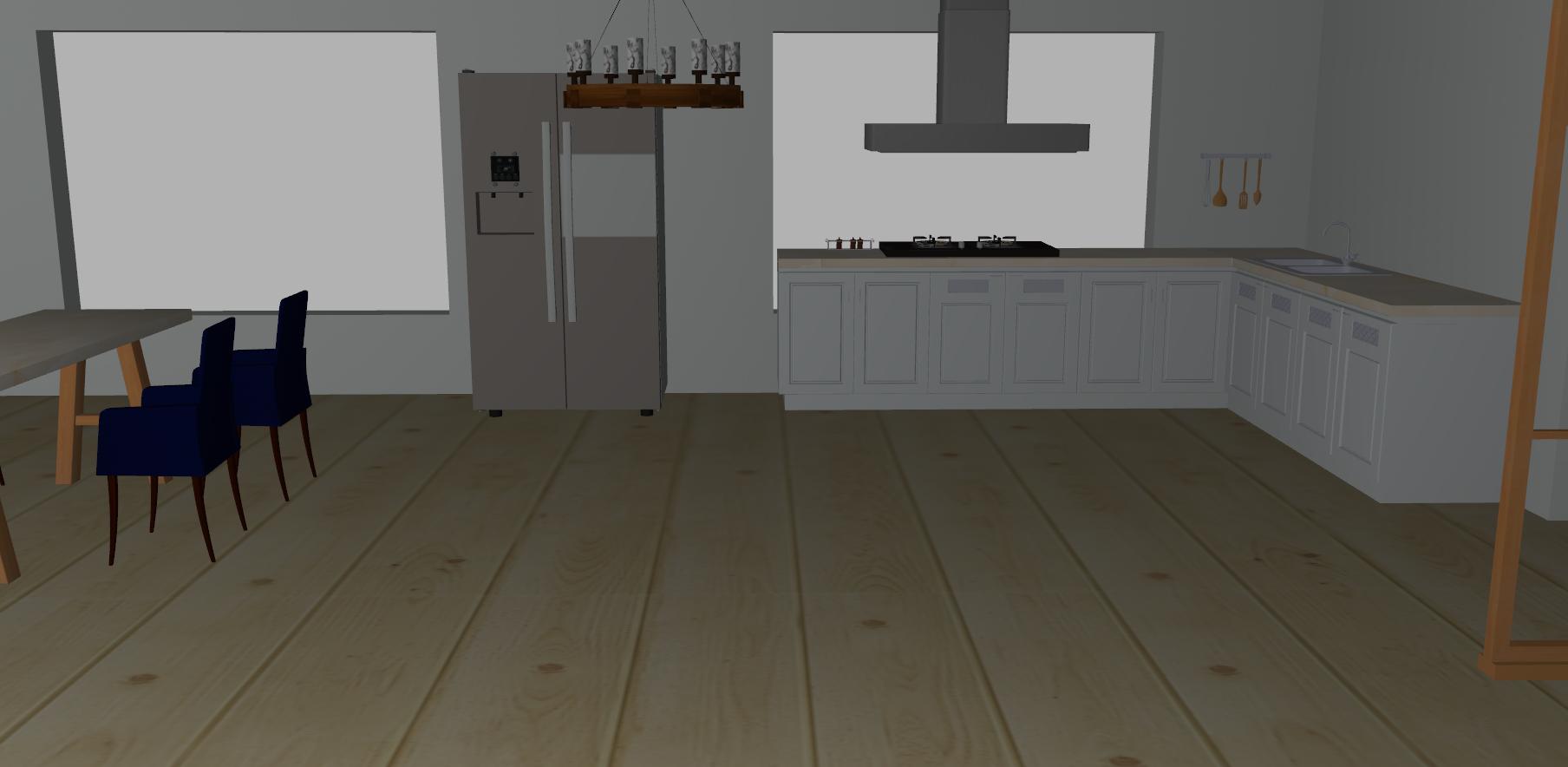}
        \caption{\centering Kitchen (low illumination)}
        \label{scenario_backgrounds:kitchen}
    \end{subfigure}%
    \caption{Different backgrounds (Gazebo worlds) used in task scenarios listed in Table \ref{task_variables}.}
    \label{scenario_backgrounds}
\end{figure}

\begin{figure}
    \centering
    \begin{subfigure}{0.25\linewidth}
        \centering
        \includegraphics[width=0.9\columnwidth]{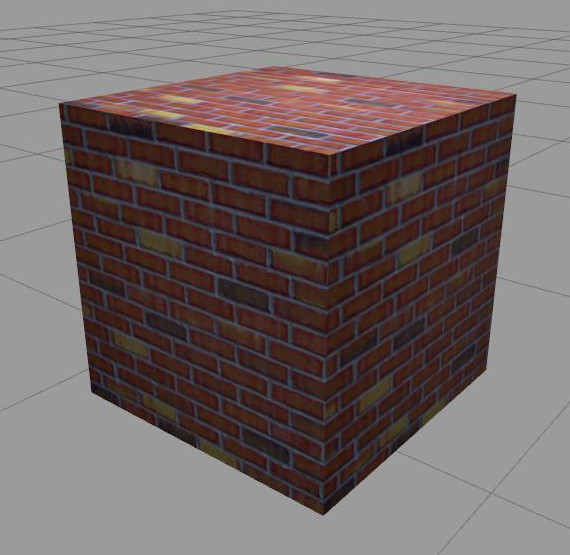}
        \caption{\centering Box: \\Brick Pattern}
        \label{scenario_obstacle_types:brick_box}
    \end{subfigure}%
    \begin{subfigure}{0.25\linewidth}
        \centering
        \includegraphics[width=0.9\columnwidth]{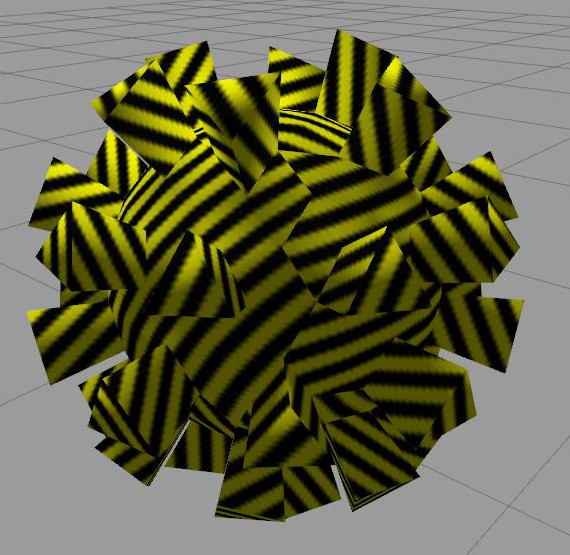}
        \caption{\centering Spiky Sphere: \\Y-B Pattern}
        \label{scenario_obstacle_types:spiky_sphere_yellow_black}
    \end{subfigure}%
    \begin{subfigure}{0.25\linewidth}
        \centering
        \includegraphics[width=0.9\columnwidth]{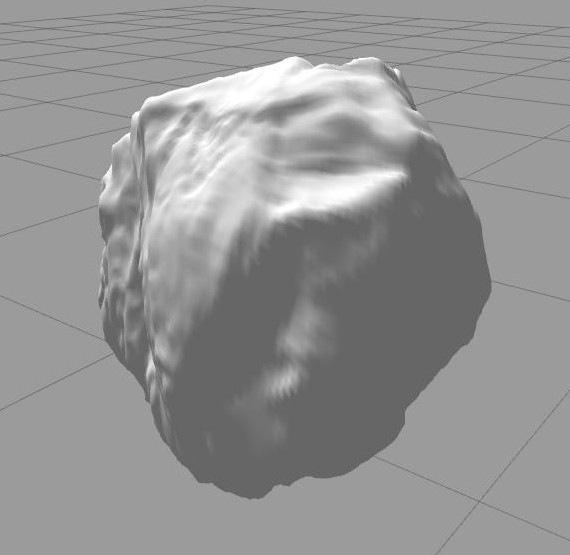}
        \caption{\centering Rock: \\White}
        \label{scenario_obstacle_types:white_rock}
    \end{subfigure}%
    \begin{subfigure}{0.25\linewidth}
        \centering
        \includegraphics[width=0.9\columnwidth]{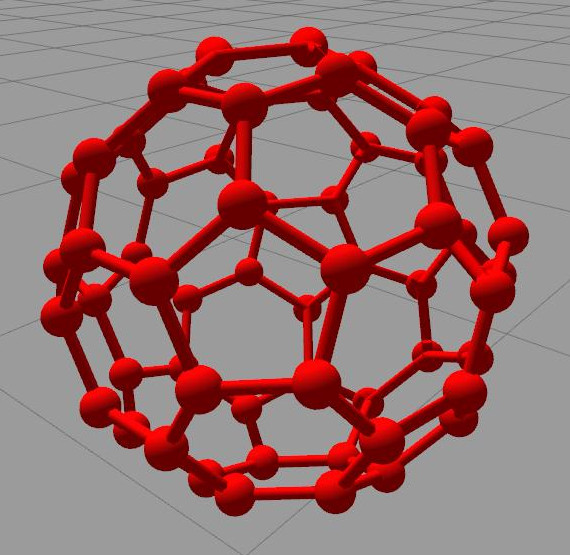}
        \caption{\centering Buckyball: \\Red}
        \label{scenario_obstacle_types:red_buckyball}
    \end{subfigure}%
    \\
    \caption{Different obstacle types used in task scenarios, listed in Table \ref{task_variables}. Each obstacle also illustrates one of the possible obstacle colors.}
    \label{scenario_obstacle_types}
\end{figure}

%\paragraph{}
Figures \ref{scenario_backgrounds} and \ref{scenario_obstacle_types} depict the different values of the background, obstacle type, and obstacle color variables presented in Table \ref{task_variables}.

\section*{Appendix B: List of Experiment Scenarios}
\label{appendix:experiment_scenarios}

%\paragraph{}
This appendix contains the lists of task scenarios that were used in the simulated and real experiments in Tables \ref{tuning_validation_testing_scenarios} and \ref{real_experiments_scenarios}.

\begin{table*}
    \small\sf
    \centering
    \caption{List of tuning, validation, and testing scenarios.}
    \begin{tabularx}{\textwidth}[t]{p{0.12\linewidth}p{0.25\linewidth}X}
        \toprule
        Scenario ID & Set & Scenario Specification\\
        \midrule
        0 & Pre-Tuning & Task 1, Empty, Cracker box\\
        1 & Tuning & Task 1, Store, Red Box\\
        2 & Tuning & Task 4, Store, Brick, Rock, Medium Speed\\
        3 & Tuning & Task 3, Empty, Brick\\
        
        4 & Validation & Task 1, Office, Red, Buckyball\\
        5 & Validation & Task 1, Office, Yellow-Black, Rock\\
        6 & Validation & Task 1, Empty, Brick, Spiky Sphere\\
        7 & Validation & Task 4, Empty, Yellow-Black, Box, Medium Speed\\
        8 & Validation & Task 4, Office, Yellow-Black, Buckyball, High Speed\\
        9 & Validation & Task 4, Store, White, Spiky Sphere, Low Speed\\
        10 & Validation & Task 3, Office, White\\
        11 & Validation & Task 3, Store, Yellow-Black\\
        
        12 & Testing & Task 1, Empty, Yellow-Black, Spiky Sphere\\
        13 & Testing & Task 1, Store, Brick, Buckyball\\
        14 & Testing & Task 1, Kitchen, White, Buckyball\\
        15 & Testing & Task 1, Empty, Yellow-Black, Box\\
        16 & Testing & Task 1, Store, Yellow-Black, Rock\\
        17 & Testing & Task 2, Office, Red, Buckyball, High Speed\\
        18 & Testing & Task 2, Office, Red, Rock, Medium Speed\\
        19 & Testing & Task 2, Store, Brick, Box, Medium Speed\\
        20 & Testing & Task 2, Kitchen, Yellow-black, Box, High Speed\\
        21 & Testing & Task 2, Empty, Red, Spiky Sphere, Medium Speed\\
        22 & Testing & Task 3, Empty, Red\\
        23 & Testing & Task 3, Empty, Yellow-Black\\
        24 & Testing & Task 3, Kitchen, White\\
        25 & Testing & Task 3, Kitchen, Red\\
        26 & Testing & Task 3, Store, Red\\
        27 & Testing & Task 4, Store, Yellow-Black, Buckyball, Low Speed\\
        28 & Testing & Task 4, Store, Red, Buckyball, High Speed\\
        29 & Testing & Task 4, Empty, Yellow-Black, Spiky Sphere, Medium Speed\\
        30 & Testing & Task 4, Empty, Yellow-Black, Box, Low Speed\\
        31 & Testing & Task 4, Empty, Brick, Buckyball, Low Speed\\
        \bottomrule
    \end{tabularx}
    \label{tuning_validation_testing_scenarios}
\end{table*}

\begin{table*}
    \small\sf
    \centering
    \caption{List of real robot experiment task scenarios.}
    \begin{tabularx}{\textwidth}[t]{p{0.12\linewidth}p{0.25\linewidth}X}
        \toprule
        Scenario ID & Set & Scenario Specification\\
        \midrule
        R1 & Testing (Real) & Task 1, Lab Background 1, Wooden Block\\
        
        R2 & Testing (Real) & Task 2, Lab Background 2, Hand\\
        R3 & Testing (Real) & Task 2, Lab Background 2, Metal Bar\\
        R4 & Testing (Real) & Task 2, Lab Background 2, Wooden Block\\
        \bottomrule
    \end{tabularx}
    \label{real_experiments_scenarios}
\end{table*}

\section*{Appendix C: Evaluation Metrics and Criteria Details}
\label{appendix:evaluation_metrics_and_criteria_details}

%\paragraph{}
This appendix contains more detailed descriptions of the quantitative metrics and qualitative evaluation criteria presented in section \ref{evaluation_metrics_and_criteria}.

\subsection*{Quantitative Metrics}

%\paragraph{}
Table \ref{quantitative_metrics} lists the quantitative metrics we use to analyze the performances of parameter sets and to evaluate our approach.

\begin{table*}
    \small\sf
    \centering
    \caption{Quantitative performance metrics.}
    \begin{tabularx}{\textwidth}[t]{p{0.18\linewidth}>{\centering\arraybackslash}p{0.15\linewidth}X}
        \toprule
        Metric & Symbol & Description\\
        \midrule
        Execution Time & $T$ & Time required to complete the task\\
        Trajectory Length & $l_{\mathbf{Y}}$ & Distance traveled by the end-effector\\
        Number of Collisions & $N_{collisions}$ & No. of instances in which the EE intersects an obstacle (in simulation)\\
        Distance to Goal & $d_G$ & Euclidean distance to the goal at the end of execution\\
        Success & S & True \textbf{iff} no collisions \textbf{and} distance to goal $< \epsilon$\\
        Velocities & $\mathbf{\dot{y}}$ & Magnitudes of instantaneous end-effector velocities\\
        Accelerations & $\mathbf{\ddot{y}}$ & Magnitudes of instantaneous end-effector accelerations\\
        \bottomrule
    \end{tabularx}
    \label{quantitative_metrics}
\end{table*}

%\paragraph{}
Execution time, $T$, refers to how long it takes to execute the task.
Relative to baseline trajectories, it indicates how much more time is expended in obstacle maneuvers.

%\paragraph{}
The trajectory length estimates the distance traveled by the end-effector and is similarly used to assess how far obstacle avoidance trajectory adaptations tend to deviate from the baseline.
It can be estimated by summing the euclidean distances between consecutive end-effector positions:
\begin{equation}
    l_{\mathbf{Y}} = \sum_{i=1}^{N_{pos}}||\mathbf{y}_{i} - \mathbf{y}_{i-1}||_2
\end{equation}
$N_{pos}$ refers to the number of recorded positions while executing trajectory $\mathbf{Y}$.
It is generally desirable to minimize both $T$ and $l_{\mathbf{Y}}$ as a secondary objective to successful obstacle avoidance and reaching the goal.

%\paragraph{}
In simulation, the number of collisions, $N_{collisions}$, represents a count of the instances in which the end-effector intersects an obstacle (given that physical collision effects are disabled). 
Intuitively, this provides an indication of how often the end-effector collides and the degree to which collisions are sustained or eventually adapted for.

%\paragraph{}
The distance-to-goal metric measures how well a given execution fulfills the goal reaching aspect of the task and is obtained from the euclidean distance from the final position, $\mathbf{y}_T$, to the commanded goal position, $g$:
\begin{equation}
    d_G = ||\mathbf{y}_T - \mathbf{g}||_2
    \label{distance_to_goal_equation}
\end{equation}

%\paragraph{}
Success is captured in binary metric $S$, which incorporates $N_{collisions}$ and $d_G$: for a success, a given trial must i) involve no collisions and ii) end with the end-effector close to the goal, according to the goal-reaching tolerance $\delta_{\mathbf{g}}$:
\begin{equation}
    S = 
    \begin{cases}
    1,	& \text{if } N_{collisions} = 0 \text{ and } d_G < \delta_{\mathbf{g}} \\
    0,	& \text{otherwise} \\
    \end{cases}
    \label{success_equation}
\end{equation}

%\paragraph{}
We also measure the magnitudes of end-effector velocities and accelerations, which provide insights on motion properties particularly with regards to qualitative criteria.

\subsection*{Qualitative Metrics}

%\paragraph{}
Table \ref{qualitative_criteria} lists our qualitative evaluation criteria, which are described along with their formalisms in the following.

%\paragraph{}
\textit{Reliability} can be defined as the degree to which a method produces consistently positive results in successive trials under reasonably consistent conditions.
We expect a reliable system to repeatedly achieve similarly positive results, as opposed to significant variances and/or failure rates, provided that testing conditions do not substantially differ from the system’s intended operating environment.
One way to estimate the reliability is through the ratio of successful trials to the total number of trials in imminent collision situations, which are provided in the task scenarios defined in section \ref{simulation_evaluation_tasks}.
A high value indicates that obstacle avoidance was consistently successful across trials, while a lower value indicates an inconsistency that shows unreliability, even if some success was achieved.

%\paragraph{}
\textit{Predictability} describes to what extent behaviour can be known or predicted in advance, for which we use angular velocities that describe changes in direction.
The results of a controller that produces fewer directional changes are easier to intuit or predict from historical observations (within or across executions).
We can quantify the local degree of direction change at every end-effector position from the angle between the vectors formed by connecting that positional point to the previous and the next point, as is done in \cite{salarpour2019direction}.
Given trajectory positions $\mathbf{y}_k$, $\mathbf{y}_{k-1}$, and $\mathbf{y}_{k+1}$, and resultant vectors $\mathbf{v}_1=(\mathbf{y}_k - \mathbf{y}_{k-1})$ and $\mathbf{v}_2=(\mathbf{y}_{k+1} - \mathbf{y}_k)$, the direction change, $\zeta$, at $\mathbf{y}_k$ can be calculated from the dot product:
\begin{equation}
    \zeta = \arccos{\left(\frac{v_1 \cdot v_2}{|v_1| |v_2|}\right)}
    \label{change_in_direction_equation}
\end{equation}
The magnitudes of $\zeta$ provide an indication of path curvature, and higher values do not necessarily indicate unpredictable motions.
On the other hand, the magnitudes of $\dot{\zeta}$, which could be expressed as an angular velocity, are related to how often and how drastically the end-effector changes the direction in which it is moving.

%\paragraph{}
\textit{Safety} refers to the degree to which a system is unlikely to cause danger, risk, or injury. 
In the case of the manipulation tasks addressed here, we can examine the magnitudes of applied velocities and accelerations of the end-effector as a measure of safety. 
Intuitively, higher velocities tend to increase the risk of collisions with other objects or people in the environment, and the magnitude of potential damage.

\section*{Appendix D: Implementation Details}
\label{appendix:implementation_details}

%\paragraph{}
The components described in section \ref{proposed_approach} were implemented as ROS nodes, using the ROS Noetic Ninjemys distribution and Python 3.8 for development.
This section summarizes details of our the ROS stack and the augmented Gazebo simulation of the Kinova Gen3.

\subsection*{ROS Components}

\begin{figure*}
\centering
\includegraphics[width=0.9\linewidth]{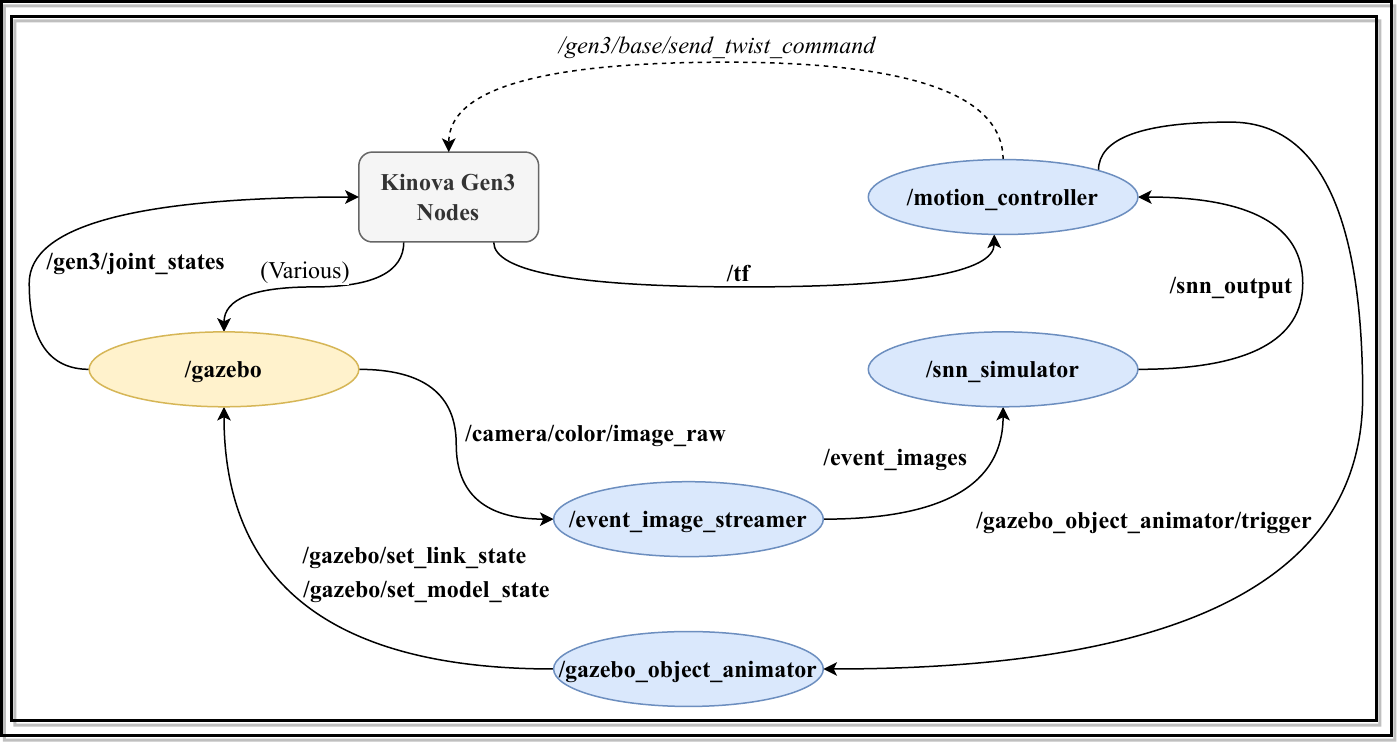}
\caption{A graph of the main ROS nodes and the topics/services they communicate on. Services are depicted with dotted arrows and italics; bold arrows represent topics.
    The multiple nodes associated with the Kinova arm drivers (from the \textit{ros\_kortex} package) have been reduced to a single “Kinova Gen3 Nodes” block.}
\label{ros_components_graph}
\end{figure*}

%\paragraph{}
Figure \ref{ros_components_graph} shows a graph of the ROS nodes and the main communication channels between them (ROS topics and services).
When running experiments on the robot, the Gazebo node is substituted by others, but the obstacle avoidance pipeline is unchanged. 
In the following, we describe each of the nodes (depicted in blue).

%\paragraph{}
Our \textit{event\_camera\_emulation} Python package provides functionalities for generating, streaming, and visualizing event data derived from RGB images from a camera or ROS publishers.
An \mintinline{Python}{event_image_streamer} ROS node handles capturing RGB image frames, creating event images, and publishing the event data.

%\paragraph{}
The \mintinline{Python}{snn_simulator} node receives event data and runs it through a simulated C-SNN to produce the neural activation maps required by the obstacle avoidance component.
We use the open-source \textit{BindsNET} package for running SNNs (presented in \cite{hazan2018bindsnet}).
In order to use the FST output representation described in section \ref{proposed_approach:obstacle_avoidance_component}, we extended the implementation to incorporate the computation and recording of first spike times.
Upon receiving event images, the node pre-processes the inputs by:
\begin{enumerate}
    \item Optionally applying an event filtering strategy
    \item Scaling the image 
    to match the input layer size
    \item Assigning $r_{ON}$ and $r_{OFF}$ rate values for Poisson encoding 
    (see section \ref{proposed_approach:convolutional_spiking_neural_networks})
    \item Encoding the data into Poisson spike trains
\end{enumerate}
The spike data is run through the C-SNN and network monitors are inspected to extract statistics such as membrane potentials, spike counts, average firing rate, and FSTs, which are published in a custom ROS message.

%\paragraph{}
The \mintinline{Python}{motion_controller} handles configuring the DMP, executing velocity commands from the PID controller to follow the trajectory plan, and adapting the plan by receiving SNN output data, decoding it, and adapting the DMP trajectory.
We use the \textit{pydmps} package to run the DMP.m, $\phi$ (see section \ref{proposed_approach:obstacle_avoidance_component}).
The ROS framework allows for the DMP execution and the processing of SNN outputs to run simultaneously, ensuring that the motion is adapted with the latest possible information on sensed obstacles.
A custom \mintinline{Python}{KinovaArmController} class interfaces with the Kinova's \textit{kortex\_driver} (within \textit{ros\_kortex}), sending twist commands computed by the PID controller (through ROS services) while the driver executes the low-level velocity commands.
This is the only component of the node and the pipeline that is robot-specific.
Therefore, the implementation can be adapted to other platforms or applications by simply integrating the associated low-level controller and adjusting parameters where necessary.

%\paragraph{}
The \mintinline{Python}{gazebo_object_animator} node executes dynamic obstacle motions by moving a given object through a specified trajectory.
During experiments, the \mintinline{Python}{motion_controller} publishes a trigger command (ROS boolean message) as execution starts to induce a controlled, relative object motion.
This is useful when statistically evaluating performance, as it eliminates variance that may arise due to slightly varying testing conditions.

\subsection*{Simulation}

%\paragraph{}
We use an existing, open-source Gazebo simulation of the Kinova Gen3 arm, and extend it to include an onboard RGB camera.
The simulation integrates with the same drivers and components that run on the real robot, which are accessible through the \textit{ros\_kortex} ROS package.
We extended the 7-DOF Gen3 arm simulation with an Intel RealSense camera (from the \textit{realsense-ros} package
mounted at the top of the end-effector and a plugin that enables configuring data capture and accessing the camera data through ROS topics.

\section*{Appendix E: Pipeline Component Parameters}
\label{appendix:component_parameters}

%\paragraph{}
Table \ref{pipeline_component_parameters} lists and describes the main parameters of our neuromorphic pipeline, divided by component.

\begin{table*}
    \small\sf
    \centering
    \caption{The parameters of each pipeline component. A (*) indicates a parameter which was not tuned in our experiments}.
    \begin{tabularx}{\textwidth}[t]{p{0.25\linewidth}>{\centering\arraybackslash}p{0.2\linewidth}X}
        \toprule
        Component & Symbol & Description \\
        \midrule
        \multirow{1}{*}{Event Camera Emulator} & \multicolumn{1}{c}{$\theta$} & Event emission threshold \\
        & \multicolumn{1}{c}{$s_{BE}$} & Binary erosion filter structure size \\
        \midrule
        \multirow{1}{*}{Convolutional SNN} & \multicolumn{1}{c}{$w_{c}$} & SNN weight initialization constant \\
        & \multicolumn{1}{c}{$T_{sim}$} & SNN simulation time period \\
        & \multicolumn{1}{c}{$v_{thresh}$} & SNN potential spiking threshold \\
        & \multicolumn{1}{c}{$T_{refrac}$} & SNN refractory spiking period \\
        & \multicolumn{1}{c}{$\tau_v$} & SNN potential decay constant \\
        & \multicolumn{1}{c}{$v_{rest}$ (*)} & SNN resting potential \\
        & \multicolumn{1}{c}{$v_{reset}$ (*)} & SNN reset potential \\
        \midrule
        \multirow{1}{*}{Obstacle Avoidance Component} & \multicolumn{1}{c}{$\phi_{max}$} & Upper limit on $\vect{\phi}$ \\
        & \multicolumn{1}{c}{$n_{\phi}$} & $\vect{\phi}$ history horizon \\
        & \multicolumn{1}{c}{$\eta$} & Constant gain (PF; Park) \\
        & \multicolumn{1}{c}{$C_{\delta}$} & Gradient constant factor (PF; Park) \\
        & \multicolumn{1}{c}{$p_0$ (*)} & Obstacle radius of influence (PF; Park) \\
        & \multicolumn{1}{c}{$t_{act}$ (*)} & FST activation threshold factor \\
        \midrule
        \multirow{1}{*}{Motion Planning and Control} & \multicolumn{1}{c}{$\delta_{\mathbf{y}}$} & Position reaching tolerance \\
        & \multicolumn{1}{c}{$\delta_{obs}$} & Obstacle avoidance distance tolerance \\
        & \multicolumn{1}{c}{$\delta_{safety, x}$} & Safety strategy distance tolerance \endnote{We use two safety strategy for the different task types (see section \ref{tuning_results}). The three safety strategy parameters have distinct values for each, which are indexed by $x$} \\
        & \multicolumn{1}{c}{$\gamma_{v, x}$} & Safety strategy velocity reduction factor \\
        & \multicolumn{1}{c}{$\gamma_{a, x}$} & Safety strategy acceleration reduction factor \\
        & \multicolumn{1}{c}{$\delta^{+}_{pos, i}$} & Upper positional limit (dimension $i$) \\
        & \multicolumn{1}{c}{$\delta^{-}_{pos, i}$} & Lower positional limit (dimension $i$) \\
        & \multicolumn{1}{c}{$K_j$ (*)} & Controller gain ($j$ $\in$ $\{P, I, D\}$) \\
        & \multicolumn{1}{c}{$\delta_{\mathbf{g}}$ (*)} & Goal reaching tolerance \\
        \bottomrule
    \end{tabularx}
    \label{pipeline_component_parameters}
\end{table*}

\section*{Appendix F: Event Emulation Comparison Visualizations}
\label{appendix:ece_comparison_visualizations}

\begin{figure}
    \centering
    \includegraphics[width=0.95\linewidth]{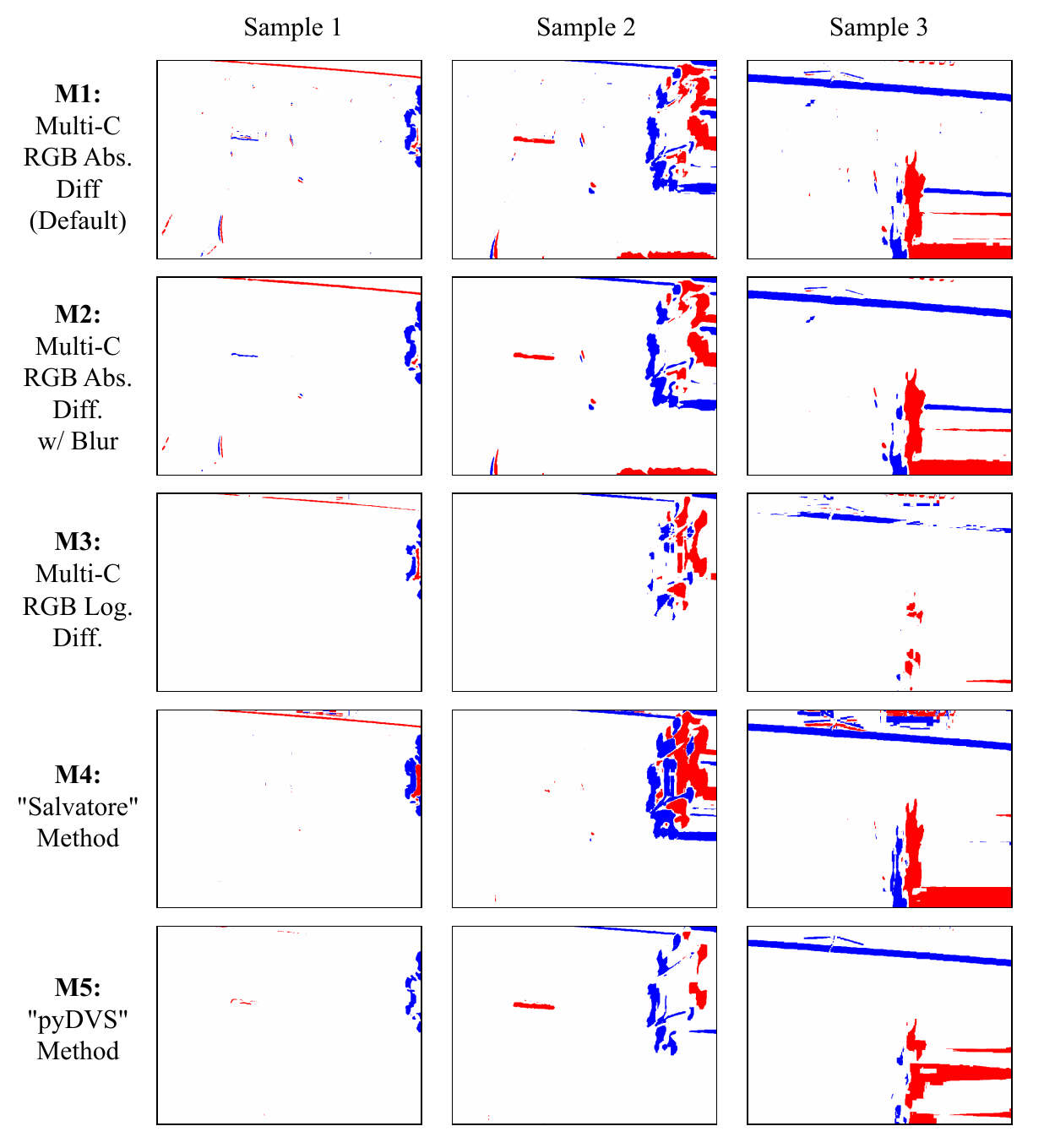}
    \caption{Event images produced by each emulation strategy described in section \ref{further_analyses:comparison_of_event_emulation_strategies}. Although the event distributions are fairly similar, a few notable differences include M2 (with blurring) producing less noisy events than M1, the log-based methods M3 and M4 producing more events in darker regions, and M1, M2, and M4 reacting more strongly to reflective surfaces.
	}
    \label{ece_strategy_comparison_events_images}
\end{figure}

\begin{figure}
    \centering
    \includegraphics[width=1.05\linewidth, trim={15pt 0pt 0pt 0pt},clip]{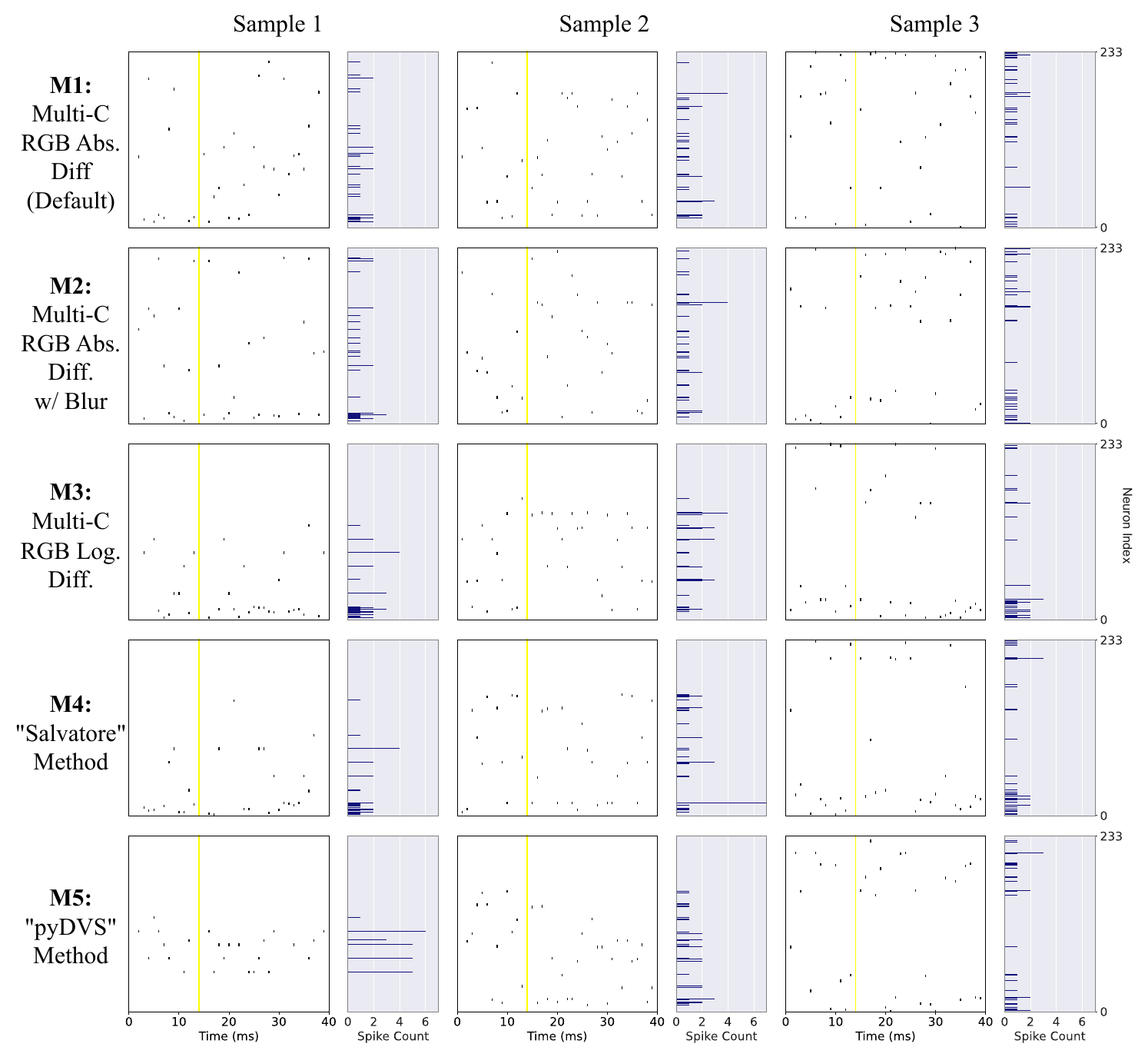}
    \caption{SNN responses to input events from each emulation strategy described in section \ref{further_analyses:comparison_of_event_emulation_strategies} for each sample. The plots show the spike trains (left) and spike counts (right) for each output neuron. The responses largely match the input events shown in Figure \ref{ece_strategy_comparison_events_images} (e.g. more spikes in the first few neurons for M1-M4, which correspond to the top regions of the event images). Despite some differences in the input event data, the SNN responses are similar (except for M5), indicating an inherent robustness to variations in events.
}
    \label{ece_strategy_comparison_spikes_plots}
\end{figure}

%\paragraph{}
This appendix contains figures that supplement the event emulation strategy comparison presented in section \ref{further_analyses:comparison_of_event_emulation_strategies}.
The event images produced by all five methods for the three sample RGB inputs and the resulting SNN responses are shown in Figures \ref{ece_strategy_comparison_events_images} and \ref{ece_strategy_comparison_spikes_plots}, respectively.
In the latter, each sub-plot depicts the recorded spike trains (left) and the total count of spikes (right) across $T_{sim}$ at each output neuron. Note that the shared y-axes are indexed by a row-major re-ordering of the neuron indices,
which are originally arranged in a two-dimensional grid, where the first and last indices are in the top-left and right-bottom positions, respectively.

\section*{Appendix G: Qualitative Evaluation Visualizations}
\label{appendix:qualitative_evaluation_visualizations}

%\paragraph{}
This appendix contains supplementary figures for the predictability qualitative evaluations.

\subsection*{Predictability: Angular Velocity Analysis}

\begin{figure}
    \centering
    \begin{subfigure}{\linewidth}
        \centering
        \includegraphics[width=\columnwidth]{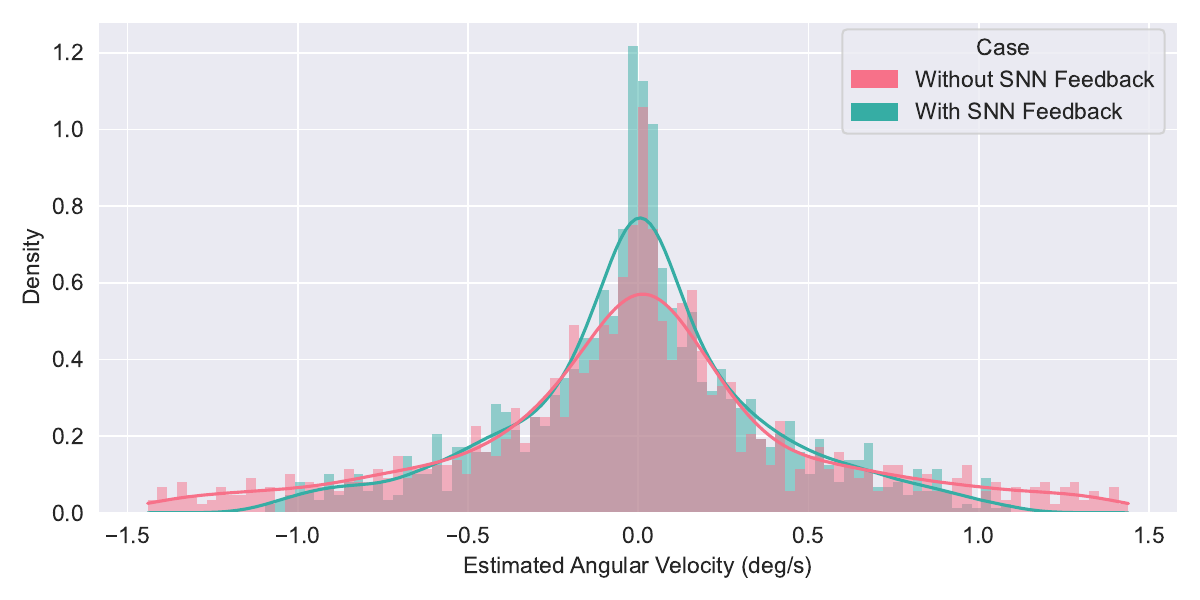}
        \caption{\centering Scenario 12 (Task 1)}
        \label{angular_vel_dist_plots_testing_phase:task_1_example}
    \end{subfigure}%
    \\
    \begin{subfigure}{\linewidth}
        \centering
        \includegraphics[width=\columnwidth]{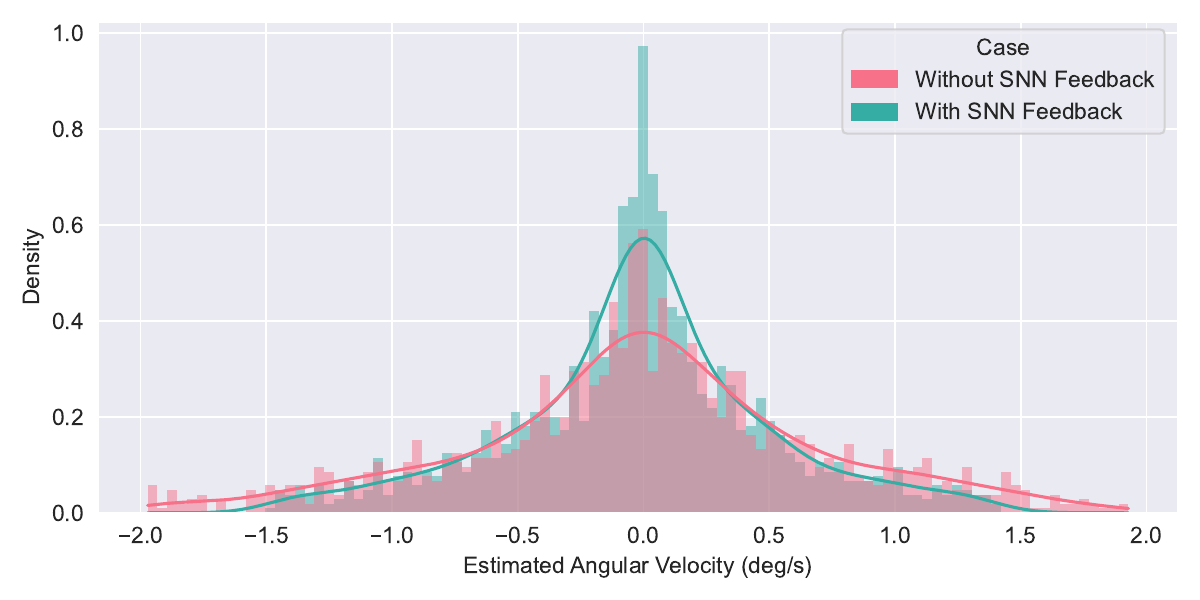}
        \caption{\centering Scenario 17 (Task 2)}
        \label{angular_vel_dist_plots_testing_phase:task_2_example}
    \end{subfigure}%
    \\
    \begin{subfigure}{\linewidth}
        \centering
        \includegraphics[width=\columnwidth]{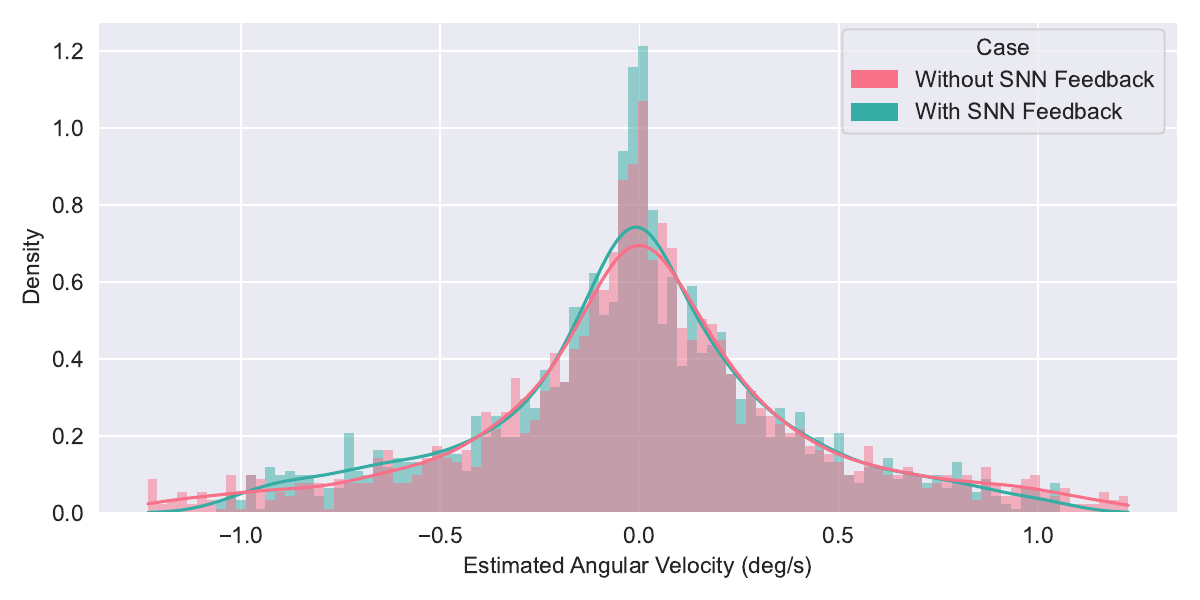}
        \caption{\centering Scenario 24 (Task 3)}
        \label{angular_vel_dist_plots_testing_phase:task_3_example}
    \end{subfigure}%
    \caption{Histograms and kernel density estimate plots depicting the distributions of estimated angular end-effector velocities during testing trials of three sample scenarios. We observe similar distributions from the two cases and small values ([-2, 2] deg/s), which indicates that the trajectories are generally predictable, as described in section \ref{testing_results}.}
    \label{angular_vel_dist_plots_testing_phase}
\end{figure}

\begin{figure}
    \centering
    \includegraphics[width=\columnwidth]{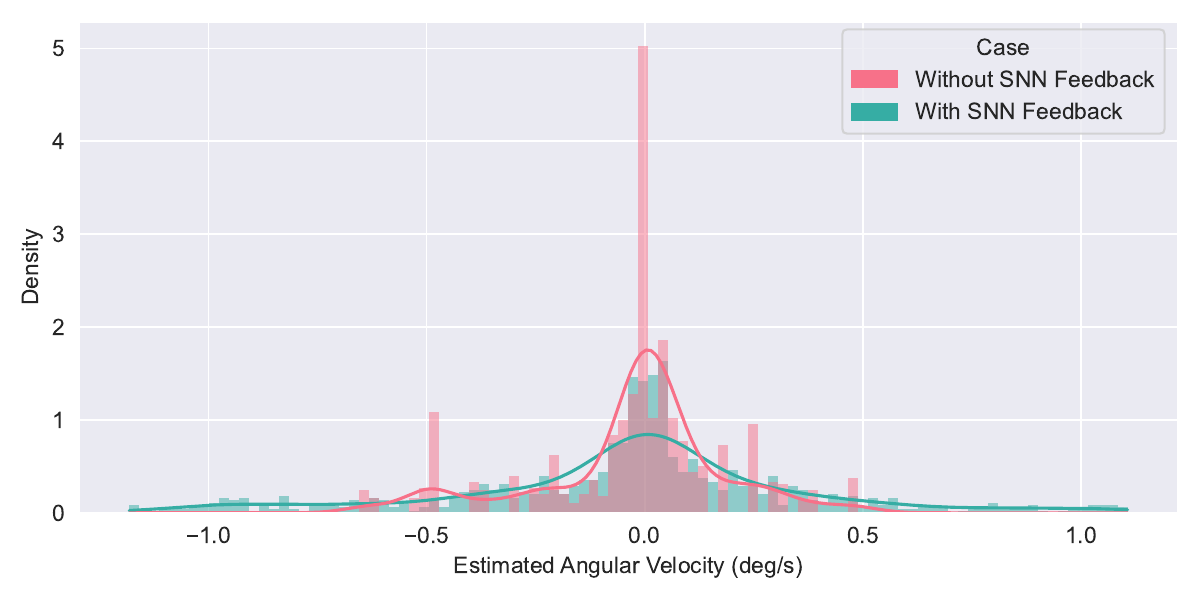}
    \caption{A histogram and kernel density estimate plot depicting the distributions of estimated angular end-effector velocities measured in real experiments (scenario R2). Similar to the simulation case, we observe low values that indicate no sudden, large changes in the end-effector's motion direction, which corresponds to generally predictable robot motions.}
    \label{angular_vel_dist_plots_real_experiments}
\end{figure}

%\paragraph{}
Figures \ref{angular_vel_dist_plots_testing_phase} and \ref{angular_vel_dist_plots_real_experiments} show examples of the estimated densities of angular velocities, $\dot{\zeta}$, that are used to evaluate predictability from simulated and real trials, respectively. Refer to the relevant discussions in sections \ref{testing_results} and \ref{results_and_discussion:real_experiments}.

\section*{Appendix H: Results Metrics and Trajectory Plots}
\label{appendix:metrics_and_trajectory_plots}

\begin{figure}
    \centering
    \begin{subfigure}{0.33\linewidth}
        \centering
        \includegraphics[width=\columnwidth, trim={90pt 10pt 83pt 25pt},clip]{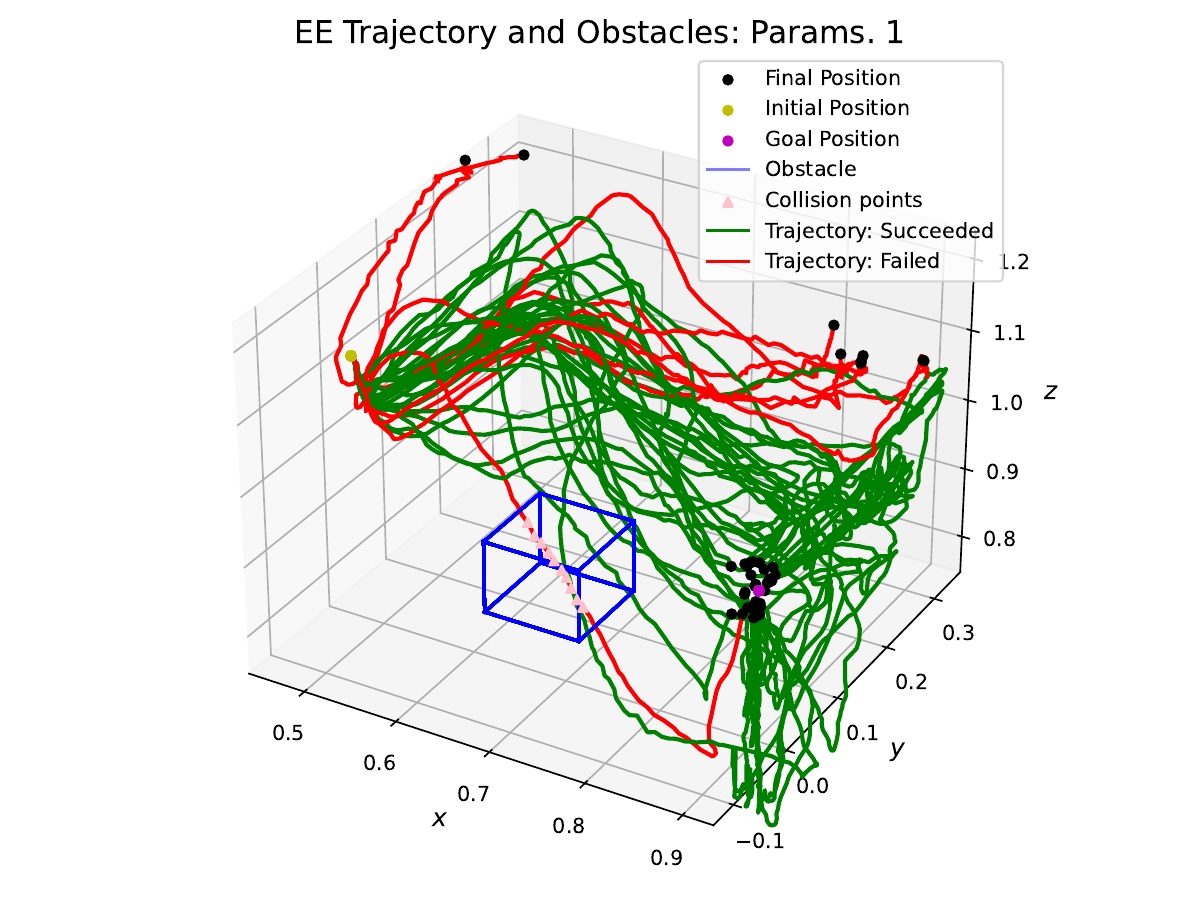}
        \caption{\centering Parameter Set 1}
        \label{tuning_scenario_1_trial_trajs_params1_to_params5:params1}
    \end{subfigure}%
    \begin{subfigure}{0.33\linewidth}
        \centering
        \includegraphics[width=\columnwidth, trim={90pt 10pt 83pt 25pt},clip]{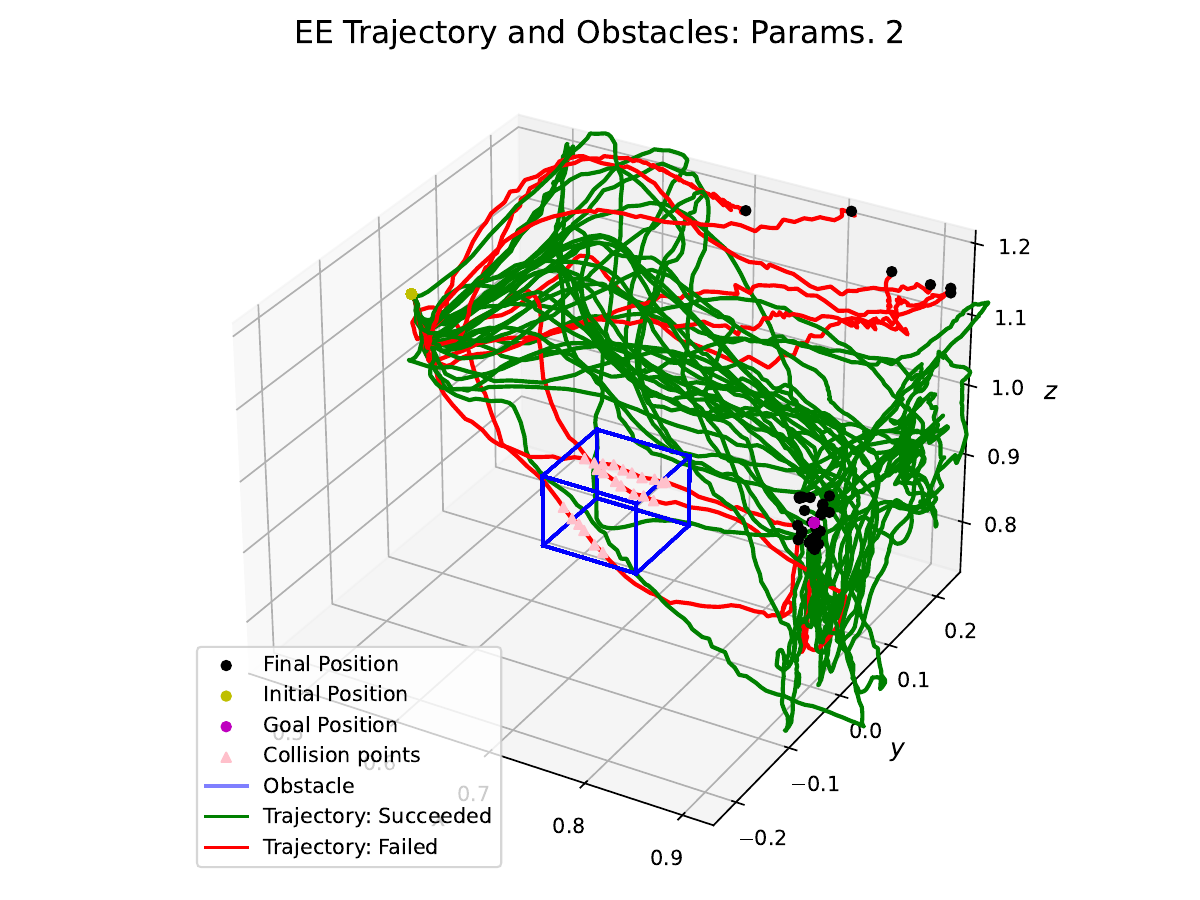}
        \caption{\centering Parameter Set 2}
        \label{tuning_scenario_1_trial_trajs_params1_to_params5:params2}
    \end{subfigure}%
    \begin{subfigure}{0.33\linewidth}
        \centering
        \includegraphics[width=\columnwidth, trim={90pt 10pt 83pt 25pt},clip]{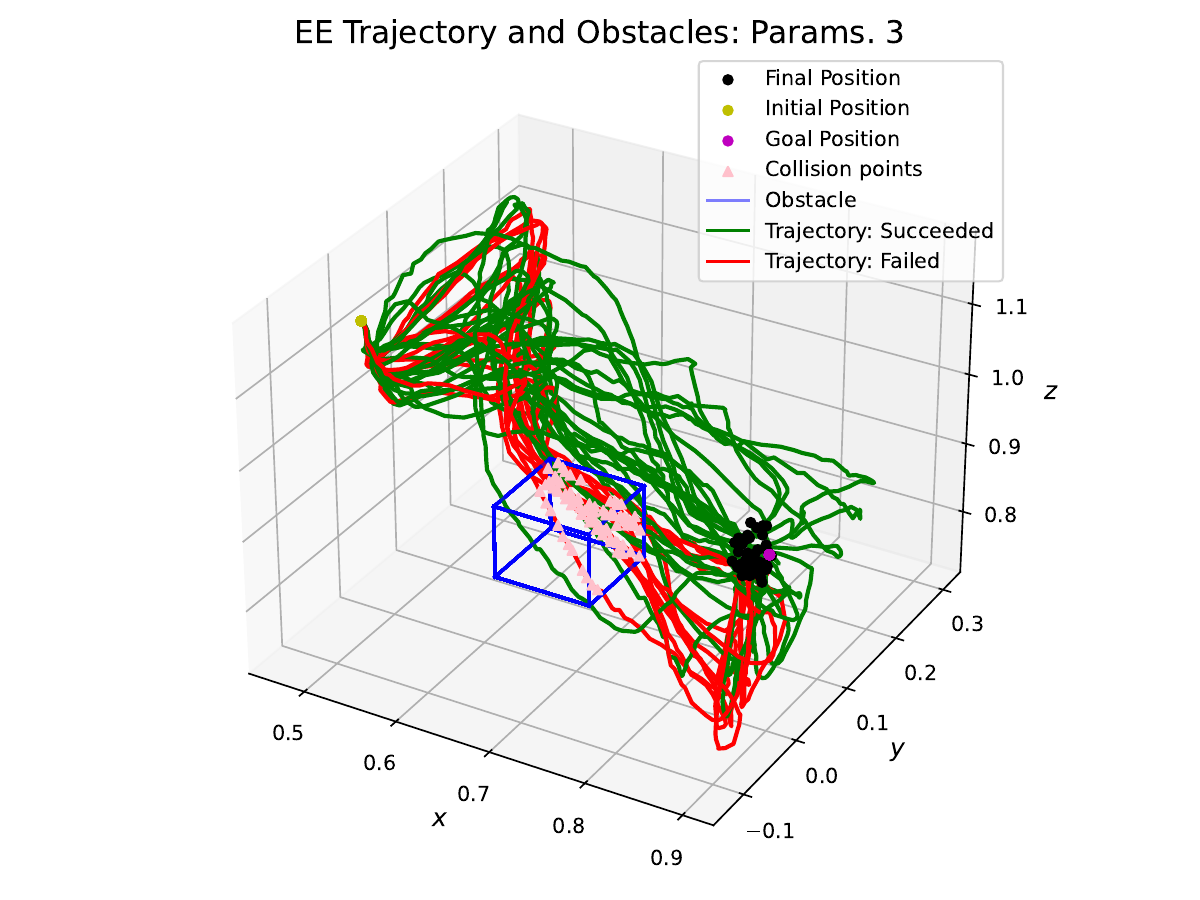}
        \caption{\centering Parameter Set 3}
        \label{tuning_scenario_1_trial_trajs_params1_to_params5:params3}
    \end{subfigure}%
    \\
    \begin{subfigure}{0.33\linewidth}
        \centering
        \includegraphics[width=\columnwidth, trim={90pt 10pt 83pt 25pt},clip]{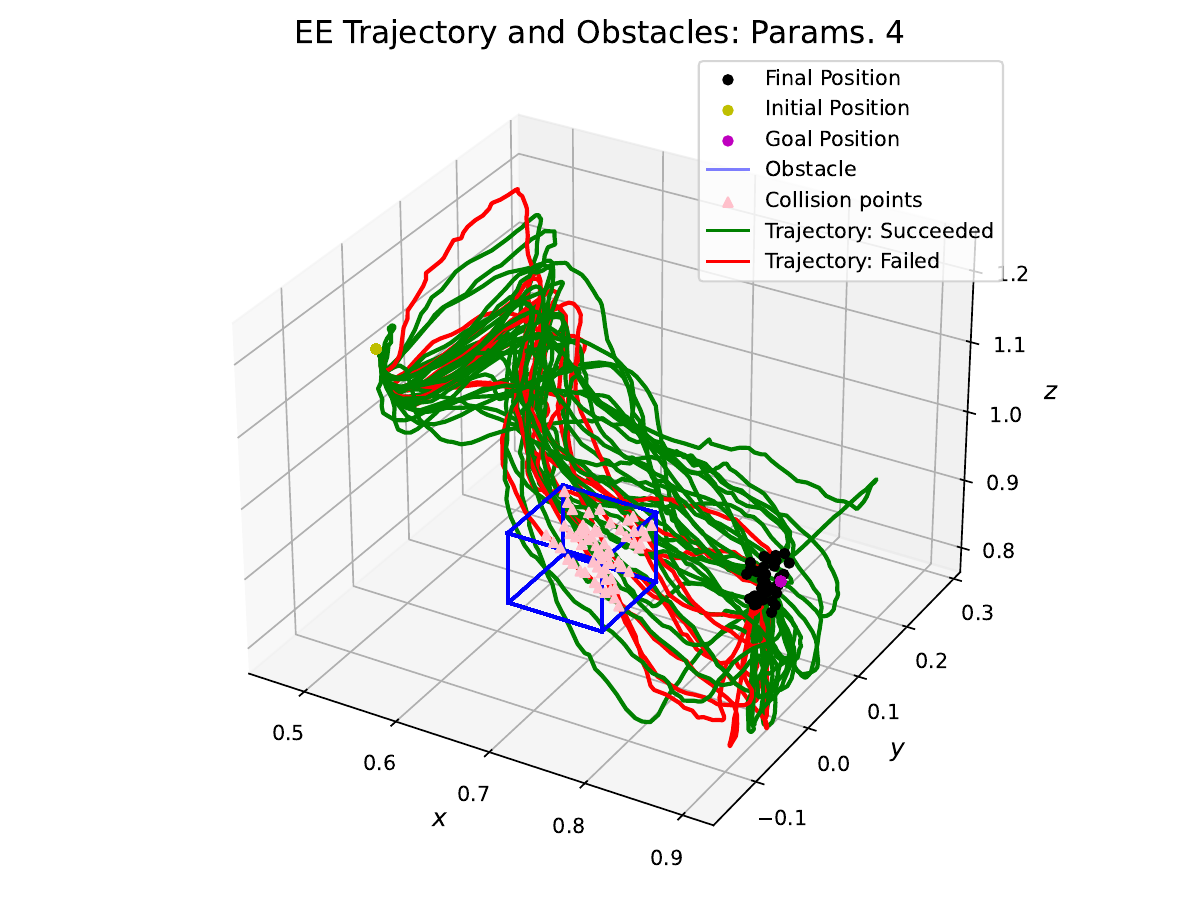}
        \caption{\centering Parameter Set 4}
        \label{tuning_scenario_1_trial_trajs_params1_to_params5:params4}
    \end{subfigure}%
    \begin{subfigure}{0.33\linewidth}
        \centering
        \includegraphics[width=\columnwidth, trim={90pt 10pt 83pt 25pt},clip]{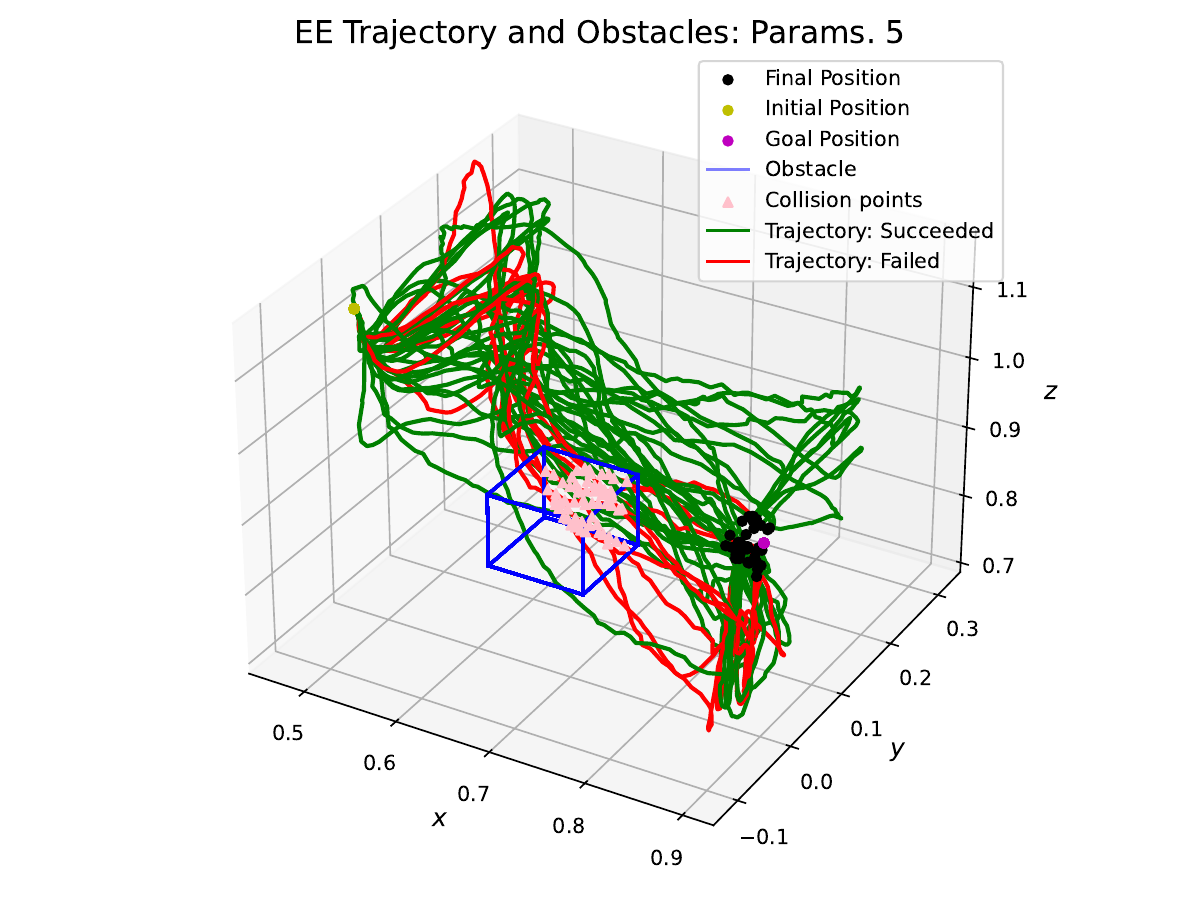}
        \caption{\centering Parameter Set 5}
        \label{tuning_scenario_1_trial_trajs_params1_to_params5:params5}
    \end{subfigure}%
    \begin{subfigure}{0.33\linewidth}
        \centering
        \includegraphics[width=\columnwidth, trim={90pt 10pt 83pt 25pt},clip]{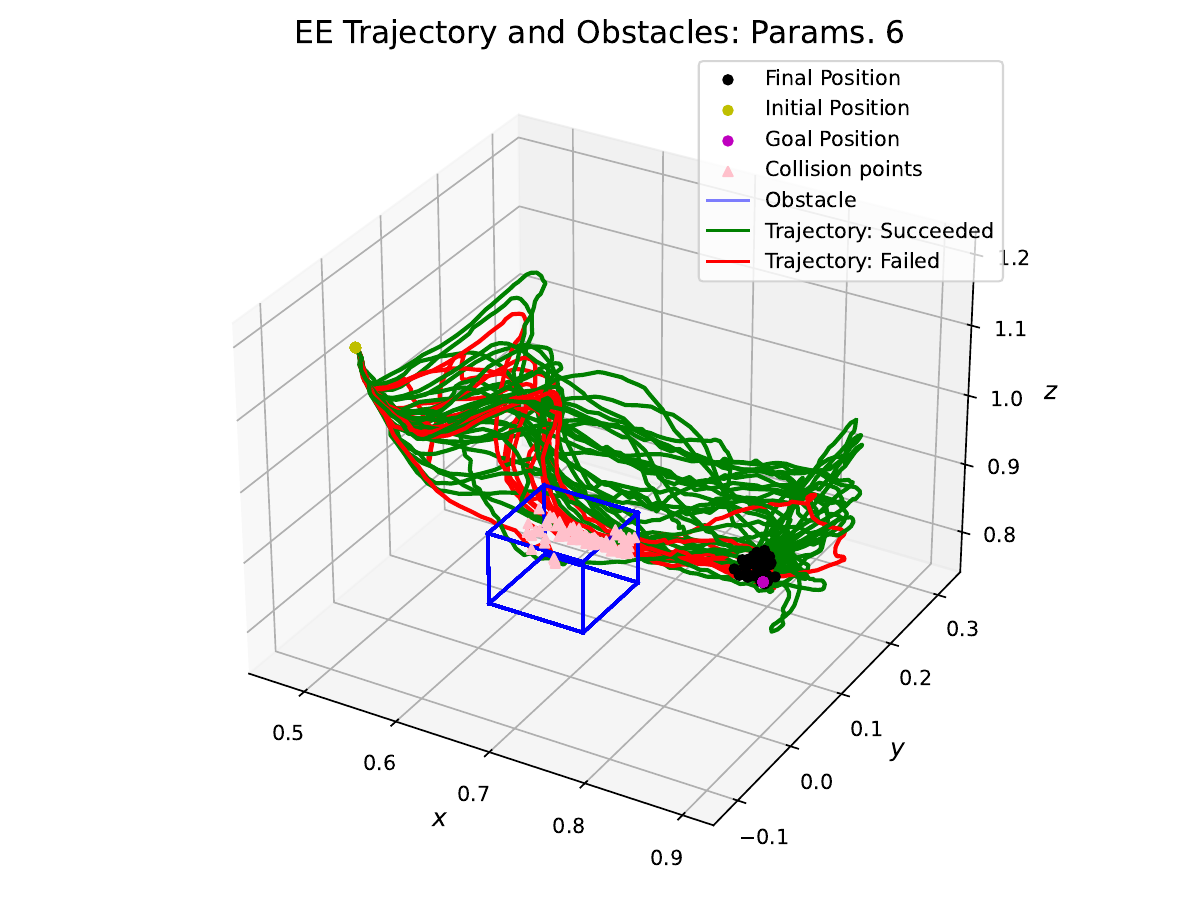}
        \caption{\centering Parameter Set 6}
        \label{tuning_scenario_1_trial_trajs_params6_to_params12:params6}
    \end{subfigure}%
    \\
    \begin{subfigure}{0.33\linewidth}
        \centering
        \includegraphics[width=\columnwidth, trim={90pt 10pt 83pt 25pt},clip]{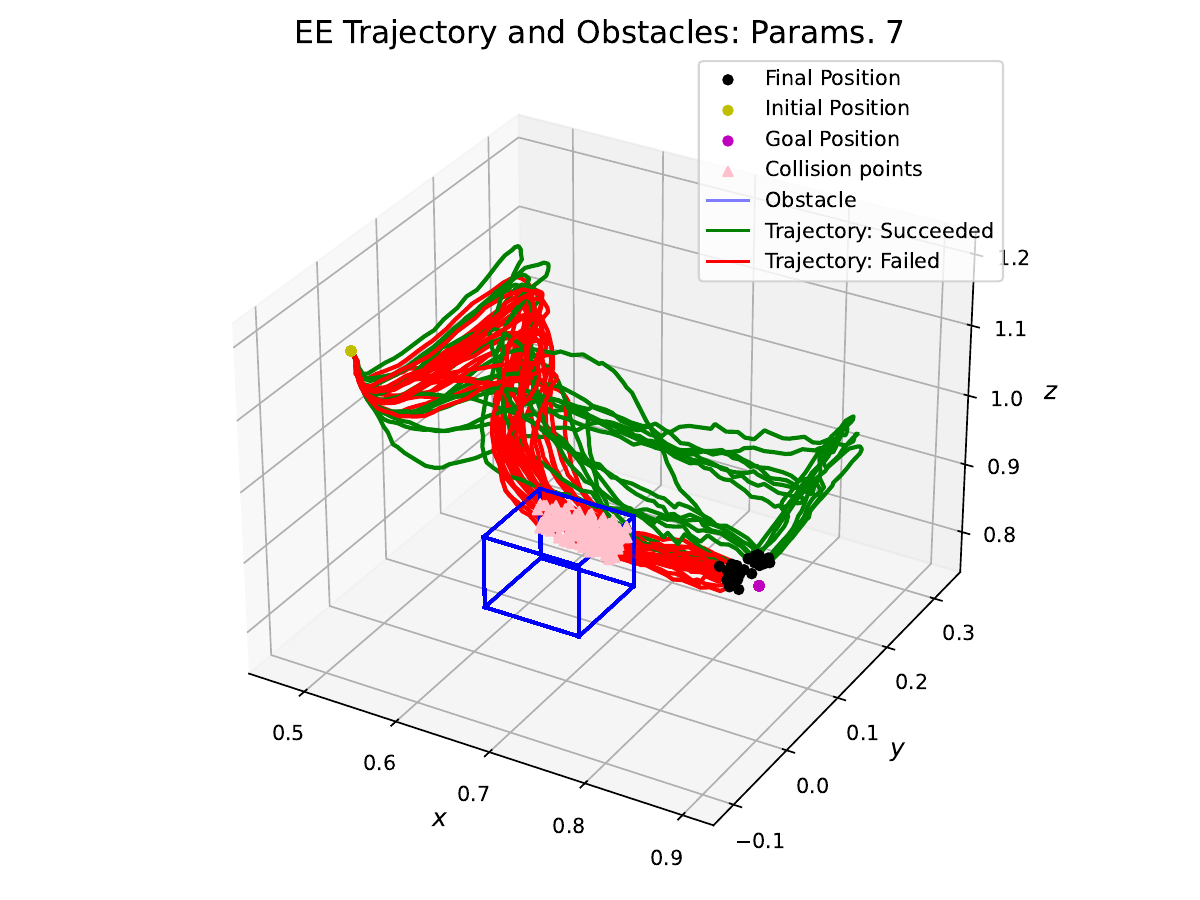}
        \caption{\centering Parameter Set 7}
        \label{tuning_scenario_1_trial_trajs_params6_to_params12:params7}
    \end{subfigure}%
    \begin{subfigure}{0.33\linewidth}
        \centering
        \includegraphics[width=\columnwidth, trim={90pt 10pt 83pt 25pt},clip]{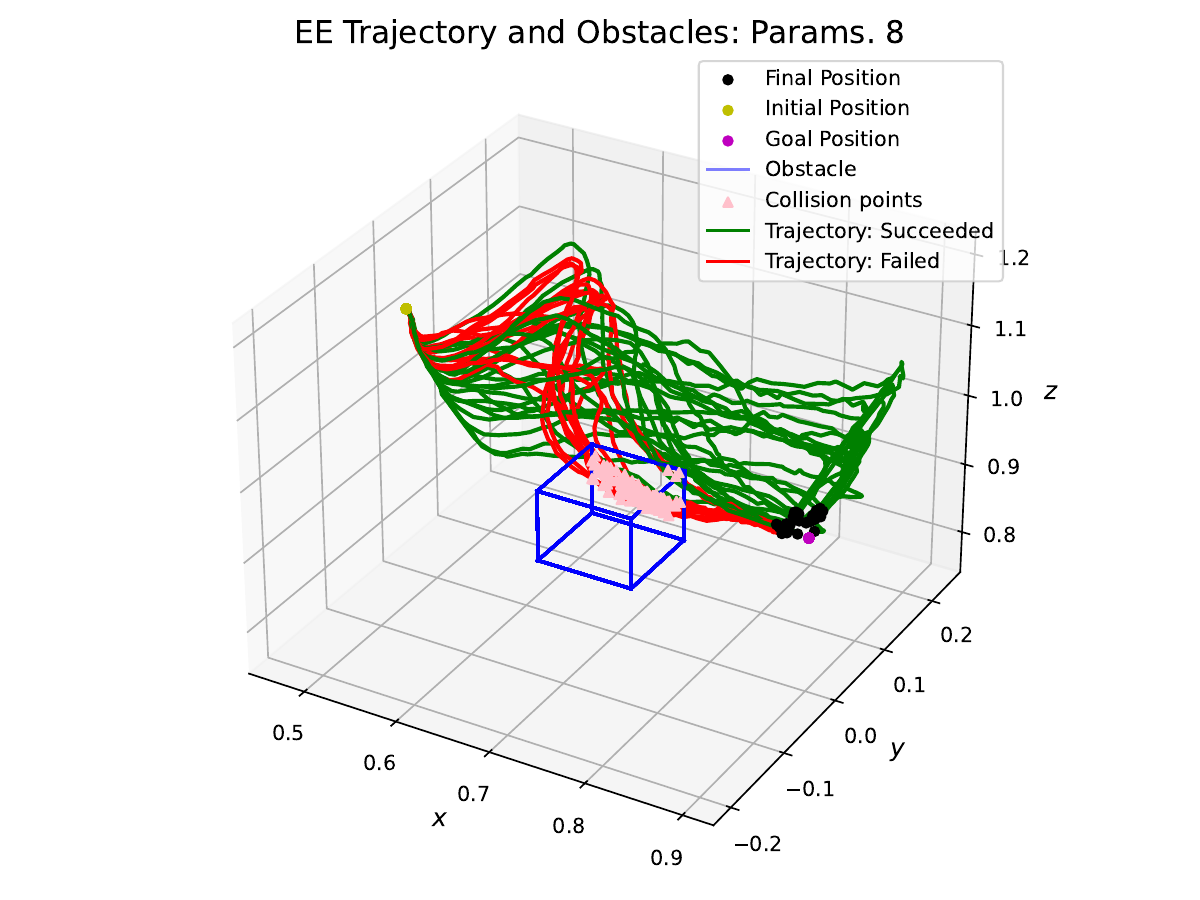}
        \caption{\centering Parameter Set 8}
        \label{tuning_scenario_1_trial_trajs_params6_to_params12:params8}
    \end{subfigure}%
    \begin{subfigure}{0.33\linewidth}
        \centering
        \includegraphics[width=\columnwidth, trim={90pt 10pt 83pt 25pt},clip]{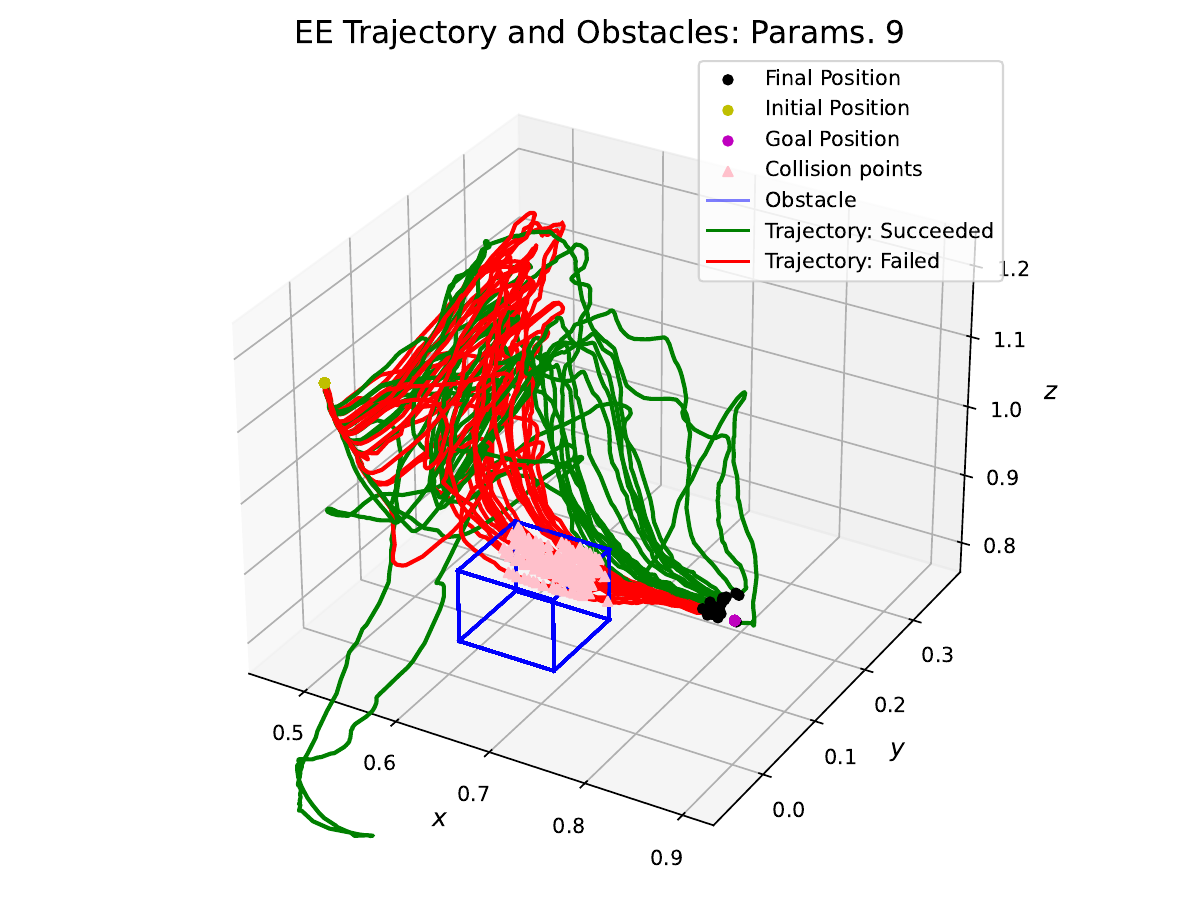}
        \caption{\centering Parameter Set 9}
        \label{tuning_scenario_1_trial_trajs_params6_to_params12:params9}
    \end{subfigure}%
    \\
    \begin{subfigure}{0.33\linewidth}
        \centering
        \includegraphics[width=\columnwidth, trim={90pt 10pt 83pt 25pt},clip]{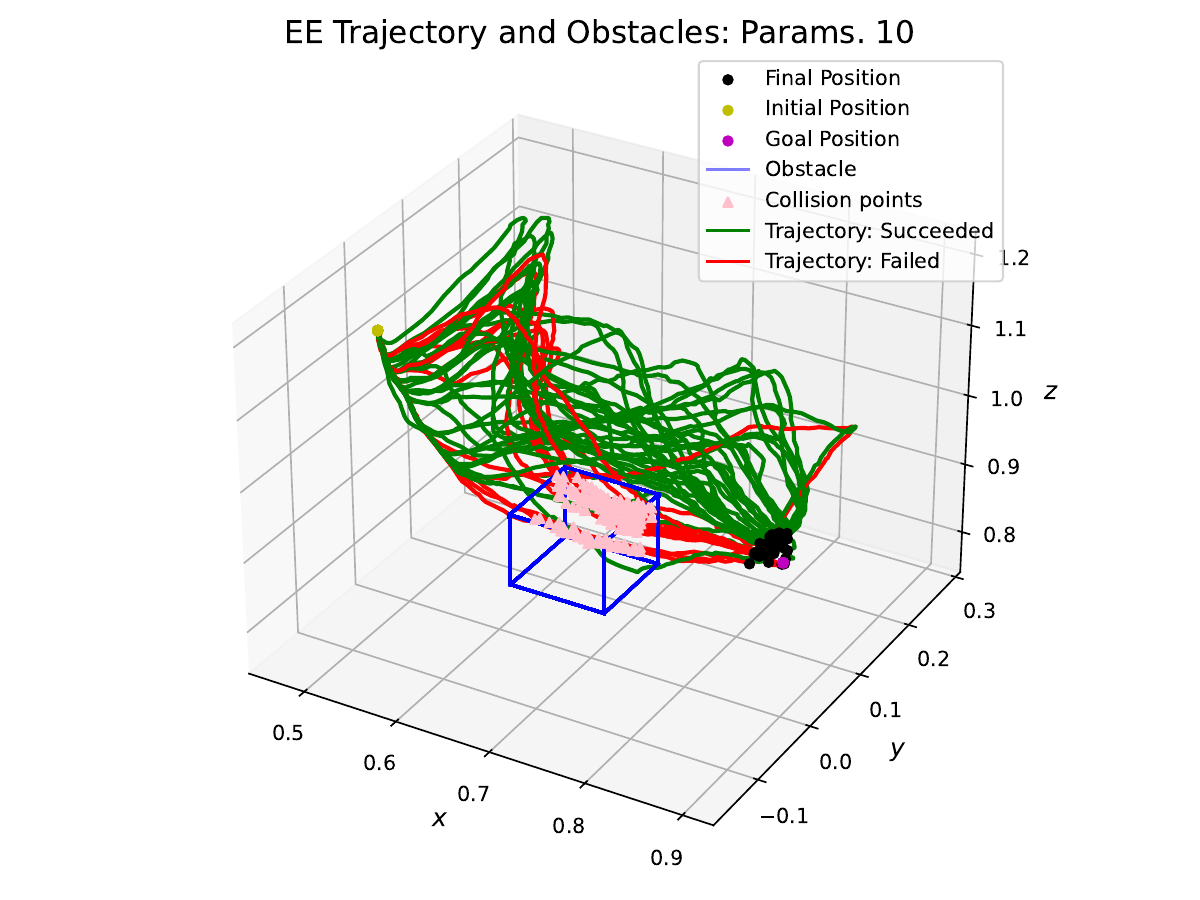}
        \caption{\centering Parameter Set 10}
        \label{tuning_scenario_1_trial_trajs_params6_to_params12:params10}
    \end{subfigure}%
    \begin{subfigure}{0.33\linewidth}
        \centering
        \includegraphics[width=\columnwidth, trim={90pt 10pt 83pt 25pt},clip]{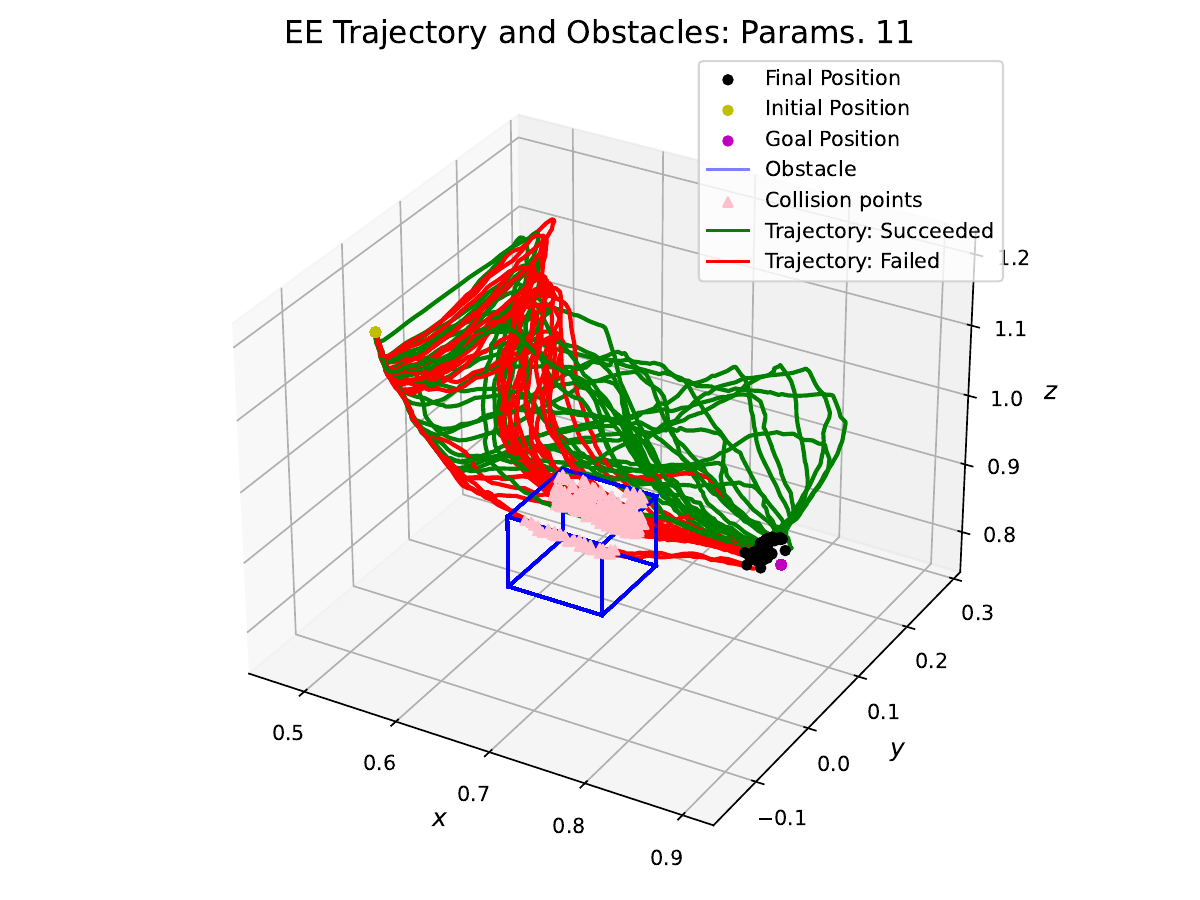}
        \caption{\centering Parameter Set 11}
        \label{tuning_scenario_1_trial_trajs_params6_to_params12:params11}
    \end{subfigure}%
    \begin{subfigure}{0.33\linewidth}
        \centering
        \includegraphics[width=\columnwidth, trim={90pt 10pt 83pt 25pt},clip]{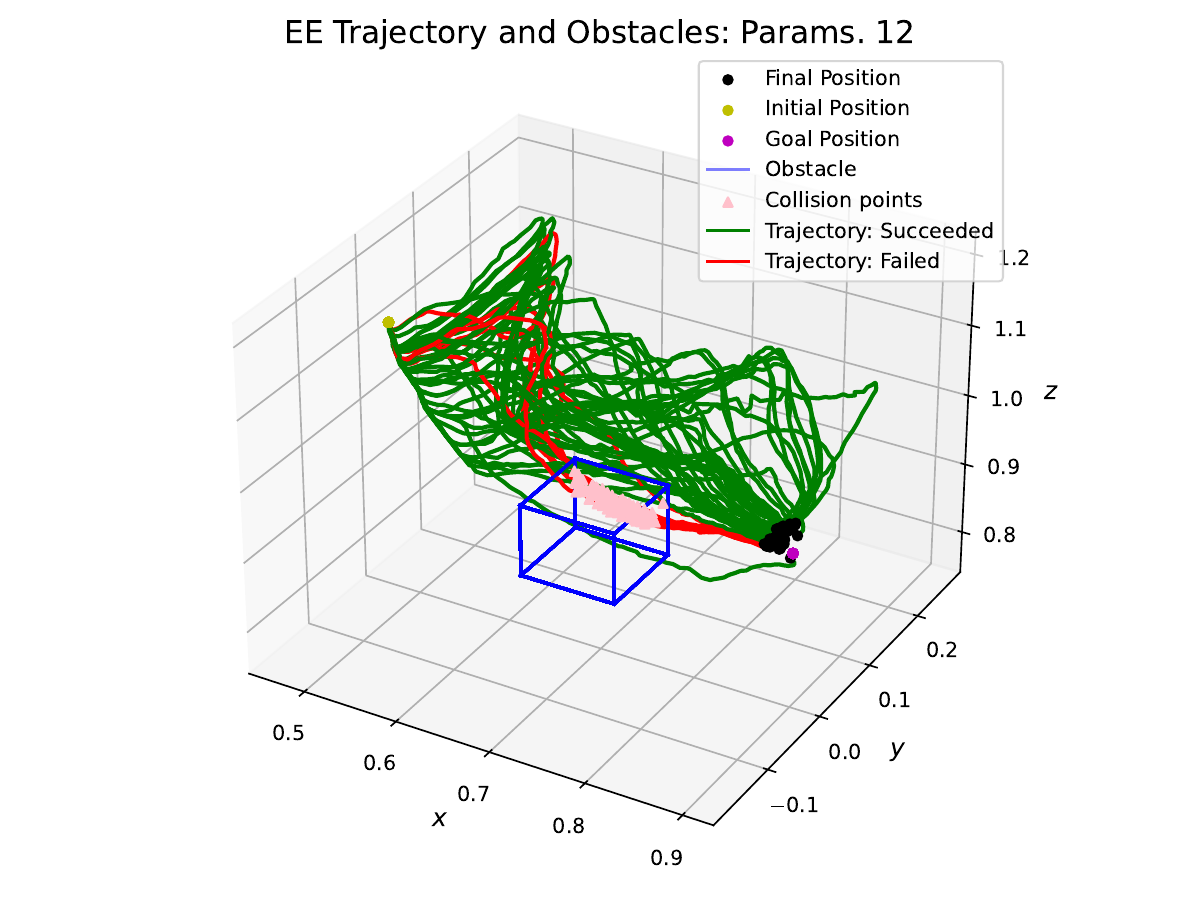}
        \caption{\centering Parameter Set 12}
        \label{tuning_scenario_1_trial_trajs_params6_to_params12:params12}
    \end{subfigure}%
    \caption{Trajectories executed in tuning scenario 1: parameter sets 1-12 (discussed in section \ref{tuning_results}).}
    \label{tuning_scenario_1_trial_trajs_params1_to_params12}
\end{figure}

\begin{figure}
    \centering
    \begin{subfigure}{0.33\linewidth}	
        \includegraphics[width=\columnwidth, trim={90pt 10pt 83pt 25pt},clip]{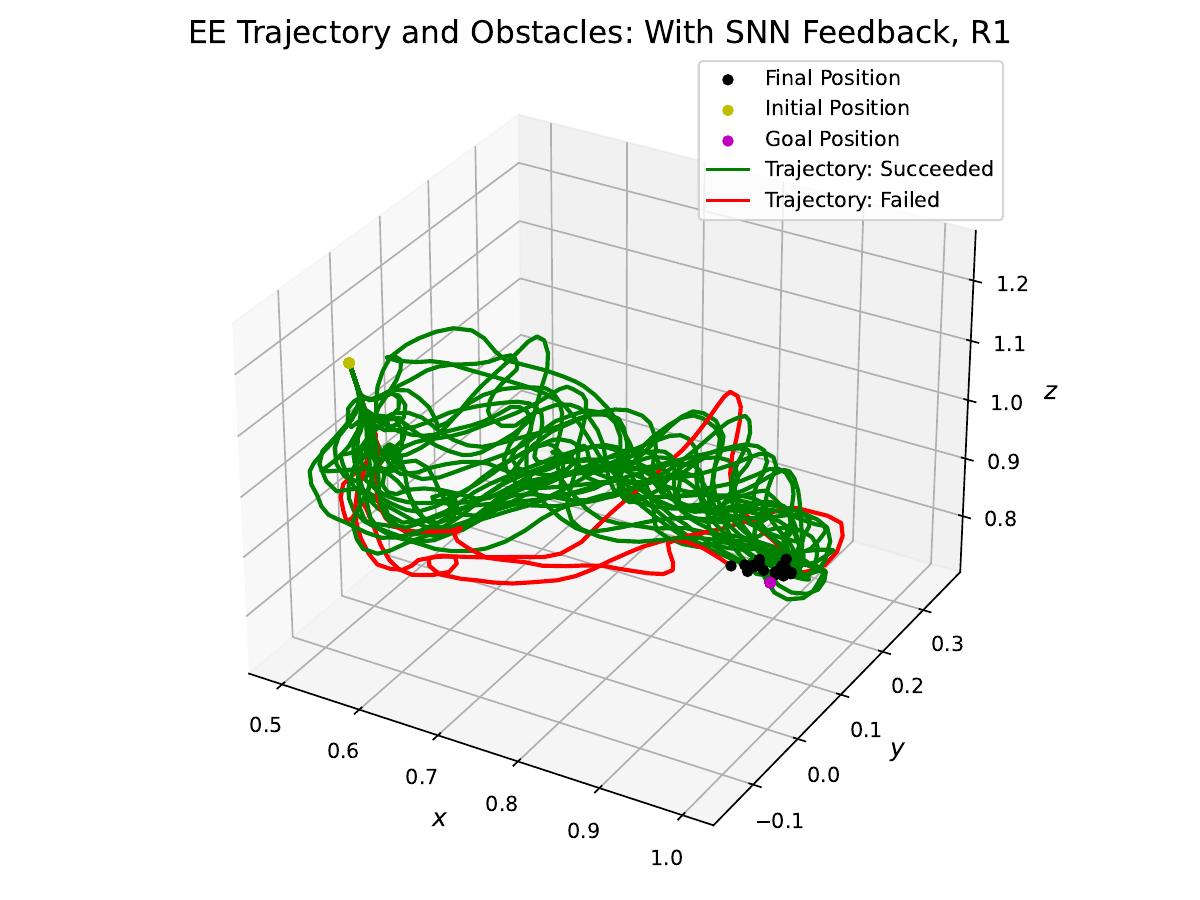}
        \caption{\centering Scenario R1}
        \label{real_robot_experiments_trial_trajs:scenario_R1}
    \end{subfigure}%
    \begin{subfigure}{0.33\linewidth}	
        \includegraphics[width=\columnwidth, trim={90pt 10pt 83pt 25pt},clip]{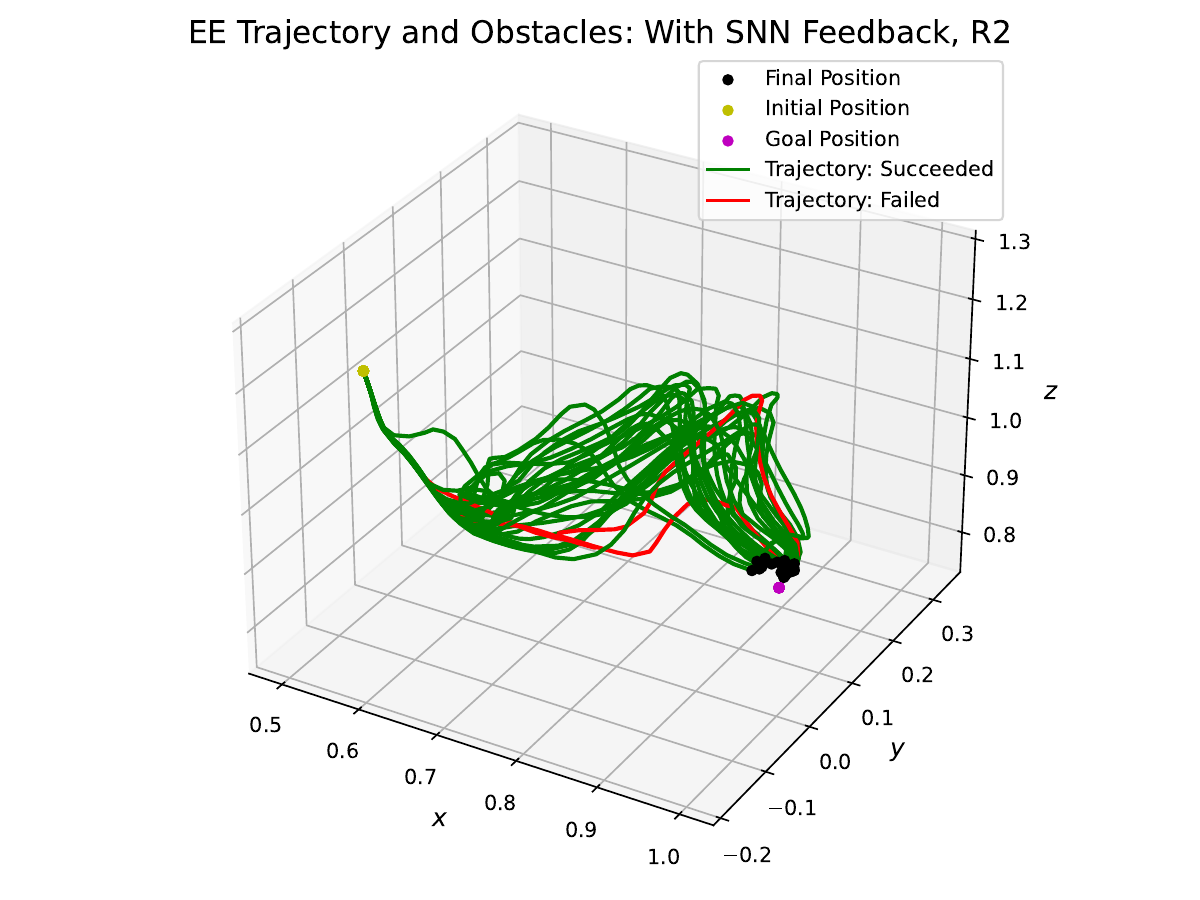}
        \caption{\centering Scenario R2}
        \label{real_robot_experiments_trial_trajs:scenario_R2}
    \end{subfigure}%
    \begin{subfigure}{0.33\linewidth}
        \centering
        \includegraphics[width=\columnwidth, trim={90pt 10pt 83pt 25pt},clip]{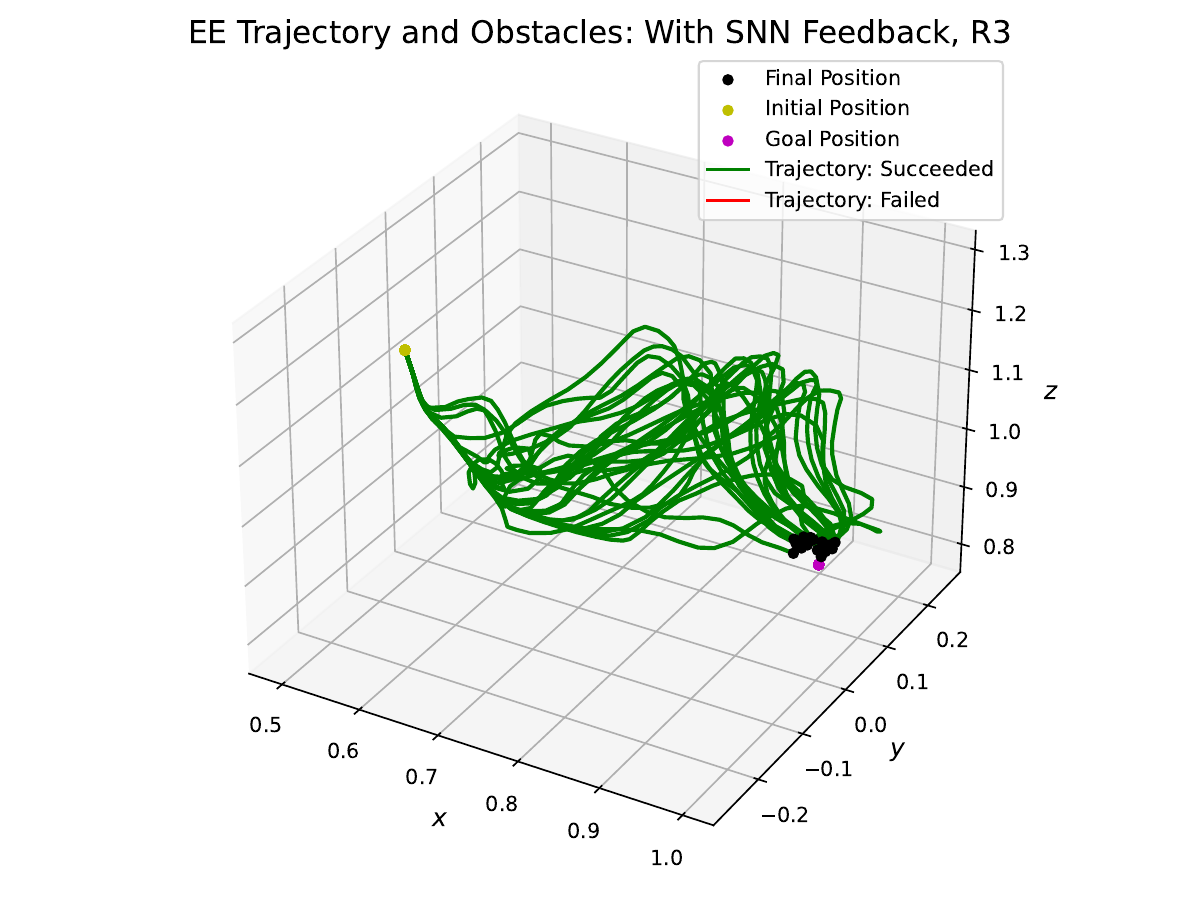}
        \caption{\centering Scenario R3}
        \label{real_robot_experiments_trial_trajs:scenario_R3}
    \end{subfigure}%
    \\
    \begin{subfigure}{0.33\linewidth}	
        \includegraphics[width=\columnwidth, trim={90pt 10pt 83pt 25pt},clip]{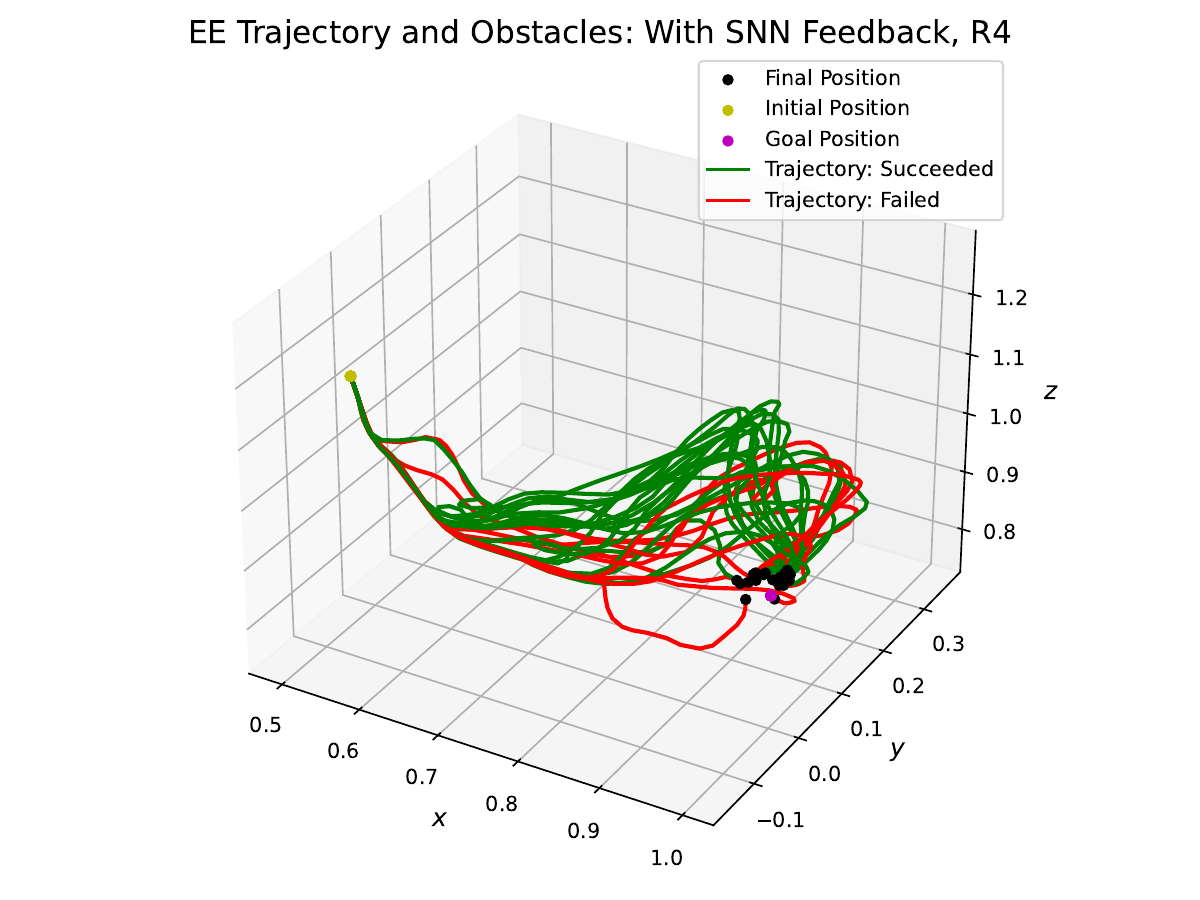}
        \caption{\centering Scenario R4}
        \label{real_robot_experiments_trial_trajs:scenario_R4}
    \end{subfigure}%
    \caption{Trajectories executed in real robot experiments with SNN feedback in scenarios R1-R4 (discussed in section \ref{results_and_discussion:real_experiments}).}
    \label{real_robot_experiments_trial_trajs}
\end{figure}

\begin{figure*}
    \centering
    \begin{subfigure}{0.49\linewidth}
        \centering
        \frame{
            \includegraphics[width=\columnwidth]{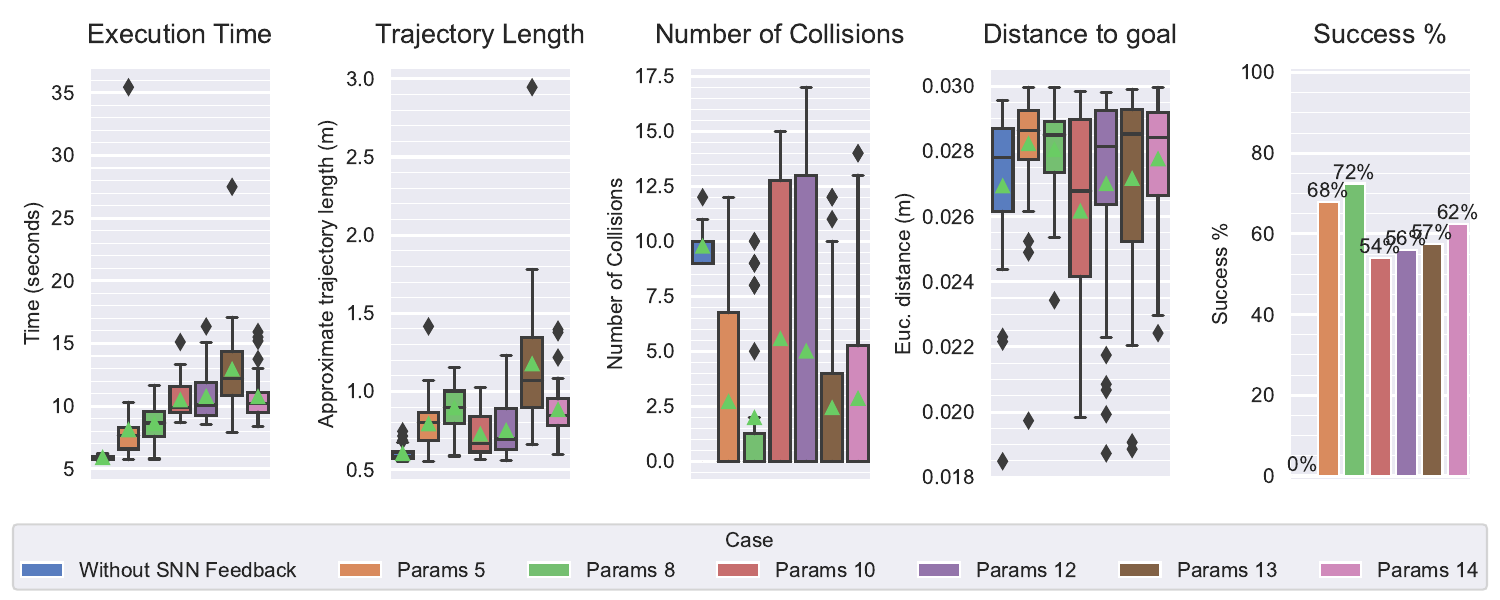}
        }
        \caption{\centering Scenario 4}
        \label{validation_scenario_metrics:scenario_4}
    \end{subfigure}%
    \hspace{0.5em}
    \begin{subfigure}{0.49\linewidth}
        \centering
        \frame{
            \includegraphics[width=\columnwidth]{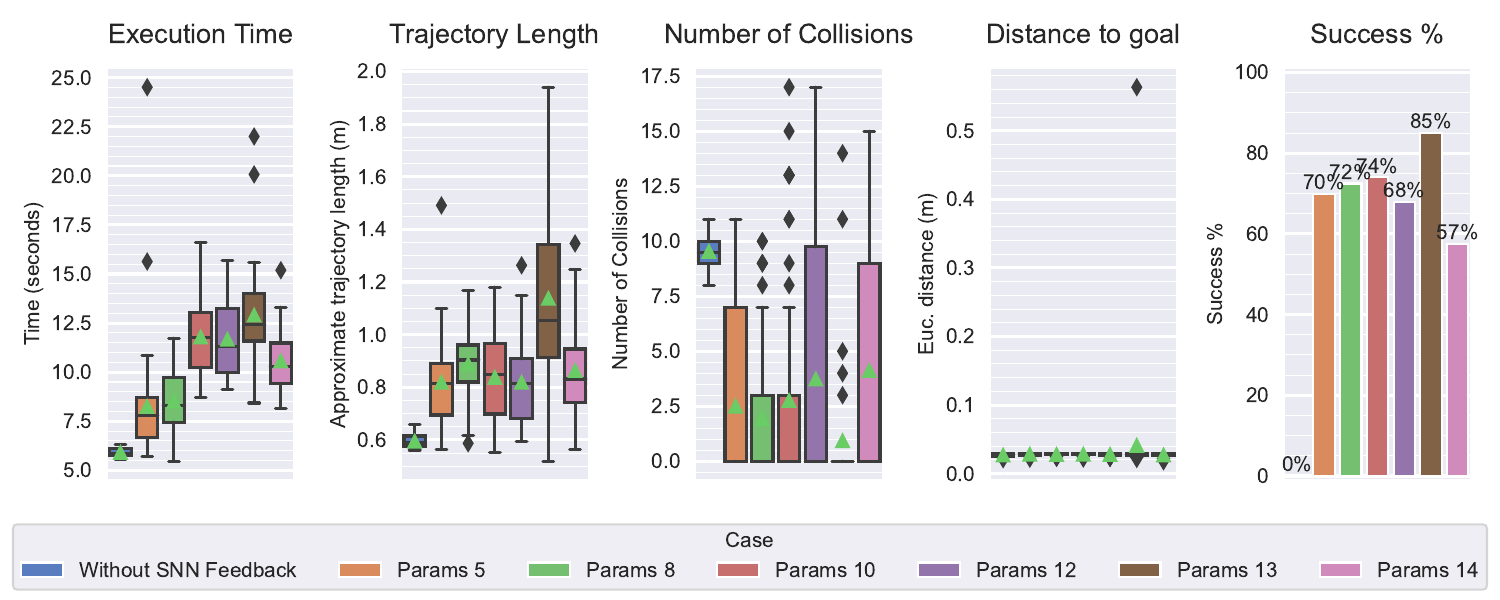}
        }
        \caption{\centering Scenario 5}
        \label{validation_scenarios_metrics:scenario_5}
    \end{subfigure}%
    \\
    \begin{subfigure}{0.49\linewidth}
        \centering
        \frame{
            \includegraphics[width=\columnwidth]{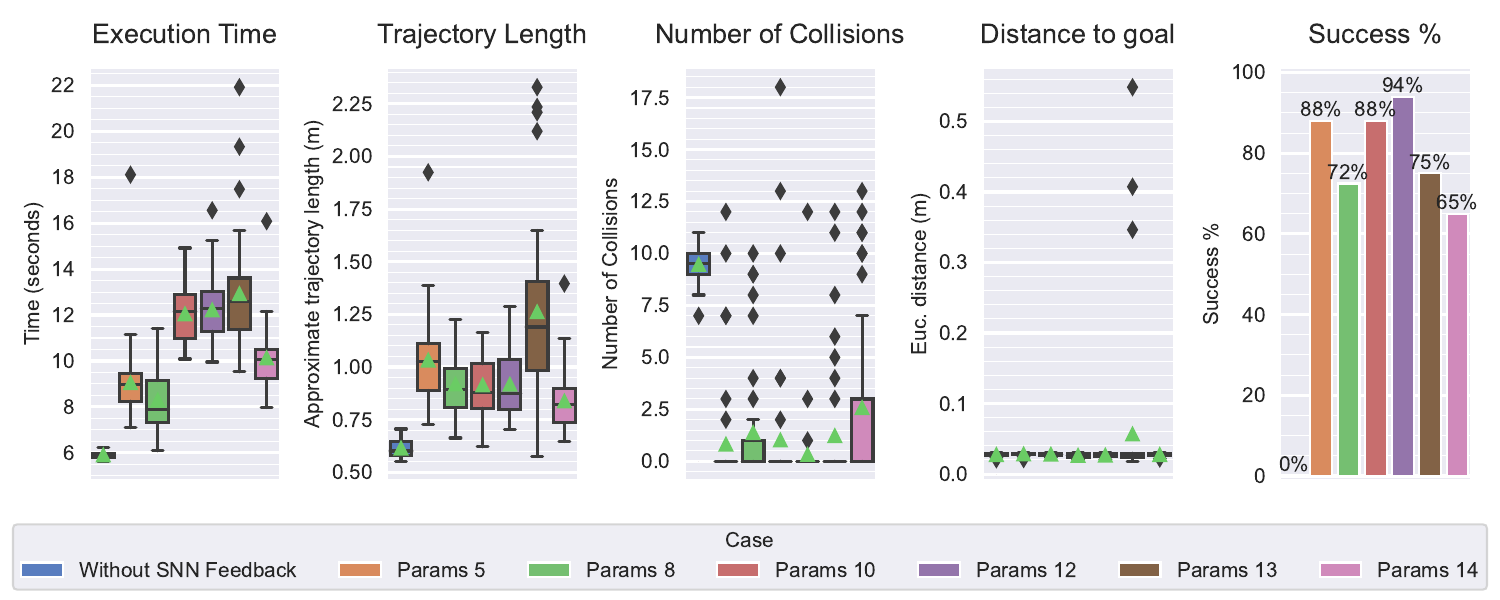}
        }
        \caption{\centering Scenario 6}
        \label{validation_scenarios_metrics:scenario_6}
    \end{subfigure}%
    \hspace{0.5em}
    \begin{subfigure}{0.49\linewidth}
        \centering
        \frame{
            \includegraphics[width=\columnwidth]{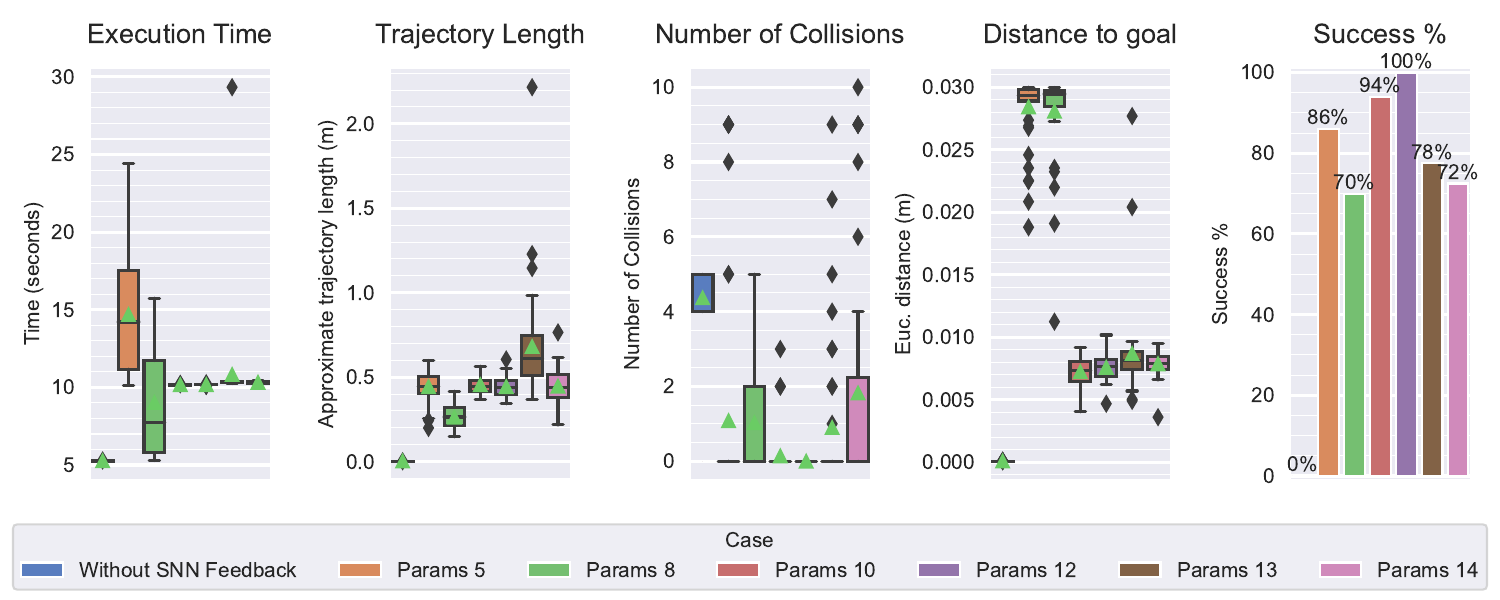}
        }
        \caption{\centering Scenario 7}
        \label{validation_scenarios_metrics:scenario_7}
    \end{subfigure}%
    \\
    \begin{subfigure}{0.49\linewidth}
        \centering
        \frame{
            \includegraphics[width=\columnwidth]{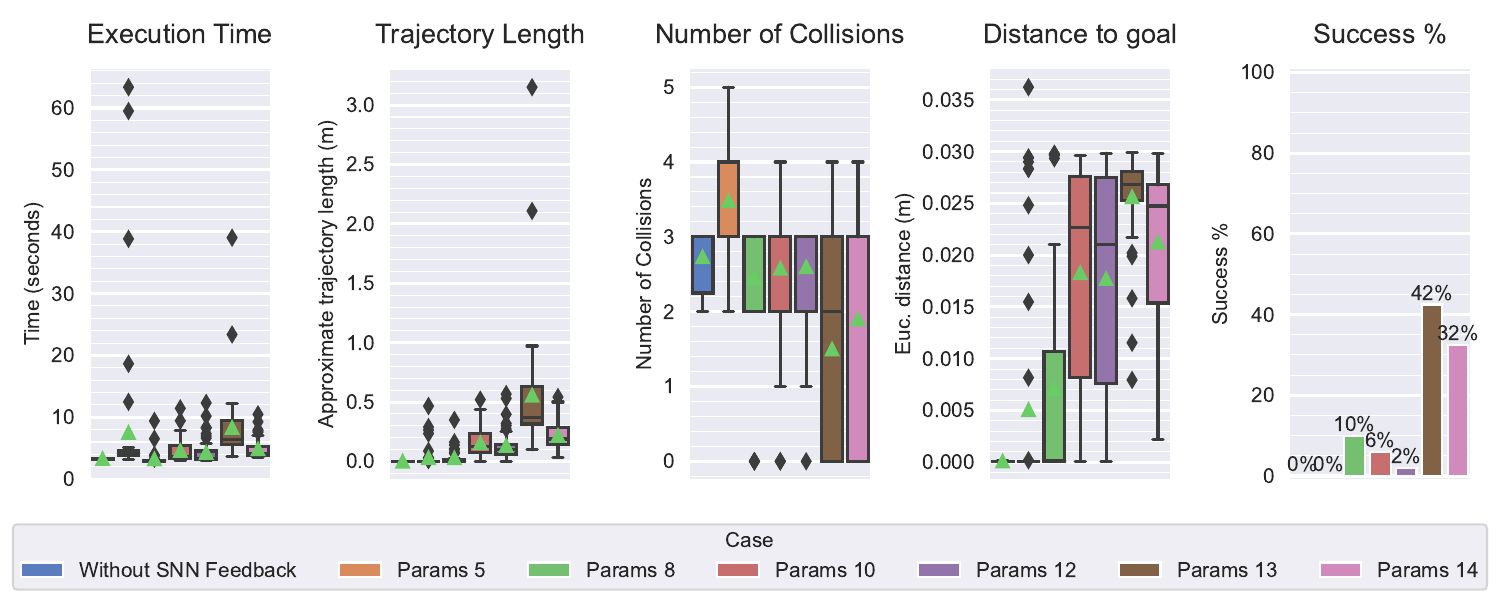}
        }
        \caption{\centering Scenario 8}
        \label{validation_scenarios_metrics:scenario_8}
    \end{subfigure}%
    \hspace{0.5em}
    \begin{subfigure}{0.49\linewidth}
        \centering
        \frame{
            \includegraphics[width=\columnwidth]{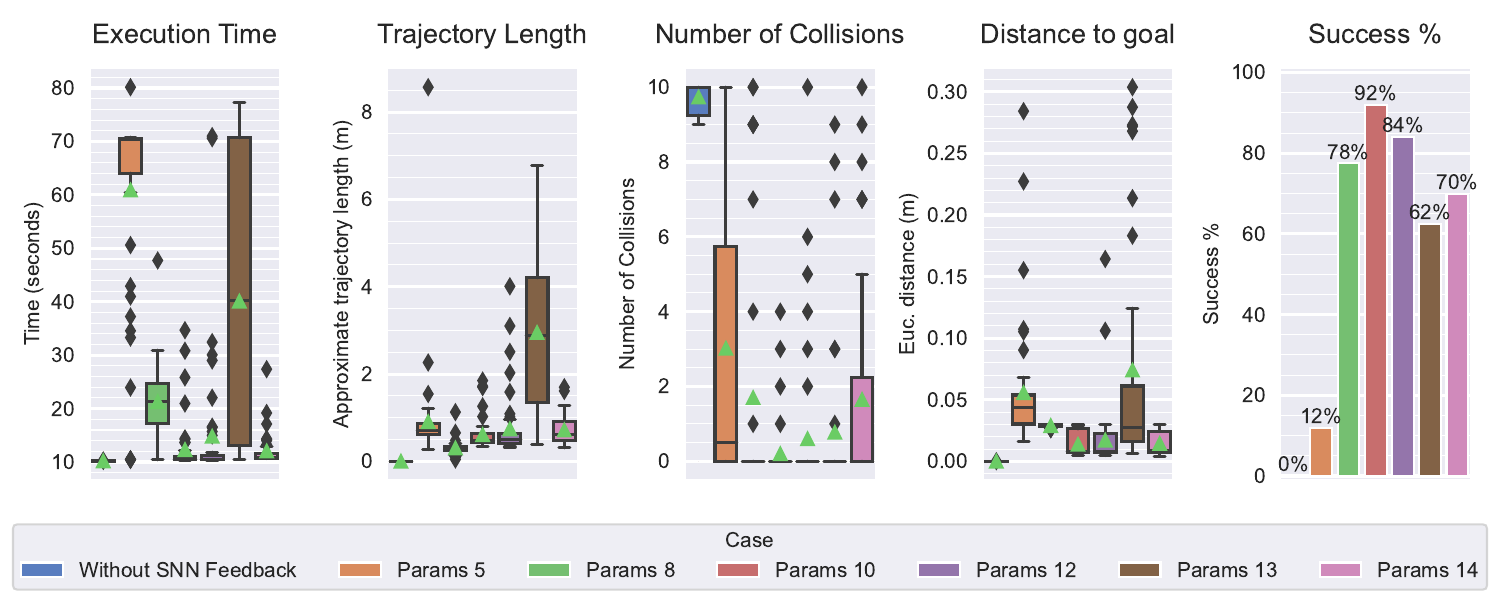}
        }
        \caption{\centering Scenario 9}
        \label{validation_scenarios_metrics:scenario_9}
    \end{subfigure}%
    \\
    \begin{subfigure}{0.49\linewidth}
        \centering
        \frame{
            \includegraphics[width=\columnwidth]{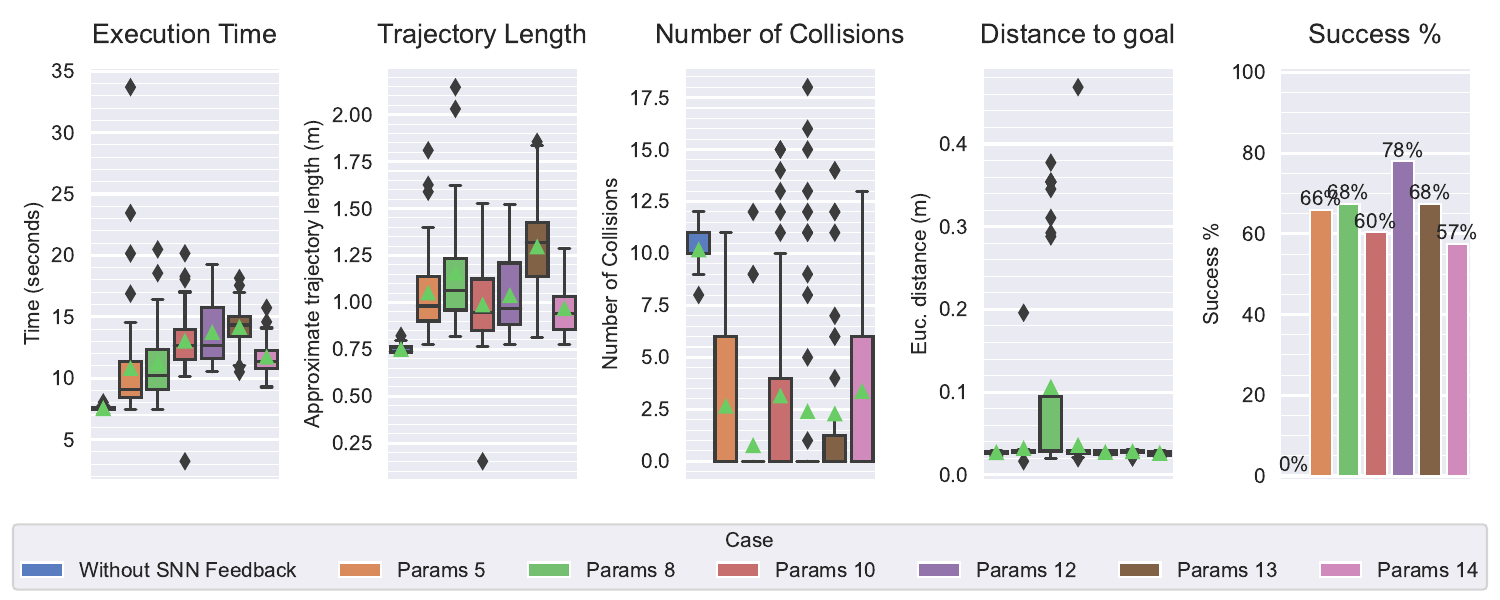}
        }
        \caption{\centering Scenario 10}
        \label{validation_scenarios_metrics:scenario_10}
    \end{subfigure}%
    \hspace{0.5em}
    \begin{subfigure}{0.49\linewidth}
        \centering
        \frame{
            \includegraphics[width=\columnwidth]{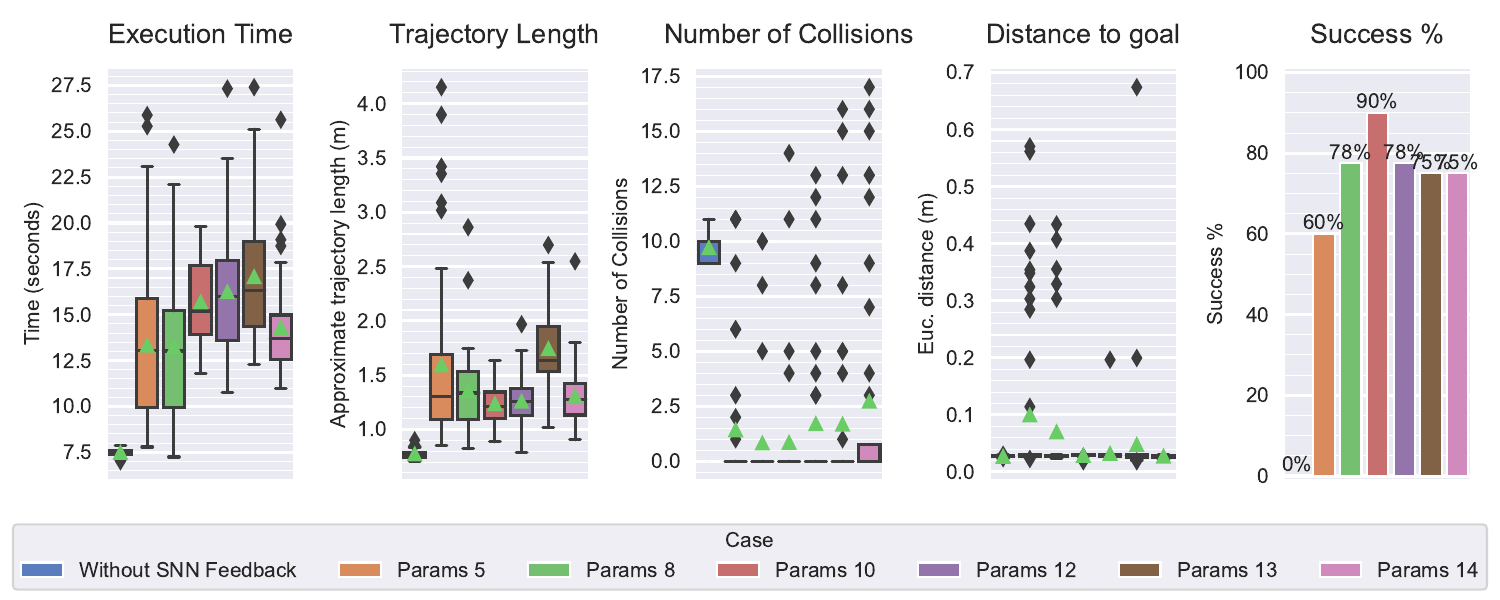}
        }
        \caption{\centering Scenario 11}
        \label{validation_scenarios_metrics:scenario_11}
    \end{subfigure}%
    \caption{Quantitative metric results for validation scenarios 4-11: without SNN feedback and with selected parameter sets.}
    \label{validation_scenarios_metrics}
\end{figure*}

%\paragraph{}
This appendix contains quantitative metric results and trajectory plots that are referred to in discussions of the tuning, validation, and testing phases of the simulation experiments in sections \ref{tuning_results}, \ref{validation_results}, and \ref{testing_results}:
\begin{itemize}
    \item Tuning phase trajectories executed in scenario 1 (parameter sets 1-12): Figure \ref{tuning_scenario_1_trial_trajs_params1_to_params12}
    \item Real experiment trajectories executed in scenarios R1-R4: Figure \ref{real_robot_experiments_trial_trajs}
%    \item Tuning phase trajectories from scenario 1 (parameter sets 1-12): Figure \ref{tuning_scenario_1_trial_trajs_params1_to_params12}
%    \item Real experiment trajectories from scenarios R1-R4: Figure \ref{real_robot_experiments_trial_trajs}
    \item Validation phase metrics in scenarios 4-11 (selected parameter sets): Figure \ref{validation_scenarios_metrics}
    \item Testing phase metrics in scenarios 12-31 (selected parameter set): Figure \ref{testing_scenarios_12_to_31_metrics}
\end{itemize}

%% Using the following clip params: 90pt 10pt 83pt 25pt

\begin{figure*}
    \centering
    \begin{subfigure}{0.49\linewidth}
        \centering
        \frame{
            \includegraphics[width=\columnwidth]{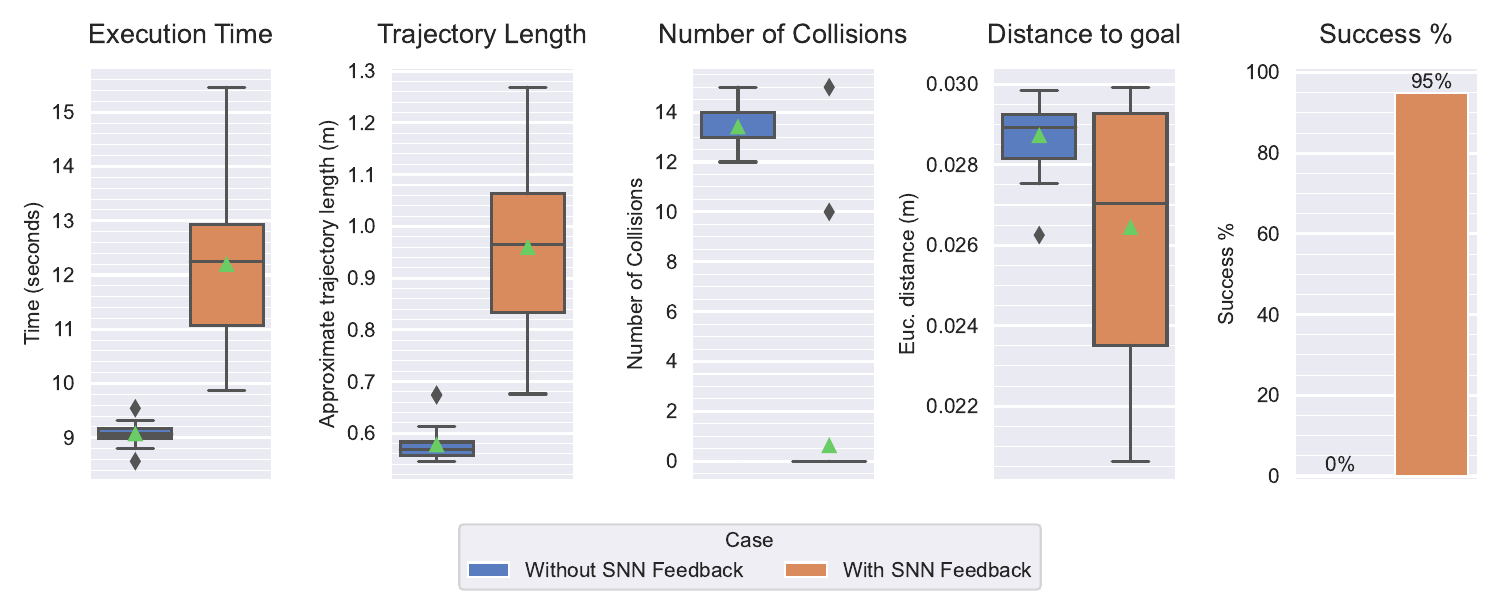}
        }
        \caption{\centering Scenario 12: \{T1, Empty, Yellow, Spiky Sphere\}}
        \label{testing_scenarios_12_to_31_metrics:scenario_12}
    \end{subfigure}%
    \hspace{0.5em}
    \begin{subfigure}{0.49\linewidth}
        \centering
        \frame{
            \includegraphics[width=\columnwidth]{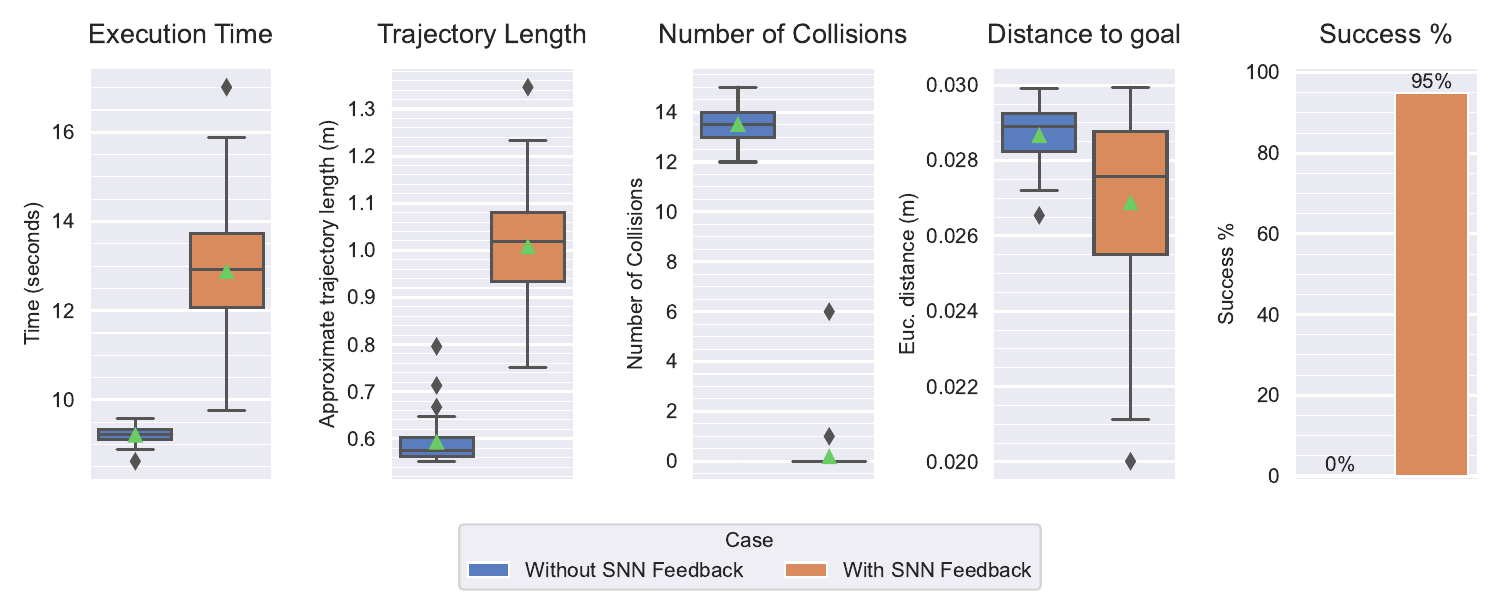}
        }
        \caption{\centering Scenario 13: \{T1, Store, Brick, Buckyball\}}
        \label{testing_scenarios_12_to_31_metrics:scenario_13}
    \end{subfigure}%
    \\
%    \vspace{1em}
    \begin{subfigure}{0.49\linewidth}
        \centering
        \frame{
            \includegraphics[width=\columnwidth]{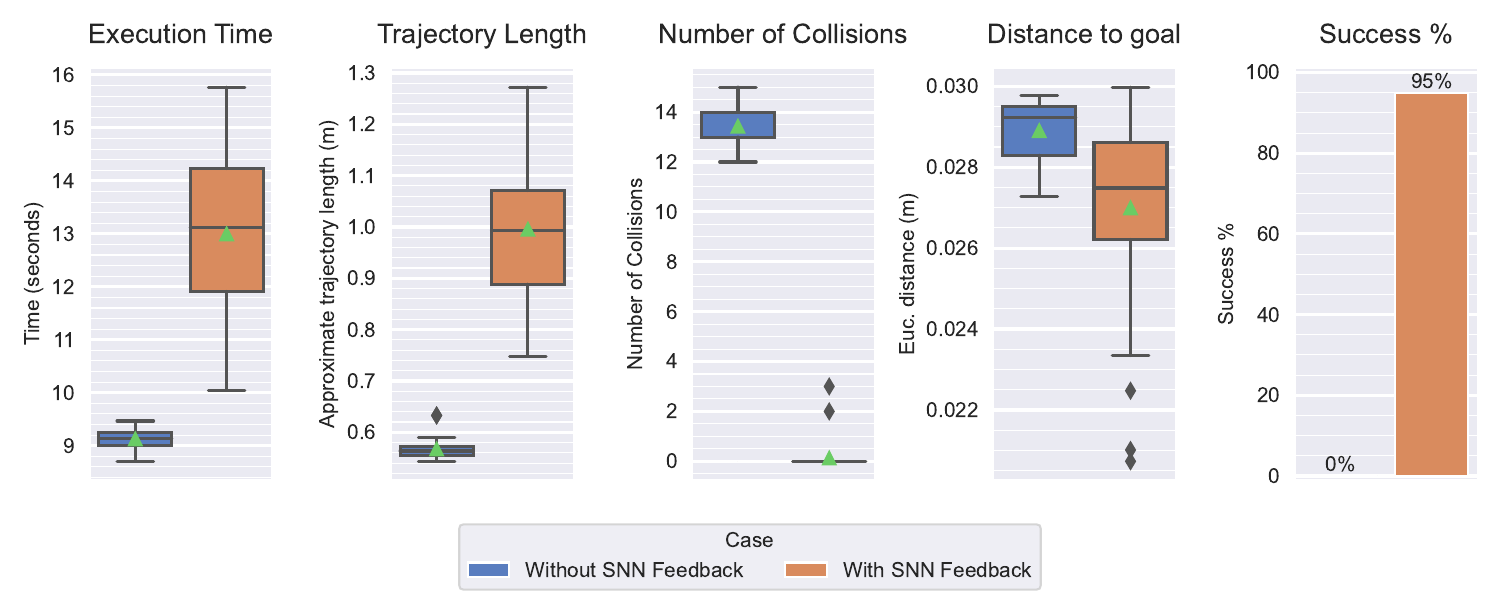}
        }
        \caption{\centering Scenario 14: \{T1, Kitchen, White, Buckyball\}}
        \label{testing_scenarios_12_to_31_metrics:scenario_14}
    \end{subfigure}%
    \hspace{0.5em}
    \begin{subfigure}{0.49\linewidth}
        \centering
        \frame{
            \includegraphics[width=\columnwidth]{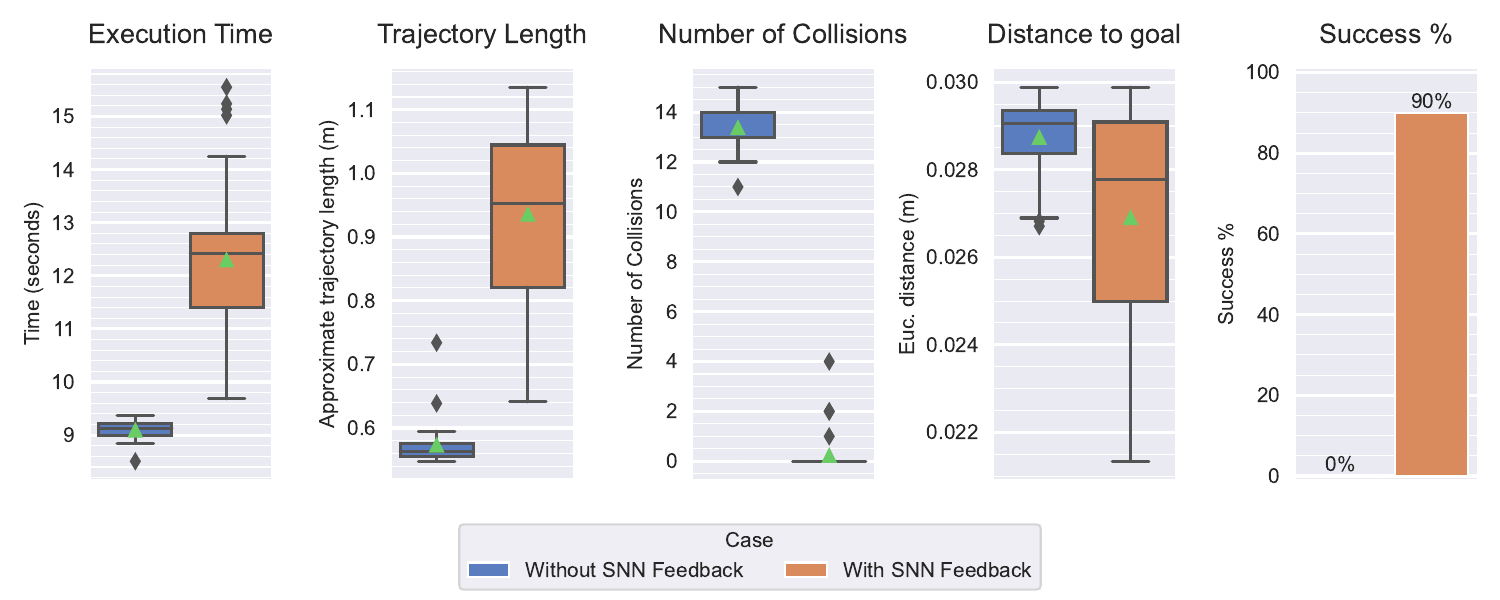}
        }
        \caption{\centering Scenario 15: \{T1, Empty, Yellow-Black, Box\}}
        \label{testing_scenarios_12_to_31_metrics:scenario_15}
    \end{subfigure}%
    \caption{Quantitative metric results without vs. with SNN feedback (best parameter set): testing scenarios 12-15.}
    \label{testing_scenarios_12_to_31_metrics}
\end{figure*}

\begin{figure*}\ContinuedFloat
    \centering
    \begin{subfigure}{0.49\linewidth}
        \centering
        \frame{
            \includegraphics[width=\columnwidth]{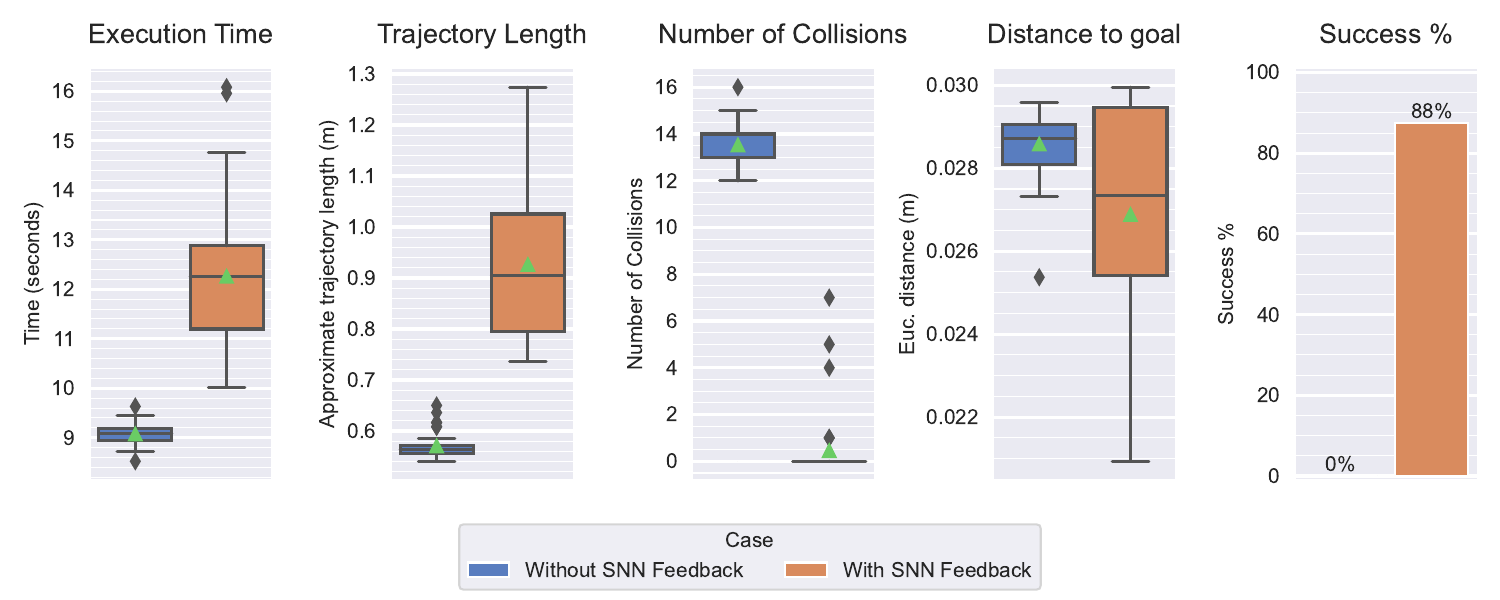}
        }
        \caption{\centering Scenario 16: \{T1, Store, Y-B, Rock\}}
        \label{testing_scenarios_12_to_31_metrics:scenario_16}
    \end{subfigure}%
    \hspace{0.5em}
    \begin{subfigure}{0.49\linewidth}
        \centering
        \frame{
            \includegraphics[width=\columnwidth]{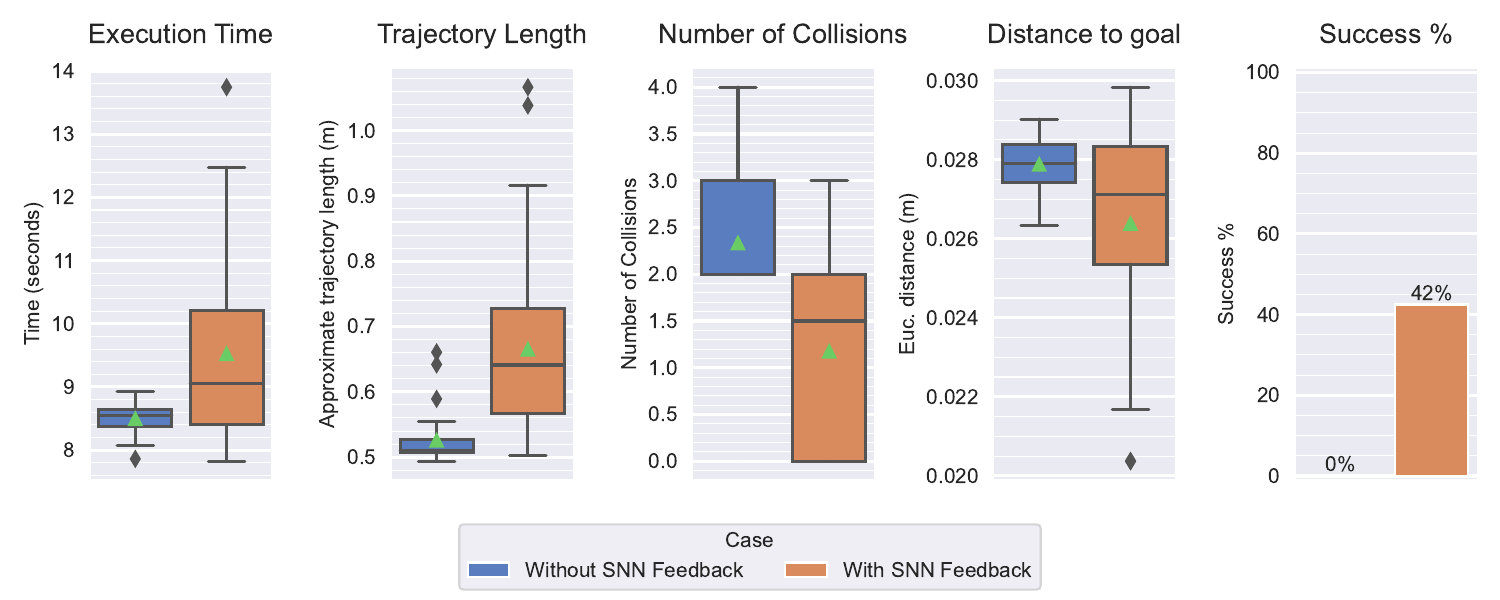}
        }
        \caption{\centering Scenario 17: \{T2, Office, Red, Buckyball, High\}}
        \label{testing_scenarios_12_to_31_metrics:scenario_17}
    \end{subfigure}%
    \\
    \begin{subfigure}{0.49\linewidth}
        \centering
        \frame{
            \includegraphics[width=\columnwidth]{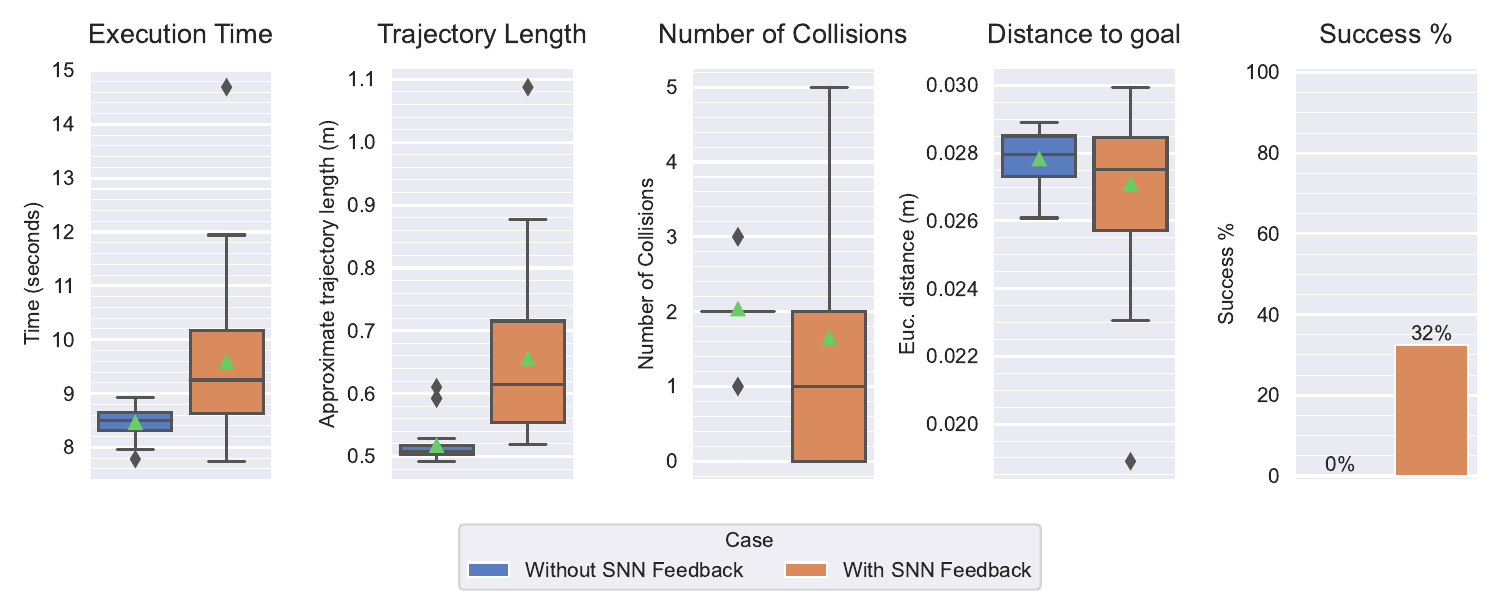}
        }
        \caption{\centering Scenario 18: \{T2, Office, Red, Rock, Med.\}}
        \label{testing_scenarios_12_to_31_metrics:scenario_18}
    \end{subfigure}%
    \hspace{0.5em}
    \begin{subfigure}{0.49\linewidth}
        \centering
        \frame{
            \includegraphics[width=\columnwidth]{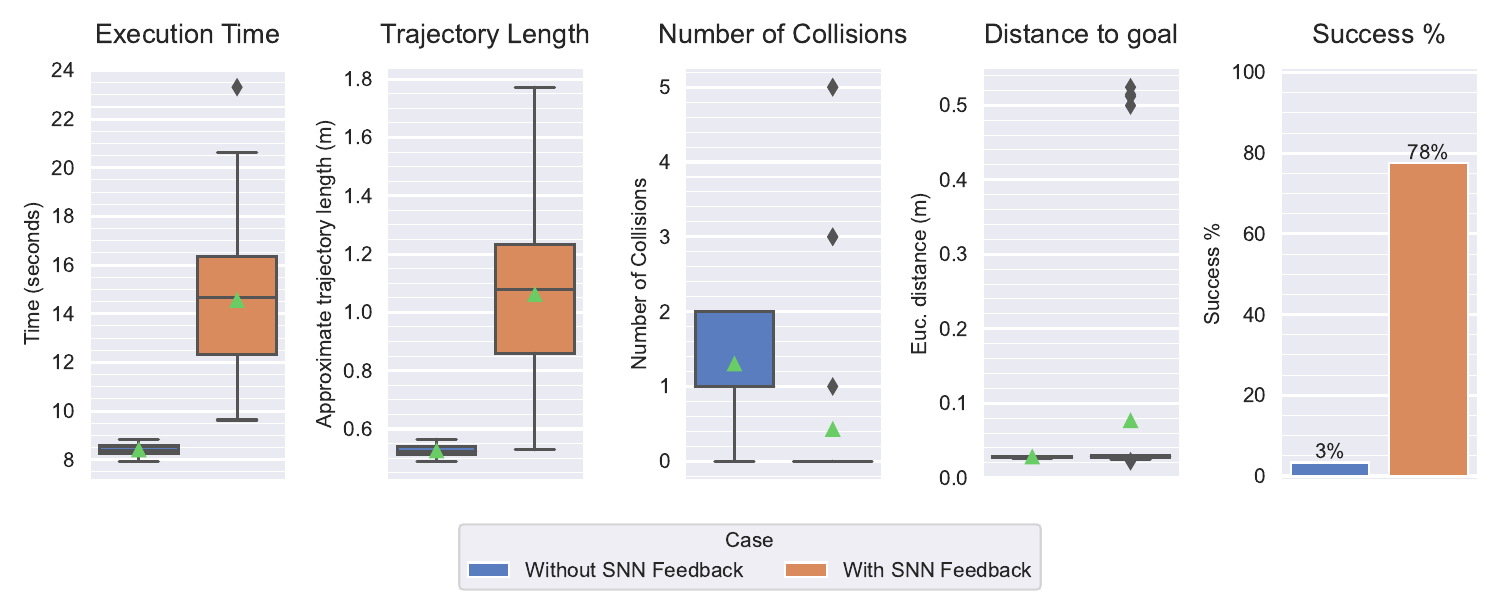}
        }
        \caption{\centering Scenario 19: \{T2, Store, Brick, Box, Med.\}}
        \label{testing_scenarios_12_to_31_metrics:scenario_19}
    \end{subfigure}%
    \\
    \begin{subfigure}{0.49\linewidth}
        \centering
        \frame{
            \includegraphics[width=\columnwidth]{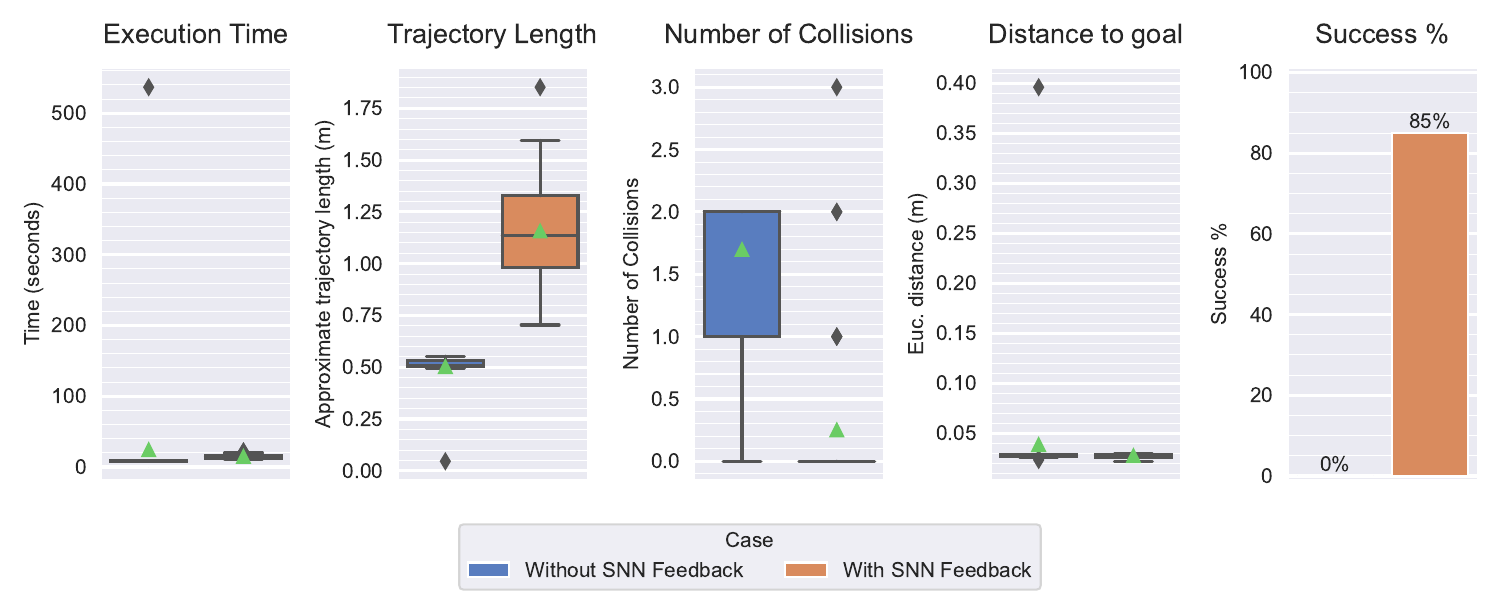}
        }
        \caption{\centering Scenario 20: \{T2, Kitchen, Y-B, Box, High\}}
        \label{testing_scenarios_12_to_31_metrics:scenario_20}
    \end{subfigure}%
    \hspace{0.5em}
    \begin{subfigure}{0.49\linewidth}
        \centering
        \frame{
            \includegraphics[width=\columnwidth]{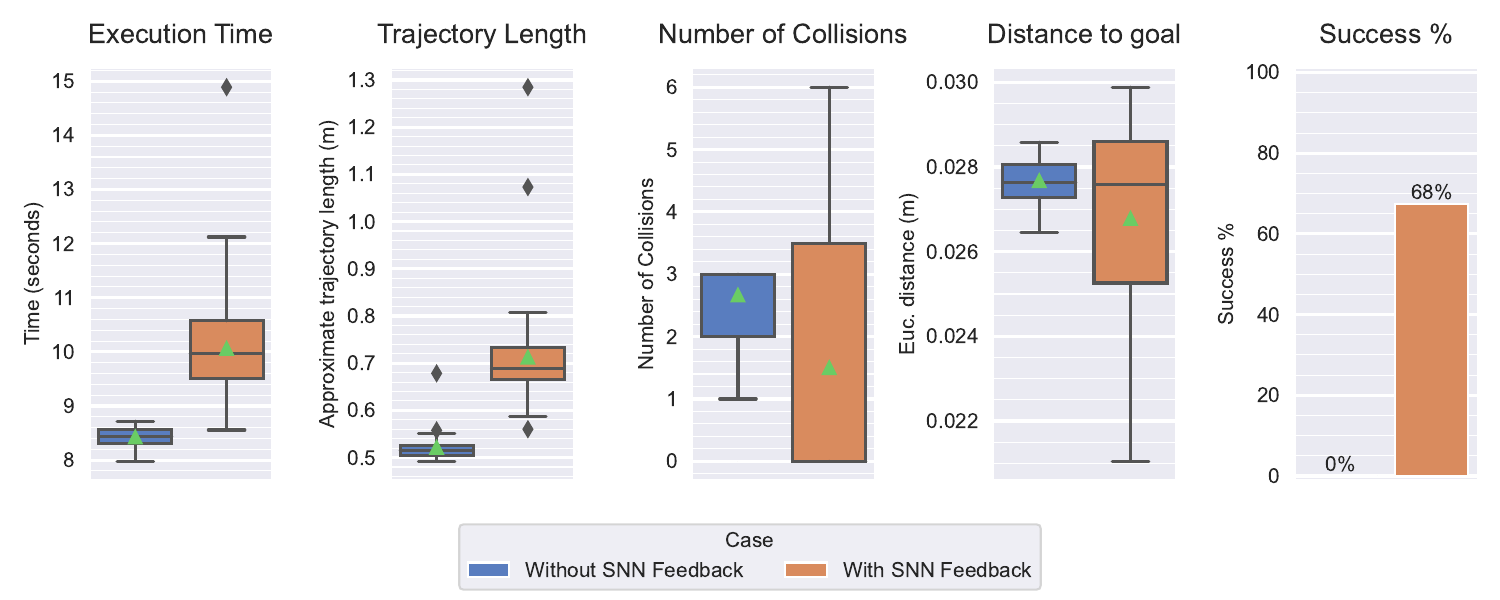}
        }
        \caption{\centering Scenario 21: \{T2, Empty, Red, Spiky Sphere, Med.\}}
        \label{testing_scenarios_12_to_31_metrics:scenario_21}
    \end{subfigure}%
    \\
    \begin{subfigure}{0.49\linewidth}
        \centering
        \frame{
            \includegraphics[width=\columnwidth]{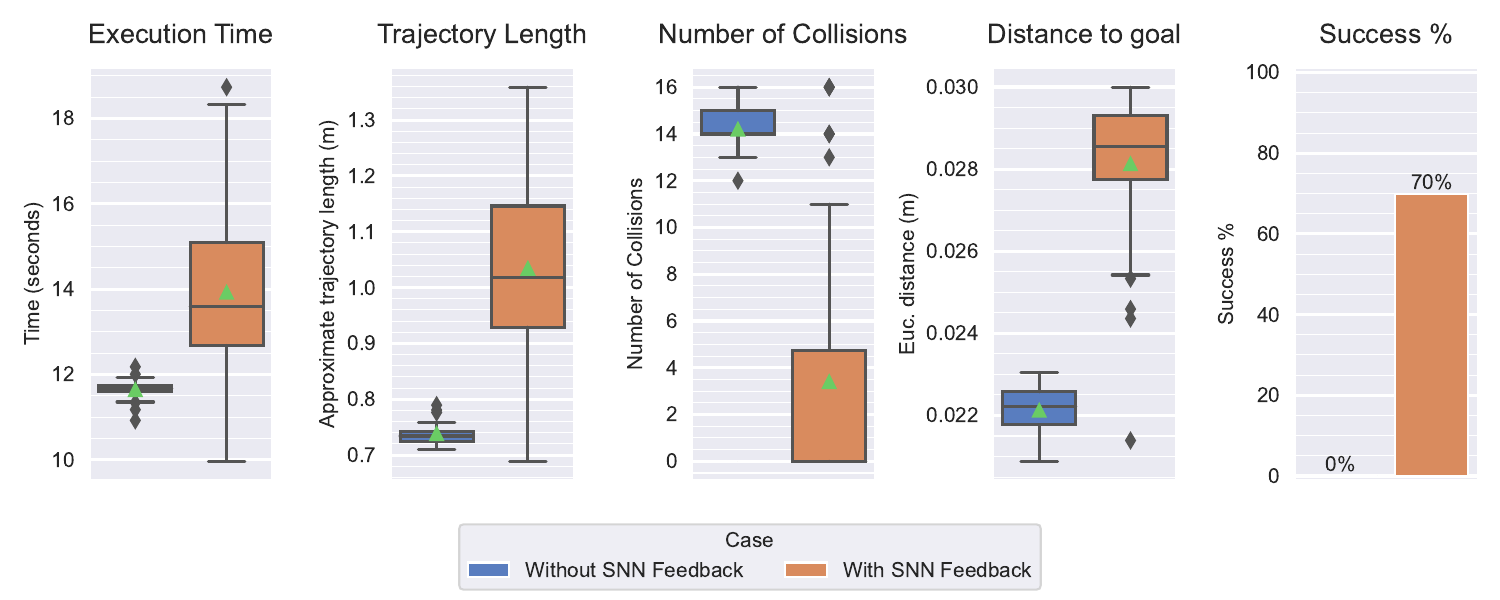}
        }
        \caption{\centering Scenario 22: \{T3, Empty, Red\}}
        \label{testing_scenarios_12_to_31_metrics:scenario_22}
    \end{subfigure}%
    \hspace{0.5em}
    \begin{subfigure}{0.49\linewidth}
        \centering
        \frame{
            \includegraphics[width=\columnwidth]{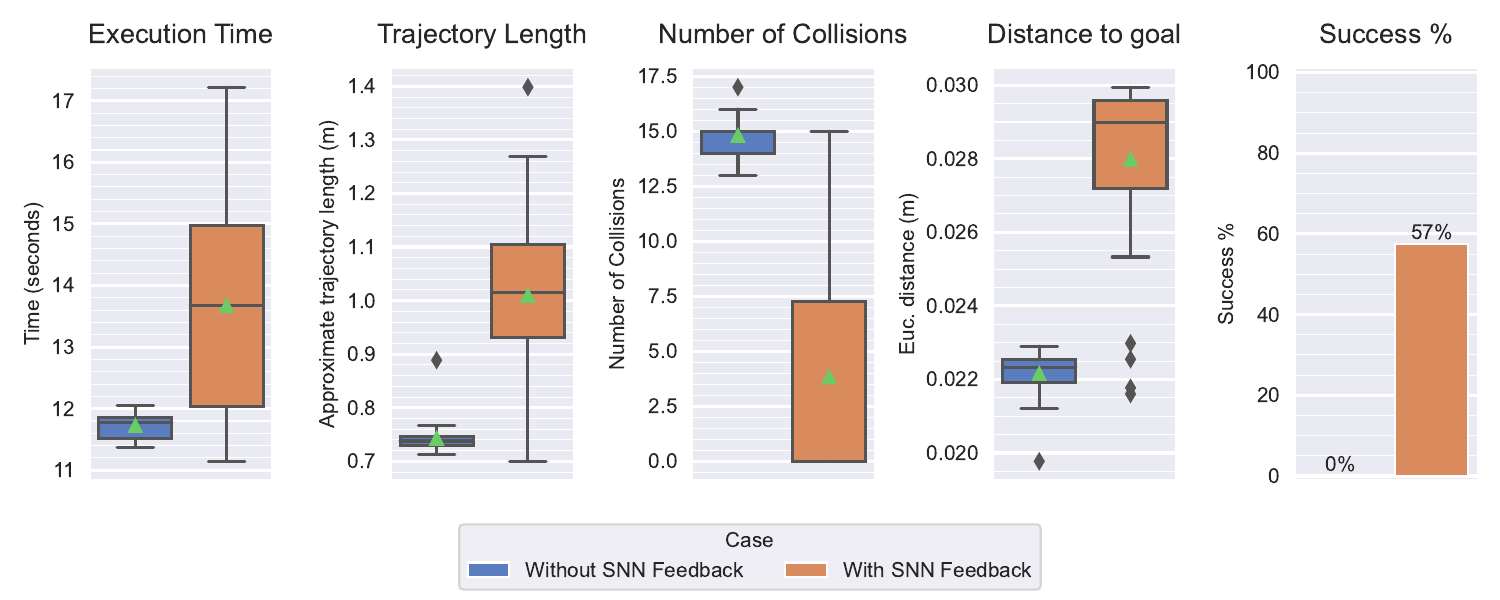}
        }
        \caption{\centering Scenario 23: \{T3, Empty, Y-B\}}
        \label{testing_scenarios_12_to_31_metrics:scenario_23}
    \end{subfigure}%
    \\
    \begin{subfigure}{0.49\linewidth}
        \centering
        \frame{
            \includegraphics[width=\columnwidth]{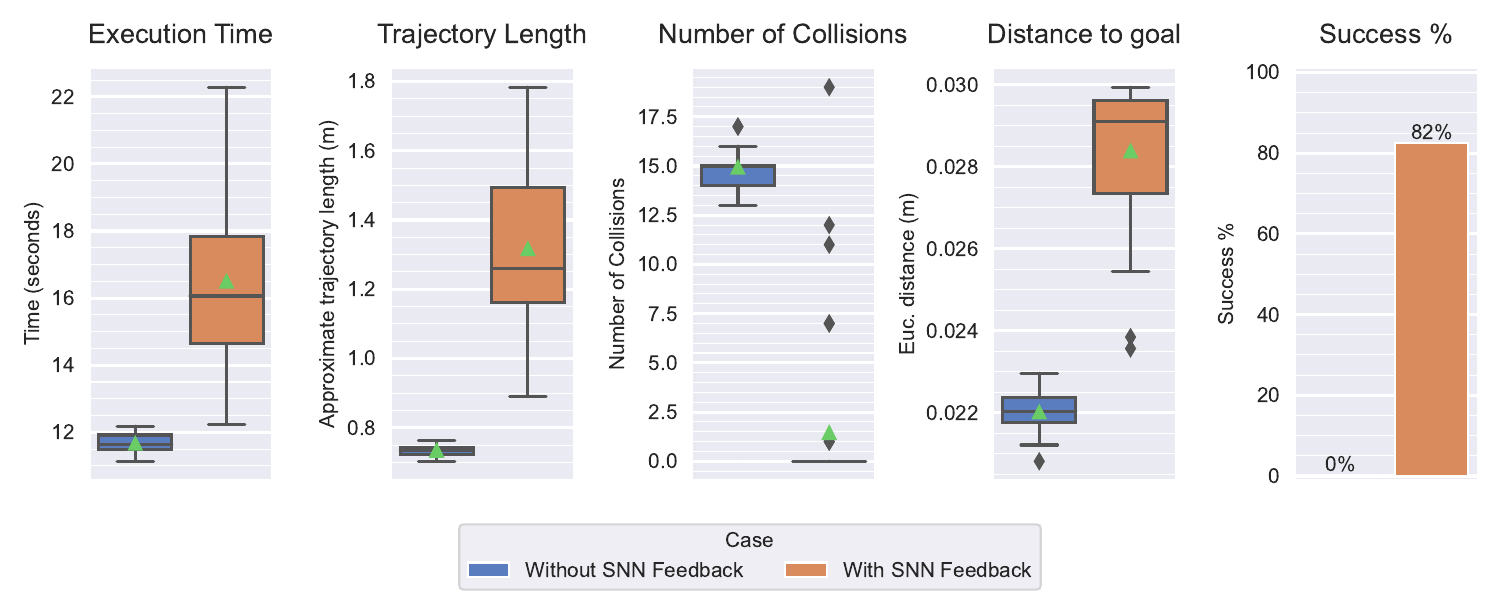}
        }
        \caption{\centering Scenario 24: \{T3, Kitchen, White\}}
        \label{testing_scenarios_12_to_31_metrics:scenario_24}
    \end{subfigure}%
    \hspace{0.5em}
    \begin{subfigure}{0.49\linewidth}
        \centering
        \frame{
            \includegraphics[width=\columnwidth]{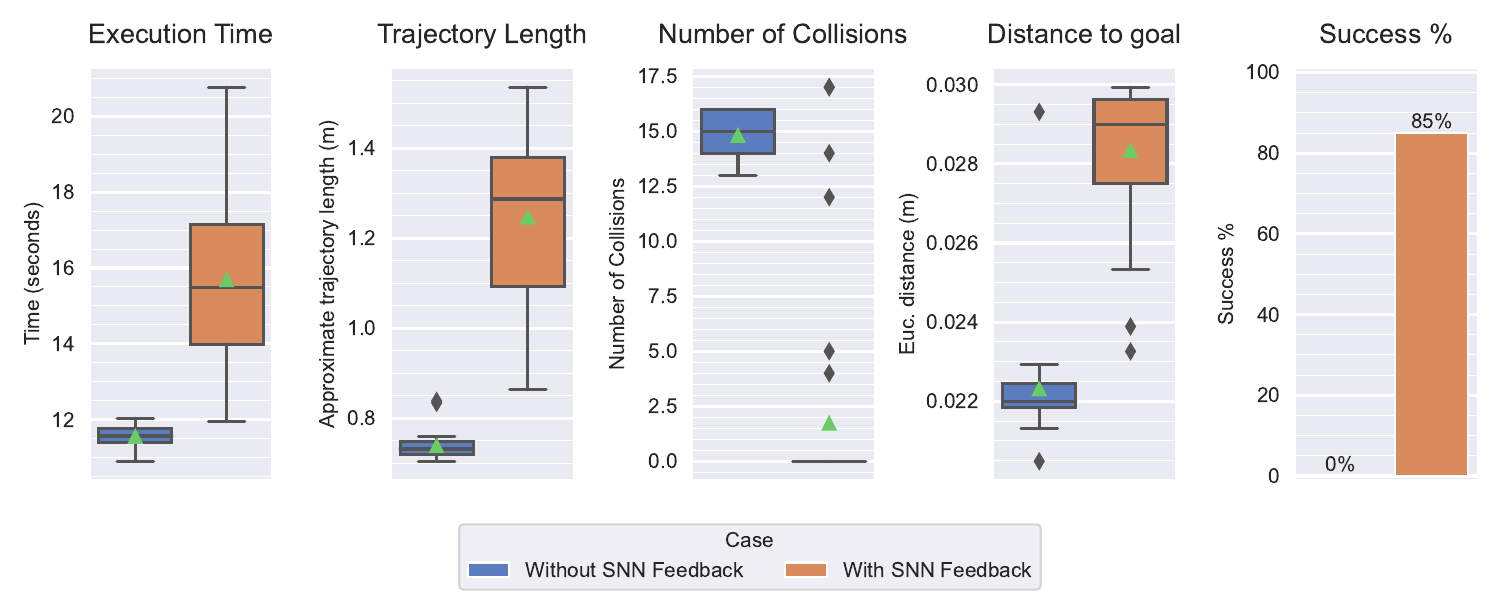}
        }
        \caption{\centering Scenario 25: \{T3, Kitchen, Red\}}
        \label{testing_scenarios_12_to_31_metrics:scenario_25}
    \end{subfigure}%
    \\
    \begin{subfigure}{0.49\linewidth}
        \centering
        \frame{
            \includegraphics[width=\columnwidth]{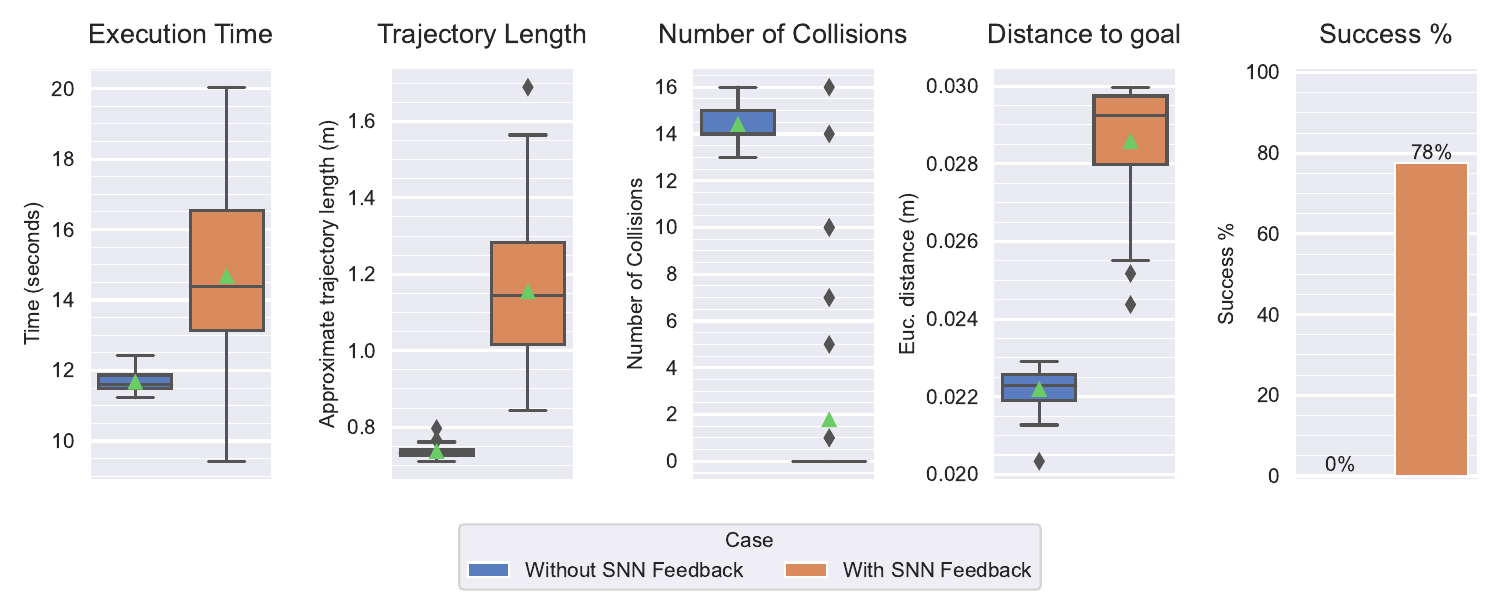}
        }
        \caption{\centering Scenario 26: \{T3, Store, Red\}}
        \label{testing_scenarios_12_to_31_metrics:scenario_26}
    \end{subfigure}%
    \hspace{0.5em}
    \begin{subfigure}{0.49\linewidth}
        \centering
        \frame{
            \includegraphics[width=\columnwidth]{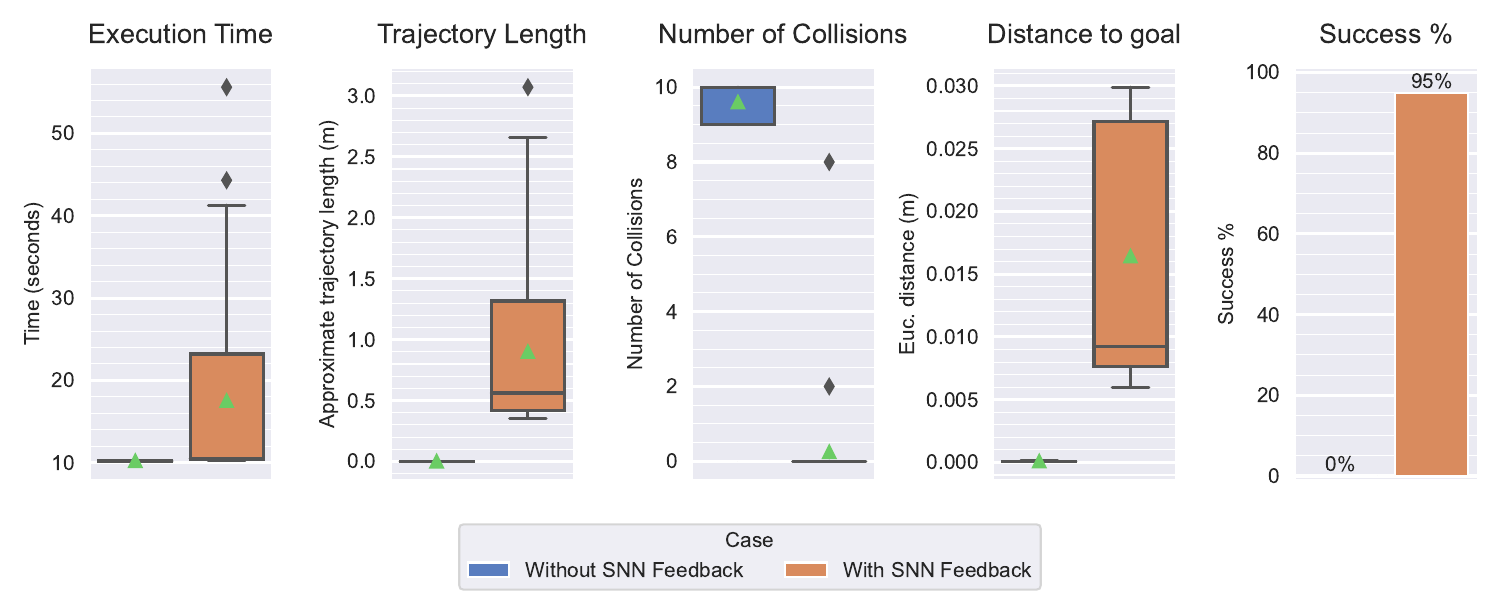}
        }
        \caption{\centering Scenario 27: \{T4, Store, Y-B, Buckyball, Low\}}
        \label{testing_scenarios_12_to_31_metrics:scenario_27}
    \end{subfigure}%
    \caption{Quantitative metric results without vs. with SNN feedback (best parameter set): testing scenarios 16-27. (cont.)}
    \label{testing_scenarios_12_to_31_metrics}
\end{figure*}

\begin{figure*}\ContinuedFloat
    \centering
        \begin{subfigure}{0.49\linewidth}
            \centering
            \frame{
                \includegraphics[width=\columnwidth]{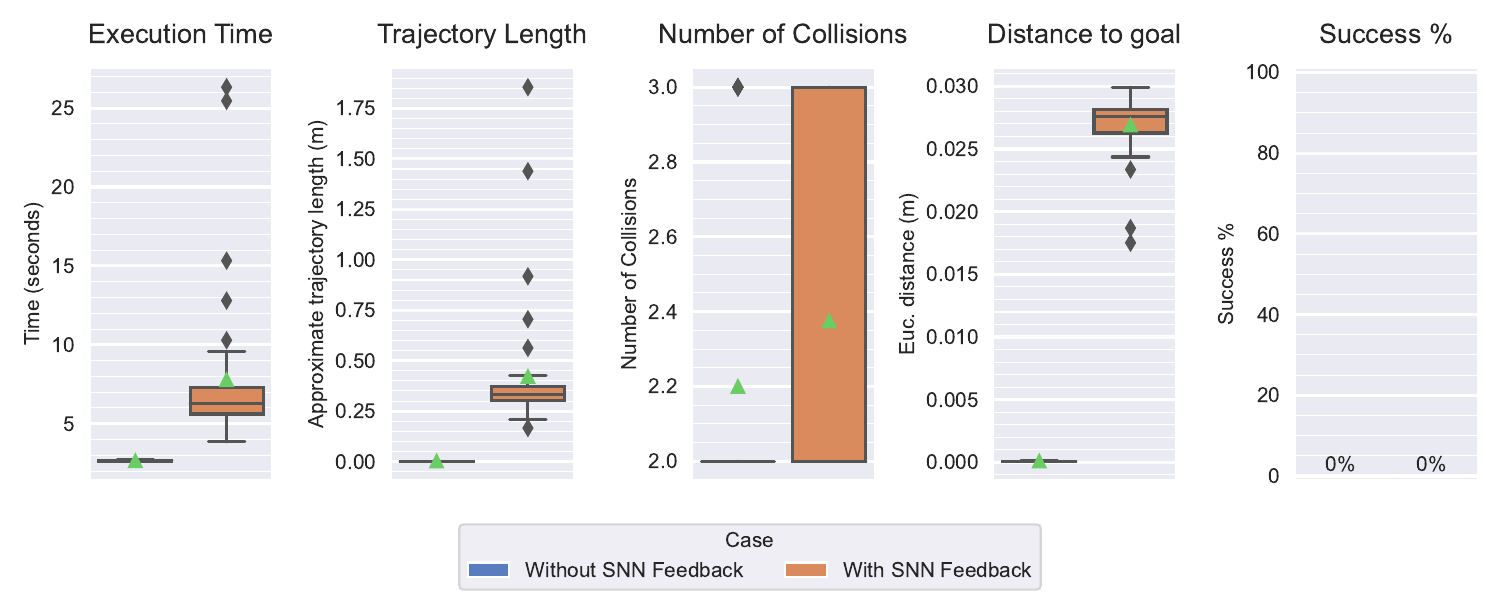}
            }
            \caption{\centering Scenario 28: \{T4, Store, Red, Buckyball, High\}}
            \label{testing_scenarios_12_to_31_metrics:scenario_28}
        \end{subfigure}%
        \hspace{0.5em}
        \begin{subfigure}{0.49\linewidth}
            \centering
            \frame{
                \includegraphics[width=\columnwidth]{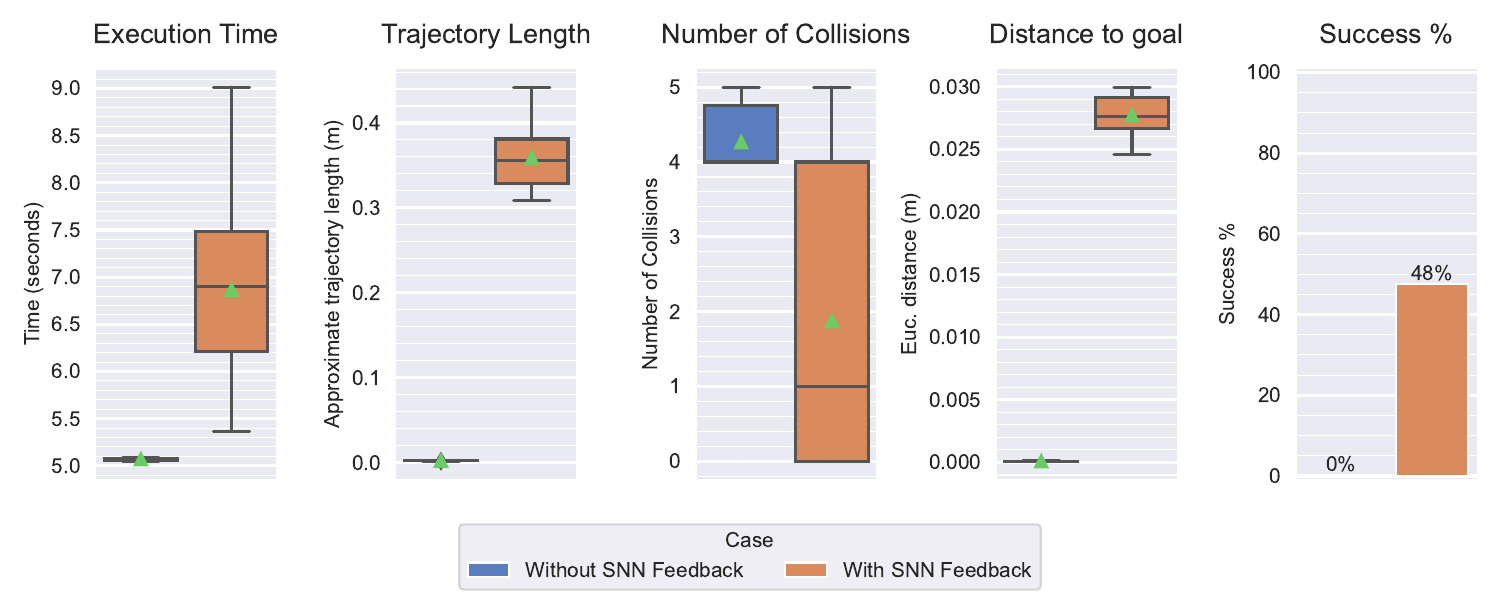}
            }
            \caption{\centering Scenario 29: \{T4, Empty, Y-B, Spiky Sphere, Med.\}}
            \label{testing_scenarios_12_to_31_metrics:scenario_29}
        \end{subfigure}%
        \\
        \begin{subfigure}{0.49\linewidth}
            \centering
            \frame{
                \includegraphics[width=\columnwidth]{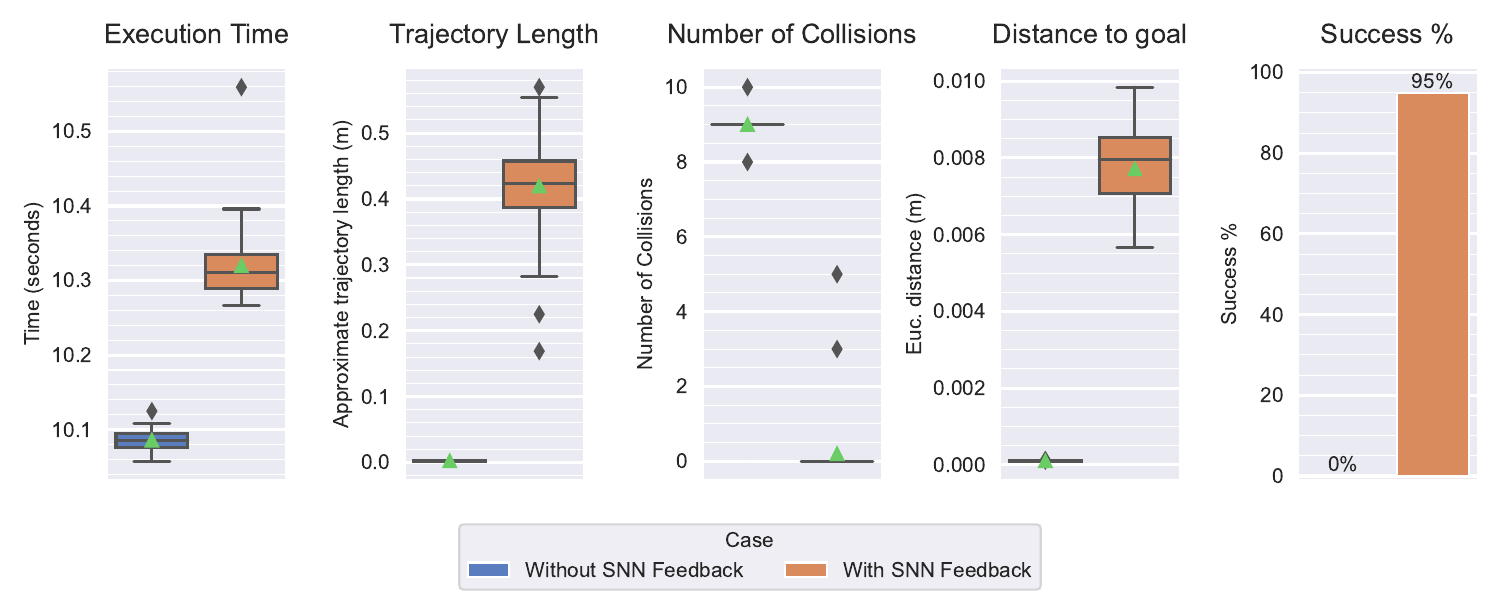}
            }
            \caption{\centering Scenario 30: \{T4, Empty, Y-B, Box, Low\}}
            \label{testing_scenarios_12_to_31_metrics:scenario_30}
        \end{subfigure}%
        \hspace{0.5em}
        \begin{subfigure}{0.49\linewidth}
            \centering
            \frame{
                \includegraphics[width=\columnwidth]{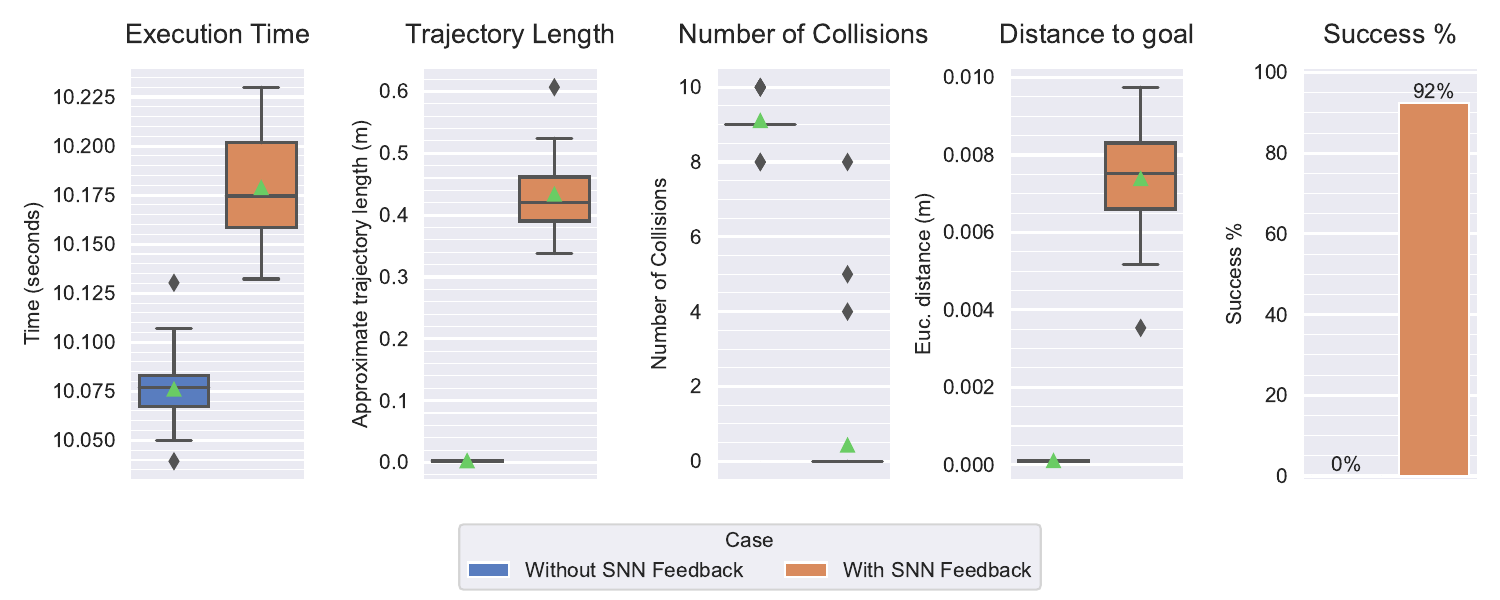}
            }
            \caption{\centering Scenario 31: \{T4, Empty, Brick, Buckyball, Low\}}
            \label{testing_scenarios_12_to_31_metrics:scenario_31}
        \end{subfigure}%
    \caption{Quantitative metric results without vs. with SNN feedback (best parameter set): testing scenarios 28-31. (cont.)}
    \label{testing_scenarios_12_to_31_metrics}
    \vspace{6.3in}
\end{figure*}

\end{document}